\documentclass{article}

\usepackage[main, final]{neurips_2026}

\usepackage[utf8]{inputenc} %
\usepackage[T1]{fontenc}    %
\usepackage{hyperref}       %
\usepackage{url}            %
\usepackage{booktabs}       %
\usepackage{amsfonts}       %
\usepackage{amssymb}        %
\usepackage{nicefrac}       %
\usepackage{microtype}      %
\usepackage{xcolor}         %
\usepackage{amsmath}
\usepackage{amsthm}
\usepackage{subcaption}
\usepackage{color,xcolor,colortbl}
\usepackage{booktabs}
\usepackage{graphicx}
\usepackage{multicol}
\usepackage{multirow}
\usepackage{cleveref}
\usepackage{paralist}

\DeclareMathOperator*{\argmax}{arg\,max}

\newcommand{\xv}{\mathbf{x}}
\newcommand{\yv}{\mathbf{y}}
\newcommand{\zv}{\mathbf{z}}
\newcommand{\uv}{\mathbf{u}}

\newcommand{\ind}{\mathbf{1}}
\newcommand{\mask}{\mathbf{M}}

\newcommand{\multirowcell}[1]{\begin{tabular}[c]{@{}c@{}}#1\end{tabular}}

\newtheorem{theorem}{Theorem}

\newtheorem{proposition}{Proposition}

\newcommand{\ecoparagraph}[1]{\noindent\textbf{#1}}

\title{Uncertainty Quantification for\\ 
Large Language Diffusion Models
}

\author{
Artem Vazhentsev\textsuperscript{1} \quad 
Vladislav Smirnov\textsuperscript{1} \quad 
David Li\textsuperscript{1} \enspace 
\\
\bf
Maxim Panov\textsuperscript{1} \quad 
Timothy Baldwin\textsuperscript{1,2} \enspace 
Artem Shelmanov\textsuperscript{1} \\
\textsuperscript{1}MBZUAI \; 
\textsuperscript{2}The University of Melbourne \; 
\\
\href{mailto:Artem.Vazhentsev@mbzuai.ac.ae}{artem.vazhentsev@mbzuai.ac.ae} ~~ 
\href{mailto:artem.shelmanov@mbzuai.ac.ae}{artem.shelmanov@mbzuai.ac.ae}
}
\begin{document}

\maketitle

\begin{abstract}
  Large Language Diffusion Models (LLDMs) are emerging as an alternative to autoregressive models, offering faster inference through higher parallelism. Similar to autoregressive LLMs, they remain prone to hallucinations, making reliable uncertainty quantification (UQ) crucial for safe deployment. However, existing UQ methods are fundamentally misaligned with this new paradigm: they assume autoregressive factorization or use expensive repeated sampling, negating the efficiency of LLDMs. In this work, we present the first systematic study of UQ for LLDMs and propose lightweight, zero-shot uncertainty signals derived from the iterative denoising process, leveraging intermediate generations, token remasking dynamics, and denoising complexity. We further adapt a state-of-the-art UQ method %
  to LLDMs by combining masked diffusion likelihoods with trajectory-based semantic dissimilarity. 
  We prove that expected trajectory dissimilarity lower bounds the masked diffusion training objective, which motivates its usage as an uncertainty score. Comprehensive experiments across three tasks, eight datasets, and four models
  show that our method achieves a great cost-performance 
  trade-off: it approaches the strongest sampling-based baselines while incurring up to 100x lower computational overhead. Our work demonstrates that LLDMs can deliver both fast inference and reliable hallucination detection simultaneously.\footnote{\url{http://lldm-uncertainty.nlpresearch.group/}}
\end{abstract}

\section{Introduction}

  Large language models (LLMs) are known to hallucinate \citep{huang2025survey}, producing fluent but factually incorrect outputs, making reliable uncertainty quantification (UQ) a critical problem \citep{kuhn2023semantic}. While most prior work on UQ has focused on autoregressive models, Large Language Diffusion Models (LLDMs) have recently emerged as a competitive alternative \citep{ye2025dream,llada,llada2}, offering faster generation through semi-autoregressive decoding, where multiple tokens within a block are generated in parallel and iteratively refined \citep{wu2025fastdllmtrainingfreeaccelerationdiffusion}.

  However, existing UQ methods are misaligned with this diffusion-based paradigm. Standard approaches are either tailored to token-by-token autoregressive generation \citep{malinin2020uncertainty} or computationally expensive due to repeated sampling \citep{duan-etal-2024-shifting,vashurin2025cocoa}, making them poorly suited for LLDMs, where efficiency is the main objective \citep{scalingdllms}.

\begin{figure*}[t!]
    \centering
    \includegraphics[trim={0.7cm 0.2cm 0.6cm 0.2cm},clip,width=\linewidth]{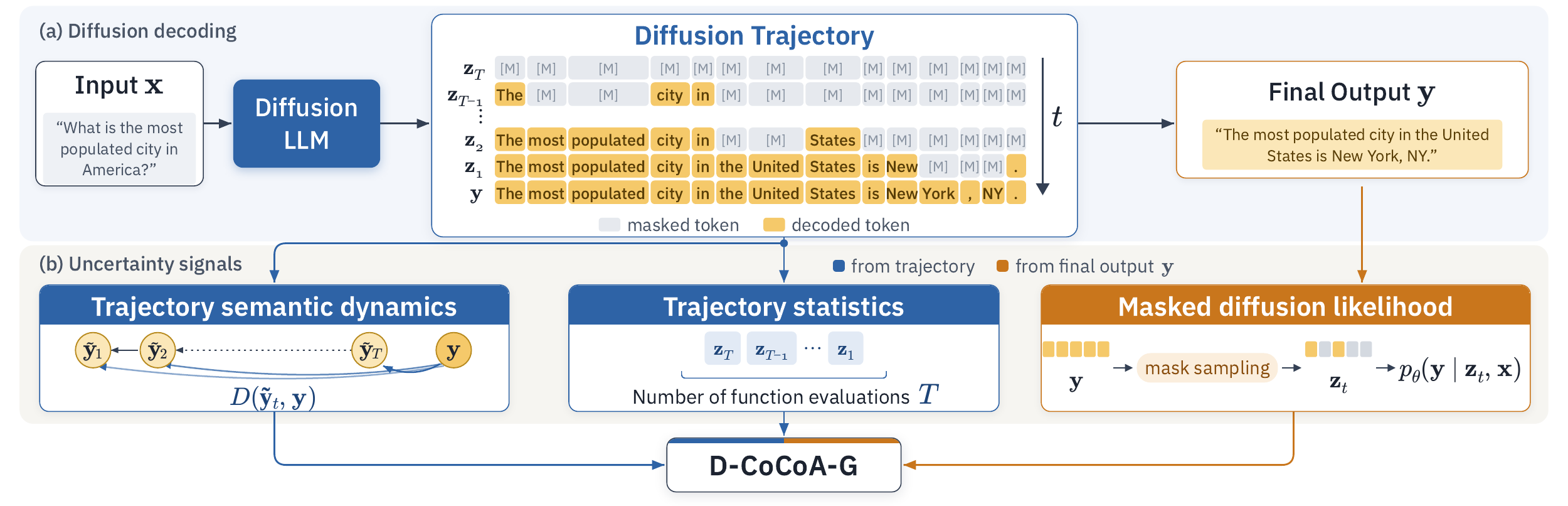}
    \caption{
    \textbf{D-CoCoA-G overview.} Given a prompt $\xv$, the LLDM generates an output $\yv$ via iterative denoising states $\zv_T, \dots, \zv_1$. D-CoCoA-G combines three diffusion-specific signals: (1) trajectory semantic instability, measuring semantic dissimilarity $D(\tilde{\yv}_t, \yv)$ between $\argmax$-decoded intermediate states and the final output; (2) trajectory statistics, capturing generation complexity through the number of function evaluations; and (3) masked diffusion likelihood, approximating the conditional log-likelihood.}
    \label{fig:d_cocoa_scheme}
  \end{figure*}

  In this work, we investigate diffusion-specific signals for UQ and introduce uncertainty measures that exploit diffusion trajectories, masking dynamics, and generation complexity, as illustrated in Figure~\ref{fig:d_cocoa_scheme}. These signals enable principled UQ that is both effective and efficient. Our results show that LLDMs offer a dual advantage: fast inference and reliable UQ, without relying on the costly sampling procedures typically required by autoregressive UQ methods \citep{vashurin-etal-2025-benchmarking}.

  We summarize our main \textbf{contributions} as follows:
\begin{compactenum}

    \item \textbf{Diffusion-specific uncertainty signals.}
    We introduce a set of novel uncertainty signals tailored to LLDMs, including trajectory semantic instability, trajectory statistics, and remasking behavior (see Section~\ref{sec:diff_signals}).%

    \item \textbf{Theoretical grounding.}
    We formally show that expected trajectory dissimilarity, one of the strongest UQ signals proposed in this work, is a principled surrogate for the discrete masked diffusion training objective, motivating its use as an uncertainty score (see Theorem~\ref{thm:dissim-lower-bound}).

    \item \textbf{D-CoCoA: an effective and efficient UQ method for LLDMs.}
    We adapt CoCoA, a state-of-the-art UQ method, to LLDMs by replacing its log-likelihood component with a diffusion-aware surrogate and its sampling-based component with intermediate states generated along the denoising trajectory (see Section~\ref{sec:d_methods}). 

    \item \textbf{First large-scale systematic empirical study of UQ for LLDMs.}
    We conduct experiments across diverse tasks and datasets, showing that our diffusion-aware D-CoCoA methods achieve a superior performance-efficiency trade-off compared to both standard sampling-based baselines and purely diffusion-specific signals (see Section~\ref{sec:experiments}). Our results show that LLDMs do not require a compromise between fast inference and reliable UQ.

\end{compactenum}

\section{Related Work}
UQ for text generation has been extensively studied for autoregressive LLMs~\citep{shelmanov-etal-2025-uncertainty}.
White-box methods leverage internal model signals, aggregating token-level probabilities, predictive entropy, or likelihoods to estimate sequence-level uncertainty~\citep{malinin2020uncertainty, fomicheva-etal-2020-unsupervised, rauq}. Some works suggest training supervised hallucination detectors based on hidden states~\citep{azaria-mitchell-2023-internal, su-etal-2024-unsupervised, vazhentsev-etal-2025-token} or attention maps~\citep{chuang-etal-2024-lookback, vazhentsev-etal-2025-unconditional, shelmanov-etal-2025-head} of LLMs. ReProbe uses lightweight internal-state probes to verify intermediate reasoning steps and guide test-time compute scaling~\citep{ni-etal-2026-reprobe}. Black-box UQ methods are usually based on the consistency of generations obtained through repetitive sampling. Approaches such as Semantic Entropy~\citep{kuhn2023semantic, farquhar2024detecting}, SAR~\citep{duan-etal-2024-shifting}, and CoCoA~\citep{vashurin2025cocoa} combine consistency with probability-based signals. Beyond output-level UQ, \citet{papamarkou2026position,papamarkou-etal-2026-bayesian} advocate Bayesian decision principles for orchestrating LLM agents and tools under uncertainty. 

Recently, diffusion models have emerged as an efficient approach to natural language generation by alleviating the autoregressive decoding bottleneck \citep{uniformdllms, sahoo2025the, sahoo2026esotericlanguagemodelsbridging}. 
Masked diffusion-based models such as LLaDA~\citep{llada}, Dream~\citep{ye2025dream}, and ReFusion~\citep{li2026refusion} already demonstrate competitive generation quality with autoregressive counterparts %
such as LLaMA-3.1~\citep{dubey2024llama}.

While UQ has been studied for diffusion-based image generation \citep{pmlr-v286-jazbec25a,kou2024bayesdiff,pmlr-v244-berry24a},
these methods
usually do not directly transfer to text generation, where intermediate states are discrete token predictions and semantic meaning is highly non-linear and non-differentiable. A few works leveraged denoising trajectories in LLDMs for hallucination detection~\citep{chang2026tracedet,qian2026dynhdhallucinationdetectiondiffusion}. However, these approaches are strictly supervised and consequently lack the generalization required to seamlessly adapt to new tasks or domains.

In summary, no prior work has introduced a systematic UQ framework for LLDMs. We address this gap by proposing a family of unsupervised methods that explicitly leverage the distinctive properties of the %
iterative denoising process for UQ. In particular, we show that intermediate denoising trajectories, diffusion statistics, masking dynamics, and denoising complexity can be repurposed into effective and efficient unsupervised uncertainty scores.

\section{Methodology}
\label{sec:methods}

\subsection{Background}

\ecoparagraph{Generation in LLDMs.}
  Let $\xv$ denote an input prompt and $\yv = (y_1, \dots, y_{|\yv|})$ a generated sequence.
  Unlike autoregressive LLMs, which generate text sequentially from left to right, LLDMs generate a sequence through an iterative denoising process. 
  Generation starts from a fully corrupted state $\zv_T = (\mask, \dots, \mask),$ where $\mask$ is a special $\left[\mathrm{MASK} \right]$ token. The model then gradually reconstructs a clean text sequence by applying a learned reverse denoising process:
  \[
    \zv_T \sim q(\zv_T),
    \qquad
    \zv_{t-1} \sim p_\theta(\zv_{t-1}\mid \zv_t,\xv), 
    \qquad 
    t=T,T-1,\ldots,1.
  \]
  Here, $\zv_t$ denotes the intermediate sequence at step $t$, $\zv_0=\yv$ is the final generated text, and $p_\theta$ is a neural denoising model parameterized by $\theta$. This non-autoregressive inference revises multiple tokens in parallel, making diffusion-based generation faster but conceptually distinct from next-token prediction implemented in standard LLMs.
  
  During training, the model learns to recover the original sequence $\yv$ from a masked corruption $\zv_t$. A common training objective is a denoising loss:
  \[
    \mathcal{L}(\theta)
    =
    - \mathbb{E}_{t,\xv,\yv,\zv_t}
    \Biggl[\frac{1}{t} \sum_{i=1}^{|\yv|}
    \ind[z_{t, i} = \mask]\log p_\theta(y_i \mid \zv_t, \xv)
    \Biggr].
  \]
\ecoparagraph{UQ methods for hallucination detection}
 of autoregressive LLMs commonly exploit (1) white-box signals such as token-level probabilities $p(y_l \mid \mathbf{y}_{<l}, \mathbf{x})$, aggregated into sequence-level scores, e.g., sequence likelihood  \mbox{$P(\mathbf{y} \mid \mathbf{x}) = \prod_{l=1}^{L} p(y_l \mid \mathbf{y}_{<l}, \mathbf{x})$}, and (2) the consistency of sampled generations $\left\{\mathbf{y}^{(i)}\right\}_{i=1}^{M}$, $\mathbf{y}^{(i)} \sim p(\mathbf{y} \mid \mathbf{x})$, where uncertainty is computed through lexical and semantic dissimilarity measures between samples~\citep{lin2024generating,zhang2024luq,nikitin2024kernel}. (3) Some methods combine information from sampling with the token probability distributions estimated by LLMs: Monte Carlo Normalized Sequence Entropy~\citep{malinin2020uncertainty}, Semantic Entropy~\citep{kuhn2023semantic,farquhar2024detecting}, and a recently proposed CoCoA method~\citep{vashurin2025cocoa}. However, two major problems arise when trying to apply these signals to UQ of LLDM generations.

  First, LLDMs do not natively use the left-to-right autoregressive factorization of standard LLMs, so token-level probabilities $p(y_l \mid \mathbf{y}_{<l}, \mathbf{x})$ are not directly available. Moreover, diffusion models generally do not admit an exact tractable likelihood and are instead optimized through a lower-bound objective, making their resulting probabilities unreliable as estimates of sequence likelihood. 
  
  Second, sampling multiple generations from LLMs is impractical in general and for LLDMs in particular, as these models are specifically designed for fast and efficient generation~\citep{wu2025fastdllmtrainingfreeaccelerationdiffusion,scalingdllms}. 
  Increasing inference cost by a large multiplicative factor through repeated sampling undermines one of their main advantages.

  We argue that the iterative denoising process provides additional signals for UQ. For example, the structured trajectory of intermediate states $\{\zv_t\}_{t=1}^{T}$ reflects how the generation evolves over time and naturally captures variation in the model's intermediate predictions. We explore these diffusion-specific signals derived from the model's generation trajectory for UQ in the following section.

\subsection{Diffusion-Specific Signals for Uncertainty Quantification}
\label{sec:diff_signals}
  We propose a set of uncertainty scores that directly exploit the denoising generation process, including generation likelihood estimates from LLDMs, trajectory statistics, and remasking dynamics.

\ecoparagraph{Negative log-likelihood estimation.}
  \citet{llada} propose a Monte Carlo estimator of an upper-bound ELBO-like surrogate for
  the negative log-likelihood based on the masked diffusion objective. This
  quantity can serve as an uncertainty score for generated sequences. Let
  $N_{mc}$ denote the number of Monte Carlo samples. 
  For each $m \in \{1,\dots,N_{mc}\}$, we obtain $\zv_m$ by sampling $l_m \sim \mathrm{Uniform}\{1,\dots,|\yv|\}$ tokens from $\yv$ without replacement for masking. The Monte Carlo masked negative log-likelihood (MCNLL) estimator is then defined as:
  \begin{equation}
    \uv_{\mathrm{MCNLL}}(\xv,\yv)
    =
    -\frac{1}{N_{mc}}
    \sum_{m=1}^{N_{mc}}
    \frac{|\yv|}{l_m}
    \sum_{i=1}^{|\yv|} \ind [z_{m, i} = \mask]
    \log p_\theta(y_i \mid \zv_m, \xv).
  \end{equation}
  A length-normalized variant,
  $\uv_{\mathrm{MCNLL}}^{\mathrm{norm}}(\xv,\yv)
  =
  \uv_{\mathrm{MCNLL}}(\xv,\yv)/|\yv|$,
  enables a more consistent comparison across outputs of varying lengths.

\ecoparagraph{Likelihood-based signals from the denoising trajectory.}
  Since the denoising process produces a trajectory of intermediate states with probability distributions, it naturally enables UQ via statistics computed along this trajectory. For semi-autoregressive LLDMs, $\yv$ is partitioned into $B$ blocks of lengths $L_b$, and $B_{\mathrm{valid}}$ denotes the number of non-empty generated blocks (i.e., blocks containing at least one non-special token). We denote $\tilde{y}_{b,t,k} = \argmax_{y'\in\mathcal{V}}\, p_\theta(y' \mid \zv_{<b,t,k}, \xv)$ as the $k$-th argmax token over the vocabulary $\mathcal{V}$ at step $t$ in block $b$, and $\zv_{<b,t,k}$ as the preceding context. Below, we construct several uncertainty scores that take advantage of the diffusion-based trajectory.

  (1) The average trajectory negative log-likelihood aggregates the log-probabilities of predicted tokens across the denoising trajectory:
  \begin{equation}
    \uv_{\mathrm{TrajNLL}}(\xv,\yv)
    =
    \frac{1}{B_{\mathrm{valid}}\,T}
    \sum_{b,t}
    \frac{1}{L_b}
    \sum_{k=1}^{L_b}
    \log p_{\theta}(\tilde{y}_{b,t,k} \mid \zv_{<b,t,k}, \xv).
  \end{equation}
  (2) The average predictive entropy captures the average uncertainty of token predictions along the denoising trajectory:
  \begin{equation}
    \uv_{\mathrm{TrajEnt}}(\xv,\yv)
    =
    \frac{1}{B_{\mathrm{valid}}\,T}
    \sum_{b,t}
    \frac{1}{L_b}
    \sum_{k=1}^{L_b}
    H\!\bigl(p_{\theta}(\cdot \mid \zv_{<b,t,k}, \xv)\bigr),
  \end{equation}
  where $H(p) = -\sum_{v \in \mathcal{V}} p(v)\log p(v)$ denotes the token-level entropy over the vocabulary $\mathcal{V}$. %

  (3) We further consider the confidence of the model at the moment each token is committed. If tokens are committed with low probability on average, it might indicate high uncertainty.
  Let $t_{k}$ denote the diffusion step at which the $k$-th token transitions from a masked to an unmasked (committed) state. The average negative log-likelihood at final commitment across all token positions is:
  \begin{equation}
    \uv_{\mathrm{CommitNLL}}(\xv,\yv)
    =
    -\frac{1}{|\yv|}\sum_{b,k}\,\log p_\theta\!\left(y_{b,k} \mid \zv_{<b,t_{k},k}, \xv\right).
  \end{equation}
\ecoparagraph{Denoising trajectory-based statistics} also provide useful uncertainty signals.

(1) The Number of Function Evaluations (NFE) denotes the number of LLDM forward passes required to generate an output. While upper-bounded by the maximum number of diffusion steps, NFE is often smaller in practice due to confidence-based remasking and early stopping. It therefore serves as a natural proxy for denoising complexity.
  Formally, let $T(\yv)$ denote the actual NFE used for generating output $\yv$:
  \begin{equation}
    \uv_{\mathrm{NFE}}(\xv,\yv)
    =
    T(\yv).
  \end{equation}
  
  (2) Remasking behavior itself captures instability at the token level. Let $\mathcal{R}_t$ be the set of tokens remasked at a diffusion step $t$ during generation. Then we consider 
  the average number of remasked tokens across all steps:
  \begin{equation}
    \uv_{\mathrm{Remask}}(\xv,\yv)
    = \frac{1}{T}\sum_{t=1}^T|\mathcal{R}_t|.
  \end{equation}
  
  (3) Token-level prediction instability is the average number of token flips, i.e., consecutive diffusion steps at which the most probable token prediction changes. Frequent prediction changes across steps reflect token-level instability during the denoising process and serve as a proxy for uncertainty. Let $T_b$ be the number of diffusion steps for the block $b$:
  \begin{equation}
    \uv_{\mathrm{FlipCount}}(\xv,\yv)
    =
    \frac{1}{|\yv|}
    \sum_{b,k}
    \sum_{t=1}^{T_b - 1}
    \ind\!\bigl[\tilde{y}_{b,t,k} \neq \tilde{y}_{b,t+1,k}\bigr].
  \end{equation}
\ecoparagraph{Denoising trajectory semantic instability.}
  We introduce a \underline{novel and highly effective} UQ signal that exploits the diffusion-based generation process for zero-shot UQ, avoiding the prohibitive cost of repeated sequence sampling. This signal measures the semantic consistency between intermediate trajectory states and the final generation, capturing the extent of semantic shift before the final output. Accordingly, we define semantic instability across the %
  denoising process via the \textbf{average dissimilarity (AD)} between intermediate states $\tilde{\yv}_t$ (obtained via $\argmax$ decoding) and the final generation $\yv$:
  \begin{equation}
    \uv_{\mathrm{AD}}(\xv,\yv)
    =
    \frac{1}{T}
    \sum_{t=1}^{T}
    D(\tilde{\yv}_t, \yv).
  \end{equation}

  Theorem~\ref{thm:dissim-lower-bound} (see below) provides theoretical support for using $\uv_{\mathrm{AD}}$ as a principled uncertainty measure by showing that its expectation lower-bounds the masked diffusion loss. This motivates $\uv_{\mathrm{AD}}$ as a proxy for generation difficulty, whose effectiveness for UQ we evaluate empirically in Section~\ref{sec:experiments}.

  \begin{theorem}[Expected Trajectory Dissimilarity as a Lower Bound on the Masking Loss]
  \label{thm:dissim-lower-bound}
    Assume that the model is perfectly calibrated, i.e., Fisher consistency~\citep{llada}, meaning it recovers the true posterior: $p_\theta(\yv \mid \zv_t, \xv) = p_{\mathrm{data}}(\yv \mid \zv_t, \xv)$ almost surely for any masked state $\zv_t$. Furthermore, assume that the dissimilarity function $D(\cdot, \cdot)$ acts as an optimal semantic measure, such that $D(\tilde{\yv}_t, \yv) = \mathbf{1}[\tilde{\yv}_t \not\equiv \yv]$, where the intermediate prediction $\tilde{\yv}_t = \argmax_{\yv'} p_\theta(\yv' \mid \zv_t, \xv)$. Under these assumptions, the expected trajectory dissimilarity is upper-bounded by the masking loss:
    \[
    \mathbb{E}_{(\xv,\yv) \sim p_{\mathrm{data}}}\!\bigl[\uv_{\mathrm{AD}}(\xv,\yv)\bigr] \le \mathcal{L}(\theta).
    \]
  \end{theorem}

  The proof of Theorem~\ref{thm:dissim-lower-bound} is in Appendix~\ref{app:proof_t1}.
  The assumptions of the theorem are practically well motivated. Perfect calibration, or Fisher consistency, is a standard assumption in diffusion analyses~\citep{llada}. Although Theorem~\ref{thm:dissim-lower-bound} assumes perfect calibration, this principled connection remains valid under model miscalibration as well: the bound includes an additive term that measures the discrepancy between the model and true data distributions (Appendix~\ref{app:miscalibration}). Importantly, our experiments do not rely on calibrated models: we evaluate the LLDM checkpoints without calibration or task-specific fine-tuning. Although the semantic measure $D(\cdot,\cdot)$ is never ideal in practice, its indicator-based definition ensures that any semantic mismatch results in nonzero dissimilarity.

  To implement AD, we need to specify how we define the trajectory generation. In practice, generation is semi-autoregressive; therefore, it can be defined in four different ways:
  (1) $\tilde{\yv}^{\mathrm{block}}_t$, the trajectory of each block is considered independently; (2) $\tilde{\yv}^{\mathrm{last}}_t$, we track the trajectory of only the last block, as it usually provides the final answer after CoT thinking; (3) $\tilde{\yv}^{\mathrm{last\_prefix}}_t$, we fix all blocks $1, ..., {B_{\mathrm{valid}}}-1$ at their final state, and only the last block varies with the prefix of all preceding blocks; (4) $\tilde{\yv}^{\mathrm{full}}_t$, we consider the whole trajectory of all blocks. We denote the corresponding AD scores by $\uv_{\mathrm{AD\text{-}Block}}$, $\uv_{\mathrm{AD\text{-}LastBlock}}$, $\uv_{\mathrm{AD\text{-}LastBlockPrefix}}$, and $\uv_{\mathrm{AD\text{-}Full}}$. We compare these variants in Appendix \ref{sec:abls}.

\ecoparagraph{Progressive trajectory dissimilarity.}
  To penalize slow convergence, we apply a progress-based weighting scheme that assigns greater importance to dissimilarities observed at earlier trajectory steps. This reflects the intuition that 
  if the denoising trajectory is already close to the final output within the first few steps, the model stabilizes quickly, indicating low uncertainty. 
  Accordingly, to capture the importance of the early convergence in the beginning of the denoising process while still accounting for later-stage changes in the trajectory, we formulate the progress-weighted score as:
  \begin{equation}
    \uv_{\mathrm{AD}}^{\mathrm{prog}}(\xv,\yv)
    =
    \frac{1}{T}
    \sum_{t=1}^{T} w_t D(\tilde{\yv}_t, \yv),
    \quad
    w_t = \frac{t}{T}, \quad t = T, T-1, ..., 1.
  \end{equation}

  \begin{proposition}[Bounds for Progressive Trajectory Dissimilarity]
  \label{propos:prog_traj_dissim}
    Assume that $D(\tilde{\yv}_t,\yv)\geq 0$ for all $t$. Then for any $(\xv,\yv)$:
    \begin{equation}
      \frac{1}{T}\uv_{\mathrm{AD}}(\xv,\yv)
      \leq
      \uv_{\mathrm{AD}}^{\mathrm{prog}}(\xv,\yv)
      \leq
      \uv_{\mathrm{AD}}(\xv,\yv).
    \end{equation}
  \end{proposition}

  The proof of Proposition~\ref{propos:prog_traj_dissim} is presented in Appendix~\ref{app:proof_p1}.

  Since $\uv_{\mathrm{AD}}^{\mathrm{prog}}(\xv,\yv)
  \leq
  \uv_{\mathrm{AD}}(\xv,\yv)$, Theorem~\ref{thm:dissim-lower-bound} implies:
  \begin{equation}
    \mathbb{E}_{(\xv, \yv)\sim p_{\mathrm{data}}}
    \bigl[
    \uv_{\mathrm{AD}}^{\mathrm{prog}}(\xv,\yv)
    \bigr]
    \leq
    \mathbb{E}_{(\xv, \yv)\sim p_{\mathrm{data}}}
    \bigl[
    \uv_{\mathrm{AD}}(\xv,\yv)
    \bigr]
    \leq
    \mathcal{L}(\theta).
  \end{equation}
  Therefore, progressive trajectory dissimilarity remains controlled by the
  masked diffusion reconstruction loss while emphasizing earlier denoising states.

\subsection{D-CoCoA: Adapting UQ Methods Developed for Autoregressive Generation to LLDMs}
\label{sec:d_methods}

  Diffusion-specific signals address the two major limitations of standard UQ methods for autoregressive generation discussed above. MCNLL estimates a variational upper bound on the sequence negative log-likelihood and can serve as a proxy in likelihood-based UQ methods~\citep{malinin2020uncertainty, kuhn2023semantic, duan-etal-2024-shifting}. Intermediate states from the denoising trajectory can replace the set of generations obtained through expensive repeated sampling in consistency-based methods, such as Lexical Similarity~\citep{fomicheva-etal-2020-unsupervised}, graph-based methods~\citep{lin2024generating}, LUQ~\citep{zhang2024luq}, and KLE~\citep{nikitin2024kernel}. Finally, white-box methods that combine likelihood- and consistency-based components, such as Semantic Entropy~\citep{kuhn2023semantic}, SAR~\citep{duan-etal-2024-shifting}, Semantic Density~\citep{qiu2024semantic}, and CoCoA~\citep{vashurin2025cocoa}, can benefit from both adaptations.

  The latter method, CoCoA, is a state-of-the-art approach in this category. It builds on Minimum Bayes Risk (MBR) estimation and naturally integrates consistency-based signals and information-based scores into a principled confidence estimation framework~\citep{vashurin2025cocoa}. Specifically, it defines a hallucination detection score as a multiplicative combination of self-consistency and information-based uncertainty scores:
\begin{equation}
\uv_{\mathrm{CoCoA}}(\xv,\yv)
=
\uv_\mathrm{inf}(\xv,\yv)
\cdot
\uv_{\mathrm{cons}}(\xv,\yv).
\end{equation} 
  To construct D-CoCoA -- a version of this method adapted to LLDMs -- we replace the self-consistency component with a trajectory-based semantic dissimilarity $\uv_{\mathrm{AD}}$ and the information-based component with a diffusion likelihood score $\uv_{\mathrm{D\text{-}inf}}$ defined in Section~\ref{sec:diff_signals}.
  \begin{equation}
    \uv_{\mathrm{D\text{-}CoCoA}}(\xv,\yv)
    =
    \uv_{\mathrm{D\text{-}inf}}(\xv,\yv)
    \cdot
    \uv_{\mathrm{AD}}(\xv,\yv).
  \end{equation}
  Both components can be naturally configured at two levels of granularity, resulting in two versions of the adapted method.

\ecoparagraph{D-CoCoA Local.}
  At the token and block level, $\uv_{\mathrm{CommitNLL}}$ provides a natural likelihood proxy unique to LLDMs: it captures model confidence at the precise moment each token is committed, reflecting local generation reliability. Block-level trajectory dissimilarity isolates semantic instability within each generation block independently, which yields:
  \begin{equation}
  \label{eq:d_cocoa_l}
    \uv_{\mathrm{D\text{-}CoCoA\text{-}L}}(\xv,\yv)
    =
    \uv_{\mathrm{CommitNLL}}(\xv,\yv)
    \cdot
    \uv_{\mathrm{AD\text{-}Block}}(\xv,\yv).
  \end{equation}

\ecoparagraph{D-CoCoA Global.}
  $\uv_{\mathrm{MCNLL}}$ and its length-normalized version provide principled approximations of the likelihood over the entire output. Full trajectory dissimilarity measures global semantic drift across all steps. This global view extends via NFE as a diffusion-specific complexity penalty, yielding:
  \begin{equation}
  \label{eq:d_cocoa_g}
    \uv_{\mathrm{D\text{-}CoCoA\text{-}G}}(\xv,\yv)
    =
    \uv_{\mathrm{MCNLL}}^{\mathrm{norm}}(\xv,\yv)
    \cdot
    \bigl(\uv_{\mathrm{NFE}}(\xv,\yv)
    \cdot
    \uv_{\mathrm{AD\text{-}Full}}(\xv,\yv)\bigr).
  \end{equation}

\section{Experiments}
\label{sec:experiments}

\subsection{Experimental Setup}
  We evaluate the effectiveness of UQ methods in identifying and filtering unreliable model outputs under a selective generation setting~\citep{vashurin-etal-2025-benchmarking}. Implementation details of the experimental setup are presented in Appendix~\ref{app:implementation}.

\ecoparagraph{Datasets.}
We evaluate UQ methods on eight benchmarks spanning question answering (MMLU, TriviaQA, CoQA, GSM8k), summarization (SamSum, XSum), and machine translation (WMT14~fr$\to$en, WMT19~de$\to$en). Detailed dataset characteristics, prompts, and generation hyperparameters are reported in Table~\ref{tab:datasets} in Appendix~\ref{app:datasets}.
  
\ecoparagraph{Models.}
  We conduct all experiments using four instruction-tuned LLDMs: the masked diffusion models LLaDA-1.5~\citep{zhu2025llada}, Dream~\citep{ye2025dream}, and ReFusion~\citep{li2026refusion}, and the uniform diffusion model DiffusionGemma~\citep{DiffusionGemma}.

\ecoparagraph{UQ baselines.}
  We evaluate all proposed and adapted diffusion-specific UQ methods in Section~\ref{sec:methods} against state-of-the-art classical likelihood-based, sampling-based, and supervised baselines developed for autoregressive generation, enabling a comprehensive comparison across paradigms. Table~\ref{tab:taxonomy} summarizes the taxonomy of the evaluated methods, including whether they require sampling or training data, and which of them are newly introduced or adapted in this work. Full details of all UQ baseline methods for autoregressive generation are provided in Appendix~\ref{app:baselines}.

  \begin{table*}[!t]
\centering
\resizebox{\textwidth}{!}{\begin{tabular}{l|c|c|l}
\toprule
\textbf{Group} & \textbf{Sampling} & \textbf{Training} & \textbf{Example methods} \\
\midrule
Information-based & -- & -- & CCP~\citep{fadeeva-etal-2024-fact}, RAUQ~\citep{rauq} \\
Sampling-based, white-box & $\checkmark$ & -- & SAR~\citep{duan-etal-2024-shifting}, CoCoA~\citep{vashurin2025cocoa} \\
Sampling-based, black-box & $\checkmark$ & -- & DegMat~\citep{lin2024generating}, KLE~\citep{nikitin2024kernel} \\
Supervised & -- & $\checkmark$ & SAPLMA~\citep{azaria-mitchell-2023-internal}, TAD~\citep{vazhentsev-etal-2025-unconditional} \\
\midrule
Supervised, diffusion-specific & -- & $\checkmark$ & TraceDet~\citep{chang2026tracedet}, DynHD~\citep{qian2026dynhdhallucinationdetectiondiffusion} \\
\midrule
Diffusion statistics & -- & -- & CommitNLL, NFE, MCNLL (Section~\ref{sec:diff_signals}) \\
Trajectory dissimilarity & -- & -- & AD (Section~\ref{sec:diff_signals}) \\
AR + MCNLL (\texttt{-MCNLL}) & $\checkmark$ & -- & SemanticEntropy-MCNLL, CoCoA-MCNLL (Section~\ref{sec:d_methods}) \\
AR + trajectory states (\texttt{D-}) & -- & -- & D-LexSim, D-KLE (Section~\ref{sec:d_methods}) \\
D-CoCoA (ours) & -- & -- & D-CoCoA-L (Eq.~\ref{eq:d_cocoa_l}), D-CoCoA-G (Eq.~\ref{eq:d_cocoa_g}) \\
\bottomrule
\end{tabular}}
\caption{\label{tab:taxonomy} \textbf{UQ method taxonomy.} Evaluated methods include baselines from standard autoregressive (AR) LLMs (top), supervised baselines for LLDMs (middle), and our diffusion-specific methods (bottom). \texttt{-MCNLL} and \texttt{D-}: AR methods using diffusion likelihoods or denoising trajectory states, respectively, instead of AR likelihoods or sampled generations. Details are provided in Appendix~\ref{app:baselines}.}
\end{table*}

\ecoparagraph{Evaluation metrics.}
  UQ performance is primarily measured using the Prediction Rejection Ratio (PRR) -- a standard metric for benchmarking UQ methods in selective classification and generation~\citep{malinin2020uncertainty,vashurin-etal-2025-benchmarking}. PRR summarizes the area under the rejection curve, which shows the average quality of the retained responses after abstaining from a given fraction of the most uncertain predictions. The metric is normalized and typically falls within a [0, 1] range: 0 corresponds to random rejection, whereas 1 corresponds to the optimal rejection order. Compared to ROC-AUC, PRR is less sensitive to class imbalance, supports continuous generation-quality metrics, and can be naturally evaluated at practically relevant rejection rates. 
  We compute PRR for rejection rates of up to 50\% (over only the first part of the curve), as rejecting more than half of the instances is typically impractical.
  We evaluate PRR using task-specific generation quality measures: accuracy for MMLU and GSM8k, COMET~\citep{rei-etal-2020-comet} for MT, and AlignScore~\citep{zha-etal-2023-alignscore} for the rest.
  We additionally report results with the ROC-AUC metric in Appendix~\ref{app:rocauc}.

  Since LLDMs target efficient generation, we also assess computational overhead. 
  We define overhead as the extra runtime beyond base generation, including auxiliary model calls (e.g., the Cross-Encoder required by CoCoA).

\subsection{Analysis of Baselines}

  \begin{table*}[!t] 
\centering
\resizebox{0.8\textwidth}{!}{\begin{tabular}{ll|c|c|c|c|c|c|c|c|c}
\toprule
\textbf{Category} & \textbf{UQ Method} & \textbf{XSum} & \textbf{SamSum} & \textbf{WMT14} & \textbf{WMT19} & \textbf{CoQA} & \textbf{TriviaQA} & \textbf{MMLU} & \textbf{GSM8k} & \textbf{Mean} \\
\midrule
\multirow{6}{*}{\shortstack[l]{Information-\\based}} & MSP & \cellcolor[rgb]{0.6594161915960784,0.7133025255607843,0.9299287241019608} -0.07 & \cellcolor[rgb]{0.8910245585529412,0.9214321063294117,0.9714899216352941} 0.11 & \cellcolor[rgb]{0.7825907906117646,0.8497192224705883,0.9983175350588236} 0.20 & \cellcolor[rgb]{0.7635661295607843,0.8323498010588236,0.9945323233411765} 0.16 & \cellcolor[rgb]{0.6309026780784314,0.6732421581529412,0.8978288875588235} -0.02 & \cellcolor[rgb]{0.6449978024117646,0.6934178380941176,0.9144631795921568} 0.05 & \cellcolor[rgb]{0.9720272867117647,0.7765767393745098,0.7177742451568627} 0.34 & \cellcolor[rgb]{0.6892991246352941,0.7519281085960785,0.9568458054235294} 0.09 & \cellcolor[rgb]{0.6569731755647059,0.7100258308470588,0.927496270517647} 0.11 \\
 & MTE & \cellcolor[rgb]{0.7635661295607843,0.8323498010588236,0.9945323233411765} -0.01 & \cellcolor[rgb]{0.7635661295607843,0.8323498010588236,0.9945323233411765} 0.02 & \cellcolor[rgb]{0.953077067017647,0.9210455325882353,0.9030752965411765} 0.34 & \cellcolor[rgb]{0.9236824528058825,0.9312362411882353,0.9427702351823529} 0.30 & \cellcolor[rgb]{0.7907430740941177,0.8567252977647059,0.9991571764705882} 0.06 & \cellcolor[rgb]{0.61490285,0.649358983,0.8768415764999999} 0.02 & \cellcolor[rgb]{0.9720272867117647,0.7765767393745098,0.7177742451568627} 0.34 & \cellcolor[rgb]{0.7048055483372548,0.7703792543686274,0.9677723612431373} 0.10 & \cellcolor[rgb]{0.7500152822588235,0.8192542337882354,0.9905350620529412} 0.15 \\
 & Perplexity & \cellcolor[rgb]{0.6286169883529411,0.6698302759882353,0.8948307002647058} -0.09 & \cellcolor[rgb]{0.7798733627843137,0.8473838640392157,0.9980376545882352} 0.03 & \cellcolor[rgb]{0.7961779297490197,0.861396014627451,0.9997169374117647} 0.21 & \cellcolor[rgb]{0.7744380861411764,0.8425517925882353,0.9971895702117648} 0.17 & \cellcolor[rgb]{0.6309026780784314,0.6732421581529412,0.8978288875588235} -0.02 & \cellcolor[rgb]{0.6331907341764705,0.6766522042117648,0.9008186597490195} 0.04 & \cellcolor[rgb]{0.9720272867117647,0.7765767393745098,0.7177742451568627} 0.34 & \cellcolor[rgb]{0.61490285,0.649358983,0.8768415764999999} 0.04 & \cellcolor[rgb]{0.61490285,0.649358983,0.8768415764999999} 0.09 \\
 & AttScore & \cellcolor[rgb]{0.852836579,0.50777808,0.575116406} 0.29 & \cellcolor[rgb]{0.9580352625941178,0.9169886544705883,0.8943464355529411} 0.17 & \cellcolor[rgb]{0.61490285,0.649358983,0.8768415764999999} 0.07 & \cellcolor[rgb]{0.61490285,0.649358983,0.8768415764999999} 0.03 & \cellcolor[rgb]{0.7500152822588235,0.8192542337882354,0.9905350620529412} 0.04 & \cellcolor[rgb]{0.6867762156470588,0.7487493527058824,0.9547336847647059} 0.09 & \cellcolor[rgb]{0.6569731755647059,0.7100258308470588,0.927496270517647} 0.05 & \cellcolor[rgb]{0.8389114905588235,0.8932732186176471,0.9955022941235294} 0.18 & \cellcolor[rgb]{0.6569731755647059,0.7100258308470588,0.927496270517647} 0.11 \\
 & CCP & \cellcolor[rgb]{0.61490285,0.649358983,0.8768415764999999} -0.10 & \cellcolor[rgb]{0.8792694128431373,0.9163932970294117,0.9792039273627451} 0.10 & \cellcolor[rgb]{0.7690021078509803,0.8374507968235294,0.9958609467764705} 0.19 & \cellcolor[rgb]{0.7152534441254902,0.7824413707294118,0.974444709590196} 0.12 & \cellcolor[rgb]{0.61490285,0.649358983,0.8768415764999999} -0.03 & \cellcolor[rgb]{0.6449978024117646,0.6934178380941176,0.9144631795921568} 0.05 & \cellcolor[rgb]{0.9720272867117647,0.7765767393745098,0.7177742451568627} 0.34 & \cellcolor[rgb]{0.6716387617176471,0.7296768173647059,0.9420609608117647} 0.08 & \cellcolor[rgb]{0.61490285,0.649358983,0.8768415764999999} 0.09 \\
 & RAUQ & \cellcolor[rgb]{0.7099953545176471,0.7764942726588235,0.9713151710941177} -0.04 & \cellcolor[rgb]{0.8517934440431373,0.9012928182607842,0.9914235664372548} 0.08 & \cellcolor[rgb]{0.8644847897843138,0.9087320678529411,0.9865938341862746} 0.26 & \cellcolor[rgb]{0.8123516188058824,0.8741592436176471,0.9993597327901961} 0.20 & \cellcolor[rgb]{0.6691882557215687,0.7264093044156863,0.9396585384392157} 0.00 & \cellcolor[rgb]{0.6449978024117646,0.6934178380941176,0.9144631795921568} 0.05 & \cellcolor[rgb]{0.9720272867117647,0.7765767393745098,0.7177742451568627} 0.34 & \cellcolor[rgb]{0.7365350864941176,0.8055387188078431,0.9853167941313725} 0.12 & \cellcolor[rgb]{0.7022106452470588,0.7673217452235295,0.966000956317647} 0.13 \\\midrule
\multirow{6}{*}{\shortstack[l]{Sampling,\\white-box}} & MCNSE & \cellcolor[rgb]{0.6594161915960784,0.7133025255607843,0.9299287241019608} -0.07 & \cellcolor[rgb]{0.8389114905588235,0.8932732186176471,0.9955022941235294} 0.07 & \cellcolor[rgb]{0.9003004236470589,0.9251791607803921,0.9650037801960785} 0.29 & \cellcolor[rgb]{0.8718771013137254,0.9125626824411764,0.9828988807745098} 0.25 & \cellcolor[rgb]{0.6691882557215687,0.7264093044156863,0.9396585384392157} -0.00 & \cellcolor[rgb]{0.6766845796941177,0.736034329145098,0.9462852021294117} 0.08 & \cellcolor[rgb]{0.9847608008647059,0.8504164338117647,0.7937540087647059} 0.30 & \cellcolor[rgb]{0.6426363887647059,0.690064711317647,0.9117342756235294} 0.06 & \cellcolor[rgb]{0.679207488682353,0.7392130850352943,0.9483973227882354} 0.12 \\
 & SemanticEntropy & \cellcolor[rgb]{0.7099953545176471,0.7764942726588235,0.9713151710941177} -0.04 & \cellcolor[rgb]{0.9377786937156862,0.9301210794313726,0.9257150330490196} 0.15 & \cellcolor[rgb]{0.9613407263117647,0.9142840690588235,0.8885271948941176} 0.35 & \cellcolor[rgb]{0.9704394715,0.9027982014117647,0.8675832782352941} 0.36 & \cellcolor[rgb]{0.7500152822588235,0.8192542337882354,0.9905350620529412} 0.04 & \cellcolor[rgb]{0.7099953545176471,0.7764942726588235,0.9713151710941177} 0.11 & \cellcolor[rgb]{0.8704786593764706,0.5611201626352941,0.5878720995529412} 0.42 & \cellcolor[rgb]{0.7554121621254901,0.8246983074117646,0.9925393881882353} 0.13 & \cellcolor[rgb]{0.8492270432274511,0.8997249420568627,0.9922887283509804} 0.19 \\
 & SAR & \cellcolor[rgb]{0.8980319349294118,0.9243466028941176,0.9667357341529412} 0.07 & \cellcolor[rgb]{0.9813541391647058,0.8767786732784314,0.8278006056137255} 0.21 & \cellcolor[rgb]{0.9830083599196078,0.8230648707941176,0.7629451741294118} 0.43 & \cellcolor[rgb]{0.9747269947588235,0.7861939559392157,0.7267214884509804} 0.46 & \cellcolor[rgb]{0.9844470791666666,0.8397397817137255,0.7814061455764706} 0.20 & \cellcolor[rgb]{0.9765268001235294,0.7926054336490196,0.7326863173137255} 0.46 & \cellcolor[rgb]{0.8704786593764706,0.5611201626352941,0.5878720995529412} 0.42 & \cellcolor[rgb]{0.8042736801705883,0.8678626149117648,0.9996769126490196} 0.16 & \cellcolor[rgb]{0.9791396989627451,0.8021675484411765,0.7416485506941177} 0.30 \\
 & EigenScore & \cellcolor[rgb]{0.9772829111803922,0.8895272809333333,0.8462654299039215} 0.14 & \cellcolor[rgb]{0.9377786937156862,0.9301210794313726,0.9257150330490196} 0.15 & \cellcolor[rgb]{0.9497671903627452,0.7203459010784313,0.6720534316176471} 0.49 & \cellcolor[rgb]{0.9575785619705882,0.7384632635882353,0.6860897073823529} 0.49 & \cellcolor[rgb]{0.9844470791666666,0.8397397817137255,0.7814061455764706} 0.20 & \cellcolor[rgb]{0.9824556940686274,0.8200795390294118,0.7599027993529412} 0.44 & \cellcolor[rgb]{0.9175136022176471,0.6568221562117647,0.6298915758705882} 0.39 & \cellcolor[rgb]{0.852836579,0.50777808,0.575116406} 0.45 & \cellcolor[rgb]{0.924020201182353,0.6691400189882353,0.6376028197294118} 0.34 \\
 & CoCoA-MTE & \cellcolor[rgb]{0.9256858185156862,0.9315626553686274,0.9405319119196078} 0.09 & \cellcolor[rgb]{0.9666105915000001,0.9077784252235295,0.8765757160705883} 0.18 & \cellcolor[rgb]{0.9305268001470588,0.6814578817647059,0.6453140635882353} 0.51 & \cellcolor[rgb]{0.924020201182353,0.6691400189882353,0.6376028197294118} 0.53 & \cellcolor[rgb]{0.9754778064901961,0.8934375166666666,0.8523803414019608} 0.17 & \cellcolor[rgb]{0.8840171821764706,0.9185176097647059,0.976244109117647} 0.26 & \cellcolor[rgb]{0.852836579,0.50777808,0.575116406} 0.43 & \cellcolor[rgb]{0.8694129974705882,0.9112858109117647,0.9841305319117647} 0.20 & \cellcolor[rgb]{0.9791396989627451,0.8021675484411765,0.7416485506941177} 0.30 \\
 & SemanticDensity & \cellcolor[rgb]{0.9671527274529412,0.762958755827451,0.7061432370215687} 0.21 & \cellcolor[rgb]{0.9747269947588235,0.7861939559392157,0.7267214884509804} 0.27 & \cellcolor[rgb]{0.9842498738333334,0.8369886898862745,0.7783246280235294} 0.42 & \cellcolor[rgb]{0.9783266054882354,0.7990169113588235,0.7386511461764707} 0.45 & \cellcolor[rgb]{0.8543598448588235,0.9028606944647058,0.9905584045235294} 0.09 & \cellcolor[rgb]{0.9824556940686274,0.8200795390294118,0.7599027993529412} 0.44 & \cellcolor[rgb]{0.7554121621254901,0.8246983074117646,0.9925393881882353} 0.11 & \cellcolor[rgb]{0.6716387617176471,0.7296768173647059,0.9420609608117647} 0.08 & \cellcolor[rgb]{0.9754778064901961,0.8934375166666666,0.8523803414019608} 0.26 \\\midrule
\multirow{3}{*}{\shortstack[l]{Sampling,\\black-box}} & LUQ & \cellcolor[rgb]{0.9575785619705882,0.7384632635882353,0.6860897073823529} 0.22 & \cellcolor[rgb]{0.9772829111803922,0.8895272809333333,0.8462654299039215} 0.20 & \cellcolor[rgb]{0.9830083599196078,0.8230648707941176,0.7629451741294118} 0.43 & \cellcolor[rgb]{0.9783266054882354,0.7990169113588235,0.7386511461764707} 0.45 & \cellcolor[rgb]{0.852836579,0.50777808,0.575116406} 0.30 & \cellcolor[rgb]{0.852836579,0.50777808,0.575116406} 0.60 & \cellcolor[rgb]{0.852836579,0.50777808,0.575116406} 0.43 & \cellcolor[rgb]{0.8543598448588235,0.9028606944647058,0.9905584045235294} 0.19 & \cellcolor[rgb]{0.9028614815235294,0.6299065681294118,0.6152808287} 0.35 \\
 & KLE & \cellcolor[rgb]{0.9575785619705882,0.7384632635882353,0.6860897073823529} 0.22 & \cellcolor[rgb]{0.9842498738333334,0.8369886898862745,0.7783246280235294} 0.24 & \cellcolor[rgb]{0.9802450306647059,0.8081382119705882,0.7477333002470588} 0.44 & \cellcolor[rgb]{0.9646785691470589,0.756126767372549,0.7003363715784314} 0.48 & \cellcolor[rgb]{0.9150932609745098,0.6523663817764705,0.6274457140333334} 0.27 & \cellcolor[rgb]{0.852836579,0.50777808,0.575116406} 0.60 & \cellcolor[rgb]{0.852836579,0.50777808,0.575116406} 0.43 & \cellcolor[rgb]{0.8230564053823529,0.8822182482647059,0.9984342312529412} 0.17 & \cellcolor[rgb]{0.8790560841823529,0.5840607685176471,0.5944135352882353} 0.36 \\
 & DegMat & \cellcolor[rgb]{0.9671527274529412,0.762958755827451,0.7061432370215687} 0.21 & \cellcolor[rgb]{0.9727701494549019,0.8993028702666667,0.8615527086490196} 0.19 & \cellcolor[rgb]{0.9830083599196078,0.8230648707941176,0.7629451741294118} 0.43 & \cellcolor[rgb]{0.9646785691470589,0.756126767372549,0.7003363715784314} 0.48 & \cellcolor[rgb]{0.9460687713941176,0.7126943685490196,0.6666446363803922} 0.25 & \cellcolor[rgb]{0.864597965917647,0.5433394684235294,0.5836202017019608} 0.59 & \cellcolor[rgb]{0.852836579,0.50777808,0.575116406} 0.43 & \cellcolor[rgb]{0.8042736801705883,0.8678626149117648,0.9996769126490196} 0.16 & \cellcolor[rgb]{0.924020201182353,0.6691400189882353,0.6376028197294118} 0.34 \\\midrule
\multirow{4}{*}{Supervised} & SAPLMA & \cellcolor[rgb]{0.852836579,0.50777808,0.575116406} 0.29 & \cellcolor[rgb]{0.9348276152529411,0.6896369097254902,0.6504705511098039} 0.32 & \cellcolor[rgb]{0.852836579,0.50777808,0.575116406} 0.57 & \cellcolor[rgb]{0.852836579,0.50777808,0.575116406} 0.59 & \cellcolor[rgb]{0.7500152822588235,0.8192542337882354,0.9905350620529412} 0.04 & \cellcolor[rgb]{0.9004151256333333,0.6254146054,0.6128478516333333} 0.56 & \cellcolor[rgb]{0.9398111318019609,0.9290874692039215,0.9229219340686274} 0.23 & \cellcolor[rgb]{0.9837721488176471,0.8654248580941176,0.812342739117647} 0.31 & \cellcolor[rgb]{0.8790560841823529,0.5840607685176471,0.5944135352882353} 0.36 \\
 & SATRMD & \cellcolor[rgb]{0.9772829111803922,0.8895272809333333,0.8462654299039215} 0.14 & \cellcolor[rgb]{0.9046643351764706,0.9264869997764706,0.9611612942352941} 0.12 & \cellcolor[rgb]{0.9337138175431372,0.9321882998862745,0.9313012310098039} 0.32 & \cellcolor[rgb]{0.8492270432274511,0.8997249420568627,0.9922887283509804} 0.23 & \cellcolor[rgb]{0.8336264621666667,0.8895882285,0.9964796065} 0.08 & \cellcolor[rgb]{0.931695915645098,0.9325418979098039,0.9338169421313726} 0.31 & \cellcolor[rgb]{0.9691631781666668,0.9044582760156863,0.8705807575137254} 0.26 & \cellcolor[rgb]{0.6426363887647059,0.690064711317647,0.9117342756235294} 0.06 & \cellcolor[rgb]{0.8492270432274511,0.8997249420568627,0.9922887283509804} 0.19 \\
 & LookBackLens & \cellcolor[rgb]{0.852836579,0.50777808,0.575116406} 0.29 & \cellcolor[rgb]{0.61490285,0.649358983,0.8768415764999999} -0.09 & \cellcolor[rgb]{0.8952807659705883,0.6156984995294118,0.6081210191470588} 0.54 & \cellcolor[rgb]{0.924020201182353,0.6691400189882353,0.6376028197294118} 0.53 & \cellcolor[rgb]{0.9841016995,0.8604220499999999,0.8061464956666666} 0.19 & \cellcolor[rgb]{0.947942297417647,0.7165372783529411,0.6693403172588235} 0.51 & \cellcolor[rgb]{0.61490285,0.649358983,0.8768415764999999} 0.02 & \cellcolor[rgb]{0.8863529743019607,0.9194891086196079,0.9746593799568628} 0.21 & \cellcolor[rgb]{0.9818859091411765,0.8745427552862746,0.824710021645098} 0.27 \\
 & TAD & \cellcolor[rgb]{0.9747269947588235,0.7861939559392157,0.7267214884509804} 0.20 & \cellcolor[rgb]{0.852836579,0.50777808,0.575116406} 0.38 & \cellcolor[rgb]{0.8952807659705883,0.6156984995294118,0.6081210191470588} 0.54 & \cellcolor[rgb]{0.9423217193470588,0.7050085489411765,0.6612532746235295} 0.51 & \cellcolor[rgb]{0.9398111318019609,0.9290874692039215,0.9229219340686274} 0.14 & \cellcolor[rgb]{0.947942297417647,0.7165372783529411,0.6693403172588235} 0.51 & \cellcolor[rgb]{0.9720272867117647,0.7765767393745098,0.7177742451568627} 0.34 & \cellcolor[rgb]{0.9774267028058823,0.7958111725039216,0.7356687317450981} 0.35 & \cellcolor[rgb]{0.852836579,0.50777808,0.575116406} 0.37 \\
\bottomrule
\end{tabular}
}\caption{\label{tab:ar_baselines_results} \textbf{AR baselines.} PRR$\uparrow$ for the LLaDA-1.5 model across various tasks for baselines from standard autoregressive LLMs (see Tables~\ref{tab:dream_ar_baselines_results}, \ref{tab:refusion_ar_baselines_results}, and~\ref{tab:diffusiongemma_ar_baselines_results} for the Dream, ReFusion, and DiffusionGemma models, respectively). Strong baselines require either $M$ additional generations (sampling-based) or labeled training data (supervised). Warmer color indicates better results.}\end{table*}
  
\ecoparagraph{UQ methods designed for autoregressive generation.}
  We first evaluate baselines designed for autoregressive LLMs and then the individual diffusion-specific signals, which serve as building blocks of D-CoCoA. We begin with %
  UQ methods designed for autoregressive LLMs (Table~\ref{tab:ar_baselines_results} for LLaDA; Tables~\ref{tab:dream_ar_baselines_results}, \ref{tab:refusion_ar_baselines_results}, and~\ref{tab:diffusiongemma_ar_baselines_results} for Dream, ReFusion, and DiffusionGemma in Appendix~\ref{app:res}, respectively).
  
  First, \emph{information-based methods that rely solely on the logits or attention maps perform poorly}, typically only marginally better than random. 
  This is the result of the mismatch between the generation process and the probability estimates used in these methods.

  Second, because information-based scores are poorly aligned with LLDMs, white-box sampling-based methods that rely on these scores also do not perform well and are generally outperformed by black-box sampling-based methods, which depend only on the consistency of the generated texts. Overall, \emph{black-box methods generally achieve the strongest performance among the unsupervised UQ methods originally designed for autoregressive generation}, with KLE standing out as one of the best-performing methods.

  Third, \emph{supervised methods that rely on engineered features derived from the model's internal states, such as SATRMD and LookBackLens, can perform poorly for LLDMs}.
  This further illustrates that the inductive biases underlying feature-engineering approaches designed for autoregressive LLMs are not well aligned with diffusion-style generation.
  However, supervised methods that use raw LLM internal states without complex engineering, such as SAPLMA and TAD, achieve strong performance, matching or exceeding black-box sampling-based methods for most models. Note that supervised methods are evaluated in-domain and require labeled training data, while their performance tends to degrade substantially in out-of-domain scenarios~\citep{vazhentsev-etal-2025-unconditional}.

\ecoparagraph{Diffusion-specific UQ methods: MCNLL, and likelihood- and statistics-based signals from the denoising trajectory.}
  Tables~\ref{tab:diff_baselines_results}, \ref{tab:dream_diff_baselines_results}, \ref{tab:refusion_diff_baselines_results}, and~\ref{tab:diffusiongemma_diff_baselines_results} show results with diffusion-specific methods (the \emph{Diffusion statistics} category)
  for LLaDA, Dream, ReFusion, and DiffusionGemma, respectively. \emph{All considered statistics typically provide useful information}, except NFE, Remask, and FlipCount on MMLU due to one-step generation. Other methods perform similarly on average. One of the strongest statistics, CommitNLL, serves as the likelihood component in D-CoCoA-L (see equation~\eqref{eq:d_cocoa_l}).

\ecoparagraph{Methods based on denoising trajectory semantic instability.}
  The \emph{Trajectory dissimilarity} category in Tables~\ref{tab:diff_baselines_results}, \ref{tab:dream_diff_baselines_results}, \ref{tab:refusion_diff_baselines_results}, and~\ref{tab:diffusiongemma_diff_baselines_results} presents the results of methods based on the semantic instability of the denoising trajectory during generation for LLaDA, Dream, ReFusion, and DiffusionGemma, respectively.
  Except for Dream, the performance on MMLU is close to random because the trajectories contain only one step. 
  For TriviaQA, all results are similar because there is only one block across all outputs. The weighted version typically outperforms the original average dissimilarity. Furthermore, all block-wise trajectory evaluations are generally worse than the raw full trajectory without specific aggregation across all datasets, except for GSM8k and XSum. Overall, results show that \emph{average trajectory dissimilarity proves to be a highly informative signal}, and therefore serves as the consistency component of both D-CoCoA variants (see equations~\eqref{eq:d_cocoa_l} and~\eqref{eq:d_cocoa_g}).

\ecoparagraph{White-box sampling-based UQ with MCNLL for likelihood estimation.}
  Tables~\ref{tab:ar_improved_results}, \ref{tab:dream_ar_improved_results}, and~\ref{tab:refusion_ar_improved_results} in Appendix~\ref{app:res} present results for sampling-based baselines originally developed for autoregressive LLMs, using MCNLL as a proxy for sequence negative log-likelihood in LLaDA, Dream, and ReFusion, respectively. As expected, \emph{MCNLL consistently and substantially improves performance across all methods} and datasets.
  CoCoA-MCNLL achieves the best average results across all considered methods, outperforming the best non-diffusion-specific method, KLE, for LLaDA-1.5.

\ecoparagraph{Black-box sampling-based UQ with denoising trajectory states.}
  Tables~\ref{tab:ar_traj_results}, \ref{tab:dream_ar_traj_results}, \ref{tab:refusion_ar_traj_results}, and~\ref{tab:diffusiongemma_ar_traj_results} in Appendix~\ref{app:res} report results for sampling-based baselines adapted to use intermediate generations from the denoising trajectory instead of independently sampled outputs.
  Most adapted methods, except D-KLE, yield meaningful uncertainty scores, but substantially underperform AD. Since both approaches rely only on the denoising trajectory, this highlights the importance of the right aggregation: D-DegMat, D-KLE, and related methods compute pairwise similarities among trajectory states, whereas AD measures the dissimilarity between each intermediate state and the final output.

\ecoparagraph{Summary.}
  This analysis yields three main findings: (1) UQ methods based on generation likelihood should use the diffusion-aware likelihood estimation, such as MCNLL; (2) denoising trajectory statistics provide useful uncertainty signals; and (3) intermediate states in the denoising process can be used instead of sampling multiple final generations, with the strongest UQ signal obtained by measuring their dissimilarity to the final generation. These findings directly motivate D-CoCoA (Section~\ref{sec:d_methods}), which combines a diffusion-aware likelihood with trajectory dissimilarity.

  \begin{table*}[!t] 
\centering
\resizebox{\textwidth}{!}{\begin{tabular}{ll|c|c|c|c|c|c|c|c|c}
\toprule
\textbf{Category} & \textbf{UQ Method} & \textbf{XSum} & \textbf{SamSum} & \textbf{WMT14} & \textbf{WMT19} & \textbf{CoQA} & \textbf{TriviaQA} & \textbf{MMLU} & \textbf{GSM8k} & \textbf{Mean} \\
\midrule
\multirow{8}{*}{\shortstack[l]{Diffusion\\statistics}} & MCNLL & \cellcolor[rgb]{0.9819030282176471,0.8170942072647058,0.7568604245764706} 0.17 & \cellcolor[rgb]{0.9756268974411765,0.7893996947941176,0.729703902882353} 0.18 & \cellcolor[rgb]{0.9691631781666668,0.9044582760156863,0.8705807575137254} 0.29 & \cellcolor[rgb]{0.953077067017647,0.9210455325882353,0.9030752965411765} 0.32 & \cellcolor[rgb]{0.8933603506784313,0.9224036051843137,0.9699051924745098} 0.24 & \cellcolor[rgb]{0.9824176791176471,0.8723068372941176,0.8216194376764705} 0.31 & \cellcolor[rgb]{0.9747269947588235,0.7861939559392157,0.7267214884509804} 0.35 & \cellcolor[rgb]{0.947942297417647,0.7165372783529411,0.6693403172588235} 0.23 & \cellcolor[rgb]{0.9802905992117648,0.8812505092627452,0.8339817735509804} 0.26 \\
 & MCNLL-Norm & \cellcolor[rgb]{0.9819030282176471,0.8170942072647058,0.7568604245764706} 0.17 & \cellcolor[rgb]{0.7662841187058823,0.8349002989411765,0.9951966350588235} 0.09 & \cellcolor[rgb]{0.7179074031529412,0.7853536515764705,0.9758586906411765} 0.19 & \cellcolor[rgb]{0.7473192431058824,0.8165111307921569,0.9894914084686275} 0.24 & \cellcolor[rgb]{0.61490285,0.649358983,0.8768415764999999} 0.18 & \cellcolor[rgb]{0.61490285,0.649358983,0.8768415764999999} 0.09 & \cellcolor[rgb]{0.9747269947588235,0.7861939559392157,0.7267214884509804} 0.35 & \cellcolor[rgb]{0.8015810339588235,0.8657637386764706,0.9997826392686274} -0.02 & \cellcolor[rgb]{0.6766845796941177,0.736034329145098,0.9462852021294117} 0.16 \\
 \cmidrule(lr){2-11}
 & NFE & \cellcolor[rgb]{0.9547297988764706,0.919693239882353,0.9001656762117647} 0.15 & \cellcolor[rgb]{0.9840526685,0.8342375980588235,0.7752431104705882} 0.17 & \cellcolor[rgb]{0.9841016995,0.8604220499999999,0.8061464956666666} 0.31 & \cellcolor[rgb]{0.931695915645098,0.9325418979098039,0.9338169421313726} 0.31 & \cellcolor[rgb]{0.8015810339588235,0.8657637386764706,0.9997826392686274} 0.22 & \cellcolor[rgb]{0.7853082184392157,0.8520545809019608,0.9985974155294117} 0.18 & \cellcolor[rgb]{0.6970208390666667,0.7612067269333334,0.9624581464666666} 0.02 & \cellcolor[rgb]{0.8675383126470588,0.5522298155294117,0.585746150627451} 0.29 & \cellcolor[rgb]{0.8569262456745098,0.9044285706686275,0.989693242609804} 0.21 \\
 & Remask & \cellcolor[rgb]{0.9808223691882353,0.8790145912705882,0.830891189582353} 0.16 & \cellcolor[rgb]{0.9840526685,0.8342375980588235,0.7752431104705882} 0.17 & \cellcolor[rgb]{0.8594926464901961,0.905996446872549,0.9888280806960784} 0.24 & \cellcolor[rgb]{0.8042736801705883,0.8678626149117648,0.9996769126490196} 0.26 & \cellcolor[rgb]{0.931695915645098,0.9325418979098039,0.9338169421313726} 0.25 & \cellcolor[rgb]{0.7258692802352942,0.794090494117647,0.9801006337941176} 0.15 & \cellcolor[rgb]{0.6970208390666667,0.7612067269333334,0.9624581464666666} 0.02 & \cellcolor[rgb]{0.8675383126470588,0.5522298155294117,0.585746150627451} 0.29 & \cellcolor[rgb]{0.7853082184392157,0.8520545809019608,0.9985974155294117} 0.19 \\
 & FlipCount & \cellcolor[rgb]{0.9547297988764706,0.919693239882353,0.9001656762117647} 0.15 & \cellcolor[rgb]{0.9824176791176471,0.8723068372941176,0.8216194376764705} 0.16 & \cellcolor[rgb]{0.7179074031529412,0.7853536515764705,0.9758586906411765} 0.19 & \cellcolor[rgb]{0.61490285,0.649358983,0.8768415764999999} 0.19 & \cellcolor[rgb]{0.931695915645098,0.9325418979098039,0.9338169421313726} 0.25 & \cellcolor[rgb]{0.6667452396901961,0.7231326097019608,0.9372260848549019} 0.12 & \cellcolor[rgb]{0.6970208390666667,0.7612067269333334,0.9624581464666666} 0.02 & \cellcolor[rgb]{0.9004151256333333,0.6254146054,0.6128478516333333} 0.27 & \cellcolor[rgb]{0.7125994850980393,0.779529089882353,0.9730307285392157} 0.17 \\
 \cmidrule(lr){2-11}
 & TrajNLL & \cellcolor[rgb]{0.9819030282176471,0.8170942072647058,0.7568604245764706} 0.17 & \cellcolor[rgb]{0.9840526685,0.8342375980588235,0.7752431104705882} 0.17 & \cellcolor[rgb]{0.7473192431058824,0.8165111307921569,0.9894914084686275} 0.20 & \cellcolor[rgb]{0.7473192431058824,0.8165111307921569,0.9894914084686275} 0.24 & \cellcolor[rgb]{0.9423217193470588,0.7050085489411765,0.6612532746235295} 0.30 & \cellcolor[rgb]{0.9176723556764705,0.9302569986470588,0.9494852049705882} 0.25 & \cellcolor[rgb]{0.852836579,0.50777808,0.575116406} 0.47 & \cellcolor[rgb]{0.9813503623666666,0.8141088755,0.7538180498} 0.18 & \cellcolor[rgb]{0.9678868848333333,0.9061183506196078,0.8735782367921568} 0.25 \\
 & TrajEntropy & \cellcolor[rgb]{0.9808223691882353,0.8790145912705882,0.830891189582353} 0.16 & \cellcolor[rgb]{0.9305268001470588,0.6814578817647059,0.6453140635882353} 0.20 & \cellcolor[rgb]{0.9112102024705881,0.9284487582705883,0.9553975652941177} 0.26 & \cellcolor[rgb]{0.9112102024705881,0.9284487582705883,0.9553975652941177} 0.30 & \cellcolor[rgb]{0.9836582578333333,0.8287354144039216,0.7690800753647059} 0.28 & \cellcolor[rgb]{0.9704394715,0.9027982014117647,0.8675832782352941} 0.29 & \cellcolor[rgb]{0.852836579,0.50777808,0.575116406} 0.47 & \cellcolor[rgb]{0.9575785619705882,0.7384632635882353,0.6860897073823529} 0.22 & \cellcolor[rgb]{0.9849255762058824,0.8479150297647058,0.7906558870392157} 0.27 \\
 & CommitNLL & \cellcolor[rgb]{0.9591408362921569,0.742086736090196,0.6888969625352941} 0.18 & \cellcolor[rgb]{0.9575785619705882,0.7384632635882353,0.6860897073823529} 0.19 & \cellcolor[rgb]{0.852836579,0.50777808,0.575116406} 0.39 & \cellcolor[rgb]{0.852836579,0.50777808,0.575116406} 0.43 & \cellcolor[rgb]{0.8015810339588235,0.8657637386764706,0.9997826392686274} 0.22 & \cellcolor[rgb]{0.852836579,0.50777808,0.575116406} 0.43 & \cellcolor[rgb]{0.852836579,0.50777808,0.575116406} 0.47 & \cellcolor[rgb]{0.852836579,0.50777808,0.575116406} 0.30 & \cellcolor[rgb]{0.852836579,0.50777808,0.575116406} 0.33 \\
\midrule
\multirow{9}{*}{\shortstack[l]{Trajectory\\dissimilarity}} & AD-Block & \cellcolor[rgb]{0.9150932609745098,0.6523663817764705,0.6274457140333334} 0.19 & \cellcolor[rgb]{0.61490285,0.649358983,0.8768415764999999} 0.05 & \cellcolor[rgb]{0.61490285,0.649358983,0.8768415764999999} 0.15 & \cellcolor[rgb]{0.953077067017647,0.9210455325882353,0.9030752965411765} 0.32 & \cellcolor[rgb]{0.8492270432274511,0.8997249420568627,0.9922887283509804} 0.23 & \cellcolor[rgb]{0.9305268001470588,0.6814578817647059,0.6453140635882353} 0.39 & \cellcolor[rgb]{0.61490285,0.649358983,0.8768415764999999} -0.05 & \cellcolor[rgb]{0.61490285,0.649358983,0.8768415764999999} -0.15 & \cellcolor[rgb]{0.61490285,0.649358983,0.8768415764999999} 0.14 \\
 & WeightedAD-Block & \cellcolor[rgb]{0.852836579,0.50777808,0.575116406} 0.20 & \cellcolor[rgb]{0.9756268974411765,0.7893996947941176,0.729703902882353} 0.18 & \cellcolor[rgb]{0.9337138175431372,0.9321882998862745,0.9313012310098039} 0.27 & \cellcolor[rgb]{0.9683898066058824,0.766374750054902,0.7090466697431372} 0.38 & \cellcolor[rgb]{0.9653342981666666,0.909438499827451,0.8795731953490196} 0.26 & \cellcolor[rgb]{0.9305268001470588,0.6814578817647059,0.6453140635882353} 0.39 & \cellcolor[rgb]{0.61490285,0.649358983,0.8768415764999999} -0.05 & \cellcolor[rgb]{0.9575785619705882,0.7384632635882353,0.6860897073823529} 0.22 & \cellcolor[rgb]{0.9196757213862745,0.930583412827451,0.9472468817078432} 0.23 \\
\cmidrule(lr){2-11}
 & AD-LastBlock & \cellcolor[rgb]{0.9808223691882353,0.8790145912705882,0.830891189582353} 0.16 & \cellcolor[rgb]{0.806966326382353,0.8699614911470589,0.9995711860294118} 0.10 & \cellcolor[rgb]{0.7744380861411764,0.8425517925882353,0.9971895702117648} 0.21 & \cellcolor[rgb]{0.9797588292352941,0.8834864272549019,0.8370723575196078} 0.34 & \cellcolor[rgb]{0.8492270432274511,0.8997249420568627,0.9922887283509804} 0.23 & \cellcolor[rgb]{0.9305268001470588,0.6814578817647059,0.6453140635882353} 0.39 & \cellcolor[rgb]{0.61490285,0.649358983,0.8768415764999999} -0.05 & \cellcolor[rgb]{0.915574114,0.9297565972666666,0.9515550793333334} 0.06 & \cellcolor[rgb]{0.7473192431058824,0.8165111307921569,0.9894914084686275} 0.18 \\
 & WeightedAD-LastBlock & \cellcolor[rgb]{0.9547297988764706,0.919693239882353,0.9001656762117647} 0.15 & \cellcolor[rgb]{0.8840171821764706,0.9185176097647059,0.976244109117647} 0.12 & \cellcolor[rgb]{0.9337138175431372,0.9321882998862745,0.9313012310098039} 0.27 & \cellcolor[rgb]{0.9683898066058824,0.766374750054902,0.7090466697431372} 0.38 & \cellcolor[rgb]{0.9653342981666666,0.909438499827451,0.8795731953490196} 0.26 & \cellcolor[rgb]{0.9305268001470588,0.6814578817647059,0.6453140635882353} 0.39 & \cellcolor[rgb]{0.61490285,0.649358983,0.8768415764999999} -0.05 & \cellcolor[rgb]{0.9276891842254902,0.9318890695490196,0.9382935886568627} 0.07 & \cellcolor[rgb]{0.8204138911862745,0.8803757532058823,0.9989228874411764} 0.20 \\
\cmidrule(lr){2-11}
 & AD-LastBlockPrefix & \cellcolor[rgb]{0.9547297988764706,0.919693239882353,0.9001656762117647} 0.15 & \cellcolor[rgb]{0.9575785619705882,0.7384632635882353,0.6860897073823529} 0.19 & \cellcolor[rgb]{0.6892991246352941,0.7519281085960785,0.9568458054235294} 0.18 & \cellcolor[rgb]{0.953077067017647,0.9210455325882353,0.9030752965411765} 0.32 & \cellcolor[rgb]{0.9696268857588235,0.769790744282353,0.7119501024647059} 0.29 & \cellcolor[rgb]{0.9305268001470588,0.6814578817647059,0.6453140635882353} 0.39 & \cellcolor[rgb]{0.61490285,0.649358983,0.8768415764999999} -0.05 & \cellcolor[rgb]{0.8492270432274511,0.8997249420568627,0.9922887283509804} 0.01 & \cellcolor[rgb]{0.7473192431058824,0.8165111307921569,0.9894914084686275} 0.18 \\
 & WeightedAD-LastBlockPrefix & \cellcolor[rgb]{0.9808223691882353,0.8790145912705882,0.830891189582353} 0.16 & \cellcolor[rgb]{0.9840526685,0.8342375980588235,0.7752431104705882} 0.17 & \cellcolor[rgb]{0.8594926464901961,0.905996446872549,0.9888280806960784} 0.24 & \cellcolor[rgb]{0.9838554631666667,0.8314865062313725,0.7721615929176471} 0.36 & \cellcolor[rgb]{0.852836579,0.50777808,0.575116406} 0.32 & \cellcolor[rgb]{0.9305268001470588,0.6814578817647059,0.6453140635882353} 0.39 & \cellcolor[rgb]{0.61490285,0.649358983,0.8768415764999999} -0.05 & \cellcolor[rgb]{0.876805309,0.9151164254999999,0.9804355785000001} 0.03 & \cellcolor[rgb]{0.8204138911862745,0.8803757532058823,0.9989228874411764} 0.20 \\
\cmidrule(lr){2-11}
 & AD-Full & \cellcolor[rgb]{0.61490285,0.649358983,0.8768415764999999} 0.09 & \cellcolor[rgb]{0.852836579,0.50777808,0.575116406} 0.22 & \cellcolor[rgb]{0.8863529743019607,0.9194891086196079,0.9746593799568628} 0.25 & \cellcolor[rgb]{0.9838554631666667,0.8314865062313725,0.7721615929176471} 0.36 & \cellcolor[rgb]{0.9696268857588235,0.769790744282353,0.7119501024647059} 0.29 & \cellcolor[rgb]{0.9305268001470588,0.6814578817647059,0.6453140635882353} 0.39 & \cellcolor[rgb]{0.61490285,0.649358983,0.8768415764999999} -0.05 & \cellcolor[rgb]{0.876805309,0.9151164254999999,0.9804355785000001} 0.03 & \cellcolor[rgb]{0.8204138911862745,0.8803757532058823,0.9989228874411764} 0.20 \\
 & WeightedAD-Full & \cellcolor[rgb]{0.61490285,0.649358983,0.8768415764999999} 0.09 & \cellcolor[rgb]{0.8952807659705883,0.6156984995294118,0.6081210191470588} 0.21 & \cellcolor[rgb]{0.953077067017647,0.9210455325882353,0.9030752965411765} 0.28 & \cellcolor[rgb]{0.9683898066058824,0.766374750054902,0.7090466697431372} 0.38 & \cellcolor[rgb]{0.852836579,0.50777808,0.575116406} 0.32 & \cellcolor[rgb]{0.9305268001470588,0.6814578817647059,0.6453140635882353} 0.39 & \cellcolor[rgb]{0.61490285,0.649358983,0.8768415764999999} -0.05 & \cellcolor[rgb]{0.8910245585529412,0.9214321063294117,0.9714899216352941} 0.04 & \cellcolor[rgb]{0.8569262456745098,0.9044285706686275,0.989693242609804} 0.21 \\
\bottomrule
\end{tabular}
}\caption{\label{tab:diff_baselines_results} \textbf{Diffusion-specific UQ signals.} PRR$\uparrow$ for the LLaDA-1.5 across various tasks for diffusion statistics and trajectory dissimilarity UQ methods (see Tables~\ref{tab:dream_diff_baselines_results}, \ref{tab:refusion_diff_baselines_results}, and~\ref{tab:diffusiongemma_diff_baselines_results} for the Dream, ReFusion, and DiffusionGemma models, respectively). Methods are defined in Section~\ref{sec:diff_signals}. CommitNLL and AD, strong signals, are the components of D-CoCoA. Warmer color indicates better results.}\end{table*}

\subsection{Main Results}

      \begin{figure*}[t!]
    \centering
    \begin{minipage}[t]{0.49\linewidth}
      \centering
      \includegraphics[trim={0.cm 0.cm 0.cm 1cm},clip,width=\linewidth]{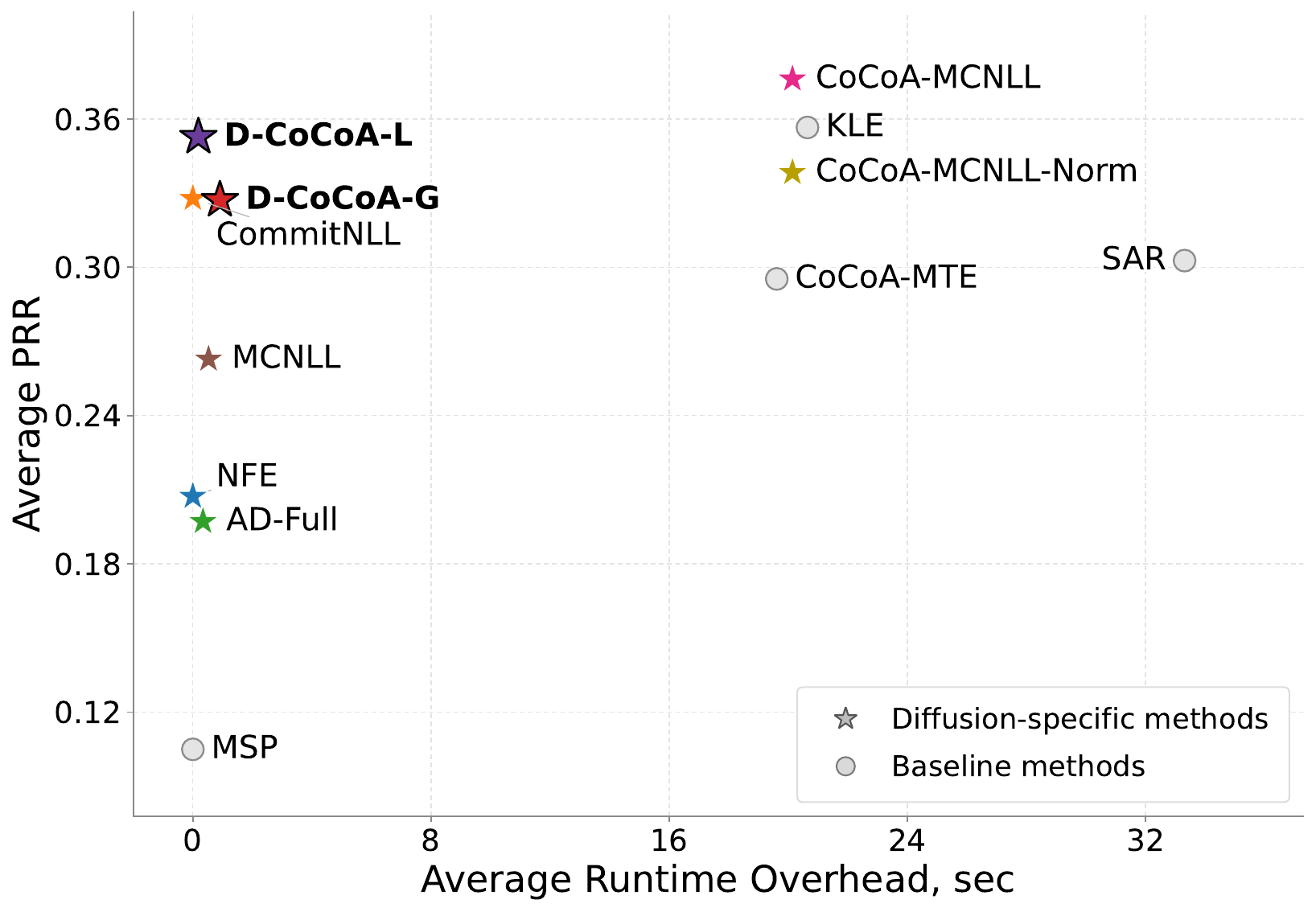}
      \subcaption{\textbf{LLaDA-1.5}}
      \label{fig:avg_prr_vs_comp_llada}
    \end{minipage}
    \hfill
    \begin{minipage}[t]{0.49\linewidth}
      \centering
      \includegraphics[trim={0.cm 0.cm 0.cm 1cm},clip,width=\linewidth]{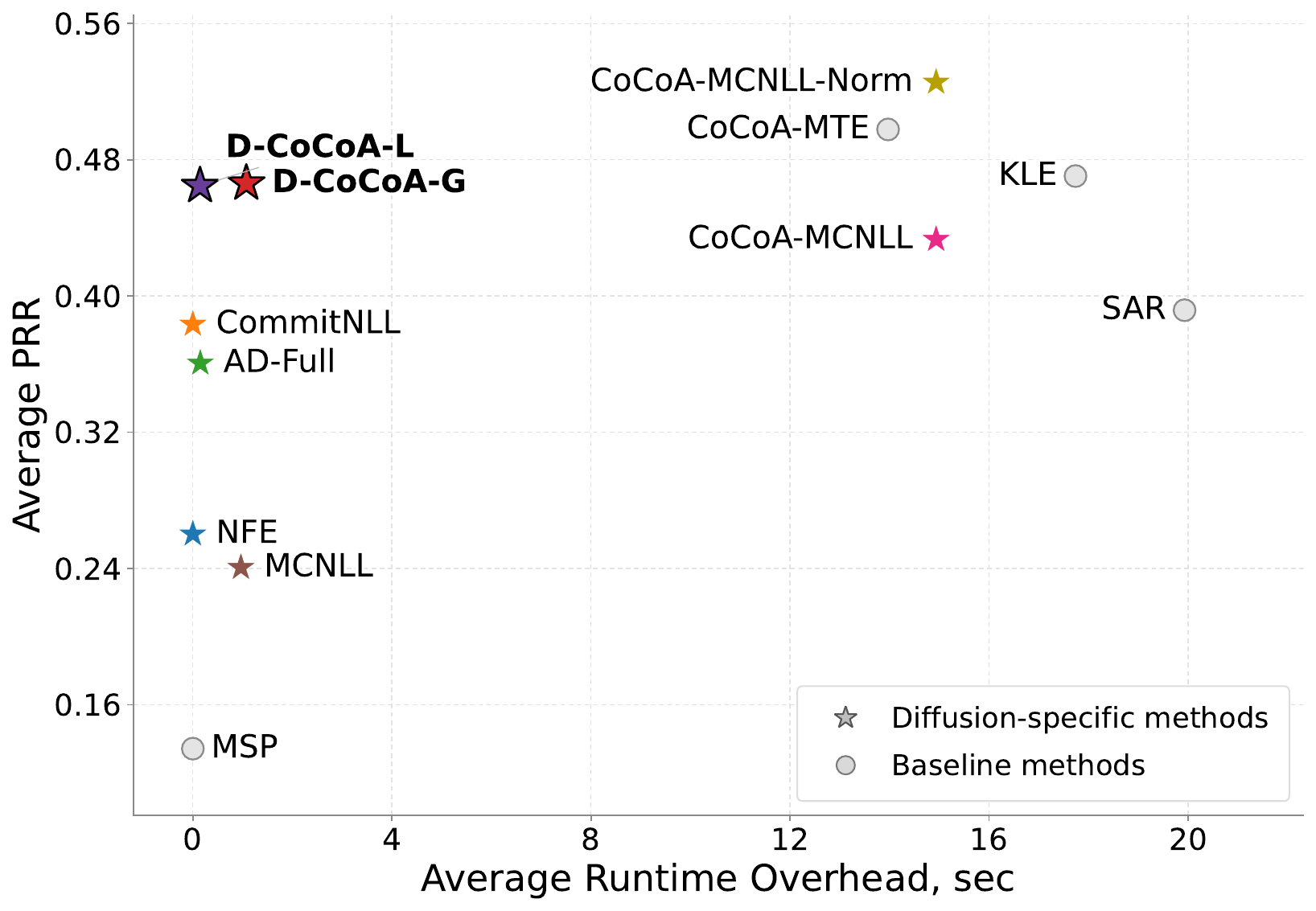}
      \subcaption{\textbf{Dream}}
      \label{fig:avg_prr_vs_comp_dream}
    \end{minipage}

    \begin{minipage}[t]{0.49\linewidth}
      \centering
      \includegraphics[trim={0.cm 0.cm 0.cm 1cm},clip,width=\linewidth]{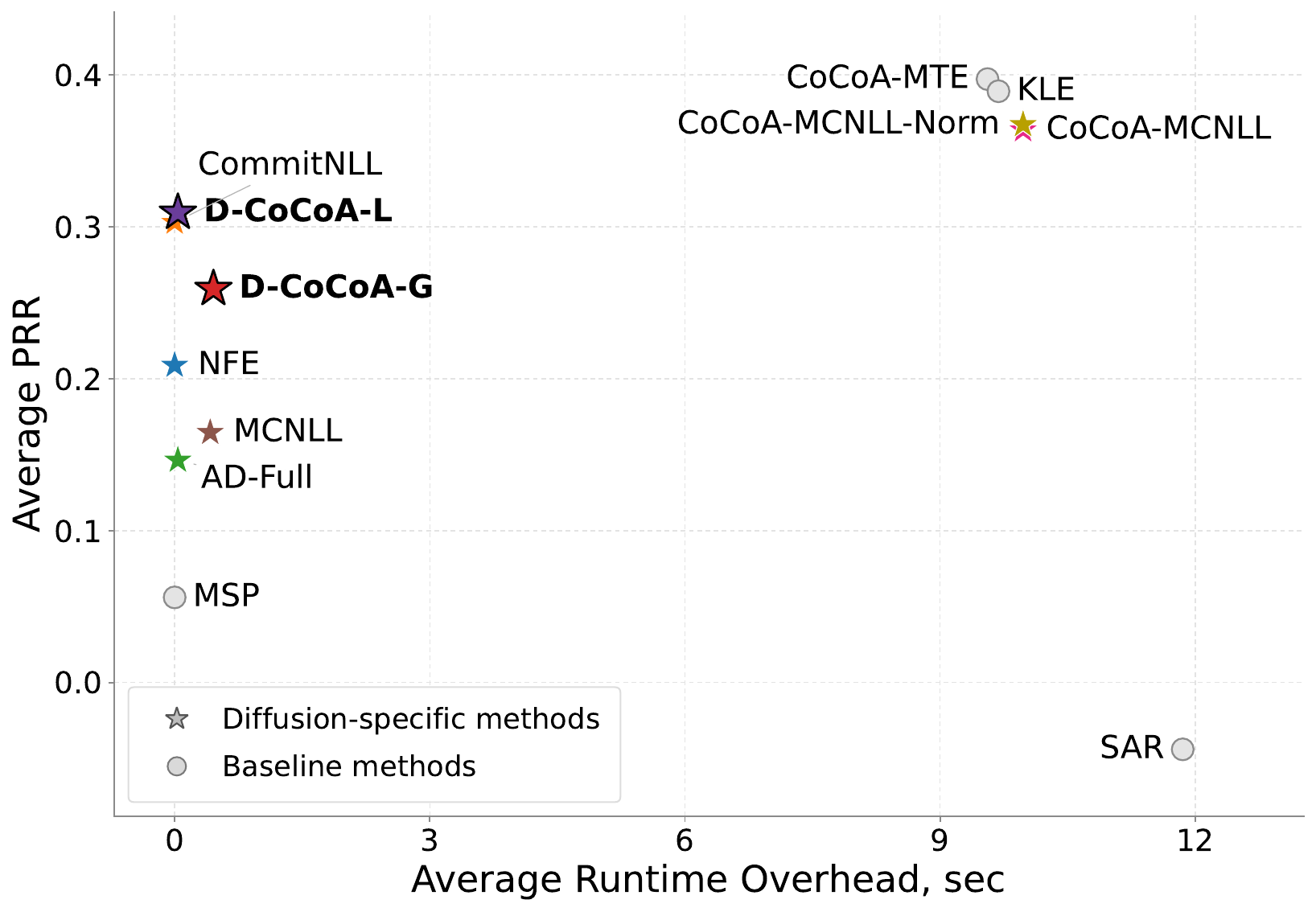}
      \subcaption{\textbf{ReFusion}}
      \label{fig:avg_prr_vs_comp_refusion}
    \end{minipage}
    \hfill
    \begin{minipage}[t]{0.49\linewidth}
      \centering
      \includegraphics[trim={0.cm 0.cm 0.cm 1cm},clip,width=\linewidth]{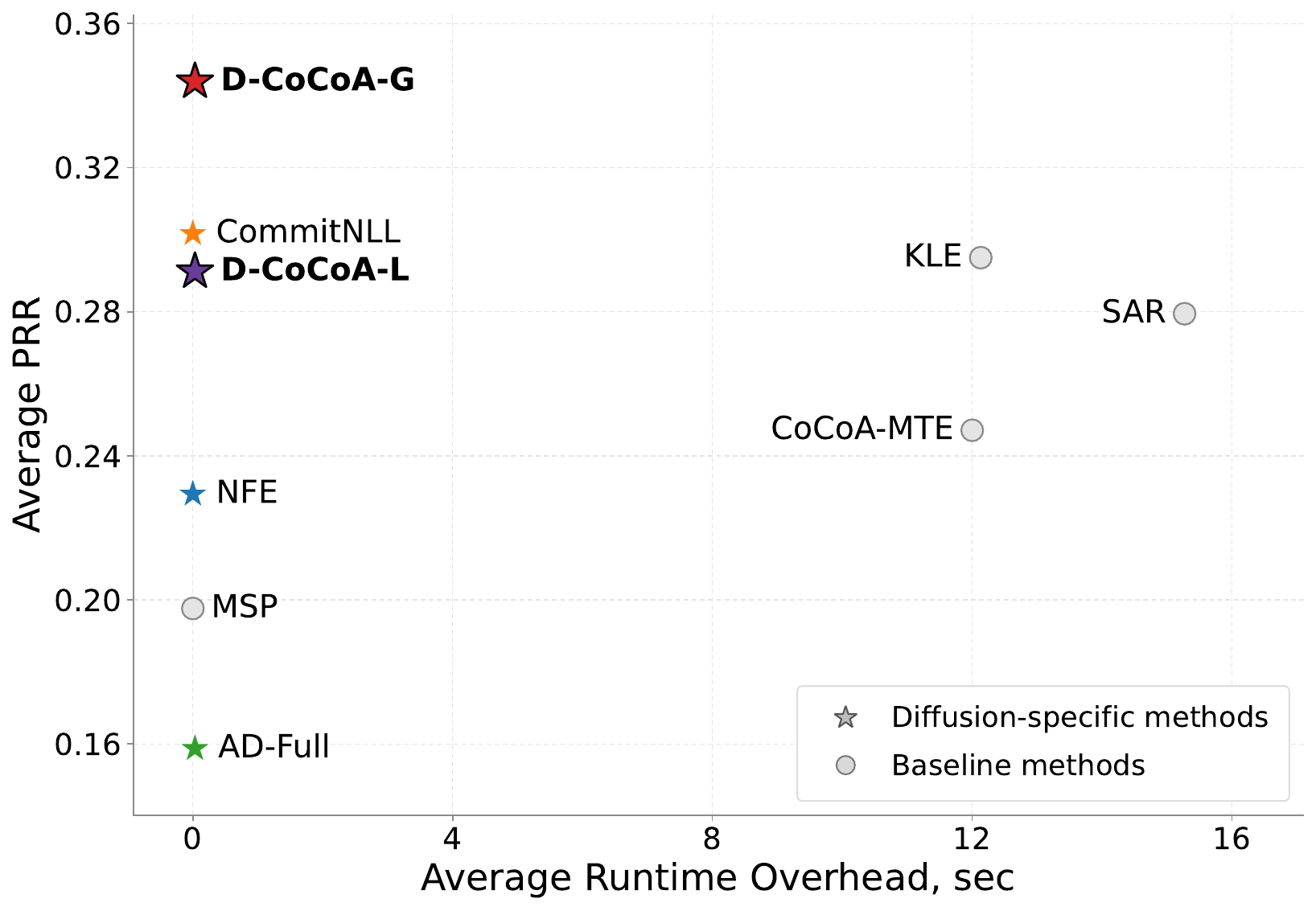}
      \subcaption{\textbf{DiffusionGemma}}
      \label{fig:avg_prr_vs_comp_diffusiongemma}
    \end{minipage}
    \caption{
    \textbf{Average performance and runtime.} UQ performance (PRR$\uparrow$) versus computational overhead (runtime in seconds$\downarrow$) averaged across all datasets. DiffusionGemma has no mask token, making MCNLL-based methods inapplicable; D-CoCoA-G instead uses CommitNLL.
    }
    \label{fig:avg_prr_vs_comp}
  \end{figure*}

  Having established the behavior of individual categories of methods, we now compare the best methods under a unified performance-efficiency framework. 
  Figure~\ref{fig:avg_prr_vs_comp} presents the trade-off between selective generation performance (measured by PRR) and computational complexity (measured as overhead time) %
  averaged across all datasets. Figures~\ref{fig:llada_prr_vs_comp}, \ref{fig:dream_prr_vs_comp}, \ref{fig:refusion_prr_vs_comp}, and~\ref{fig:diffusiongemma_prr_vs_comp} in Appendix~\ref{app:res} present the detailed trade-off between selective generation performance and computational complexity for all individual datasets.
  The detailed PRR scores are presented in Tables~\ref{tab:detailed_main_results}, \ref{tab:dream_detailed_main_results}, \ref{tab:refusion_detailed_main_results}, and~\ref{tab:diffusiongemma_detailed_main_results} in Appendix~\ref{app:res}. 
  
  Overall, we observe a clear efficiency-performance frontier. Extensive sampling methods, such as basic CoCoA variants and KLE, achieve higher PRR for LLaDA, Dream, and ReFusion but incur substantially greater computational overhead. In contrast, lightweight diffusion-specific signals, such as NFE and trajectory-based statistics, provide very low-cost uncertainty estimates but are less accurate and generally insufficient for high-quality UQ. 
  
  Our proposed methods, D-CoCoA-G and D-CoCoA-L, consistently achieve a strong balance between performance and efficiency. They clearly outperform other unsupervised low- to medium-complexity methods and approach the performance of substantially more expensive sampling-based methods. This trade-off is highly favorable: for example, \mbox{D-CoCoA-L} requires roughly \textbf{one hundred times less} additional computation time than the strongest sampling-based methods, such as CoCoA-MCNLL and KLE, while these methods typically provide only modest additional gains (e.g., 2.3\% for LLaDA). This trend holds both on average and across individual datasets. 

  \underline{\textbf{Key take-away:}} Experimental results show that by leveraging the distinctive properties of the LLDM 
  denoising process, we can obtain strong uncertainty estimates at almost no additional cost while preserving computational efficiency -- one of the main motivations for using LLDMs.

\ecoparagraph{Ablation summary.}
We ablate four D-CoCoA components: source of likelihood, trajectory aggregation strategy, similarity function, and the optional NFE term (Tables~\ref{tab:abl_cocoa_inf_results}, \ref{tab:abl_cocoa_traj_results}, \ref{tab:abl_cocoa_sim_results}, and~\ref{tab:abl_cocoa_nfe_results}). We find that the default configuration (normalized MCNLL and the full trajectory for D-CoCoA-G, CommitNLL and per-block averaging for D-CoCoA-L, and RoBERTa Cross-Encoder similarity for both) yields the best average performance. Including NFE further benefits D-CoCoA-G but slightly hurts D-CoCoA-L. The full ablation tables and per-component discussion are reported in Appendix~\ref{sec:abls}.

\ecoparagraph{Analysis summary.}
We further analyze D-CoCoA with LLaDA-1.5. We find that adding $M \ge 2$ sampled generations to D-CoCoA-L matches the performance of CoCoA-MCNLL, while adding more generations does not significantly improve performance (Table~\ref{tab:abl_cocoa_hybrid_results}). We also find that generative perplexity from GPT-2~\citep{radford2019language} is not a reliable uncertainty signal (Table~\ref{tab:abl_genppl_results}), diffusion-specific methods are among the best calibrated by ECE~\citep{pmlr-v70-guo17a} (Table~\ref{tab:abl_ece_results}), and D-CoCoA-L requires 100--500 times fewer additional TFLOPs than KLE (Table~\ref{tab:abl_tflops_results}). Details are in Appendix~\ref{app:analyses}.

\section{Conclusion}

We presented the first systematic study of unsupervised UQ for LLDMs. We showed that UQ methods developed for autoregressive LLMs are often poorly aligned with diffusion-based generation, either because they rely on left-to-right sequence likelihoods or because they require repeated sampling that undermines the efficiency advantages of LLDMs. To address this gap, we introduced a family of diffusion-specific uncertainty signals derived directly from the denoising process, including masked diffusion likelihoods, trajectory statistics, remasking dynamics, generation complexity, and semantic instability along the denoising trajectory. Our theoretical analysis connects semantic instability to the masked diffusion training objective, providing support for its use as a principled uncertainty measure. 

Building on these signals, we proposed D-CoCoA, a diffusion-aware adaptation of a state-of-the-art hybrid UQ method. 
Our experiments show that D-CoCoA achieves a favorable performance-efficiency trade-off: it approaches or matches strong sampling-based baselines while requiring substantially less computation. For D-CoCoA-L, the runtime overhead is roughly 100 times lower. Overall, our results demonstrate that LLDMs can support effective hallucination detection while retaining their inference efficiency.

\begin{ack}
This work received no third-party funding. The authors declare no competing interests.
\end{ack}

\bibliographystyle{abbrvnat_no_doi_no_url}
\bibliography{ref}

\begin{thebibliography}{58}
\providecommand{\natexlab}[1]{#1}
\providecommand{\url}[1]{\texttt{#1}}
\expandafter\ifx\csname urlstyle\endcsname\relax
  \providecommand{\doi}[1]{doi: #1}\else
  \providecommand{\doi}{doi: \begingroup \urlstyle{rm}\Url}\fi

\bibitem[Azaria and Mitchell(2023)]{azaria-mitchell-2023-internal}
A.~Azaria and T.~Mitchell.
\newblock The internal state of an {LLM} knows when it`s lying.
\newblock In H.~Bouamor, J.~Pino, and K.~Bali, editors, \emph{Findings of the
  Association for Computational Linguistics: EMNLP 2023}, pages 967--976,
  Singapore, Dec. 2023. Association for Computational Linguistics.

\bibitem[Barrault et~al.(2019)Barrault, Bojar, Costa-juss{\`a}, Federmann,
  Fishel, Graham, Haddow, Huck, Koehn, Malmasi, Monz, M{\"u}ller, Pal, Post,
  and Zampieri]{barrault-etal-2019-findings}
L.~Barrault, O.~Bojar, M.~R. Costa-juss{\`a}, C.~Federmann, M.~Fishel,
  Y.~Graham, B.~Haddow, M.~Huck, P.~Koehn, S.~Malmasi, C.~Monz, M.~M{\"u}ller,
  S.~Pal, M.~Post, and M.~Zampieri.
\newblock Findings of the 2019 {Conference on Machine Translation} ({WMT}19).
\newblock In O.~Bojar, R.~Chatterjee, C.~Federmann, M.~Fishel, Y.~Graham,
  B.~Haddow, M.~Huck, A.~J. Yepes, P.~Koehn, A.~Martins, C.~Monz, M.~Negri,
  A.~N{\'e}v{\'e}ol, M.~Neves, M.~Post, M.~Turchi, and K.~Verspoor, editors,
  \emph{Proceedings of the Fourth Conference on Machine Translation (Volume 2:
  Shared Task Papers, Day 1)}, pages 1--61, Florence, Italy, Aug. 2019.
  Association for Computational Linguistics.

\bibitem[Berry et~al.(2024)Berry, Brando, and Meger]{pmlr-v244-berry24a}
L.~Berry, A.~Brando, and D.~Meger.
\newblock Shedding light on large generative networks: Estimating epistemic
  uncertainty in diffusion models.
\newblock In N.~Kiyavash and J.~M. Mooij, editors, \emph{Uncertainty in
  Artificial Intelligence, 15-19 July 2024, Universitat Pompeu Fabra,
  Barcelona, Spain}, Proceedings of Machine Learning Research, pages 360--376.
  {PMLR}, 2024.

\bibitem[Bie et~al.(2025)Bie, Cao, Chen, Du, Gong, Gong, Gu, Hu, Huang, Lan,
  Li, Li, Li, Li, Liu, Liu, Lu, Lu, Ma, Tan, Wei, Wen, Xing, Zhang, Zhao,
  Zheng, Zhou, Zhou, Zhou, Zhu, and Zhuang]{llada2}
T.~Bie, M.~Cao, K.~Chen, L.~Du, M.~Gong, Z.~Gong, Y.~Gu, J.~Hu, Z.~Huang,
  Z.~Lan, C.~Li, C.~Li, J.~Li, Z.~Li, H.~Liu, L.~Liu, G.~Lu, X.~Lu, Y.~Ma,
  J.~Tan, L.~Wei, J.~Wen, Y.~Xing, X.~Zhang, J.~Zhao, D.~Zheng, J.~Zhou,
  J.~Zhou, Z.~Zhou, L.~Zhu, and Y.~Zhuang.
\newblock {LLaDA}2.0: Scaling up diffusion language models to 100b.
\newblock \emph{CoRR}, abs/2512.15745, 2025.

\bibitem[Bojar et~al.(2014)Bojar, Buck, Federmann, Haddow, Koehn, Leveling,
  Monz, Pecina, Post, Saint-Amand, Soricut, Specia, and
  Tamchyna]{bojar-etal-2014-findings}
O.~Bojar, C.~Buck, C.~Federmann, B.~Haddow, P.~Koehn, J.~Leveling, C.~Monz,
  P.~Pecina, M.~Post, H.~Saint-Amand, R.~Soricut, L.~Specia, and A.~Tamchyna.
\newblock Findings of the 2014 workshop on statistical machine translation.
\newblock In O.~Bojar, C.~Buck, C.~Federmann, B.~Haddow, P.~Koehn, C.~Monz,
  M.~Post, and L.~Specia, editors, \emph{Proceedings of the Ninth Workshop on
  Statistical Machine Translation}, pages 12--58, Baltimore, Maryland, USA,
  June 2014. Association for Computational Linguistics.

\bibitem[Chang et~al.(2026)Chang, Yu, Wang, Chen, Yu, Torr, and
  Gu]{chang2026tracedet}
S.~Chang, J.~Yu, W.~Wang, Y.~Chen, J.~Yu, P.~Torr, and J.~Gu.
\newblock {TRACEDET}: {Hallucination} {Detection} {From} {the} {Decoding}
  {Trace} {of} {Diffusion} {Large} {Language} {Models}.
\newblock In \emph{The Fourteenth International Conference on Learning
  Representations}, 2026.

\bibitem[Chen et~al.(2024)Chen, Liu, Chen, Gu, Wu, Tao, Fu, and
  Ye]{chen2024inside}
C.~Chen, K.~Liu, Z.~Chen, Y.~Gu, Y.~Wu, M.~Tao, Z.~Fu, and J.~Ye.
\newblock {INSIDE}: {LLM}s' internal states retain the power of hallucination
  detection.
\newblock In \emph{The Twelfth International Conference on Learning
  Representations}, 2024.

\bibitem[Chuang et~al.(2024)Chuang, Qiu, Hsieh, Krishna, Kim, and
  Glass]{chuang-etal-2024-lookback}
Y.-S. Chuang, L.~Qiu, C.-Y. Hsieh, R.~Krishna, Y.~Kim, and J.~R. Glass.
\newblock Lookback lens: Detecting and mitigating contextual hallucinations in
  large language models using only attention maps.
\newblock In Y.~Al-Onaizan, M.~Bansal, and Y.-N. Chen, editors,
  \emph{Proceedings of the 2024 Conference on Empirical Methods in Natural
  Language Processing}, pages 1419--1436, Miami, Florida, USA, Nov. 2024.
  Association for Computational Linguistics.

\bibitem[Cobbe et~al.(2021)Cobbe, Kosaraju, Bavarian, Chen, Jun, Kaiser,
  Plappert, Tworek, Hilton, Nakano, Hesse, and Schulman]{cobbe2021training}
K.~Cobbe, V.~Kosaraju, M.~Bavarian, M.~Chen, H.~Jun, L.~Kaiser, M.~Plappert,
  J.~Tworek, J.~Hilton, R.~Nakano, C.~Hesse, and J.~Schulman.
\newblock Training verifiers to solve math word problems.
\newblock \emph{CoRR}, abs/2110.14168, 2021.

\bibitem[{DiffusionGemma Team}(2026)]{DiffusionGemma}
{DiffusionGemma Team}.
\newblock {DiffusionGemma} technical report.
\newblock \emph{CoRR}, abs/2608.00146, 2026.

\bibitem[Duan et~al.(2024)Duan, Cheng, Wang, Zavalny, Wang, Xu, Kailkhura, and
  Xu]{duan-etal-2024-shifting}
J.~Duan, H.~Cheng, S.~Wang, A.~Zavalny, C.~Wang, R.~Xu, B.~Kailkhura, and
  K.~Xu.
\newblock Shifting attention to relevance: Towards the predictive uncertainty
  quantification of free-form large language models.
\newblock In L.-W. Ku, A.~Martins, and V.~Srikumar, editors, \emph{Proceedings
  of the 62nd Annual Meeting of the Association for Computational Linguistics
  (Volume 1: Long Papers)}, pages 5050--5063, Bangkok, Thailand, Aug. 2024.
  Association for Computational Linguistics.

\bibitem[Fadeeva et~al.(2023)Fadeeva, Vashurin, Tsvigun, Vazhentsev, Petrakov,
  Fedyanin, Vasilev, Goncharova, Panchenko, Panov, Baldwin, and
  Shelmanov]{fadeeva2023lm}
E.~Fadeeva, R.~Vashurin, A.~Tsvigun, A.~Vazhentsev, S.~Petrakov, K.~Fedyanin,
  D.~Vasilev, E.~Goncharova, A.~Panchenko, M.~Panov, T.~Baldwin, and
  A.~Shelmanov.
\newblock {LM}-polygraph: Uncertainty estimation for language models.
\newblock In Y.~Feng and E.~Lefever, editors, \emph{Proceedings of the 2023
  Conference on Empirical Methods in Natural Language Processing: System
  Demonstrations}, pages 446--461, Singapore, Dec. 2023. Association for
  Computational Linguistics.

\bibitem[Fadeeva et~al.(2024)Fadeeva, Rubashevskii, Shelmanov, Petrakov, Li,
  Mubarak, Tsymbalov, Kuzmin, Panchenko, Baldwin, Nakov, and
  Panov]{fadeeva-etal-2024-fact}
E.~Fadeeva, A.~Rubashevskii, A.~Shelmanov, S.~Petrakov, H.~Li, H.~Mubarak,
  E.~Tsymbalov, G.~Kuzmin, A.~Panchenko, T.~Baldwin, P.~Nakov, and M.~Panov.
\newblock Fact-checking the output of large language models via token-level
  uncertainty quantification.
\newblock In L.-W. Ku, A.~Martins, and V.~Srikumar, editors, \emph{Findings of
  the Association for Computational Linguistics: ACL 2024}, pages 9367--9385,
  Bangkok, Thailand, Aug. 2024. Association for Computational Linguistics.

\bibitem[Farquhar et~al.(2024)Farquhar, Kossen, Kuhn, and
  Gal]{farquhar2024detecting}
S.~Farquhar, J.~Kossen, L.~Kuhn, and Y.~Gal.
\newblock Detecting hallucinations in large language models using semantic
  entropy.
\newblock \emph{Nature}, 630\penalty0 (8017):\penalty0 625--630, 2024.

\bibitem[Fomicheva et~al.(2020)Fomicheva, Sun, Yankovskaya, Blain, Guzm{\'a}n,
  Fishel, Aletras, Chaudhary, and Specia]{fomicheva-etal-2020-unsupervised}
M.~Fomicheva, S.~Sun, L.~Yankovskaya, F.~Blain, F.~Guzm{\'a}n, M.~Fishel,
  N.~Aletras, V.~Chaudhary, and L.~Specia.
\newblock Unsupervised quality estimation for neural machine translation.
\newblock \emph{Transactions of the Association for Computational Linguistics},
  8:\penalty0 539--555, 2020.

\bibitem[Gliwa et~al.(2019)Gliwa, Mochol, Biesek, and
  Wawer]{gliwa-etal-2019-samsum}
B.~Gliwa, I.~Mochol, M.~Biesek, and A.~Wawer.
\newblock {SAMS}um corpus: A human-annotated dialogue dataset for abstractive
  summarization.
\newblock In L.~Wang, J.~C.~K. Cheung, G.~Carenini, and F.~Liu, editors,
  \emph{Proceedings of the 2nd Workshop on New Frontiers in Summarization},
  pages 70--79, Hong Kong, China, Nov. 2019. Association for Computational
  Linguistics.

\bibitem[Grattafiori et~al.(2024)Grattafiori, Dubey, Jauhri, Pandey, Kadian,
  Al-Dahle, Letman, Mathur, Schelten, Vaughan, Yang, Fan, Goyal,
  et~al.]{dubey2024llama}
A.~Grattafiori, A.~Dubey, A.~Jauhri, A.~Pandey, A.~Kadian, A.~Al-Dahle,
  A.~Letman, A.~Mathur, A.~Schelten, A.~Vaughan, A.~Yang, A.~Fan, A.~Goyal,
  et~al.
\newblock The {Llama 3} herd of models.
\newblock \emph{arXiv preprint arXiv:2407.21783}, 2024.

\bibitem[Guo et~al.(2017)Guo, Pleiss, Sun, and Weinberger]{pmlr-v70-guo17a}
C.~Guo, G.~Pleiss, Y.~Sun, and K.~Q. Weinberger.
\newblock On calibration of modern neural networks.
\newblock In D.~Precup and Y.~W. Teh, editors, \emph{Proceedings of the 34th
  International Conference on Machine Learning}, volume~70 of \emph{Proceedings
  of Machine Learning Research}, pages 1321--1330. PMLR, 06--11 Aug 2017.

\bibitem[Hendrycks et~al.(2021)Hendrycks, Burns, Basart, Zou, Mazeika, Song,
  and Steinhardt]{hendryckstest2021}
D.~Hendrycks, C.~Burns, S.~Basart, A.~Zou, M.~Mazeika, D.~Song, and
  J.~Steinhardt.
\newblock Measuring massive multitask language understanding.
\newblock In \emph{9th International Conference on Learning Representations,
  {ICLR} 2021, Virtual Event, Austria, May 3-7, 2021}. OpenReview.net, 2021.

\bibitem[Huang et~al.(2025)Huang, Yu, Ma, Zhong, Feng, Wang, Chen, Peng, Feng,
  Qin, et~al.]{huang2025survey}
L.~Huang, W.~Yu, W.~Ma, W.~Zhong, Z.~Feng, H.~Wang, Q.~Chen, W.~Peng, X.~Feng,
  B.~Qin, et~al.
\newblock A survey on hallucination in large language models: Principles,
  taxonomy, challenges, and open questions.
\newblock \emph{ACM Transactions on Information Systems}, 43\penalty0
  (2):\penalty0 1--55, 2025.

\bibitem[Jazbec et~al.(2025)Jazbec, Wong{-}Toi, Xia, Zhang, Nalisnick, and
  Mandt]{pmlr-v286-jazbec25a}
M.~Jazbec, E.~Wong{-}Toi, G.~Xia, D.~Zhang, E.~T. Nalisnick, and S.~Mandt.
\newblock Generative uncertainty in diffusion models.
\newblock In S.~Chiappa and S.~Magliacane, editors, \emph{Conference on
  Uncertainty in Artificial Intelligence, Rio Othon Palace, Rio de Janeiro,
  Brazil, 21-25 July 2025}, Proceedings of Machine Learning Research, pages
  1837--1858. {PMLR}, 2025.

\bibitem[Joshi et~al.(2017)Joshi, Choi, Weld, and
  Zettlemoyer]{joshi-etal-2017-triviaqa}
M.~Joshi, E.~Choi, D.~Weld, and L.~Zettlemoyer.
\newblock {T}rivia{QA}: A large scale distantly supervised challenge dataset
  for reading comprehension.
\newblock In R.~Barzilay and M.-Y. Kan, editors, \emph{Proceedings of the 55th
  Annual Meeting of the Association for Computational Linguistics (Volume 1:
  Long Papers)}, pages 1601--1611, Vancouver, Canada, July 2017. Association
  for Computational Linguistics.

\bibitem[Kou et~al.(2024)Kou, Gan, Wang, Li, and Deng]{kou2024bayesdiff}
S.~Kou, L.~Gan, D.~Wang, C.~Li, and Z.~Deng.
\newblock Bayesdiff: Estimating pixel-wise uncertainty in diffusion via
  bayesian inference.
\newblock In \emph{The Twelfth International Conference on Learning
  Representations}, 2024.

\bibitem[Kuhn et~al.(2023)Kuhn, Gal, and Farquhar]{kuhn2023semantic}
L.~Kuhn, Y.~Gal, and S.~Farquhar.
\newblock Semantic uncertainty: Linguistic invariances for uncertainty
  estimation in natural language generation.
\newblock In \emph{The Eleventh International Conference on Learning
  Representations, {ICLR} 2023, Kigali, Rwanda, May 1-5, 2023}. OpenReview.net,
  2023.

\bibitem[Li et~al.(2026)Li, Guan, Wu, and Li]{li2026refusion}
J.-N. Li, J.~Guan, W.~Wu, and C.~Li.
\newblock Refusion: A diffusion large language model with parallel
  autoregressive decoding.
\newblock In \emph{The Fourteenth International Conference on Learning
  Representations}, 2026.

\bibitem[Lin et~al.(2024)Lin, Trivedi, and Sun]{lin2024generating}
Z.~Lin, S.~Trivedi, and J.~Sun.
\newblock Generating with confidence: Uncertainty quantification for black-box
  large language models.
\newblock \emph{Transactions on Machine Learning Research}, 2024.
\newblock ISSN 2835-8856.

\bibitem[Malinin and Gales(2021)]{malinin2020uncertainty}
A.~Malinin and M.~J.~F. Gales.
\newblock Uncertainty estimation in autoregressive structured prediction.
\newblock In \emph{9th International Conference on Learning Representations,
  {ICLR} 2021, Virtual Event, Austria, May 3-7, 2021}. OpenReview.net, 2021.

\bibitem[Narayan et~al.(2018)Narayan, Cohen, and
  Lapata]{narayan-etal-2018-dont}
S.~Narayan, S.~B. Cohen, and M.~Lapata.
\newblock Don`t give me the details, just the summary! topic-aware
  convolutional neural networks for extreme summarization.
\newblock In E.~Riloff, D.~Chiang, J.~Hockenmaier, and J.~Tsujii, editors,
  \emph{Proceedings of the 2018 Conference on Empirical Methods in Natural
  Language Processing}, pages 1797--1807, Brussels, Belgium, Oct.-Nov. 2018.
  Association for Computational Linguistics.

\bibitem[Ni et~al.(2026)Ni, Fadeeva, Wu, Akhtar, Zhang, Ash, Leippold, Baldwin,
  Ng, Shelmanov, and Sachan]{ni-etal-2026-reprobe}
J.~Ni, E.~Fadeeva, T.~Wu, M.~Akhtar, J.~Zhang, E.~Ash, M.~Leippold, T.~Baldwin,
  S.-K. Ng, A.~Shelmanov, and M.~Sachan.
\newblock {R}e{P}robe: Efficient test-time scaling of multi-step reasoning by
  probing internal states of large language models.
\newblock In M.~Liakata, V.~P. Moreira, J.~Zhang, and D.~Jurgens, editors,
  \emph{Proceedings of the 64th Annual Meeting of the {A}ssociation for
  {C}omputational {L}inguistics (Volume 1: Long Papers)}, pages 11667--11689,
  San Diego, California, United States, July 2026. Association for
  Computational Linguistics.
\newblock ISBN 979-8-89176-390-6.

\bibitem[Nie et~al.(2025)Nie, Zhu, You, Zhang, Ou, Hu, Zhou, Lin, Wen, and
  Li]{llada}
S.~Nie, F.~Zhu, Z.~You, X.~Zhang, J.~Ou, J.~Hu, J.~Zhou, Y.~Lin, J.-R. Wen, and
  C.~Li.
\newblock Large language diffusion models.
\newblock In \emph{Advances in Neural Information Processing Systems},
  volume~38, 2025.

\bibitem[Nikitin et~al.(2024)Nikitin, Kossen, Gal, and
  Marttinen]{nikitin2024kernel}
A.~Nikitin, J.~Kossen, Y.~Gal, and P.~Marttinen.
\newblock Kernel language entropy: Fine-grained uncertainty quantification for
  {LLM}s from semantic similarities.
\newblock In \emph{Advances in Neural Information Processing Systems},
  volume~37, pages 8901--8929, 2024.

\bibitem[Papamarkou et~al.(2026{\natexlab{a}})Papamarkou, Alquier, Bauer,
  Buntine, Davison, Dziugaite, Filippone, Foong, Fortuin, Fouskakis, Frellsen,
  H\"{u}llermeier, Karaletsos, Khan, Kotelevskii, Lahlou, Li, Liu, Lyle,
  M\"{o}llenhoff, Palla, Panov, Sale, Schweighofer, Shelmanov, Swaroop, Trapp,
  Waegeman, Wilson, and Zaytsev]{papamarkou2026position}
T.~Papamarkou, P.~Alquier, M.~Bauer, W.~Buntine, A.~J. Davison, G.~K.
  Dziugaite, M.~Filippone, A.~Y.~K. Foong, V.~Fortuin, D.~Fouskakis,
  J.~Frellsen, E.~H\"{u}llermeier, T.~Karaletsos, M.~E. Khan, N.~Kotelevskii,
  S.~Lahlou, Y.~Li, F.~Liu, C.~Lyle, T.~M\"{o}llenhoff, K.~Palla, M.~Panov,
  Y.~Sale, K.~Schweighofer, A.~Shelmanov, S.~Swaroop, M.~Trapp, W.~Waegeman,
  A.~G. Wilson, and A.~Zaytsev.
\newblock Position: Agentic {AI} orchestration should be {B}ayes-consistent.
\newblock In T.~Zhang, M.~Dudik, M.~Jaggi, A.~Agarwal, S.~Li, D.~Schuurmans,
  J.~Zhu, F.~Berkenkamp, H.~Dong, and A.~Bietti, editors, \emph{Proceedings of
  the 43rd International Conference on Machine Learning}, volume 306 of
  \emph{Proceedings of Machine Learning Research}, pages 171622--171639. PMLR,
  06--11 Jul 2026{\natexlab{a}}.

\bibitem[Papamarkou et~al.(2026{\natexlab{b}})Papamarkou, Smirnov, Mazanov,
  Vazhentsev, Nakov, Baldwin, and Shelmanov]{papamarkou-etal-2026-bayesian}
T.~Papamarkou, V.~Smirnov, V.~Mazanov, A.~Vazhentsev, P.~Nakov, T.~Baldwin, and
  A.~Shelmanov.
\newblock {B}ayesian control for coding agents.
\newblock In H.-Y. Lee, S.~Dev, J.~J. Li, and D.~Elliott, editors,
  \emph{Proceedings of the 2026 Conference on Empirical Methods in Natural
  Language Processing}, pages 35299--35341, Budapest, Hungary, Oct.
  2026{\natexlab{b}}. Association for Computational Linguistics.
\newblock ISBN 979-8-89176-459-0.

\bibitem[Paszke et~al.(2019)Paszke, Gross, Massa, Lerer, Bradbury, Chanan,
  Killeen, Lin, Gimelshein, Antiga, Desmaison, Kopf, Yang, DeVito, Raison,
  Tejani, Chilamkurthy, Steiner, Fang, Bai, and Chintala]{NEURIPS2019_bdbca288}
A.~Paszke, S.~Gross, F.~Massa, A.~Lerer, J.~Bradbury, G.~Chanan, T.~Killeen,
  Z.~Lin, N.~Gimelshein, L.~Antiga, A.~Desmaison, A.~Kopf, E.~Yang, Z.~DeVito,
  M.~Raison, A.~Tejani, S.~Chilamkurthy, B.~Steiner, L.~Fang, J.~Bai, and
  S.~Chintala.
\newblock Pytorch: An imperative style, high-performance deep learning library.
\newblock In H.~Wallach, H.~Larochelle, A.~Beygelzimer, F.~d\textquotesingle
  Alch\'{e}-Buc, E.~Fox, and R.~Garnett, editors, \emph{Advances in Neural
  Information Processing Systems}, volume~32. Curran Associates, Inc., 2019.

\bibitem[Qian et~al.(2026)Qian, Tan, Liu, Yu, and
  Pan]{qian2026dynhdhallucinationdetectiondiffusion}
Y.~Qian, Y.~Tan, Y.~Liu, W.~Yu, and S.~Pan.
\newblock {D}yn{HD}: Hallucination detection for diffusion large language
  models via denoising dynamics deviation learning.
\newblock In H.-Y. Lee, S.~Dev, J.~Li, and D.~Elliott, editors, \emph{Findings
  of the {A}ssociation for {C}omputational {L}inguistics: {EMNLP} 2026}, pages
  7712--7727, Budapest, Hungary, Oct. 2026. Association for Computational
  Linguistics.
\newblock ISBN 979-8-89176-462-0.

\bibitem[Qiu and Miikkulainen(2024)]{qiu2024semantic}
X.~Qiu and R.~Miikkulainen.
\newblock Semantic density: Uncertainty quantification for large language
  models through confidence measurement in semantic space.
\newblock In A.~Globerson, L.~Mackey, D.~Belgrave, A.~Fan, U.~Paquet,
  J.~Tomczak, and C.~Zhang, editors, \emph{Advances in Neural Information
  Processing Systems}, volume~37, pages 134507--134533. Curran Associates,
  Inc., 2024.

\bibitem[Radford et~al.(2019)Radford, Wu, Child, Luan, Amodei, and
  Sutskever]{radford2019language}
A.~Radford, J.~Wu, R.~Child, D.~Luan, D.~Amodei, and I.~Sutskever.
\newblock Language models are unsupervised multitask learners.
\newblock \emph{OpenAI blog}, 2019.

\bibitem[Reddy et~al.(2019)Reddy, Chen, and Manning]{reddy-etal-2019-coqa}
S.~Reddy, D.~Chen, and C.~D. Manning.
\newblock {C}o{QA}: A conversational question answering challenge.
\newblock \emph{Transactions of the Association for Computational Linguistics},
  7:\penalty0 249--266, 2019.

\bibitem[Rei et~al.(2020)Rei, Stewart, Farinha, and Lavie]{rei-etal-2020-comet}
R.~Rei, C.~Stewart, A.~C. Farinha, and A.~Lavie.
\newblock {COMET}: A neural framework for {MT} evaluation.
\newblock In B.~Webber, T.~Cohn, Y.~He, and Y.~Liu, editors, \emph{Proceedings
  of the 2020 Conference on Empirical Methods in Natural Language Processing
  (EMNLP)}, pages 2685--2702, Online, Nov. 2020. Association for Computational
  Linguistics.

\bibitem[Sahoo et~al.(2024)Sahoo, Arriola, Schiff, Gokaslan, Marroquin, Chiu,
  Rush, and Kuleshov]{sahoo2024simple}
S.~S. Sahoo, M.~Arriola, Y.~Schiff, A.~Gokaslan, E.~Marroquin, J.~T. Chiu,
  A.~Rush, and V.~Kuleshov.
\newblock Simple and effective masked diffusion language models.
\newblock In \emph{Advances in Neural Information Processing Systems},
  volume~37, pages 130136--130184, 2024.

\bibitem[Sahoo et~al.(2025)Sahoo, Deschenaux, Gokaslan, Wang, Chiu, and
  Kuleshov]{sahoo2025the}
S.~S. Sahoo, J.~Deschenaux, A.~Gokaslan, G.~Wang, J.~T. Chiu, and V.~Kuleshov.
\newblock The diffusion duality.
\newblock In \emph{Proceedings of the 42nd International Conference on Machine
  Learning}, volume 267 of \emph{Proceedings of Machine Learning Research},
  pages 52584--52619. PMLR, 2025.

\bibitem[Sahoo et~al.(2026{\natexlab{a}})Sahoo, Lemercier, Yang, Deschenaux,
  Liu, Thickstun, and Juki\'{c}]{scalingdllms}
S.~S. Sahoo, J.-M. Lemercier, Z.~Yang, J.~Deschenaux, J.~Liu, J.~Thickstun, and
  A.~Juki\'{c}.
\newblock Scaling beyond masked diffusion language models.
\newblock In T.~Zhang, M.~Dudik, M.~Jaggi, A.~Agarwal, S.~Li, D.~Schuurmans,
  J.~Zhu, F.~Berkenkamp, H.~Dong, and A.~Bietti, editors, \emph{Proceedings of
  the 43rd International Conference on Machine Learning}, volume 306 of
  \emph{Proceedings of Machine Learning Research}, pages 106997--107014. PMLR,
  06--11 Jul 2026{\natexlab{a}}.

\bibitem[Sahoo et~al.(2026{\natexlab{b}})Sahoo, Yang, Akhauri, Liu, Singh,
  Cheng, Liu, Xing, Thickstun, and
  Vahdat]{sahoo2026esotericlanguagemodelsbridging}
S.~S. Sahoo, Z.~Yang, Y.~Akhauri, J.~Liu, D.~Singh, Z.~Cheng, Z.~Liu, E.~P.
  Xing, J.~Thickstun, and A.~Vahdat.
\newblock Esoteric language models: A family of any-order diffusion {LLM}s.
\newblock In T.~Zhang, M.~Dudik, M.~Jaggi, A.~Agarwal, S.~Li, D.~Schuurmans,
  J.~Zhu, F.~Berkenkamp, H.~Dong, and A.~Bietti, editors, \emph{Proceedings of
  the 43rd International Conference on Machine Learning}, volume 306 of
  \emph{Proceedings of Machine Learning Research}, pages 107015--107054. PMLR,
  06--11 Jul 2026{\natexlab{b}}.

\bibitem[Shelmanov et~al.(2025{\natexlab{a}})Shelmanov, Fadeeva, Tsvigun,
  Tsvigun, Xie, Kiselev, Daheim, Zhang, Vazhentsev, Sachan, Nakov, and
  Baldwin]{shelmanov-etal-2025-head}
A.~Shelmanov, E.~Fadeeva, A.~Tsvigun, I.~Tsvigun, Z.~Xie, I.~Kiselev,
  N.~Daheim, C.~Zhang, A.~Vazhentsev, M.~Sachan, P.~Nakov, and T.~Baldwin.
\newblock A head to predict and a head to question: Pre-trained uncertainty
  quantification heads for hallucination detection in {LLM} outputs.
\newblock In C.~Christodoulopoulos, T.~Chakraborty, C.~Rose, and V.~Peng,
  editors, \emph{Proceedings of the 2025 Conference on Empirical Methods in
  Natural Language Processing}, pages 35712--35731, Suzhou, China, Nov.
  2025{\natexlab{a}}. Association for Computational Linguistics.
\newblock ISBN 979-8-89176-332-6.

\bibitem[Shelmanov et~al.(2025{\natexlab{b}})Shelmanov, Panov, Vashurin,
  Vazhentsev, Fadeeva, and Baldwin]{shelmanov-etal-2025-uncertainty}
A.~Shelmanov, M.~Panov, R.~Vashurin, A.~Vazhentsev, E.~Fadeeva, and T.~Baldwin.
\newblock Uncertainty quantification for large language models.
\newblock In Y.~Arase, D.~Jurgens, and F.~Xia, editors, \emph{Proceedings of
  the 63rd Annual Meeting of the {A}ssociation for {C}omputational
  {L}inguistics (Volume 5: Tutorial Abstracts)}, pages 3--4, Vienna, Austria,
  July 2025{\natexlab{b}}. Association for Computational Linguistics.
\newblock ISBN 979-8-89176-255-8.

\bibitem[Sriramanan et~al.(2024)Sriramanan, Bharti, Sadasivan, Saha,
  Kattakinda, and Feizi]{NEURIPS2024_LLM}
G.~Sriramanan, S.~Bharti, V.~S. Sadasivan, S.~Saha, P.~Kattakinda, and
  S.~Feizi.
\newblock {LLM}-check: Investigating detection of hallucinations in large
  language models.
\newblock In A.~Globerson, L.~Mackey, D.~Belgrave, A.~Fan, U.~Paquet,
  J.~Tomczak, and C.~Zhang, editors, \emph{Advances in Neural Information
  Processing Systems}, volume~37, pages 34188--34216. Curran Associates, Inc.,
  2024.

\bibitem[Su et~al.(2024)Su, Wang, Ai, Hu, Wu, Zhou, and
  Liu]{su-etal-2024-unsupervised}
W.~Su, C.~Wang, Q.~Ai, Y.~Hu, Z.~Wu, Y.~Zhou, and Y.~Liu.
\newblock Unsupervised real-time hallucination detection based on the internal
  states of large language models.
\newblock In L.-W. Ku, A.~Martins, and V.~Srikumar, editors, \emph{Findings of
  the Association for Computational Linguistics: ACL 2024}, pages 14379--14391,
  Bangkok, Thailand, Aug. 2024. Association for Computational Linguistics.

\bibitem[Vashurin et~al.(2025{\natexlab{a}})Vashurin, Fadeeva, Vazhentsev,
  Rvanova, Vasilev, Tsvigun, Petrakov, Xing, Sadallah, Grishchenkov, Panchenko,
  Baldwin, Nakov, Panov, and Shelmanov]{vashurin-etal-2025-benchmarking}
R.~Vashurin, E.~Fadeeva, A.~Vazhentsev, L.~Rvanova, D.~Vasilev, A.~Tsvigun,
  S.~Petrakov, R.~Xing, A.~Sadallah, K.~Grishchenkov, A.~Panchenko, T.~Baldwin,
  P.~Nakov, M.~Panov, and A.~Shelmanov.
\newblock Benchmarking uncertainty quantification methods for large language
  models with {LM}-polygraph.
\newblock \emph{Transactions of the Association for Computational Linguistics},
  13:\penalty0 220--248, 2025{\natexlab{a}}.

\bibitem[Vashurin et~al.(2025{\natexlab{b}})Vashurin, Goloburda, Ilina,
  Rubashevskii, Nakov, Shelmanov, and Panov]{vashurin2025cocoa}
R.~Vashurin, M.~Goloburda, A.~Ilina, A.~Rubashevskii, P.~Nakov, A.~Shelmanov,
  and M.~Panov.
\newblock {CoCoA}: A minimum bayes risk framework bridging confidence and
  consistency for uncertainty quantification in {LLM}s.
\newblock In \emph{Advances in Neural Information Processing Systems},
  volume~38, 2025{\natexlab{b}}.

\bibitem[Vazhentsev et~al.(2025{\natexlab{a}})Vazhentsev, Fadeeva, Xing,
  Kuzmin, Lazichny, Panchenko, Nakov, Baldwin, Panov, and
  Shelmanov]{vazhentsev-etal-2025-unconditional}
A.~Vazhentsev, E.~Fadeeva, R.~Xing, G.~Kuzmin, I.~Lazichny, A.~Panchenko,
  P.~Nakov, T.~Baldwin, M.~Panov, and A.~Shelmanov.
\newblock Unconditional truthfulness: Learning unconditional uncertainty of
  large language models.
\newblock In C.~Christodoulopoulos, T.~Chakraborty, C.~Rose, and V.~Peng,
  editors, \emph{Proceedings of the 2025 Conference on Empirical Methods in
  Natural Language Processing}, pages 35673--35694, Suzhou, China, Nov.
  2025{\natexlab{a}}. Association for Computational Linguistics.
\newblock ISBN 979-8-89176-332-6.

\bibitem[Vazhentsev et~al.(2025{\natexlab{b}})Vazhentsev, Rvanova, Lazichny,
  Panchenko, Panov, Baldwin, and Shelmanov]{vazhentsev-etal-2025-token}
A.~Vazhentsev, L.~Rvanova, I.~Lazichny, A.~Panchenko, M.~Panov, T.~Baldwin, and
  A.~Shelmanov.
\newblock Token-level density-based uncertainty quantification methods for
  eliciting truthfulness of large language models.
\newblock In L.~Chiruzzo, A.~Ritter, and L.~Wang, editors, \emph{Proceedings of
  the 2025 Conference of the Nations of the Americas Chapter of the Association
  for Computational Linguistics: Human Language Technologies (Volume 1: Long
  Papers)}, pages 2246--2262, Albuquerque, New Mexico, Apr. 2025{\natexlab{b}}.
  Association for Computational Linguistics.
\newblock ISBN 979-8-89176-189-6.

\bibitem[Vazhentsev et~al.(2026)Vazhentsev, Rvanova, Kuzmin, Fadeeva, Lazichny,
  Panchenko, Panov, Sachan, Nakov, Baldwin, and Shelmanov]{rauq}
A.~Vazhentsev, L.~Rvanova, G.~Kuzmin, E.~Fadeeva, I.~Lazichny, A.~Panchenko,
  M.~Panov, M.~Sachan, P.~Nakov, T.~Baldwin, and A.~Shelmanov.
\newblock Efficient hallucination detection for {LLM}s using uncertainty-aware
  attention heads.
\newblock In T.~Zhang, M.~Dudik, M.~Jaggi, A.~Agarwal, S.~Li, D.~Schuurmans,
  J.~Zhu, F.~Berkenkamp, H.~Dong, and A.~Bietti, editors, \emph{Proceedings of
  the 43rd International Conference on Machine Learning}, volume 306 of
  \emph{Proceedings of Machine Learning Research}, pages 123596--123627. PMLR,
  06--11 Jul 2026.

\bibitem[Wu et~al.(2026)Wu, Zhang, Xue, Liu, Diao, Zhu, Luo, Han, and
  Xie]{wu2025fastdllmtrainingfreeaccelerationdiffusion}
C.~Wu, H.~Zhang, S.~Xue, Z.~Liu, S.~Diao, L.~Zhu, P.~Luo, S.~Han, and E.~Xie.
\newblock Fast-d{LLM}: Training-free acceleration of diffusion {LLM} by
  enabling {KV} cache and parallel decoding.
\newblock In \emph{The Fourteenth International Conference on Learning
  Representations}, 2026.

\bibitem[Xu et~al.(2025)Xu, Geffner, Kreis, Nie, Xu, Leskovec, Ermon, and
  Vahdat]{uniformdllms}
M.~Xu, T.~Geffner, K.~Kreis, W.~Nie, Y.~Xu, J.~Leskovec, S.~Ermon, and
  A.~Vahdat.
\newblock Energy-based diffusion language models for text generation.
\newblock In \emph{The Thirteenth International Conference on Learning
  Representations, {ICLR} 2025, Singapore, April 24-28, 2025}. OpenReview.net,
  2025.

\bibitem[Ye et~al.(2025)Ye, Xie, Zheng, Gao, Wu, Jiang, Li, and
  Kong]{ye2025dream}
J.~Ye, Z.~Xie, L.~Zheng, J.~Gao, Z.~Wu, X.~Jiang, Z.~Li, and L.~Kong.
\newblock Dream {7B}: Diffusion large language models.
\newblock \emph{arXiv preprint arXiv:2508.15487}, 2025.

\bibitem[Zha et~al.(2023)Zha, Yang, Li, and Hu]{zha-etal-2023-alignscore}
Y.~Zha, Y.~Yang, R.~Li, and Z.~Hu.
\newblock {A}lign{S}core: Evaluating factual consistency with a unified
  alignment function.
\newblock In A.~Rogers, J.~Boyd-Graber, and N.~Okazaki, editors,
  \emph{Proceedings of the 61st Annual Meeting of the Association for
  Computational Linguistics (Volume 1: Long Papers)}, pages 11328--11348,
  Toronto, Canada, July 2023. Association for Computational Linguistics.

\bibitem[Zhang et~al.(2024)Zhang, Liu, Basaldella, and Collier]{zhang2024luq}
C.~Zhang, F.~Liu, M.~Basaldella, and N.~Collier.
\newblock {LUQ}: Long-text uncertainty quantification for {LLM}s.
\newblock In Y.~Al-Onaizan, M.~Bansal, and Y.-N. Chen, editors,
  \emph{Proceedings of the 2024 Conference on Empirical Methods in Natural
  Language Processing}, pages 5244--5262, Miami, Florida, USA, Nov. 2024.
  Association for Computational Linguistics.

\bibitem[Zhu et~al.(2026)Zhu, Wang, Nie, Zhang, Wu, Zhou, Lin, Wen, and
  Li]{zhu2025llada}
F.~Zhu, R.~Wang, S.~Nie, X.~Zhang, C.~Wu, J.~Zhou, Y.~Lin, J.-R. Wen, and
  C.~Li.
\newblock {LLaDA} 1.5: Variance-reduced preference optimization for large
  language diffusion models.
\newblock In M.~Liakata, V.~P. Moreira, J.~Zhang, and D.~Jurgens, editors,
  \emph{Proceedings of the 64th Annual Meeting of the {A}ssociation for
  {C}omputational {L}inguistics (Volume 1: Long Papers)}, pages 11425--11460,
  San Diego, California, United States, July 2026. Association for
  Computational Linguistics.
\newblock ISBN 979-8-89176-390-6.

\end{thebibliography}

\clearpage
\appendix

\section{Proofs}
\label{app:proofs}

\subsection{Proof of Theorem~\ref{thm:dissim-lower-bound}}
\label{app:proof_t1}
  Let the reverse process be uniformly discretized into $T$ steps, and let $\zv_t$, $t \in \{1,\dots,T\}$, denote the sequence $\yv$ where each token is independently replaced by the $\left[ \mathrm{MASK} \right]$ token with probability $t/T \in (0,1]$. Let $\mathcal{M}_t$ be the set of masked positions, such that $\mathbb{E}[|\mathcal{M}_t|] = \frac{t}{T}|\yv|$. The full masking loss is $\mathcal{L}(\theta)$.
  In this case, the discretized loss is $\mathcal{L}^{\text{discr.}}(\theta) = \frac{1}{T} \sum_{t=1}^T \mathcal{L}^{\text{discr.}}_t(\theta)$ where:
  \begin{equation}
    \label{eq:loss}
    \mathcal{L}^{\text{discr.}}_t(\theta) = \mathbb{E}_{\yv, \zv_t, \xv}\!\left[\frac{T}{t}\sum_{i \in \mathcal{M}_t} -\log p_\theta(y_i \mid \zv_t, \xv)\right].
  \end{equation}
  Also, we have the property that $\lim_{T\to\infty}\mathcal{L}^{\text{discr.}}(\theta) = \mathcal{L}(\theta)$~\citep{sahoo2024simple}.
  
  The trajectory dissimilarity metric is defined as $\uv_{\mathrm{AD}}(\xv,\yv) = \frac{1}{T}\sum_{t=1}^{T} \mathbf{1}[\tilde{\yv}_t \not\equiv \yv]$.
  By linearity of expectation,
  \begin{equation}
    \mathbb{E}_{(\xv, \yv) \sim p_{\mathrm{data}}}\!\left[\uv_{\mathrm{AD}}(\xv,\yv)\right] = \frac{1}{T}\sum_{t=1}^{T} \mathbb{E}_{(\xv, \yv) \sim p_{\mathrm{data}}}[\mathbf{1}[\tilde{\yv}_t \not\equiv \yv]] = \frac{1}{T}\sum_{t=1}^{T} \mathbb{P}\bigl(\tilde{\yv}_t \not\equiv \yv\bigr).
  \end{equation}

  Conditioning on $(\zv_t, \xv)$ and using the calibration assumption,
  $\mathbb{P}(\tilde{\yv}_t \not\equiv \yv \mid \zv_t, \xv) = 1 - p_\theta(\tilde{\yv}_t \mid \zv_t, \xv)$.
  Marginalizing over $(\zv_t, \xv)$, we obtain:
  \begin{equation}
    \label{eq:p_ad}
    \mathbb{P}(\tilde{\yv}_t \not\equiv \yv) = \mathbb{E}_{(\zv_t, \xv)}\!\left[1 - p_\theta(\tilde{\yv}_t \mid \zv_t, \xv)\right].
  \end{equation}

  Since $\tilde{\yv}_t = \argmax_{\yv'} p_\theta(\yv' \mid \zv_t, \xv)$, we have $p_\theta(\tilde{\yv}_t \mid \zv_t, \xv) \ge p_\theta(\yv' \mid \zv_t, \xv)$ for any $\yv'$. Thus, employing the standard inequality $1 - p \le -\log p$, we get:
  \begin{equation}
  \label{eq:p_ineq}
    1 - p_\theta(\tilde{\yv}_{t} \mid \zv_t, \xv) \le 1 - p_\theta(\yv' \mid \zv_t, \xv) \le -\log p_\theta(\yv' \mid \zv_t, \xv).
  \end{equation}

  Since unmasked positions are observed, only masked positions contribute, and by the subadditivity of entropy, in expectation over $\yv' \sim p_\theta(\cdot \mid \zv_t, \xv)$:
  \begin{equation}
  \label{eq:p_mask}
    \mathbb{E}_{\yv'}\!\left[-\log p_\theta(\yv' \mid \zv_t, \xv)\right] \le \mathbb{E}_{\yv'}\!\left[\sum_{i \in \mathcal{M}_t} -\log p_\theta(y_i' \mid \zv_t, \xv)\right].
  \end{equation}

  Taking expectations over $(\zv_t, \xv, \yv')$ and applying equations~\eqref{eq:p_ineq} and~\eqref{eq:p_mask},
  \begin{equation}
    \mathbb{E}_{\zv_t, \xv} [1 - p_{\theta}(\tilde{\yv}_t \mid \zv_t, \xv)]
    \le \mathbb{E}_{\yv', \zv_t, \xv}\!\left[-\log p_\theta(\yv' \mid \zv_t, \xv)\right]
    \le \mathbb{E}_{\yv', \zv_t, \xv}\!\left[\sum_{i \in \mathcal{M}_t} -\log p_\theta(y_i' \mid \zv_t, \xv)\right]. \nonumber
  \end{equation}
  Combining equations~\eqref{eq:loss} and~\eqref{eq:p_ad} and changing notation $\yv' = \yv$, we get:
  \begin{equation}
      \mathbb{P}(\tilde{\yv}_t \not\equiv \yv) \leq \mathbb{E}_{\yv, \zv_t, \xv}\!\left[\sum_{i \in \mathcal{M}_t} -\log p_\theta(y_i \mid \zv_t, \xv)\right] = \frac{t}{T} \mathcal{L}_t^{\text{discr.}}(\theta)
  \end{equation}
  
  Plugging this bound back into the expectation of the metric:
  \begin{equation}
    \mathbb{E}[\uv_{\mathrm{AD}}(\xv,\yv)] \le \frac{1}{T}\sum_{t=1}^{T} \frac{t}{T} \mathcal{L}_{t}^{\text{discr.}}(\theta)
  \end{equation}

  Since $\frac{t}{T} \le1, \forall t,$ we have:
  \begin{equation}
    \mathbb{E}[\uv_{\mathrm{AD}}(\xv,\yv)] \le \frac{1}{T}\sum_{t=1}^{T} \frac{t}{T} \mathcal{L}_{t}^{\text{discr.}}(\theta) \le \frac{1}{T}\sum_{t=1}^{T} \mathcal{L}_{t}^{\text{discr.}}(\theta) = 
    \mathcal{L}^{\text{discr.}}(\theta)
  \end{equation}
  
  Taking $T\to \infty$ gives us:
  \begin{equation}
    \mathbb{E}[\uv_{\mathrm{AD}}(\xv,\yv)] \le \mathcal{L}(\theta), 
  \end{equation}
  which completes the proof.

\subsection{Theorem~\ref{thm:dissim-lower-bound} under Model Miscalibration}
\label{app:miscalibration}

  Although Theorem~\ref{thm:dissim-lower-bound} assumes perfect calibration, this principled connection remains valid under model miscalibration as well.

  Let $p_\theta$ and $q$ denote the model and true data distributions, respectively. We define the pointwise calibration error as
  \begin{equation}
    \epsilon_t(\xv, \zv_t) = \bigl| p_\theta(\tilde{\yv}_t \mid \zv_t, \xv) - q(\tilde{\yv}_t \mid \zv_t, \xv) \bigr|.
  \end{equation}
  The expected posterior-mismatch term is then
  \begin{equation}
    \epsilon = \mathbb{E}_{t, \xv, \zv_t}\bigl[\epsilon_t(\xv, \zv_t)\bigr].
  \end{equation}
  Specifically,
  \begin{equation}
    1 - q(\tilde{\yv}_t \mid \zv_t, \xv) \le -\log p_\theta(\tilde{\yv}_t \mid \zv_t, \xv) + \epsilon_t(\xv, \zv_t).
  \end{equation}
  Therefore, following the remaining steps of the proof of Theorem~\ref{thm:dissim-lower-bound},
  \begin{equation}
    \mathbb{E}[\uv_{\mathrm{AD}}(\xv,\yv)] \le \mathcal{L}(\theta) + \epsilon.
  \end{equation}
  Thus, under miscalibration, the bound includes an additive term $\epsilon$ that measures the discrepancy between the model and true data distributions. This term loosens the bound by at most $\epsilon$ but does not invalidate the theoretical connection. Its exact practical tightness cannot be estimated in advance for a given model because $q$ is unknown.

\subsection{Proof of Proposition~\ref{propos:prog_traj_dissim}}
\label{app:proof_p1}
  Since $w_t=t/T$, we have $1/T \leq w_t \leq 1$ for all $t$.
  Because $D(\tilde{\yv}_t,\yv)\geq 0$, it follows that
  \begin{equation}
    \frac{1}{T}D(\tilde{\yv}_t,\yv)
    \leq
    w_tD(\tilde{\yv}_t,\yv)
    \leq
    D(\tilde{\yv}_t,\yv).
  \end{equation}
  Averaging over $t=1,\dots,T$ gives
  \begin{equation}
    \frac{1}{T}
    \left(
    \frac{1}{T}
    \sum_{t=1}^{T}
    D(\tilde{\yv}_t,\yv)
    \right)
    \leq
    \frac{1}{T}
    \sum_{t=1}^{T}
    w_tD(\tilde{\yv}_t,\yv)
    \leq
    \frac{1}{T}
    \sum_{t=1}^{T}
    D(\tilde{\yv}_t,\yv).
  \end{equation}
  Using the definitions of $\uv_{\mathrm{AD}}$ and
  $\uv_{\mathrm{AD}}^{\mathrm{prog}}$ completes the proof.

\clearpage
\section{Detailed Results}
\label{app:res}

\subsection{Results for LLaDA-1.5}
\label{app:res_llada}
\begin{figure*}[h!]
    \centering
    \includegraphics[trim={0.cm 0.cm 0.cm 0.cm},clip,width=0.99\linewidth]{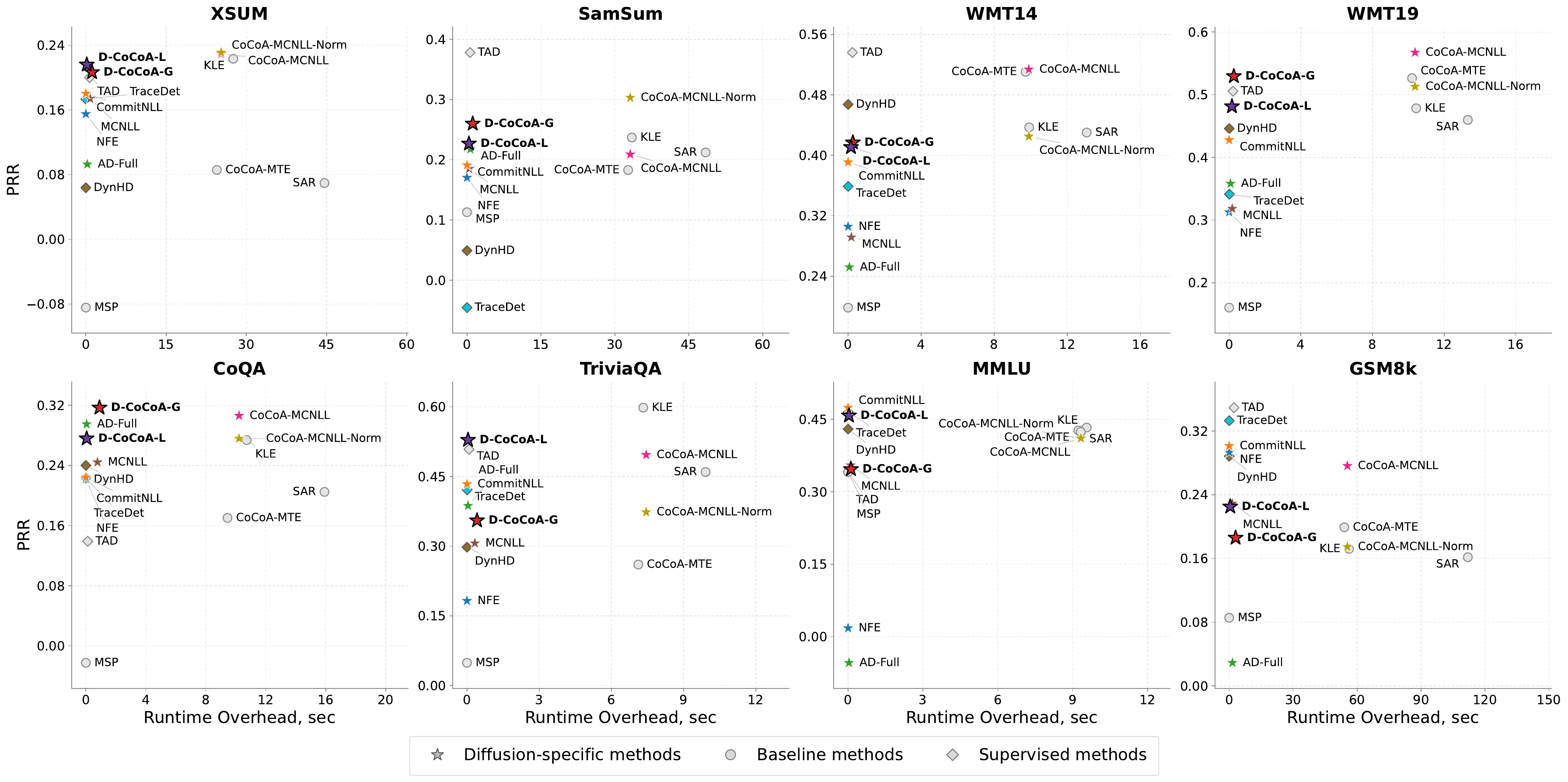}
    \caption{
    \textbf{LLaDA-1.5 performance and runtime.} Comparison of uncertainty quantification performance (PRR$\uparrow$) versus computational complexity (Time in seconds$\downarrow$) across eight benchmark datasets for the LLaDA-1.5 model (see Figures~\ref{fig:dream_prr_vs_comp}, \ref{fig:refusion_prr_vs_comp}, and~\ref{fig:diffusiongemma_prr_vs_comp} for the Dream, ReFusion, and DiffusionGemma models, respectively).
    }
    \label{fig:llada_prr_vs_comp}
  \end{figure*}

  Tables~\ref{tab:ar_improved_results}, \ref{tab:ar_traj_results}, and~\ref{tab:detailed_main_results} present the results for LLaDA-1.5 with sampling-based baselines using MCNLL, sampling-based baselines using denoising trajectory states, and the best method from each category, respectively. Figure~\ref{fig:llada_prr_vs_comp} presents the performance-efficiency trade-off for each dataset.

  \begin{table*}[!ht] \resizebox{\textwidth}{!}{\begin{tabular}{ll|c|c|c|c|c|c|c|c|c}
\toprule
\textbf{Category} & \textbf{UQ Method} & \textbf{XSum} & \textbf{SamSum} & \textbf{WMT14} & \textbf{WMT19} & \textbf{CoQA} & \textbf{TriviaQA} & \textbf{MMLU} & \textbf{GSM8k} & \textbf{Mean} \\
\midrule
\multirow{4}{*}{AR} & MCNSE & \cellcolor[rgb]{0.61490285,0.649358983,0.8768415764999999} -0.07 & \cellcolor[rgb]{0.61490285,0.649358983,0.8768415764999999} 0.07 & \cellcolor[rgb]{0.61490285,0.649358983,0.8768415764999999} 0.29 & \cellcolor[rgb]{0.61490285,0.649358983,0.8768415764999999} 0.25 & \cellcolor[rgb]{0.61490285,0.649358983,0.8768415764999999} -0.00 & \cellcolor[rgb]{0.61490285,0.649358983,0.8768415764999999} 0.08 & \cellcolor[rgb]{0.9813503623666666,0.8141088755,0.7538180498} 0.30 & \cellcolor[rgb]{0.6355521478235294,0.6800053309882352,0.9035475637176471} 0.06 & \cellcolor[rgb]{0.61490285,0.649358983,0.8768415764999999} 0.12 \\
 & SemanticEntropy & \cellcolor[rgb]{0.6741616707058824,0.7328555732549019,0.9441730814705882} -0.04 & \cellcolor[rgb]{0.8644847897843138,0.9087320678529411,0.9865938341862746} 0.15 & \cellcolor[rgb]{0.7907430740941177,0.8567252977647059,0.9991571764705882} 0.35 & \cellcolor[rgb]{0.8389114905588235,0.8932732186176471,0.9955022941235294} 0.36 & \cellcolor[rgb]{0.6944259359764706,0.7581492177882353,0.9606867415411765} 0.04 & \cellcolor[rgb]{0.6569731755647059,0.7100258308470588,0.927496270517647} 0.11 & \cellcolor[rgb]{0.852836579,0.50777808,0.575116406} 0.42 & \cellcolor[rgb]{0.815044265017647,0.8762581198529411,0.9992540061705882} 0.13 & \cellcolor[rgb]{0.7880256462666666,0.8543899393333334,0.9988772960000001} 0.19 \\
 & SemanticDensity & \cellcolor[rgb]{0.9004151256333333,0.6254146054,0.6128478516333333} 0.21 & \cellcolor[rgb]{0.8871684250764706,0.5998796340235294,0.6012672772176471} 0.27 & \cellcolor[rgb]{0.9717157648333333,0.9011381268078431,0.8645857989568628} 0.42 & \cellcolor[rgb]{0.9797588292352941,0.8834864272549019,0.8370723575196078} 0.45 & \cellcolor[rgb]{0.8042736801705883,0.8678626149117648,0.9996769126490196} 0.09 & \cellcolor[rgb]{0.9423217193470588,0.7050085489411765,0.6612532746235295} 0.44 & \cellcolor[rgb]{0.8177369112294117,0.8783569960882354,0.9991482795509804} 0.11 & \cellcolor[rgb]{0.6842533066588236,0.7455705968156863,0.9526215641058824} 0.08 & \cellcolor[rgb]{0.9514243351588236,0.9223978252941176,0.9059849168705882} 0.26 \\
 & CoCoA-MSP & \cellcolor[rgb]{0.7635661295607843,0.8323498010588236,0.9945323233411765} 0.00 & \cellcolor[rgb]{0.8933603506784313,0.9224036051843137,0.9699051924745098} 0.16 & \cellcolor[rgb]{0.6970208390666667,0.7612067269333334,0.9624581464666666} 0.32 & \cellcolor[rgb]{0.7771559349568627,0.8450485056078432,0.9977577741176471} 0.33 & \cellcolor[rgb]{0.7825907906117646,0.8497192224705883,0.9983175350588236} 0.08 & \cellcolor[rgb]{0.7500152822588235,0.8192542337882354,0.9905350620529412} 0.17 & \cellcolor[rgb]{0.852836579,0.50777808,0.575116406} 0.42 & \cellcolor[rgb]{0.9133921582352942,0.9291026777686274,0.9534763223137255} 0.17 & \cellcolor[rgb]{0.8415278407803921,0.8950213134450979,0.9948842140921569} 0.21 \\
\midrule
\multirow{4}{*}{\shortstack[l]{AR +\\MCNLL}} & MCNSE-MCNLL & \cellcolor[rgb]{0.9528917390058824,0.727592846082353,0.6776679419235294} 0.18 & \cellcolor[rgb]{0.8336264621666667,0.8895882285,0.9964796065} 0.14 & \cellcolor[rgb]{0.8844643114450981,0.5946066788549019,0.5989826965745098} 0.50 & \cellcolor[rgb]{0.852836579,0.50777808,0.575116406} 0.57 & \cellcolor[rgb]{0.9563825307352941,0.9183409471764705,0.8972560558823529} 0.17 & \cellcolor[rgb]{0.8792694128431373,0.9163932970294117,0.9792039273627451} 0.25 & \cellcolor[rgb]{0.61490285,0.649358983,0.8768415764999999} -0.03 & \cellcolor[rgb]{0.9514243351588236,0.9223978252941176,0.9059849168705882} 0.19 & \cellcolor[rgb]{0.931695915645098,0.9325418979098039,0.9338169421313726} 0.25 \\
 & SemanticEntropy-MCNLL & \cellcolor[rgb]{0.9196824685392158,0.6609281104705882,0.632461990490196} 0.20 & \cellcolor[rgb]{0.9196757213862745,0.930583412827451,0.9472468817078432} 0.17 & \cellcolor[rgb]{0.9090282467058823,0.927794838772549,0.9573188082745099} 0.39 & \cellcolor[rgb]{0.9797588292352941,0.8834864272549019,0.8370723575196078} 0.45 & \cellcolor[rgb]{0.9838554631666667,0.8314865062313725,0.7721615929176471} 0.22 & \cellcolor[rgb]{0.9653342981666666,0.909438499827451,0.8795731953490196} 0.32 & \cellcolor[rgb]{0.7880256462666666,0.8543899393333334,0.9988772960000001} 0.09 & \cellcolor[rgb]{0.852836579,0.50777808,0.575116406} 0.31 & \cellcolor[rgb]{0.9666105915000001,0.9077784252235295,0.8765757160705883} 0.27 \\
 & SemanticDensity-MCNLL & \cellcolor[rgb]{0.9004151256333333,0.6254146054,0.6128478516333333} 0.21 & \cellcolor[rgb]{0.852836579,0.50777808,0.575116406} 0.28 & \cellcolor[rgb]{0.9717157648333333,0.9011381268078431,0.8645857989568628} 0.42 & \cellcolor[rgb]{0.9797588292352941,0.8834864272549019,0.8370723575196078} 0.45 & \cellcolor[rgb]{0.8466606424117646,0.8981570658529412,0.9931538902647059} 0.11 & \cellcolor[rgb]{0.9528917390058824,0.727592846082353,0.6776679419235294} 0.43 & \cellcolor[rgb]{0.8492270432274511,0.8997249420568627,0.9922887283509804} 0.13 & \cellcolor[rgb]{0.61490285,0.649358983,0.8768415764999999} 0.05 & \cellcolor[rgb]{0.9514243351588236,0.9223978252941176,0.9059849168705882} 0.26 \\
 & CoCoA-MCNLL & \cellcolor[rgb]{0.852836579,0.50777808,0.575116406} 0.23 & \cellcolor[rgb]{0.9841016995,0.8604220499999999,0.8061464956666666} 0.21 & \cellcolor[rgb]{0.852836579,0.50777808,0.575116406} 0.51 & \cellcolor[rgb]{0.852836579,0.50777808,0.575116406} 0.57 & \cellcolor[rgb]{0.852836579,0.50777808,0.575116406} 0.31 & \cellcolor[rgb]{0.852836579,0.50777808,0.575116406} 0.50 & \cellcolor[rgb]{0.8675383126470588,0.5522298155294117,0.585746150627451} 0.41 & \cellcolor[rgb]{0.9283579338254901,0.6773519275058824,0.6427436489686275} 0.28 & \cellcolor[rgb]{0.852836579,0.50777808,0.575116406} 0.38 \\
\bottomrule
\end{tabular}
}\caption{\label{tab:ar_improved_results} \textbf{Baselines with diffusion likelihoods.} PRR$\uparrow$ for the LLaDA-1.5 model across various tasks for sampling-based baselines from standard autoregressive LLMs, where the diffusion likelihood (MCNLL) is used instead of the AR likelihood (see Tables~\ref{tab:dream_ar_improved_results} and \ref{tab:refusion_ar_improved_results} for the Dream and ReFusion models, respectively). Warmer color indicates better results.}\end{table*}
  \begin{table*}[!ht] \resizebox{\textwidth}{!}{\begin{tabular}{ll|c|c|c|c|c|c|c|c|c}
\toprule
\textbf{Category} & \textbf{UQ Method} & \textbf{XSum} & \textbf{SamSum} & \textbf{WMT14} & \textbf{WMT19} & \textbf{CoQA} & \textbf{TriviaQA} & \textbf{MMLU} & \textbf{GSM8k} & \textbf{Mean} \\
\midrule
\multirow{6}{*}{\shortstack[l]{AR +\\trajectory\\states}} & D-LexSim & \cellcolor[rgb]{0.8952807659705883,0.6156984995294118,0.6081210191470588} 0.09 & \cellcolor[rgb]{0.9024823794117647,0.9258330802784314,0.9630825372156863} -0.03 & \cellcolor[rgb]{0.9367011412764705,0.6934798195294117,0.6531662319882353} 0.10 & \cellcolor[rgb]{0.9175136022176471,0.6568221562117647,0.6298915758705882} 0.18 & \cellcolor[rgb]{0.8979687697431373,0.6209226426705883,0.6104148745666667} 0.21 & \cellcolor[rgb]{0.61490285,0.649358983,0.8768415764999999} 0.16 & \cellcolor[rgb]{0.931695915645098,0.9325418979098039,0.9338169421313726} 0.02 & \cellcolor[rgb]{0.9846442845,0.8424908735411765,0.7844876631294118} -0.02 & \cellcolor[rgb]{0.9791396989627451,0.8021675484411765,0.7416485506941177} 0.09 \\
 & D-DegMat & \cellcolor[rgb]{0.8952807659705883,0.6156984995294118,0.6081210191470588} 0.09 & \cellcolor[rgb]{0.9261890675039215,0.6732459732470588,0.6401732343490196} 0.01 & \cellcolor[rgb]{0.9175136022176471,0.6568221562117647,0.6298915758705882} 0.11 & \cellcolor[rgb]{0.8704786593764706,0.5611201626352941,0.5878720995529412} 0.21 & \cellcolor[rgb]{0.9150932609745098,0.6523663817764705,0.6274457140333334} 0.20 & \cellcolor[rgb]{0.924020201182353,0.6691400189882353,0.6376028197294118} 0.41 & \cellcolor[rgb]{0.61490285,0.649358983,0.8768415764999999} 0.00 & \cellcolor[rgb]{0.947942297417647,0.7165372783529411,0.6693403172588235} -0.00 & \cellcolor[rgb]{0.8871684250764706,0.5998796340235294,0.6012672772176471} 0.13 \\
 & D-EigVal & \cellcolor[rgb]{0.8952807659705883,0.6156984995294118,0.6081210191470588} 0.09 & \cellcolor[rgb]{0.852836579,0.50777808,0.575116406} 0.02 & \cellcolor[rgb]{0.8979687697431373,0.6209226426705883,0.6104148745666667} 0.12 & \cellcolor[rgb]{0.8871684250764706,0.5998796340235294,0.6012672772176471} 0.20 & \cellcolor[rgb]{0.8979687697431373,0.6209226426705883,0.6104148745666667} 0.21 & \cellcolor[rgb]{0.852836579,0.50777808,0.575116406} 0.44 & \cellcolor[rgb]{0.852836579,0.50777808,0.575116406} 0.04 & \cellcolor[rgb]{0.947942297417647,0.7165372783529411,0.6693403172588235} 0.00 & \cellcolor[rgb]{0.852836579,0.50777808,0.575116406} 0.14 \\
 & D-Ecc & \cellcolor[rgb]{0.852836579,0.50777808,0.575116406} 0.10 & \cellcolor[rgb]{0.9261890675039215,0.6732459732470588,0.6401732343490196} 0.01 & \cellcolor[rgb]{0.852836579,0.50777808,0.575116406} 0.14 & \cellcolor[rgb]{0.852836579,0.50777808,0.575116406} 0.22 & \cellcolor[rgb]{0.852836579,0.50777808,0.575116406} 0.23 & \cellcolor[rgb]{0.9653342981666666,0.909438499827451,0.8795731953490196} 0.32 & \cellcolor[rgb]{0.931695915645098,0.9325418979098039,0.9338169421313726} 0.02 & \cellcolor[rgb]{0.9053078374137256,0.6343985308588236,0.6177138057666667} 0.01 & \cellcolor[rgb]{0.8871684250764706,0.5998796340235294,0.6012672772176471} 0.13 \\
 & D-KLE & \cellcolor[rgb]{0.61490285,0.649358983,0.8768415764999999} -0.07 & \cellcolor[rgb]{0.61490285,0.649358983,0.8768415764999999} -0.07 & \cellcolor[rgb]{0.61490285,0.649358983,0.8768415764999999} -0.17 & \cellcolor[rgb]{0.61490285,0.649358983,0.8768415764999999} -0.20 & \cellcolor[rgb]{0.61490285,0.649358983,0.8768415764999999} -0.09 & \cellcolor[rgb]{0.7258692802352942,0.794090494117647,0.9801006337941176} 0.21 & \cellcolor[rgb]{0.931695915645098,0.9325418979098039,0.9338169421313726} 0.02 & \cellcolor[rgb]{0.61490285,0.649358983,0.8768415764999999} -0.11 & \cellcolor[rgb]{0.61490285,0.649358983,0.8768415764999999} -0.06 \\
 & D-LUQ & \cellcolor[rgb]{0.9305268001470588,0.6814578817647059,0.6453140635882353} 0.08 & \cellcolor[rgb]{0.852836579,0.50777808,0.575116406} 0.02 & \cellcolor[rgb]{0.9634414899941177,0.752710773145098,0.6974329388568627} 0.08 & \cellcolor[rgb]{0.9175136022176471,0.6568221562117647,0.6298915758705882} 0.18 & \cellcolor[rgb]{0.9622044108411765,0.7492947789176471,0.6945295061352941} 0.17 & \cellcolor[rgb]{0.9423217193470588,0.7050085489411765,0.6612532746235295} 0.40 & \cellcolor[rgb]{0.931695915645098,0.9325418979098039,0.9338169421313726} 0.02 & \cellcolor[rgb]{0.852836579,0.50777808,0.575116406} 0.02 & \cellcolor[rgb]{0.9196824685392158,0.6609281104705882,0.632461990490196} 0.12 \\
\bottomrule
\end{tabular}
}\caption{\label{tab:ar_traj_results} \textbf{Baselines with trajectory states.} PRR$\uparrow$ for the LLaDA-1.5 model across various tasks for sampling-based baselines from standard autoregressive LLMs, where denoising trajectory states are used instead of sampled generations (see Tables~\ref{tab:dream_ar_traj_results}, \ref{tab:refusion_ar_traj_results}, and~\ref{tab:diffusiongemma_ar_traj_results} for the Dream, ReFusion, and DiffusionGemma models, respectively). Warmer color indicates better results.}\end{table*}
    
  \begin{table*}[!ht] \resizebox{\textwidth}{!}{\begin{tabular}{ll|c|c|c|c|c|c|c|c|c}
\toprule
\textbf{Category} & \textbf{UQ Method} & \textbf{XSum} & \textbf{SamSum} & \textbf{WMT14} & \textbf{WMT19} & \textbf{CoQA} & \textbf{TriviaQA} & \textbf{MMLU} & \textbf{GSM8k} & \textbf{Mean} \\
\midrule
\multirow{5}{*}{\shortstack[l]{AR\\baselines}} & MSP & \cellcolor[rgb]{0.61490285,0.649358983,0.8768415764999999} -0.08 & \cellcolor[rgb]{0.8594926464901961,0.905996446872549,0.9888280806960784} 0.11 & \cellcolor[rgb]{0.61490285,0.649358983,0.8768415764999999} 0.20 & \cellcolor[rgb]{0.61490285,0.649358983,0.8768415764999999} 0.16 & \cellcolor[rgb]{0.61490285,0.649358983,0.8768415764999999} -0.02 & \cellcolor[rgb]{0.61490285,0.649358983,0.8768415764999999} 0.05 & \cellcolor[rgb]{0.9791396989627451,0.8021675484411765,0.7416485506941177} 0.34 & \cellcolor[rgb]{0.7311771982901961,0.7999150558117647,0.9829285958960784} 0.09 & \cellcolor[rgb]{0.61490285,0.649358983,0.8768415764999999} 0.10 \\
 & SAR & \cellcolor[rgb]{0.9236824528058825,0.9312362411882353,0.9427702351823529} 0.07 & \cellcolor[rgb]{0.974575254145098,0.8953926345333334,0.8554377971509803} 0.21 & \cellcolor[rgb]{0.9845960255235294,0.8529178378588235,0.7968521304901961} 0.43 & \cellcolor[rgb]{0.9813503623666666,0.8141088755,0.7538180498} 0.46 & \cellcolor[rgb]{0.9824176791176471,0.8723068372941176,0.8216194376764705} 0.20 & \cellcolor[rgb]{0.9796923648137255,0.8051528802058823,0.7446909254705882} 0.46 & \cellcolor[rgb]{0.9175136022176471,0.6568221562117647,0.6298915758705882} 0.42 & \cellcolor[rgb]{0.8792694128431373,0.9163932970294117,0.9792039273627451} 0.16 & \cellcolor[rgb]{0.9836582578333333,0.8287354144039216,0.7690800753647059} 0.30 \\
 & CoCoA-MTE & \cellcolor[rgb]{0.9563825307352941,0.9183409471764705,0.8972560558823529} 0.09 & \cellcolor[rgb]{0.9497716033000001,0.923750118,0.9088945372} 0.18 & \cellcolor[rgb]{0.9126469050843138,0.6478744190470588,0.6250127369666667} 0.51 & \cellcolor[rgb]{0.9175136022176471,0.6568221562117647,0.6298915758705882} 0.53 & \cellcolor[rgb]{0.9613407263117647,0.9142840690588235,0.8885271948941176} 0.17 & \cellcolor[rgb]{0.8644847897843138,0.9087320678529411,0.9865938341862746} 0.26 & \cellcolor[rgb]{0.9053078374137256,0.6343985308588236,0.6177138057666667} 0.43 & \cellcolor[rgb]{0.9497716033000001,0.923750118,0.9088945372} 0.20 & \cellcolor[rgb]{0.9836582578333333,0.8287354144039216,0.7690800753647059} 0.30 \\
 & KLE & \cellcolor[rgb]{0.8763519705509804,0.5787878133490196,0.5921289546450981} 0.22 & \cellcolor[rgb]{0.9844312501823529,0.8554192419058824,0.7999502522156863} 0.24 & \cellcolor[rgb]{0.9840526685,0.8342375980588235,0.7752431104705882} 0.44 & \cellcolor[rgb]{0.9720272867117647,0.7765767393745098,0.7177742451568627} 0.48 & \cellcolor[rgb]{0.9441952453705882,0.708851458745098,0.6639489555019608} 0.27 & \cellcolor[rgb]{0.852836579,0.50777808,0.575116406} 0.60 & \cellcolor[rgb]{0.9053078374137256,0.6343985308588236,0.6177138057666667} 0.43 & \cellcolor[rgb]{0.9003004236470589,0.9251791607803921,0.9650037801960785} 0.17 & \cellcolor[rgb]{0.9028614815235294,0.6299065681294118,0.6152808287} 0.36 \\
 & TAD & \cellcolor[rgb]{0.9175136022176471,0.6568221562117647,0.6298915758705882} 0.20 & \cellcolor[rgb]{0.852836579,0.50777808,0.575116406} 0.38 & \cellcolor[rgb]{0.852836579,0.50777808,0.575116406} 0.54 & \cellcolor[rgb]{0.9441952453705882,0.708851458745098,0.6639489555019608} 0.51 & \cellcolor[rgb]{0.9176723556764705,0.9302569986470588,0.9494852049705882} 0.14 & \cellcolor[rgb]{0.9513294646843138,0.7239693735803922,0.6748606867705882} 0.51 & \cellcolor[rgb]{0.9791396989627451,0.8021675484411765,0.7416485506941177} 0.34 & \cellcolor[rgb]{0.852836579,0.50777808,0.575116406} 0.35 & \cellcolor[rgb]{0.8790560841823529,0.5840607685176471,0.5944135352882353} 0.37 \\\midrule
\multirow{2}{*}{\shortstack[l]{Diffusion\\baselines}} & TraceDet & \cellcolor[rgb]{0.9634414899941177,0.752710773145098,0.6974329388568627} 0.17 & \cellcolor[rgb]{0.61490285,0.649358983,0.8768415764999999} -0.05 & \cellcolor[rgb]{0.9176723556764705,0.9302569986470588,0.9494852049705882} 0.36 & \cellcolor[rgb]{0.9003004236470589,0.9251791607803921,0.9650037801960785} 0.34 & \cellcolor[rgb]{0.9840526685,0.8342375980588235,0.7752431104705882} 0.22 & \cellcolor[rgb]{0.9844312501823529,0.8554192419058824,0.7999502522156863} 0.42 & \cellcolor[rgb]{0.852836579,0.50777808,0.575116406} 0.47 & \cellcolor[rgb]{0.8952807659705883,0.6156984995294118,0.6081210191470588} 0.33 & \cellcolor[rgb]{0.9818859091411765,0.8745427552862746,0.824710021645098} 0.28 \\
 & DynHD & \cellcolor[rgb]{0.9068462909411765,0.9271409192745098,0.959240051254902} 0.06 & \cellcolor[rgb]{0.7635661295607843,0.8323498010588236,0.9945323233411765} 0.05 & \cellcolor[rgb]{0.9671527274529412,0.762958755827451,0.7061432370215687} 0.47 & \cellcolor[rgb]{0.9838554631666667,0.8314865062313725,0.7721615929176471} 0.45 & \cellcolor[rgb]{0.9756268974411765,0.7893996947941176,0.729703902882353} 0.24 & \cellcolor[rgb]{0.9090282467058823,0.927794838772549,0.9573188082745099} 0.30 & \cellcolor[rgb]{0.9053078374137256,0.6343985308588236,0.6177138057666667} 0.43 & \cellcolor[rgb]{0.9607031106137255,0.7457102085921569,0.6917042176882353} 0.29 & \cellcolor[rgb]{0.9845960255235294,0.8529178378588235,0.7968521304901961} 0.29 \\\midrule
\multirow{4}{*}{\shortstack[l]{Diffusion\\signals}} & MCNLL & \cellcolor[rgb]{0.9634414899941177,0.752710773145098,0.6974329388568627} 0.17 & \cellcolor[rgb]{0.9497716033000001,0.923750118,0.9088945372} 0.18 & \cellcolor[rgb]{0.7853082184392157,0.8520545809019608,0.9985974155294117} 0.29 & \cellcolor[rgb]{0.8694129974705882,0.9112858109117647,0.9841305319117647} 0.32 & \cellcolor[rgb]{0.9756268974411765,0.7893996947941176,0.729703902882353} 0.24 & \cellcolor[rgb]{0.9196757213862745,0.930583412827451,0.9472468817078432} 0.31 & \cellcolor[rgb]{0.9747269947588235,0.7861939559392157,0.7267214884509804} 0.35 & \cellcolor[rgb]{0.9797588292352941,0.8834864272549019,0.8370723575196078} 0.23 & \cellcolor[rgb]{0.9653342981666666,0.909438499827451,0.8795731953490196} 0.26 \\
 & NFE & \cellcolor[rgb]{0.9802450306647059,0.8081382119705882,0.7477333002470588} 0.15 & \cellcolor[rgb]{0.9377786937156862,0.9301210794313726,0.9257150330490196} 0.17 & \cellcolor[rgb]{0.8256989195784314,0.8840607433235295,0.9979455750647059} 0.31 & \cellcolor[rgb]{0.8543598448588235,0.9028606944647058,0.9905584045235294} 0.31 & \cellcolor[rgb]{0.9840526685,0.8342375980588235,0.7752431104705882} 0.22 & \cellcolor[rgb]{0.7662841187058823,0.8349002989411765,0.9951966350588235} 0.18 & \cellcolor[rgb]{0.6970208390666667,0.7612067269333334,0.9624581464666666} 0.02 & \cellcolor[rgb]{0.9607031106137255,0.7457102085921569,0.6917042176882353} 0.29 & \cellcolor[rgb]{0.8718771013137254,0.9125626824411764,0.9828988807745098} 0.21 \\
 & AD-Full & \cellcolor[rgb]{0.9563825307352941,0.9183409471764705,0.8972560558823529} 0.09 & \cellcolor[rgb]{0.9797588292352941,0.8834864272549019,0.8370723575196078} 0.22 & \cellcolor[rgb]{0.7048055483372548,0.7703792543686274,0.9677723612431373} 0.25 & \cellcolor[rgb]{0.9256858185156862,0.9315626553686274,0.9405319119196078} 0.36 & \cellcolor[rgb]{0.9126469050843138,0.6478744190470588,0.6250127369666667} 0.29 & \cellcolor[rgb]{0.9781854635254902,0.8875721630666666,0.8432079741549019} 0.39 & \cellcolor[rgb]{0.61490285,0.649358983,0.8768415764999999} -0.05 & \cellcolor[rgb]{0.61490285,0.649358983,0.8768415764999999} 0.03 & \cellcolor[rgb]{0.8492270432274511,0.8997249420568627,0.9922887283509804} 0.20 \\
 & CommitNLL & \cellcolor[rgb]{0.9513294646843138,0.7239693735803922,0.6748606867705882} 0.18 & \cellcolor[rgb]{0.9596879944529412,0.9156363617647059,0.8914368152235295} 0.19 & \cellcolor[rgb]{0.9613407263117647,0.9142840690588235,0.8885271948941176} 0.39 & \cellcolor[rgb]{0.9837721488176471,0.8654248580941176,0.812342739117647} 0.43 & \cellcolor[rgb]{0.9840526685,0.8342375980588235,0.7752431104705882} 0.22 & \cellcolor[rgb]{0.9848414898333333,0.8452419653686274,0.7875691806823529} 0.43 & \cellcolor[rgb]{0.852836579,0.50777808,0.575116406} 0.47 & \cellcolor[rgb]{0.947942297417647,0.7165372783529411,0.6693403172588235} 0.30 & \cellcolor[rgb]{0.9575785619705882,0.7384632635882353,0.6860897073823529} 0.33 \\\midrule
\multirow{2}{*}{\shortstack[l]{AR +\\MCNLL}} & CoCoA-MCNLL & \cellcolor[rgb]{0.852836579,0.50777808,0.575116406} 0.23 & \cellcolor[rgb]{0.974575254145098,0.8953926345333334,0.8554377971509803} 0.21 & \cellcolor[rgb]{0.9126469050843138,0.6478744190470588,0.6250127369666667} 0.51 & \cellcolor[rgb]{0.852836579,0.50777808,0.575116406} 0.57 & \cellcolor[rgb]{0.8734190061058824,0.5700105097411765,0.5899980484784314} 0.31 & \cellcolor[rgb]{0.9591408362921569,0.742086736090196,0.6888969625352941} 0.50 & \cellcolor[rgb]{0.9283579338254901,0.6773519275058824,0.6427436489686275} 0.41 & \cellcolor[rgb]{0.9708639649117647,0.7732067385098039,0.7148535351862745} 0.28 & \cellcolor[rgb]{0.852836579,0.50777808,0.575116406} 0.38 \\
 & CoCoA-MCNLL-Norm & \cellcolor[rgb]{0.852836579,0.50777808,0.575116406} 0.23 & \cellcolor[rgb]{0.9607031106137255,0.7457102085921569,0.6917042176882353} 0.30 & \cellcolor[rgb]{0.9824176791176471,0.8723068372941176,0.8216194376764705} 0.42 & \cellcolor[rgb]{0.9441952453705882,0.708851458745098,0.6639489555019608} 0.51 & \cellcolor[rgb]{0.9305268001470588,0.6814578817647059,0.6453140635882353} 0.28 & \cellcolor[rgb]{0.9678868848333333,0.9061183506196078,0.8735782367921568} 0.37 & \cellcolor[rgb]{0.9283579338254901,0.6773519275058824,0.6427436489686275} 0.41 & \cellcolor[rgb]{0.9003004236470589,0.9251791607803921,0.9650037801960785} 0.17 & \cellcolor[rgb]{0.9423217193470588,0.7050085489411765,0.6612532746235295} 0.34 \\\midrule
\multirow{2}{*}{\shortstack[l]{\textbf{D-CoCoA}\\\textbf{(ours)}}} & D-CoCoA-G & \cellcolor[rgb]{0.8979687697431373,0.6209226426705883,0.6104148745666667} 0.21 & \cellcolor[rgb]{0.9830083599196078,0.8230648707941176,0.7629451741294118} 0.26 & \cellcolor[rgb]{0.9824176791176471,0.8723068372941176,0.8216194376764705} 0.42 & \cellcolor[rgb]{0.9175136022176471,0.6568221562117647,0.6298915758705882} 0.53 & \cellcolor[rgb]{0.852836579,0.50777808,0.575116406} 0.32 & \cellcolor[rgb]{0.9627817115,0.9127586490352941,0.8855681539058824} 0.36 & \cellcolor[rgb]{0.9747269947588235,0.7861939559392157,0.7267214884509804} 0.35 & \cellcolor[rgb]{0.9337138175431372,0.9321882998862745,0.9313012310098039} 0.19 & \cellcolor[rgb]{0.9575785619705882,0.7384632635882353,0.6860897073823529} 0.33 \\
 & D-CoCoA-L & \cellcolor[rgb]{0.8763519705509804,0.5787878133490196,0.5921289546450981} 0.22 & \cellcolor[rgb]{0.9829494490941176,0.8700709193019608,0.8185288537078431} 0.23 & \cellcolor[rgb]{0.9781854635254902,0.8875721630666666,0.8432079741549019} 0.41 & \cellcolor[rgb]{0.9720272867117647,0.7765767393745098,0.7177742451568627} 0.48 & \cellcolor[rgb]{0.9305268001470588,0.6814578817647059,0.6453140635882353} 0.28 & \cellcolor[rgb]{0.9348276152529411,0.6896369097254902,0.6504705511098039} 0.53 & \cellcolor[rgb]{0.864597965917647,0.5433394684235294,0.5836202017019608} 0.46 & \cellcolor[rgb]{0.9797588292352941,0.8834864272549019,0.8370723575196078} 0.23 & \cellcolor[rgb]{0.924020201182353,0.6691400189882353,0.6376028197294118} 0.35 \\
\bottomrule
\end{tabular}}\caption{\label{tab:detailed_main_results} \textbf{D-CoCoA comparison.} PRR$\uparrow$ for the LLaDA-1.5 model across various tasks for D-CoCoA methods, comparing the best method from each category (see Tables~\ref{tab:dream_detailed_main_results}, \ref{tab:refusion_detailed_main_results}, and~\ref{tab:diffusiongemma_detailed_main_results} for the Dream, ReFusion, and DiffusionGemma models, respectively). Warmer color indicates better results.}\end{table*}

\clearpage
\subsection{Results for Dream}
\label{app:res_dream}
\begin{figure*}[h!]
    \centering
    \includegraphics[trim={0.cm 0.cm 0.cm 0.cm},clip,width=0.99\linewidth]{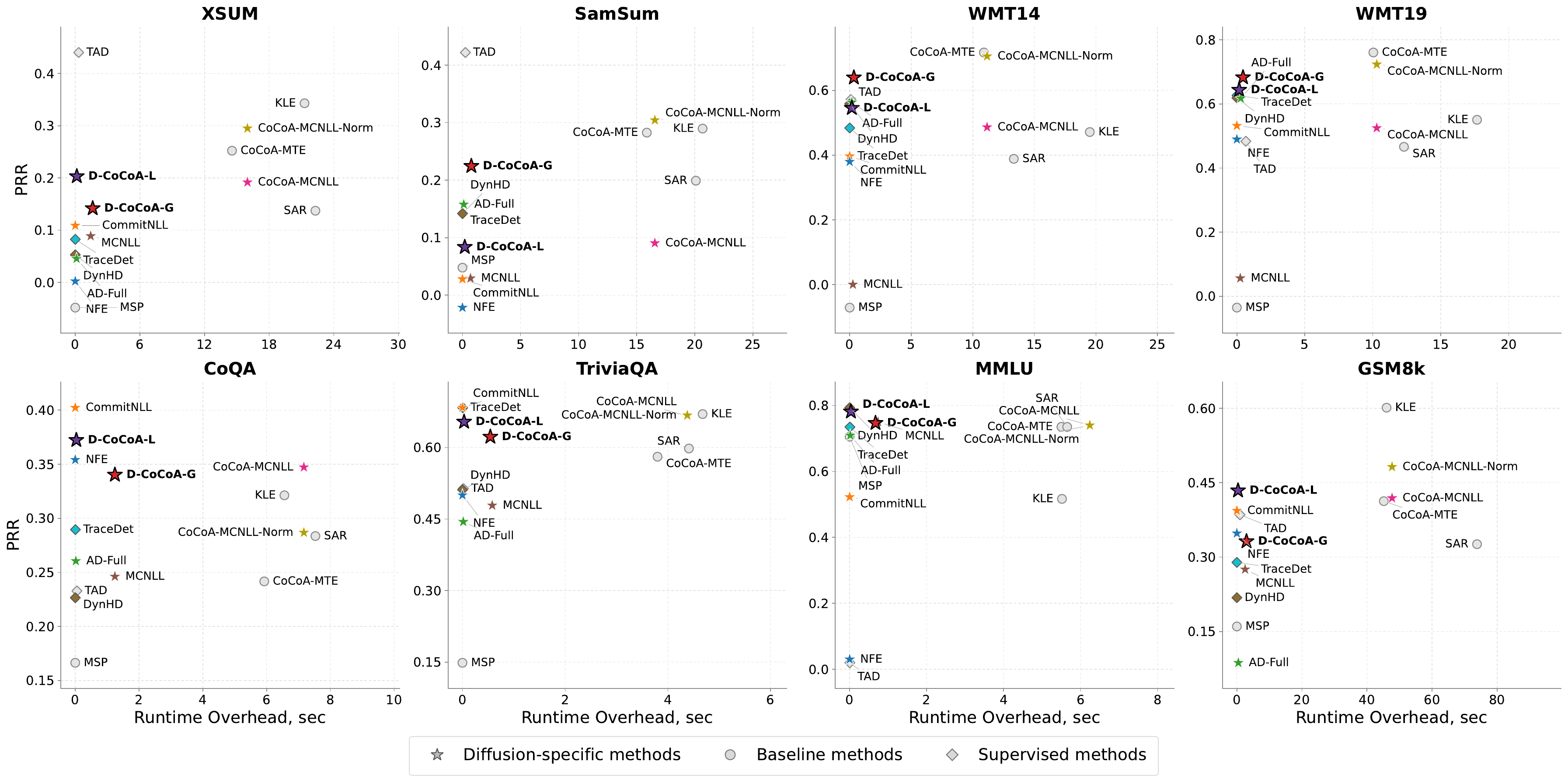}
    \caption{
    \textbf{Dream performance and runtime.} Comparison of uncertainty quantification performance (PRR$\uparrow$) versus computational complexity (Time in seconds$\downarrow$) across eight benchmark datasets for the Dream model (see Figures~\ref{fig:llada_prr_vs_comp}, \ref{fig:refusion_prr_vs_comp}, and~\ref{fig:diffusiongemma_prr_vs_comp} for the LLaDA-1.5, ReFusion, and DiffusionGemma models, respectively).
    }
    \label{fig:dream_prr_vs_comp}
  \end{figure*}

  Tables~\ref{tab:dream_ar_baselines_results}, \ref{tab:dream_diff_baselines_results}, \ref{tab:dream_ar_improved_results}, \ref{tab:dream_ar_traj_results}, and~\ref{tab:dream_detailed_main_results} present the results for Dream with baselines from standard autoregressive LLMs, diffusion-specific methods, sampling-based baselines using MCNLL, sampling-based baselines using denoising trajectory states, and the best method from each category, respectively. Figure~\ref{fig:dream_prr_vs_comp} presents the performance-efficiency trade-off for each dataset.

  \begin{table*}[!h] 
\centering
\resizebox{\textwidth}{!}{\begin{tabular}{ll|c|c|c|c|c|c|c|c|c}
\toprule
\textbf{Category} & \textbf{UQ Method} & \textbf{XSum} & \textbf{SamSum} & \textbf{WMT14} & \textbf{WMT19} & \textbf{CoQA} & \textbf{TriviaQA} & \textbf{MMLU} & \textbf{GSM8k} & \textbf{Mean} \\
\midrule

\multirow{6}{*}{\shortstack[l]{Information-\\based}} & MSP & \cellcolor[rgb]{0.61490285,0.649358983,0.8768415764999999} -0.05 & \cellcolor[rgb]{0.6594161915960784,0.7133025255607843,0.9299287241019608} 0.05 & \cellcolor[rgb]{0.6594161915960784,0.7133025255607843,0.9299287241019608} -0.07 & \cellcolor[rgb]{0.6691882557215687,0.7264093044156863,0.9396585384392157} -0.04 & \cellcolor[rgb]{0.9256858185156862,0.9315626553686274,0.9405319119196078} 0.17 & \cellcolor[rgb]{0.7311771982901961,0.7999150558117647,0.9829285958960784} 0.15 & \cellcolor[rgb]{0.9004151256333333,0.6254146054,0.6128478516333333} 0.71 & \cellcolor[rgb]{0.7690021078509803,0.8374507968235294,0.9958609467764705} 0.16 & \cellcolor[rgb]{0.7608481404156863,0.8297993031764705,0.9938680116235294} 0.13 \\
 & MTE & \cellcolor[rgb]{0.8816813900509803,0.917546110909804,0.9778288382784314} 0.15 & \cellcolor[rgb]{0.9666105915000001,0.9077784252235295,0.8765757160705883} 0.25 & \cellcolor[rgb]{0.974575254145098,0.8953926345333334,0.8554377971509803} 0.41 & \cellcolor[rgb]{0.9802905992117648,0.8812505092627452,0.8339817735509804} 0.44 & \cellcolor[rgb]{0.6892991246352941,0.7519281085960785,0.9568458054235294} 0.05 & \cellcolor[rgb]{0.7744380861411764,0.8425517925882353,0.9971895702117648} 0.19 & \cellcolor[rgb]{0.9622044108411765,0.7492947789176471,0.6945295061352941} 0.62 & \cellcolor[rgb]{0.7446232039529412,0.8137680277960784,0.9884477548843138} 0.14 & \cellcolor[rgb]{0.953077067017647,0.9210455325882353,0.9030752965411765} 0.28 \\
 & Perplexity & \cellcolor[rgb]{0.7152534441254902,0.7824413707294118,0.974444709590196} 0.03 & \cellcolor[rgb]{0.9666105915000001,0.9077784252235295,0.8765757160705883} 0.25 & \cellcolor[rgb]{0.8230564053823529,0.8822182482647059,0.9984342312529412} 0.15 & \cellcolor[rgb]{0.8362689763627451,0.8914307235588235,0.9959909503117648} 0.18 & \cellcolor[rgb]{0.8543598448588235,0.9028606944647058,0.9905584045235294} 0.13 & \cellcolor[rgb]{0.6996157421568627,0.7642642360784314,0.9642295513921568} 0.12 & \cellcolor[rgb]{0.9819030282176471,0.8170942072647058,0.7568604245764706} 0.56 & \cellcolor[rgb]{0.6996157421568627,0.7642642360784314,0.9642295513921568} 0.10 & \cellcolor[rgb]{0.8466606424117646,0.8981570658529412,0.9931538902647059} 0.19 \\
 & AttScore & \cellcolor[rgb]{0.661859207627451,0.7165792202745098,0.9323611776862746} -0.01 & \cellcolor[rgb]{0.6594161915960784,0.7133025255607843,0.9299287241019608} 0.05 & \cellcolor[rgb]{0.61490285,0.649358983,0.8768415764999999} -0.14 & \cellcolor[rgb]{0.61490285,0.649358983,0.8768415764999999} -0.12 & \cellcolor[rgb]{0.6892991246352941,0.7519281085960785,0.9568458054235294} 0.05 & \cellcolor[rgb]{0.7099953545176471,0.7764942726588235,0.9713151710941177} 0.13 & \cellcolor[rgb]{0.6217599191764706,0.6595946294941176,0.885836138382353} 0.03 & \cellcolor[rgb]{0.806966326382353,0.8699614911470589,0.9995711860294118} 0.19 & \cellcolor[rgb]{0.61490285,0.649358983,0.8768415764999999} 0.02 \\
 & CCP & \cellcolor[rgb]{0.6379135614705882,0.6833584577647058,0.9062764676862745} -0.03 & \cellcolor[rgb]{0.7232153212078432,0.7911782132705882,0.9786866527431373} 0.09 & \cellcolor[rgb]{0.6594161915960784,0.7133025255607843,0.9299287241019608} -0.07 & \cellcolor[rgb]{0.6741616707058824,0.7328555732549019,0.9441730814705882} -0.03 & \cellcolor[rgb]{0.9398111318019609,0.9290874692039215,0.9229219340686274} 0.18 & \cellcolor[rgb]{0.7311771982901961,0.7999150558117647,0.9829285958960784} 0.15 & \cellcolor[rgb]{0.9754778064901961,0.8934375166666666,0.8523803414019608} 0.47 & \cellcolor[rgb]{0.7446232039529412,0.8137680277960784,0.9884477548843138} 0.14 & \cellcolor[rgb]{0.7311771982901961,0.7999150558117647,0.9829285958960784} 0.11 \\
 & RAUQ & \cellcolor[rgb]{0.7022106452470588,0.7673217452235295,0.966000956317647} 0.02 & \cellcolor[rgb]{0.7744380861411764,0.8425517925882353,0.9971895702117648} 0.12 & \cellcolor[rgb]{0.7690021078509803,0.8374507968235294,0.9958609467764705} 0.08 & \cellcolor[rgb]{0.7825907906117646,0.8497192224705883,0.9983175350588236} 0.11 & \cellcolor[rgb]{0.9256858185156862,0.9315626553686274,0.9405319119196078} 0.17 & \cellcolor[rgb]{0.61490285,0.649358983,0.8768415764999999} 0.03 & \cellcolor[rgb]{0.9834610525,0.8259843225764706,0.7659985578117647} 0.55 & \cellcolor[rgb]{0.6867762156470588,0.7487493527058824,0.9547336847647059} 0.09 & \cellcolor[rgb]{0.7907430740941177,0.8567252977647059,0.9991571764705882} 0.15 \\\midrule
\multirow{6}{*}{\shortstack[l]{Sampling,\\white-box}} & MCNSE & \cellcolor[rgb]{0.7581301512705882,0.8272488052941176,0.9932036999058824} 0.06 & \cellcolor[rgb]{0.7934605019215686,0.8590606561960784,0.9994370569411765} 0.13 & \cellcolor[rgb]{0.9176723556764705,0.9302569986470588,0.9494852049705882} 0.29 & \cellcolor[rgb]{0.953077067017647,0.9210455325882353,0.9030752965411765} 0.36 & \cellcolor[rgb]{0.9754778064901961,0.8934375166666666,0.8523803414019608} 0.21 & \cellcolor[rgb]{0.9756268974411765,0.7893996947941176,0.729703902882353} 0.52 & \cellcolor[rgb]{0.9077541933039216,0.6388904935882354,0.6201467828333334} 0.70 & \cellcolor[rgb]{0.61490285,0.649358983,0.8768415764999999} 0.02 & \cellcolor[rgb]{0.9613407263117647,0.9142840690588235,0.8885271948941176} 0.29 \\
 & SemanticEntropy & \cellcolor[rgb]{0.6379135614705882,0.6833584577647058,0.9062764676862745} -0.03 & \cellcolor[rgb]{0.61490285,0.649358983,0.8768415764999999} 0.02 & \cellcolor[rgb]{0.7771559349568627,0.8450485056078432,0.9977577741176471} 0.09 & \cellcolor[rgb]{0.8569262456745098,0.9044285706686275,0.989693242609804} 0.21 & \cellcolor[rgb]{0.9696268857588235,0.769790744282353,0.7119501024647059} 0.27 & \cellcolor[rgb]{0.9423217193470588,0.7050085489411765,0.6612532746235295} 0.58 & \cellcolor[rgb]{0.852836579,0.50777808,0.575116406} 0.76 & \cellcolor[rgb]{0.8177369112294117,0.8783569960882354,0.9991482795509804} 0.20 & \cellcolor[rgb]{0.9337138175431372,0.9321882998862745,0.9313012310098039} 0.26 \\
 & SAR & \cellcolor[rgb]{0.8694129974705882,0.9112858109117647,0.9841305319117647} 0.14 & \cellcolor[rgb]{0.9068462909411765,0.9271409192745098,0.959240051254902} 0.20 & \cellcolor[rgb]{0.9691631781666668,0.9044582760156863,0.8705807575137254} 0.39 & \cellcolor[rgb]{0.9839369241588236,0.8629234540470588,0.8092446173921568} 0.47 & \cellcolor[rgb]{0.9591408362921569,0.742086736090196,0.6888969625352941} 0.28 & \cellcolor[rgb]{0.9261890675039215,0.6732459732470588,0.6401732343490196} 0.60 & \cellcolor[rgb]{0.8817601978137255,0.5893337236862745,0.5966981159313726} 0.73 & \cellcolor[rgb]{0.9497716033000001,0.923750118,0.9088945372} 0.33 & \cellcolor[rgb]{0.9738270920764706,0.7829882170843137,0.7237390740196079} 0.39 \\
 & EigenScore & \cellcolor[rgb]{0.7853082184392157,0.8520545809019608,0.9985974155294117} 0.08 & \cellcolor[rgb]{0.8096589725941177,0.8720603673823529,0.9994654594098039} 0.14 & \cellcolor[rgb]{0.9842664748411765,0.8579206459529412,0.8030483739411765} 0.47 & \cellcolor[rgb]{0.9756268974411765,0.7893996947941176,0.729703902882353} 0.56 & \cellcolor[rgb]{0.852836579,0.50777808,0.575116406} 0.34 & \cellcolor[rgb]{0.8763519705509804,0.5787878133490196,0.5921289546450981} 0.65 & \cellcolor[rgb]{0.8620206859411765,0.9074551963235293,0.9878254853235294} 0.30 & \cellcolor[rgb]{0.8763519705509804,0.5787878133490196,0.5921289546450981} 0.58 & \cellcolor[rgb]{0.9738270920764706,0.7829882170843137,0.7237390740196079} 0.39 \\
 & CoCoA-MTE & \cellcolor[rgb]{0.9763803588352942,0.8914823988,0.8493228856529411} 0.25 & \cellcolor[rgb]{0.9829494490941176,0.8700709193019608,0.8185288537078431} 0.28 & \cellcolor[rgb]{0.8925766523392157,0.6104255443607843,0.6058364385039215} 0.72 & \cellcolor[rgb]{0.8587172724588235,0.5255587742117647,0.5793683038509804} 0.76 & \cellcolor[rgb]{0.9844470791666666,0.8397397817137255,0.7814061455764706} 0.24 & \cellcolor[rgb]{0.9423217193470588,0.7050085489411765,0.6612532746235295} 0.58 & \cellcolor[rgb]{0.8704786593764706,0.5611201626352941,0.5878720995529412} 0.74 & \cellcolor[rgb]{0.9844312501823529,0.8554192419058824,0.7999502522156863} 0.41 & \cellcolor[rgb]{0.852836579,0.50777808,0.575116406} 0.50 \\
 & SemanticDensity & \cellcolor[rgb]{0.9046643351764706,0.9264869997764706,0.9611612942352941} 0.17 & \cellcolor[rgb]{0.9196757213862745,0.930583412827451,0.9472468817078432} 0.21 & \cellcolor[rgb]{0.9607031106137255,0.7457102085921569,0.6917042176882353} 0.60 & \cellcolor[rgb]{0.9261890675039215,0.6732459732470588,0.6401732343490196} 0.67 & \cellcolor[rgb]{0.8336264621666667,0.8895882285,0.9964796065} 0.12 & \cellcolor[rgb]{0.9175136022176471,0.6568221562117647,0.6298915758705882} 0.61 & \cellcolor[rgb]{0.6379135614705882,0.6833584577647058,0.9062764676862745} 0.05 & \cellcolor[rgb]{0.6867762156470588,0.7487493527058824,0.9547336847647059} 0.09 & \cellcolor[rgb]{0.974575254145098,0.8953926345333334,0.8554377971509803} 0.31 \\\midrule
\multirow{3}{*}{\shortstack[l]{Sampling,\\black-box}} & LUQ & \cellcolor[rgb]{0.9774267028058823,0.7958111725039216,0.7356687317450981} 0.32 & \cellcolor[rgb]{0.9842498738333334,0.8369886898862745,0.7783246280235294} 0.30 & \cellcolor[rgb]{0.9837721488176471,0.8654248580941176,0.812342739117647} 0.46 & \cellcolor[rgb]{0.9844470791666666,0.8397397817137255,0.7814061455764706} 0.50 & \cellcolor[rgb]{0.9150932609745098,0.6523663817764705,0.6274457140333334} 0.31 & \cellcolor[rgb]{0.864597965917647,0.5433394684235294,0.5836202017019608} 0.66 & \cellcolor[rgb]{0.9513294646843138,0.7239693735803922,0.6748606867705882} 0.64 & \cellcolor[rgb]{0.864597965917647,0.5433394684235294,0.5836202017019608} 0.59 & \cellcolor[rgb]{0.8979687697431373,0.6209226426705883,0.6104148745666667} 0.47 \\
 & KLE & \cellcolor[rgb]{0.9671527274529412,0.762958755827451,0.7061432370215687} 0.34 & \cellcolor[rgb]{0.9844312501823529,0.8554192419058824,0.7999502522156863} 0.29 & \cellcolor[rgb]{0.9842664748411765,0.8579206459529412,0.8030483739411765} 0.47 & \cellcolor[rgb]{0.9783266054882354,0.7990169113588235,0.7386511461764707} 0.55 & \cellcolor[rgb]{0.8952807659705883,0.6156984995294118,0.6081210191470588} 0.32 & \cellcolor[rgb]{0.852836579,0.50777808,0.575116406} 0.67 & \cellcolor[rgb]{0.9844312501823529,0.8554192419058824,0.7999502522156863} 0.52 & \cellcolor[rgb]{0.852836579,0.50777808,0.575116406} 0.60 & \cellcolor[rgb]{0.8979687697431373,0.6209226426705883,0.6104148745666667} 0.47 \\
 & DegMat & \cellcolor[rgb]{0.9653342981666666,0.909438499827451,0.8795731953490196} 0.23 & \cellcolor[rgb]{0.9563825307352941,0.9183409471764705,0.8972560558823529} 0.24 & \cellcolor[rgb]{0.9846442845,0.8424908735411765,0.7844876631294118} 0.49 & \cellcolor[rgb]{0.9729271893941176,0.7797824782294118,0.7207566595882353} 0.57 & \cellcolor[rgb]{0.8952807659705883,0.6156984995294118,0.6081210191470588} 0.32 & \cellcolor[rgb]{0.852836579,0.50777808,0.575116406} 0.67 & \cellcolor[rgb]{0.9640580048333334,0.9110985744313725,0.882570674627451} 0.44 & \cellcolor[rgb]{0.8763519705509804,0.5787878133490196,0.5921289546450981} 0.58 & \cellcolor[rgb]{0.9348276152529411,0.6896369097254902,0.6504705511098039} 0.44 \\\midrule
\multirow{4}{*}{Supervised} & SAPLMA & \cellcolor[rgb]{0.9717157648333333,0.9011381268078431,0.8645857989568628} 0.24 & \cellcolor[rgb]{0.9736727018,0.8973477524,0.8584952529000001} 0.26 & \cellcolor[rgb]{0.852836579,0.50777808,0.575116406} 0.77 & \cellcolor[rgb]{0.852836579,0.50777808,0.575116406} 0.77 & \cellcolor[rgb]{0.61490285,0.649358983,0.8768415764999999} 0.01 & \cellcolor[rgb]{0.8979687697431373,0.6209226426705883,0.6104148745666667} 0.63 & \cellcolor[rgb]{0.9398111318019609,0.9290874692039215,0.9229219340686274} 0.40 & \cellcolor[rgb]{0.8415278407803921,0.8950213134450979,0.9948842140921569} 0.22 & \cellcolor[rgb]{0.9622044108411765,0.7492947789176471,0.6945295061352941} 0.41 \\
 & SATRMD & \cellcolor[rgb]{0.9653342981666666,0.909438499827451,0.8795731953490196} 0.23 & \cellcolor[rgb]{0.9337138175431372,0.9321882998862745,0.9313012310098039} 0.22 & \cellcolor[rgb]{0.9791396989627451,0.8021675484411765,0.7416485506941177} 0.54 & \cellcolor[rgb]{0.9622044108411765,0.7492947789176471,0.6945295061352941} 0.60 & \cellcolor[rgb]{0.9547297988764706,0.919693239882353,0.9001656762117647} 0.19 & \cellcolor[rgb]{0.915574114,0.9297565972666666,0.9515550793333334} 0.33 & \cellcolor[rgb]{0.8616576191882352,0.5344491213176471,0.5814942527764706} 0.75 & \cellcolor[rgb]{0.7934605019215686,0.8590606561960784,0.9994370569411765} 0.18 & \cellcolor[rgb]{0.9791396989627451,0.8021675484411765,0.7416485506941177} 0.38 \\
 & LookBackLens & \cellcolor[rgb]{0.9671527274529412,0.762958755827451,0.7061432370215687} 0.34 & \cellcolor[rgb]{0.9844312501823529,0.8554192419058824,0.7999502522156863} 0.29 & \cellcolor[rgb]{0.8925766523392157,0.6104255443607843,0.6058364385039215} 0.72 & \cellcolor[rgb]{0.8763519705509804,0.5787878133490196,0.5921289546450981} 0.74 & \cellcolor[rgb]{0.8952807659705883,0.6156984995294118,0.6081210191470588} 0.32 & \cellcolor[rgb]{0.8871684250764706,0.5998796340235294,0.6012672772176471} 0.64 & \cellcolor[rgb]{0.9708639649117647,0.7732067385098039,0.7148535351862745} 0.60 & \cellcolor[rgb]{0.931695915645098,0.9325418979098039,0.9338169421313726} 0.31 & \cellcolor[rgb]{0.8675383126470588,0.5522298155294117,0.585746150627451} 0.49 \\
 & TAD & \cellcolor[rgb]{0.852836579,0.50777808,0.575116406} 0.44 & \cellcolor[rgb]{0.852836579,0.50777808,0.575116406} 0.42 & \cellcolor[rgb]{0.9720272867117647,0.7765767393745098,0.7177742451568627} 0.57 & \cellcolor[rgb]{0.9844312501823529,0.8554192419058824,0.7999502522156863} 0.48 & \cellcolor[rgb]{0.9841016995,0.8604220499999999,0.8061464956666666} 0.23 & \cellcolor[rgb]{0.9791396989627451,0.8021675484411765,0.7416485506941177} 0.51 & \cellcolor[rgb]{0.61490285,0.649358983,0.8768415764999999} 0.02 & \cellcolor[rgb]{0.9813541391647058,0.8767786732784314,0.8278006056137255} 0.39 & \cellcolor[rgb]{0.9791396989627451,0.8021675484411765,0.7416485506941177} 0.38 \\

\bottomrule
\end{tabular}
}\caption{\label{tab:dream_ar_baselines_results} \textbf{AR baselines.} PRR$\uparrow$ for the Dream model across various tasks for baselines from standard autoregressive LLMs (see Tables~\ref{tab:ar_baselines_results}, \ref{tab:refusion_ar_baselines_results}, and~\ref{tab:diffusiongemma_ar_baselines_results} for the LLaDA-1.5, ReFusion, and DiffusionGemma models, respectively). Warmer color indicates better results.}\end{table*}
  \begin{table*}[!h] 
\centering
\resizebox{\textwidth}{!}{\begin{tabular}{ll|c|c|c|c|c|c|c|c|c}
\toprule
\textbf{Category} & \textbf{UQ Method} & \textbf{XSum} & \textbf{SamSum} & \textbf{WMT14} & \textbf{WMT19} & \textbf{CoQA} & \textbf{TriviaQA} & \textbf{MMLU} & \textbf{GSM8k} & \textbf{Mean} \\
\midrule
\multirow{8}{*}{\shortstack[l]{Diffusion\\statistics}} & MCNLL & \cellcolor[rgb]{0.931695915645098,0.9325418979098039,0.9338169421313726} 0.09 & \cellcolor[rgb]{0.7419271648,0.8110249248,0.9874041013} 0.03 & \cellcolor[rgb]{0.61490285,0.649358983,0.8768415764999999} 0.00 & \cellcolor[rgb]{0.6716387617176471,0.7296768173647059,0.9420609608117647} 0.06 & \cellcolor[rgb]{0.9418435698882353,0.9280538589764706,0.9201288350882353} 0.25 & \cellcolor[rgb]{0.8840171821764706,0.9185176097647059,0.976244109117647} 0.48 & \cellcolor[rgb]{0.852836579,0.50777808,0.575116406} 0.75 & \cellcolor[rgb]{0.9813503623666666,0.8141088755,0.7538180498} 0.28 & \cellcolor[rgb]{0.8594926464901961,0.905996446872549,0.9888280806960784} 0.24 \\
 & MCNLL-Norm & \cellcolor[rgb]{0.852836579,0.50777808,0.575116406} 0.22 & \cellcolor[rgb]{0.852836579,0.50777808,0.575116406} 0.31 & \cellcolor[rgb]{0.9646785691470589,0.756126767372549,0.7003363715784314} 0.45 & \cellcolor[rgb]{0.9802450306647059,0.8081382119705882,0.7477333002470588} 0.46 & \cellcolor[rgb]{0.61490285,0.649358983,0.8768415764999999} 0.09 & \cellcolor[rgb]{0.6667452396901961,0.7231326097019608,0.9372260848549019} 0.37 & \cellcolor[rgb]{0.852836579,0.50777808,0.575116406} 0.75 & \cellcolor[rgb]{0.9763803588352942,0.8914823988,0.8493228856529411} 0.23 & \cellcolor[rgb]{0.9348276152529411,0.6896369097254902,0.6504705511098039} 0.36 \\
\cmidrule(lr){2-11}
 & NFE & \cellcolor[rgb]{0.7099953545176471,0.7764942726588235,0.9713151710941177} 0.00 & \cellcolor[rgb]{0.6473592160588235,0.6967709648705882,0.9171920835607843} -0.02 & \cellcolor[rgb]{0.9845960255235294,0.8529178378588235,0.7968521304901961} 0.38 & \cellcolor[rgb]{0.9683898066058824,0.766374750054902,0.7090466697431372} 0.49 & \cellcolor[rgb]{0.9513294646843138,0.7239693735803922,0.6748606867705882} 0.35 & \cellcolor[rgb]{0.9176723556764705,0.9302569986470588,0.9494852049705882} 0.50 & \cellcolor[rgb]{0.61490285,0.649358983,0.8768415764999999} 0.03 & \cellcolor[rgb]{0.9175136022176471,0.6568221562117647,0.6298915758705882} 0.35 & \cellcolor[rgb]{0.9112102024705881,0.9284487582705883,0.9553975652941177} 0.26 \\
 & Remask & \cellcolor[rgb]{0.7099953545176471,0.7764942726588235,0.9713151710941177} 0.00 & \cellcolor[rgb]{0.6309026780784314,0.6732421581529412,0.8978288875588235} -0.03 & \cellcolor[rgb]{0.9704394715,0.9027982014117647,0.8675832782352941} 0.33 & \cellcolor[rgb]{0.9846442845,0.8424908735411765,0.7844876631294118} 0.43 & \cellcolor[rgb]{0.9729271893941176,0.7797824782294118,0.7207566595882353} 0.33 & \cellcolor[rgb]{0.8256989195784314,0.8840607433235295,0.9979455750647059} 0.45 & \cellcolor[rgb]{0.61490285,0.649358983,0.8768415764999999} 0.03 & \cellcolor[rgb]{0.9028614815235294,0.6299065681294118,0.6152808287} 0.36 & \cellcolor[rgb]{0.8594926464901961,0.905996446872549,0.9888280806960784} 0.24 \\
 & FlipCount & \cellcolor[rgb]{0.6594161915960784,0.7133025255607843,0.9299287241019608} -0.02 & \cellcolor[rgb]{0.61490285,0.649358983,0.8768415764999999} -0.04 & \cellcolor[rgb]{0.6667452396901961,0.7231326097019608,0.9372260848549019} 0.05 & \cellcolor[rgb]{0.7258692802352942,0.794090494117647,0.9801006337941176} 0.11 & \cellcolor[rgb]{0.9418435698882353,0.9280538589764706,0.9201288350882353} 0.25 & \cellcolor[rgb]{0.61490285,0.649358983,0.8768415764999999} 0.34 & \cellcolor[rgb]{0.7554121621254901,0.8246983074117646,0.9925393881882353} 0.19 & \cellcolor[rgb]{0.9813503623666666,0.8141088755,0.7538180498} 0.28 & \cellcolor[rgb]{0.61490285,0.649358983,0.8768415764999999} 0.15 \\
\cmidrule(lr){2-11}
 & TrajNLL & \cellcolor[rgb]{0.9133921582352942,0.9291026777686274,0.9534763223137255} 0.08 & \cellcolor[rgb]{0.7022106452470588,0.7673217452235295,0.966000956317647} 0.01 & \cellcolor[rgb]{0.8933603506784313,0.9224036051843137,0.9699051924745098} 0.24 & \cellcolor[rgb]{0.9727701494549019,0.8993028702666667,0.8615527086490196} 0.37 & \cellcolor[rgb]{0.9845960255235294,0.8529178378588235,0.7968521304901961} 0.30 & \cellcolor[rgb]{0.9807976965156863,0.8111235437352942,0.7507756750235295} 0.59 & \cellcolor[rgb]{0.9847608008647059,0.8504164338117647,0.7937540087647059} 0.52 & \cellcolor[rgb]{0.9175136022176471,0.6568221562117647,0.6298915758705882} 0.35 & \cellcolor[rgb]{0.9841016995,0.8604220499999999,0.8061464956666666} 0.31 \\
 & TrajEntropy & \cellcolor[rgb]{0.9666105915000001,0.9077784252235295,0.8765757160705883} 0.11 & \cellcolor[rgb]{0.7419271648,0.8110249248,0.9874041013} 0.03 & \cellcolor[rgb]{0.9418435698882353,0.9280538589764706,0.9201288350882353} 0.29 & \cellcolor[rgb]{0.9824176791176471,0.8723068372941176,0.8216194376764705} 0.40 & \cellcolor[rgb]{0.9845960255235294,0.8529178378588235,0.7968521304901961} 0.30 & \cellcolor[rgb]{0.9781854635254902,0.8875721630666666,0.8432079741549019} 0.55 & \cellcolor[rgb]{0.8817601978137255,0.5893337236862745,0.5966981159313726} 0.72 & \cellcolor[rgb]{0.8704786593764706,0.5611201626352941,0.5878720995529412} 0.38 & \cellcolor[rgb]{0.9528917390058824,0.727592846082353,0.6776679419235294} 0.35 \\
 & CommitNLL & \cellcolor[rgb]{0.9666105915000001,0.9077784252235295,0.8765757160705883} 0.11 & \cellcolor[rgb]{0.7419271648,0.8110249248,0.9874041013} 0.03 & \cellcolor[rgb]{0.9836582578333333,0.8287354144039216,0.7690800753647059} 0.40 & \cellcolor[rgb]{0.9441952453705882,0.708851458745098,0.6639489555019608} 0.53 & \cellcolor[rgb]{0.852836579,0.50777808,0.575116406} 0.40 & \cellcolor[rgb]{0.852836579,0.50777808,0.575116406} 0.68 & \cellcolor[rgb]{0.9847608008647059,0.8504164338117647,0.7937540087647059} 0.52 & \cellcolor[rgb]{0.852836579,0.50777808,0.575116406} 0.39 & \cellcolor[rgb]{0.8817601978137255,0.5893337236862745,0.5966981159313726} 0.38 \\
\midrule
\multirow{8}{*}{\shortstack[l]{Trajectory\\dissimilarity}} & AD-Block & \cellcolor[rgb]{0.947942297417647,0.7165372783529411,0.6693403172588235} 0.18 & \cellcolor[rgb]{0.9653342981666666,0.909438499827451,0.8795731953490196} 0.16 & \cellcolor[rgb]{0.9126469050843138,0.6478744190470588,0.6250127369666667} 0.51 & \cellcolor[rgb]{0.9077541933039216,0.6388904935882354,0.6201467828333334} 0.57 & \cellcolor[rgb]{0.9563825307352941,0.9183409471764705,0.8972560558823529} 0.26 & \cellcolor[rgb]{0.806966326382353,0.8699614911470589,0.9995711860294118} 0.44 & \cellcolor[rgb]{0.8925766523392157,0.6104255443607843,0.6058364385039215} 0.71 & \cellcolor[rgb]{0.6426363887647059,0.690064711317647,0.9117342756235294} -0.00 & \cellcolor[rgb]{0.9528917390058824,0.727592846082353,0.6776679419235294} 0.35 \\
 & WeightedAD-Block & \cellcolor[rgb]{0.9846442845,0.8424908735411765,0.7844876631294118} 0.14 & \cellcolor[rgb]{0.8933603506784313,0.9224036051843137,0.9699051924745098} 0.11 & \cellcolor[rgb]{0.8898725387078432,0.6051525891921569,0.6035518578607844} 0.53 & \cellcolor[rgb]{0.8979687697431373,0.6209226426705883,0.6104148745666667} 0.58 & \cellcolor[rgb]{0.9838554631666667,0.8314865062313725,0.7721615929176471} 0.31 & \cellcolor[rgb]{0.9003004236470589,0.9251791607803921,0.9650037801960785} 0.49 & \cellcolor[rgb]{0.9004151256333333,0.6254146054,0.6128478516333333} 0.70 & \cellcolor[rgb]{0.9837721488176471,0.8654248580941176,0.812342739117647} 0.25 & \cellcolor[rgb]{0.852836579,0.50777808,0.575116406} 0.39 \\
\cmidrule(lr){2-11}
 & AD-LastBlock & \cellcolor[rgb]{0.8910245585529412,0.9214321063294117,0.9714899216352941} 0.07 & \cellcolor[rgb]{0.8015810339588235,0.8657637386764706,0.9997826392686274} 0.06 & \cellcolor[rgb]{0.61490285,0.649358983,0.8768415764999999} 0.00 & \cellcolor[rgb]{0.61490285,0.649358983,0.8768415764999999} -0.00 & \cellcolor[rgb]{0.9563825307352941,0.9183409471764705,0.8972560558823529} 0.26 & \cellcolor[rgb]{0.806966326382353,0.8699614911470589,0.9995711860294118} 0.44 & \cellcolor[rgb]{0.8925766523392157,0.6104255443607843,0.6058364385039215} 0.71 & \cellcolor[rgb]{0.61490285,0.649358983,0.8768415764999999} -0.02 & \cellcolor[rgb]{0.7179074031529412,0.7853536515764705,0.9758586906411765} 0.19 \\
 & WeightedAD-LastBlock & \cellcolor[rgb]{0.9514243351588236,0.9223978252941176,0.9059849168705882} 0.10 & \cellcolor[rgb]{0.7798733627843137,0.8473838640392157,0.9980376545882352} 0.05 & \cellcolor[rgb]{0.61490285,0.649358983,0.8768415764999999} 0.00 & \cellcolor[rgb]{0.61490285,0.649358983,0.8768415764999999} -0.00 & \cellcolor[rgb]{0.9838554631666667,0.8314865062313725,0.7721615929176471} 0.31 & \cellcolor[rgb]{0.9003004236470589,0.9251791607803921,0.9650037801960785} 0.49 & \cellcolor[rgb]{0.9004151256333333,0.6254146054,0.6128478516333333} 0.70 & \cellcolor[rgb]{0.6716387617176471,0.7296768173647059,0.9420609608117647} 0.02 & \cellcolor[rgb]{0.7744380861411764,0.8425517925882353,0.9971895702117648} 0.21 \\
\cmidrule(lr){2-11}
 & AD-LastBlockPrefix & \cellcolor[rgb]{0.7099953545176471,0.7764942726588235,0.9713151710941177} 0.00 & \cellcolor[rgb]{0.8594926464901961,0.905996446872549,0.9888280806960784} 0.09 & \cellcolor[rgb]{0.9844470791666666,0.8397397817137255,0.7814061455764706} 0.39 & \cellcolor[rgb]{0.9763803588352942,0.8914823988,0.8493228856529411} 0.38 & \cellcolor[rgb]{0.9563825307352941,0.9183409471764705,0.8972560558823529} 0.26 & \cellcolor[rgb]{0.806966326382353,0.8699614911470589,0.9995711860294118} 0.44 & \cellcolor[rgb]{0.8925766523392157,0.6104255443607843,0.6058364385039215} 0.71 & \cellcolor[rgb]{0.7880256462666666,0.8543899393333334,0.9988772960000001} 0.09 & \cellcolor[rgb]{0.9691631781666668,0.9044582760156863,0.8705807575137254} 0.29 \\
 & WeightedAD-LastBlockPrefix & \cellcolor[rgb]{0.61490285,0.649358983,0.8768415764999999} -0.04 & \cellcolor[rgb]{0.8594926464901961,0.905996446872549,0.9888280806960784} 0.09 & \cellcolor[rgb]{0.9844470791666666,0.8397397817137255,0.7814061455764706} 0.39 & \cellcolor[rgb]{0.9763803588352942,0.8914823988,0.8493228856529411} 0.38 & \cellcolor[rgb]{0.9838554631666667,0.8314865062313725,0.7721615929176471} 0.31 & \cellcolor[rgb]{0.9003004236470589,0.9251791607803921,0.9650037801960785} 0.49 & \cellcolor[rgb]{0.9004151256333333,0.6254146054,0.6128478516333333} 0.70 & \cellcolor[rgb]{0.8543598448588235,0.9028606944647058,0.9905584045235294} 0.13 & \cellcolor[rgb]{0.9790880158705882,0.8856170452000001,0.8401505184058824} 0.30 \\
\cmidrule(lr){2-11}
 & AD-Full & \cellcolor[rgb]{0.8415278407803921,0.8950213134450979,0.9948842140921569} 0.05 & \cellcolor[rgb]{0.9653342981666666,0.909438499827451,0.8795731953490196} 0.16 & \cellcolor[rgb]{0.852836579,0.50777808,0.575116406} 0.56 & \cellcolor[rgb]{0.852836579,0.50777808,0.575116406} 0.62 & \cellcolor[rgb]{0.9563825307352941,0.9183409471764705,0.8972560558823529} 0.26 & \cellcolor[rgb]{0.806966326382353,0.8699614911470589,0.9995711860294118} 0.44 & \cellcolor[rgb]{0.8925766523392157,0.6104255443607843,0.6058364385039215} 0.71 & \cellcolor[rgb]{0.7880256462666666,0.8543899393333334,0.9988772960000001} 0.09 & \cellcolor[rgb]{0.9348276152529411,0.6896369097254902,0.6504705511098039} 0.36 \\
 & WeightedAD-Full & \cellcolor[rgb]{0.8910245585529412,0.9214321063294117,0.9714899216352941} 0.07 & \cellcolor[rgb]{0.9653342981666666,0.909438499827451,0.8795731953490196} 0.16 & \cellcolor[rgb]{0.864597965917647,0.5433394684235294,0.5836202017019608} 0.55 & \cellcolor[rgb]{0.864597965917647,0.5433394684235294,0.5836202017019608} 0.61 & \cellcolor[rgb]{0.9838554631666667,0.8314865062313725,0.7721615929176471} 0.31 & \cellcolor[rgb]{0.9003004236470589,0.9251791607803921,0.9650037801960785} 0.49 & \cellcolor[rgb]{0.9004151256333333,0.6254146054,0.6128478516333333} 0.70 & \cellcolor[rgb]{0.7717200969960785,0.8400012947058824,0.9965252584941177} 0.08 & \cellcolor[rgb]{0.9102005491941176,0.643382456317647,0.6225797599} 0.37 \\
\bottomrule
\end{tabular}
}\caption{\label{tab:dream_diff_baselines_results} \textbf{Diffusion-specific UQ signals.} PRR$\uparrow$ for the Dream model across various tasks for diffusion statistics and trajectory dissimilarity UQ methods (see Tables~\ref{tab:diff_baselines_results}, \ref{tab:refusion_diff_baselines_results}, and~\ref{tab:diffusiongemma_diff_baselines_results} for the LLaDA-1.5, ReFusion, and DiffusionGemma models, respectively). Warmer color indicates better results.}\end{table*}

  \begin{table*}[!h] \resizebox{\textwidth}{!}{\begin{tabular}{ll|c|c|c|c|c|c|c|c|c}
\toprule
\textbf{Category} & \textbf{UQ Method} & \textbf{XSum} & \textbf{SamSum} & \textbf{WMT14} & \textbf{WMT19} & \textbf{CoQA} & \textbf{TriviaQA} & \textbf{MMLU} & \textbf{GSM8k} & \textbf{Mean} \\
\midrule
\multirow{4}{*}{AR} & MCNSE & \cellcolor[rgb]{0.9112102024705881,0.9284487582705883,0.9553975652941177} 0.06 & \cellcolor[rgb]{0.9653342981666666,0.909438499827451,0.8795731953490196} 0.13 & \cellcolor[rgb]{0.9563825307352941,0.9183409471764705,0.8972560558823529} 0.29 & \cellcolor[rgb]{0.9666105915000001,0.9077784252235295,0.8765757160705883} 0.36 & \cellcolor[rgb]{0.9848414898333333,0.8452419653686274,0.7875691806823529} 0.21 & \cellcolor[rgb]{0.9836582578333333,0.8287354144039216,0.7690800753647059} 0.52 & \cellcolor[rgb]{0.8952807659705883,0.6156984995294118,0.6081210191470588} 0.70 & \cellcolor[rgb]{0.6817303976705882,0.7423918409254902,0.9505094434470589} 0.02 & \cellcolor[rgb]{0.9736727018,0.8973477524,0.8584952529000001} 0.29 \\
 & SemanticEntropy & \cellcolor[rgb]{0.6643022236588235,0.7198559149882353,0.9347936312705882} -0.03 & \cellcolor[rgb]{0.7258692802352942,0.794090494117647,0.9801006337941176} 0.02 & \cellcolor[rgb]{0.7825907906117646,0.8497192224705883,0.9983175350588236} 0.09 & \cellcolor[rgb]{0.8569262456745098,0.9044285706686275,0.989693242609804} 0.21 & \cellcolor[rgb]{0.9575785619705882,0.7384632635882353,0.6860897073823529} 0.27 & \cellcolor[rgb]{0.9560162876490197,0.7348397910862745,0.6832824522294118} 0.58 & \cellcolor[rgb]{0.852836579,0.50777808,0.575116406} 0.76 & \cellcolor[rgb]{0.9377786937156862,0.9301210794313726,0.9257150330490196} 0.20 & \cellcolor[rgb]{0.9398111318019609,0.9290874692039215,0.9229219340686274} 0.26 \\
 & SemanticDensity & \cellcolor[rgb]{0.9102005491941176,0.643382456317647,0.6225797599} 0.17 & \cellcolor[rgb]{0.9423217193470588,0.7050085489411765,0.6612532746235295} 0.21 & \cellcolor[rgb]{0.852836579,0.50777808,0.575116406} 0.60 & \cellcolor[rgb]{0.852836579,0.50777808,0.575116406} 0.67 & \cellcolor[rgb]{0.9276891842254902,0.9318890695490196,0.9382935886568627} 0.12 & \cellcolor[rgb]{0.9283579338254901,0.6773519275058824,0.6427436489686275} 0.61 & \cellcolor[rgb]{0.8042736801705883,0.8678626149117648,0.9996769126490196} 0.05 & \cellcolor[rgb]{0.7880256462666666,0.8543899393333334,0.9988772960000001} 0.09 & \cellcolor[rgb]{0.9837721488176471,0.8654248580941176,0.812342739117647} 0.31 \\
 & CoCoA-MSP & \cellcolor[rgb]{0.8863529743019607,0.9194891086196079,0.9746593799568628} 0.05 & \cellcolor[rgb]{0.9133921582352942,0.9291026777686274,0.9534763223137255} 0.10 & \cellcolor[rgb]{0.9627817115,0.9127586490352941,0.8855681539058824} 0.30 & \cellcolor[rgb]{0.9613407263117647,0.9142840690588235,0.8885271948941176} 0.35 & \cellcolor[rgb]{0.947942297417647,0.7165372783529411,0.6693403172588235} 0.28 & \cellcolor[rgb]{0.9727701494549019,0.8993028702666667,0.8615527086490196} 0.46 & \cellcolor[rgb]{0.852836579,0.50777808,0.575116406} 0.76 & \cellcolor[rgb]{0.9813503623666666,0.8141088755,0.7538180498} 0.30 & \cellcolor[rgb]{0.9849255762058824,0.8479150297647058,0.7906558870392157} 0.32 \\
\midrule
\multirow{4}{*}{\shortstack[l]{AR +\\MCNLL}} & MCNSE-MCNLL & \cellcolor[rgb]{0.9683898066058824,0.766374750054902,0.7090466697431372} 0.14 & \cellcolor[rgb]{0.8256989195784314,0.8840607433235295,0.9979455750647059} 0.06 & \cellcolor[rgb]{0.9813541391647058,0.8767786732784314,0.8278006056137255} 0.35 & \cellcolor[rgb]{0.9438760079745099,0.9270202487490196,0.9173357361078431} 0.32 & \cellcolor[rgb]{0.61490285,0.649358983,0.8768415764999999} -0.10 & \cellcolor[rgb]{0.61490285,0.649358983,0.8768415764999999} 0.15 & \cellcolor[rgb]{0.61490285,0.649358983,0.8768415764999999} -0.24 & \cellcolor[rgb]{0.61490285,0.649358983,0.8768415764999999} -0.03 & \cellcolor[rgb]{0.61490285,0.649358983,0.8768415764999999} 0.08 \\
 & SemanticEntropy-MCNLL & \cellcolor[rgb]{0.61490285,0.649358983,0.8768415764999999} -0.05 & \cellcolor[rgb]{0.61490285,0.649358983,0.8768415764999999} -0.03 & \cellcolor[rgb]{0.61490285,0.649358983,0.8768415764999999} -0.09 & \cellcolor[rgb]{0.61490285,0.649358983,0.8768415764999999} -0.06 & \cellcolor[rgb]{0.9736727018,0.8973477524,0.8584952529000001} 0.17 & \cellcolor[rgb]{0.7880256462666666,0.8543899393333334,0.9988772960000001} 0.29 & \cellcolor[rgb]{0.9004151256333333,0.6254146054,0.6128478516333333} 0.69 & \cellcolor[rgb]{0.9790880158705882,0.8856170452000001,0.8401505184058824} 0.25 & \cellcolor[rgb]{0.7419271648,0.8110249248,0.9874041013} 0.15 \\
 & SemanticDensity-MCNLL & \cellcolor[rgb]{0.9528917390058824,0.727592846082353,0.6776679419235294} 0.15 & \cellcolor[rgb]{0.852836579,0.50777808,0.575116406} 0.25 & \cellcolor[rgb]{0.852836579,0.50777808,0.575116406} 0.60 & \cellcolor[rgb]{0.8616576191882352,0.5344491213176471,0.5814942527764706} 0.66 & \cellcolor[rgb]{0.9836582578333333,0.8287354144039216,0.7690800753647059} 0.22 & \cellcolor[rgb]{0.9727701494549019,0.8993028702666667,0.8615527086490196} 0.46 & \cellcolor[rgb]{0.9004151256333333,0.6254146054,0.6128478516333333} 0.69 & \cellcolor[rgb]{0.915574114,0.9297565972666666,0.9515550793333334} 0.18 & \cellcolor[rgb]{0.9102005491941176,0.643382456317647,0.6225797599} 0.40 \\
 & CoCoA-MCNLL & \cellcolor[rgb]{0.852836579,0.50777808,0.575116406} 0.19 & \cellcolor[rgb]{0.8933603506784313,0.9224036051843137,0.9699051924745098} 0.09 & \cellcolor[rgb]{0.9497671903627452,0.7203459010784313,0.6720534316176471} 0.49 & \cellcolor[rgb]{0.9671527274529412,0.762958755827451,0.7061432370215687} 0.52 & \cellcolor[rgb]{0.852836579,0.50777808,0.575116406} 0.35 & \cellcolor[rgb]{0.852836579,0.50777808,0.575116406} 0.67 & \cellcolor[rgb]{0.8675383126470588,0.5522298155294117,0.585746150627451} 0.74 & \cellcolor[rgb]{0.852836579,0.50777808,0.575116406} 0.42 & \cellcolor[rgb]{0.852836579,0.50777808,0.575116406} 0.43 \\
\bottomrule
\end{tabular}
}\caption{\label{tab:dream_ar_improved_results} \textbf{Baselines with diffusion likelihoods.} PRR$\uparrow$ for the Dream model across various tasks for sampling-based baselines from standard autoregressive LLMs, where the diffusion likelihood (MCNLL) is used instead of the AR likelihood (see Tables~\ref{tab:ar_improved_results} and \ref{tab:refusion_ar_improved_results} for the LLaDA-1.5 and ReFusion models, respectively). Warmer color indicates better results.}\end{table*}
  \begin{table*}[!h] \resizebox{\textwidth}{!}{\begin{tabular}{ll|c|c|c|c|c|c|c|c|c}
\toprule
\textbf{Category} & \textbf{UQ Method} & \textbf{XSum} & \textbf{SamSum} & \textbf{WMT14} & \textbf{WMT19} & \textbf{CoQA} & \textbf{TriviaQA} & \textbf{MMLU} & \textbf{GSM8k} & \textbf{Mean} \\
\midrule
\multirow{6}{*}{\shortstack[l]{AR +\\trajectory\\states}} & D-LexSim & \cellcolor[rgb]{0.8336264621666667,0.8895882285,0.9964796065} -0.03 & \cellcolor[rgb]{0.9497716033000001,0.923750118,0.9088945372} 0.04 & \cellcolor[rgb]{0.6402749751176471,0.6867115845411765,0.909005371654902} 0.02 & \cellcolor[rgb]{0.61490285,0.649358983,0.8768415764999999} 0.02 & \cellcolor[rgb]{0.9607031106137255,0.7457102085921569,0.6917042176882353} 0.21 & \cellcolor[rgb]{0.61490285,0.649358983,0.8768415764999999} 0.12 & \cellcolor[rgb]{0.9004151256333333,0.6254146054,0.6128478516333333} 0.66 & \cellcolor[rgb]{0.9790880158705882,0.8856170452000001,0.8401505184058824} 0.03 & \cellcolor[rgb]{0.61490285,0.649358983,0.8768415764999999} 0.13 \\
 & D-DegMat & \cellcolor[rgb]{0.7179074031529412,0.7853536515764705,0.9758586906411765} -0.04 & \cellcolor[rgb]{0.9497716033000001,0.923750118,0.9088945372} 0.04 & \cellcolor[rgb]{0.9261890675039215,0.6732459732470588,0.6401732343490196} 0.40 & \cellcolor[rgb]{0.9513294646843138,0.7239693735803922,0.6748606867705882} 0.38 & \cellcolor[rgb]{0.8817601978137255,0.5893337236862745,0.5966981159313726} 0.28 & \cellcolor[rgb]{0.8734190061058824,0.5700105097411765,0.5899980484784314} 0.45 & \cellcolor[rgb]{0.61490285,0.649358983,0.8768415764999999} 0.39 & \cellcolor[rgb]{0.9326956664686274,0.6855638360235294,0.6478844782078431} 0.05 & \cellcolor[rgb]{0.9849255762058824,0.8479150297647058,0.7906558870392157} 0.24 \\
 & D-EigVal & \cellcolor[rgb]{0.8336264621666667,0.8895882285,0.9964796065} -0.03 & \cellcolor[rgb]{0.61490285,0.649358983,0.8768415764999999} -0.04 & \cellcolor[rgb]{0.61490285,0.649358983,0.8768415764999999} -0.00 & \cellcolor[rgb]{0.6402749751176471,0.6867115845411765,0.909005371654902} 0.04 & \cellcolor[rgb]{0.852836579,0.50777808,0.575116406} 0.30 & \cellcolor[rgb]{0.852836579,0.50777808,0.575116406} 0.46 & \cellcolor[rgb]{0.852836579,0.50777808,0.575116406} 0.68 & \cellcolor[rgb]{0.9790880158705882,0.8856170452000001,0.8401505184058824} 0.03 & \cellcolor[rgb]{0.8177369112294117,0.8783569960882354,0.9991482795509804} 0.18 \\
 & D-Ecc & \cellcolor[rgb]{0.61490285,0.649358983,0.8768415764999999} -0.05 & \cellcolor[rgb]{0.9497716033000001,0.923750118,0.9088945372} 0.04 & \cellcolor[rgb]{0.61490285,0.649358983,0.8768415764999999} 0.00 & \cellcolor[rgb]{0.61490285,0.649358983,0.8768415764999999} 0.02 & \cellcolor[rgb]{0.8675383126470588,0.5522298155294117,0.585746150627451} 0.29 & \cellcolor[rgb]{0.8952807659705883,0.6156984995294118,0.6081210191470588} 0.44 & \cellcolor[rgb]{0.852836579,0.50777808,0.575116406} 0.68 & \cellcolor[rgb]{0.9791396989627451,0.8021675484411765,0.7416485506941177} 0.04 & \cellcolor[rgb]{0.8177369112294117,0.8783569960882354,0.9991482795509804} 0.18 \\
 & D-KLE & \cellcolor[rgb]{0.852836579,0.50777808,0.575116406} 0.01 & \cellcolor[rgb]{0.9736727018,0.8973477524,0.8584952529000001} 0.05 & \cellcolor[rgb]{0.9261890675039215,0.6732459732470588,0.6401732343490196} 0.40 & \cellcolor[rgb]{0.9513294646843138,0.7239693735803922,0.6748606867705882} 0.38 & \cellcolor[rgb]{0.61490285,0.649358983,0.8768415764999999} -0.18 & \cellcolor[rgb]{0.6309026780784314,0.6732421581529412,0.8978288875588235} 0.13 & \cellcolor[rgb]{0.8763519705509804,0.5787878133490196,0.5921289546450981} 0.67 & \cellcolor[rgb]{0.61490285,0.649358983,0.8768415764999999} -0.02 & \cellcolor[rgb]{0.8177369112294117,0.8783569960882354,0.9991482795509804} 0.18 \\
 & D-LUQ & \cellcolor[rgb]{0.931695915645098,0.9325418979098039,0.9338169421313726} -0.02 & \cellcolor[rgb]{0.852836579,0.50777808,0.575116406} 0.11 & \cellcolor[rgb]{0.852836579,0.50777808,0.575116406} 0.45 & \cellcolor[rgb]{0.852836579,0.50777808,0.575116406} 0.45 & \cellcolor[rgb]{0.9348276152529411,0.6896369097254902,0.6504705511098039} 0.24 & \cellcolor[rgb]{0.9671527274529412,0.762958755827451,0.7061432370215687} 0.39 & \cellcolor[rgb]{0.9560162876490197,0.7348397910862745,0.6832824522294118} 0.63 & \cellcolor[rgb]{0.852836579,0.50777808,0.575116406} 0.06 & \cellcolor[rgb]{0.852836579,0.50777808,0.575116406} 0.29 \\
\bottomrule
\end{tabular}
}\caption{\label{tab:dream_ar_traj_results} \textbf{Baselines with trajectory states.} PRR$\uparrow$ for the Dream model across various tasks for sampling-based baselines from standard autoregressive LLMs, where denoising trajectory states are used instead of sampled generations (see Tables~\ref{tab:ar_traj_results}, \ref{tab:refusion_ar_traj_results}, and~\ref{tab:diffusiongemma_ar_traj_results} for the LLaDA-1.5, ReFusion, and DiffusionGemma models, respectively). Warmer color indicates better results.}\end{table*}

  \begin{table*}[!h] \resizebox{\textwidth}{!}{\begin{tabular}{ll|c|c|c|c|c|c|c|c|c}
\toprule
\textbf{Category} & \textbf{UQ Method} & \textbf{XSum} & \textbf{SamSum} & \textbf{WMT14} & \textbf{WMT19} & \textbf{CoQA} & \textbf{TriviaQA} & \textbf{MMLU} & \textbf{GSM8k} & \textbf{Mean} \\
\midrule
\multirow{5}{*}{\shortstack[l]{AR\\baselines}} & MSP & \cellcolor[rgb]{0.61490285,0.649358983,0.8768415764999999} -0.05 & \cellcolor[rgb]{0.7125994850980393,0.779529089882353,0.9730307285392157} 0.05 & \cellcolor[rgb]{0.61490285,0.649358983,0.8768415764999999} -0.07 & \cellcolor[rgb]{0.61490285,0.649358983,0.8768415764999999} -0.04 & \cellcolor[rgb]{0.61490285,0.649358983,0.8768415764999999} 0.17 & \cellcolor[rgb]{0.61490285,0.649358983,0.8768415764999999} 0.15 & \cellcolor[rgb]{0.9218513348607843,0.6650340647294117,0.635032405109804} 0.71 & \cellcolor[rgb]{0.6996157421568627,0.7642642360784314,0.9642295513921568} 0.16 & \cellcolor[rgb]{0.61490285,0.649358983,0.8768415764999999} 0.13 \\
 & SAR & \cellcolor[rgb]{0.8694129974705882,0.9112858109117647,0.9841305319117647} 0.14 & \cellcolor[rgb]{0.931695915645098,0.9325418979098039,0.9338169421313726} 0.20 & \cellcolor[rgb]{0.9691631781666668,0.9044582760156863,0.8705807575137254} 0.39 & \cellcolor[rgb]{0.9813541391647058,0.8767786732784314,0.8278006056137255} 0.47 & \cellcolor[rgb]{0.9216790870960785,0.9309098270078431,0.945008558445098} 0.28 & \cellcolor[rgb]{0.9460687713941176,0.7126943685490196,0.6666446363803922} 0.60 & \cellcolor[rgb]{0.9053078374137256,0.6343985308588236,0.6177138057666667} 0.73 & \cellcolor[rgb]{0.9176723556764705,0.9302569986470588,0.9494852049705882} 0.33 & \cellcolor[rgb]{0.9829494490941176,0.8700709193019608,0.8185288537078431} 0.39 \\
 & CoCoA-MTE & \cellcolor[rgb]{0.9763803588352942,0.8914823988,0.8493228856529411} 0.25 & \cellcolor[rgb]{0.9847608008647059,0.8504164338117647,0.7937540087647059} 0.28 & \cellcolor[rgb]{0.852836579,0.50777808,0.575116406} 0.72 & \cellcolor[rgb]{0.852836579,0.50777808,0.575116406} 0.76 & \cellcolor[rgb]{0.8123516188058824,0.8741592436176471,0.9993597327901961} 0.24 & \cellcolor[rgb]{0.9622044108411765,0.7492947789176471,0.6945295061352941} 0.58 & \cellcolor[rgb]{0.8979687697431373,0.6209226426705883,0.6104148745666667} 0.74 & \cellcolor[rgb]{0.9797588292352941,0.8834864272549019,0.8370723575196078} 0.41 & \cellcolor[rgb]{0.9053078374137256,0.6343985308588236,0.6177138057666667} 0.50 \\
 & KLE & \cellcolor[rgb]{0.9671527274529412,0.762958755827451,0.7061432370215687} 0.34 & \cellcolor[rgb]{0.9840526685,0.8342375980588235,0.7752431104705882} 0.29 & \cellcolor[rgb]{0.9847608008647059,0.8504164338117647,0.7937540087647059} 0.47 & \cellcolor[rgb]{0.9807976965156863,0.8111235437352942,0.7507756750235295} 0.55 & \cellcolor[rgb]{0.9829494490941176,0.8700709193019608,0.8185288537078431} 0.32 & \cellcolor[rgb]{0.864597965917647,0.5433394684235294,0.5836202017019608} 0.67 & \cellcolor[rgb]{0.9829494490941176,0.8700709193019608,0.8185288537078431} 0.52 & \cellcolor[rgb]{0.852836579,0.50777808,0.575116406} 0.60 & \cellcolor[rgb]{0.9460687713941176,0.7126943685490196,0.6666446363803922} 0.47 \\
 & TAD & \cellcolor[rgb]{0.852836579,0.50777808,0.575116406} 0.44 & \cellcolor[rgb]{0.852836579,0.50777808,0.575116406} 0.42 & \cellcolor[rgb]{0.9622044108411765,0.7492947789176471,0.6945295061352941} 0.57 & \cellcolor[rgb]{0.9829494490941176,0.8700709193019608,0.8185288537078431} 0.48 & \cellcolor[rgb]{0.7825907906117646,0.8497192224705883,0.9983175350588236} 0.23 & \cellcolor[rgb]{0.9845960255235294,0.8529178378588235,0.7968521304901961} 0.51 & \cellcolor[rgb]{0.61490285,0.649358983,0.8768415764999999} 0.02 & \cellcolor[rgb]{0.9704394715,0.9027982014117647,0.8675832782352941} 0.39 & \cellcolor[rgb]{0.9790880158705882,0.8856170452000001,0.8401505184058824} 0.38 \\\midrule
\multirow{2}{*}{\shortstack[l]{Diffusion\\baselines}} & TraceDet & \cellcolor[rgb]{0.7853082184392157,0.8520545809019608,0.9985974155294117} 0.08 & \cellcolor[rgb]{0.8543598448588235,0.9028606944647058,0.9905584045235294} 0.14 & \cellcolor[rgb]{0.9844470791666666,0.8397397817137255,0.7814061455764706} 0.48 & \cellcolor[rgb]{0.9513294646843138,0.7239693735803922,0.6748606867705882} 0.63 & \cellcolor[rgb]{0.9438760079745099,0.9270202487490196,0.9173357361078431} 0.29 & \cellcolor[rgb]{0.852836579,0.50777808,0.575116406} 0.68 & \cellcolor[rgb]{0.9053078374137256,0.6343985308588236,0.6177138057666667} 0.73 & \cellcolor[rgb]{0.8718771013137254,0.9125626824411764,0.9828988807745098} 0.29 & \cellcolor[rgb]{0.9824556940686274,0.8200795390294118,0.7599027993529412} 0.42 \\
 & DynHD & \cellcolor[rgb]{0.7446232039529412,0.8137680277960784,0.9884477548843138} 0.05 & \cellcolor[rgb]{0.8543598448588235,0.9028606944647058,0.9905584045235294} 0.14 & \cellcolor[rgb]{0.9659156483,0.7595427616,0.7032398043} 0.56 & \cellcolor[rgb]{0.9560162876490197,0.7348397910862745,0.6832824522294118} 0.62 & \cellcolor[rgb]{0.7825907906117646,0.8497192224705883,0.9983175350588236} 0.23 & \cellcolor[rgb]{0.9845960255235294,0.8529178378588235,0.7968521304901961} 0.51 & \cellcolor[rgb]{0.852836579,0.50777808,0.575116406} 0.79 & \cellcolor[rgb]{0.7798733627843137,0.8473838640392157,0.9980376545882352} 0.22 & \cellcolor[rgb]{0.9829494490941176,0.8700709193019608,0.8185288537078431} 0.39 \\\midrule
\multirow{4}{*}{\shortstack[l]{Diffusion\\signals}} & MCNLL & \cellcolor[rgb]{0.8015810339588235,0.8657637386764706,0.9997826392686274} 0.09 & \cellcolor[rgb]{0.6842533066588236,0.7455705968156863,0.9526215641058824} 0.03 & \cellcolor[rgb]{0.6667452396901961,0.7231326097019608,0.9372260848549019} 0.00 & \cellcolor[rgb]{0.6892991246352941,0.7519281085960785,0.9568458054235294} 0.06 & \cellcolor[rgb]{0.8440942415960784,0.8965891896490196,0.9940190521784313} 0.25 & \cellcolor[rgb]{0.9790880158705882,0.8856170452000001,0.8401505184058824} 0.48 & \cellcolor[rgb]{0.8898725387078432,0.6051525891921569,0.6035518578607844} 0.75 & \cellcolor[rgb]{0.8594926464901961,0.905996446872549,0.9888280806960784} 0.28 & \cellcolor[rgb]{0.7934605019215686,0.8590606561960784,0.9994370569411765} 0.24 \\
 & NFE & \cellcolor[rgb]{0.6766845796941177,0.736034329145098,0.9462852021294117} 0.00 & \cellcolor[rgb]{0.61490285,0.649358983,0.8768415764999999} -0.02 & \cellcolor[rgb]{0.9640580048333334,0.9110985744313725,0.882570674627451} 0.38 & \cellcolor[rgb]{0.9839369241588236,0.8629234540470588,0.8092446173921568} 0.49 & \cellcolor[rgb]{0.9708639649117647,0.7732067385098039,0.7148535351862745} 0.35 & \cellcolor[rgb]{0.9839369241588236,0.8629234540470588,0.8092446173921568} 0.50 & \cellcolor[rgb]{0.6217599191764706,0.6595946294941176,0.885836138382353} 0.03 & \cellcolor[rgb]{0.9377786937156862,0.9301210794313726,0.9257150330490196} 0.35 & \cellcolor[rgb]{0.8283414337745099,0.8859032383823529,0.9974569188764706} 0.26 \\
 & AD-Full & \cellcolor[rgb]{0.7446232039529412,0.8137680277960784,0.9884477548843138} 0.05 & \cellcolor[rgb]{0.8816813900509803,0.917546110909804,0.9778288382784314} 0.16 & \cellcolor[rgb]{0.9659156483,0.7595427616,0.7032398043} 0.56 & \cellcolor[rgb]{0.9560162876490197,0.7348397910862745,0.6832824522294118} 0.62 & \cellcolor[rgb]{0.8718771013137254,0.9125626824411764,0.9828988807745098} 0.26 & \cellcolor[rgb]{0.9563825307352941,0.9183409471764705,0.8972560558823529} 0.44 & \cellcolor[rgb]{0.9218513348607843,0.6650340647294117,0.635032405109804} 0.71 & \cellcolor[rgb]{0.61490285,0.649358983,0.8768415764999999} 0.09 & \cellcolor[rgb]{0.9666105915000001,0.9077784252235295,0.8765757160705883} 0.36 \\
 & CommitNLL & \cellcolor[rgb]{0.8283414337745099,0.8859032383823529,0.9974569188764706} 0.11 & \cellcolor[rgb]{0.6842533066588236,0.7455705968156863,0.9526215641058824} 0.03 & \cellcolor[rgb]{0.9727701494549019,0.8993028702666667,0.8615527086490196} 0.40 & \cellcolor[rgb]{0.9836582578333333,0.8287354144039216,0.7690800753647059} 0.53 & \cellcolor[rgb]{0.852836579,0.50777808,0.575116406} 0.40 & \cellcolor[rgb]{0.852836579,0.50777808,0.575116406} 0.68 & \cellcolor[rgb]{0.9829494490941176,0.8700709193019608,0.8185288537078431} 0.52 & \cellcolor[rgb]{0.9704394715,0.9027982014117647,0.8675832782352941} 0.39 & \cellcolor[rgb]{0.9790880158705882,0.8856170452000001,0.8401505184058824} 0.38 \\\midrule
\multirow{2}{*}{\shortstack[l]{AR +\\MCNLL}} & CoCoA-MCNLL & \cellcolor[rgb]{0.9276891842254902,0.9318890695490196,0.9382935886568627} 0.19 & \cellcolor[rgb]{0.7744380861411764,0.8425517925882353,0.9971895702117648} 0.09 & \cellcolor[rgb]{0.9838554631666667,0.8314865062313725,0.7721615929176471} 0.49 & \cellcolor[rgb]{0.9842498738333334,0.8369886898862745,0.7783246280235294} 0.52 & \cellcolor[rgb]{0.9708639649117647,0.7732067385098039,0.7148535351862745} 0.35 & \cellcolor[rgb]{0.864597965917647,0.5433394684235294,0.5836202017019608} 0.67 & \cellcolor[rgb]{0.8979687697431373,0.6209226426705883,0.6104148745666667} 0.74 & \cellcolor[rgb]{0.9824176791176471,0.8723068372941176,0.8216194376764705} 0.42 & \cellcolor[rgb]{0.9791396989627451,0.8021675484411765,0.7416485506941177} 0.43 \\
 & CoCoA-MCNLL-Norm & \cellcolor[rgb]{0.9846442845,0.8424908735411765,0.7844876631294118} 0.29 & \cellcolor[rgb]{0.9819030282176471,0.8170942072647058,0.7568604245764706} 0.30 & \cellcolor[rgb]{0.8616576191882352,0.5344491213176471,0.5814942527764706} 0.71 & \cellcolor[rgb]{0.8871684250764706,0.5998796340235294,0.6012672772176471} 0.72 & \cellcolor[rgb]{0.9438760079745099,0.9270202487490196,0.9173357361078431} 0.29 & \cellcolor[rgb]{0.864597965917647,0.5433394684235294,0.5836202017019608} 0.67 & \cellcolor[rgb]{0.8979687697431373,0.6209226426705883,0.6104148745666667} 0.74 & \cellcolor[rgb]{0.9756268974411765,0.7893996947941176,0.729703902882353} 0.48 & \cellcolor[rgb]{0.852836579,0.50777808,0.575116406} 0.53 \\\midrule
\multirow{2}{*}{\shortstack[l]{\textbf{D-CoCoA}\\\textbf{(ours)}}} & D-CoCoA-G & \cellcolor[rgb]{0.8694129974705882,0.9112858109117647,0.9841305319117647} 0.14 & \cellcolor[rgb]{0.9547297988764706,0.919693239882353,0.9001656762117647} 0.22 & \cellcolor[rgb]{0.9196824685392158,0.6609281104705882,0.632461990490196} 0.64 & \cellcolor[rgb]{0.9196824685392158,0.6609281104705882,0.632461990490196} 0.68 & \cellcolor[rgb]{0.9802450306647059,0.8081382119705882,0.7477333002470588} 0.34 & \cellcolor[rgb]{0.9261890675039215,0.6732459732470588,0.6401732343490196} 0.62 & \cellcolor[rgb]{0.8898725387078432,0.6051525891921569,0.6035518578607844} 0.75 & \cellcolor[rgb]{0.9176723556764705,0.9302569986470588,0.9494852049705882} 0.33 & \cellcolor[rgb]{0.9460687713941176,0.7126943685490196,0.6666446363803922} 0.47 \\
 & D-CoCoA-L & \cellcolor[rgb]{0.9377786937156862,0.9301210794313726,0.9257150330490196} 0.20 & \cellcolor[rgb]{0.7608481404156863,0.8297993031764705,0.9938680116235294} 0.08 & \cellcolor[rgb]{0.9708639649117647,0.7732067385098039,0.7148535351862745} 0.55 & \cellcolor[rgb]{0.9460687713941176,0.7126943685490196,0.6666446363803922} 0.64 & \cellcolor[rgb]{0.9367011412764705,0.6934798195294117,0.6531662319882353} 0.37 & \cellcolor[rgb]{0.8925766523392157,0.6104255443607843,0.6058364385039215} 0.65 & \cellcolor[rgb]{0.8616576191882352,0.5344491213176471,0.5814942527764706} 0.78 & \cellcolor[rgb]{0.9841016995,0.8604220499999999,0.8061464956666666} 0.43 & \cellcolor[rgb]{0.9560162876490197,0.7348397910862745,0.6832824522294118} 0.46 \\
\bottomrule
\end{tabular}}\caption{\label{tab:dream_detailed_main_results} \textbf{D-CoCoA comparison.} PRR$\uparrow$ for the Dream model across various tasks for D-CoCoA methods, comparing the best method from each category (see Tables~\ref{tab:detailed_main_results}, \ref{tab:refusion_detailed_main_results}, and~\ref{tab:diffusiongemma_detailed_main_results} for the LLaDA-1.5, ReFusion, and DiffusionGemma models, respectively). Warmer color indicates better results.}\end{table*}

\clearpage
\subsection{Results for ReFusion}
\label{app:res_refusion}
\begin{figure*}[h!]
    \centering
    \includegraphics[trim={0.cm 0.cm 0.cm 0.cm},clip,width=0.99\linewidth]{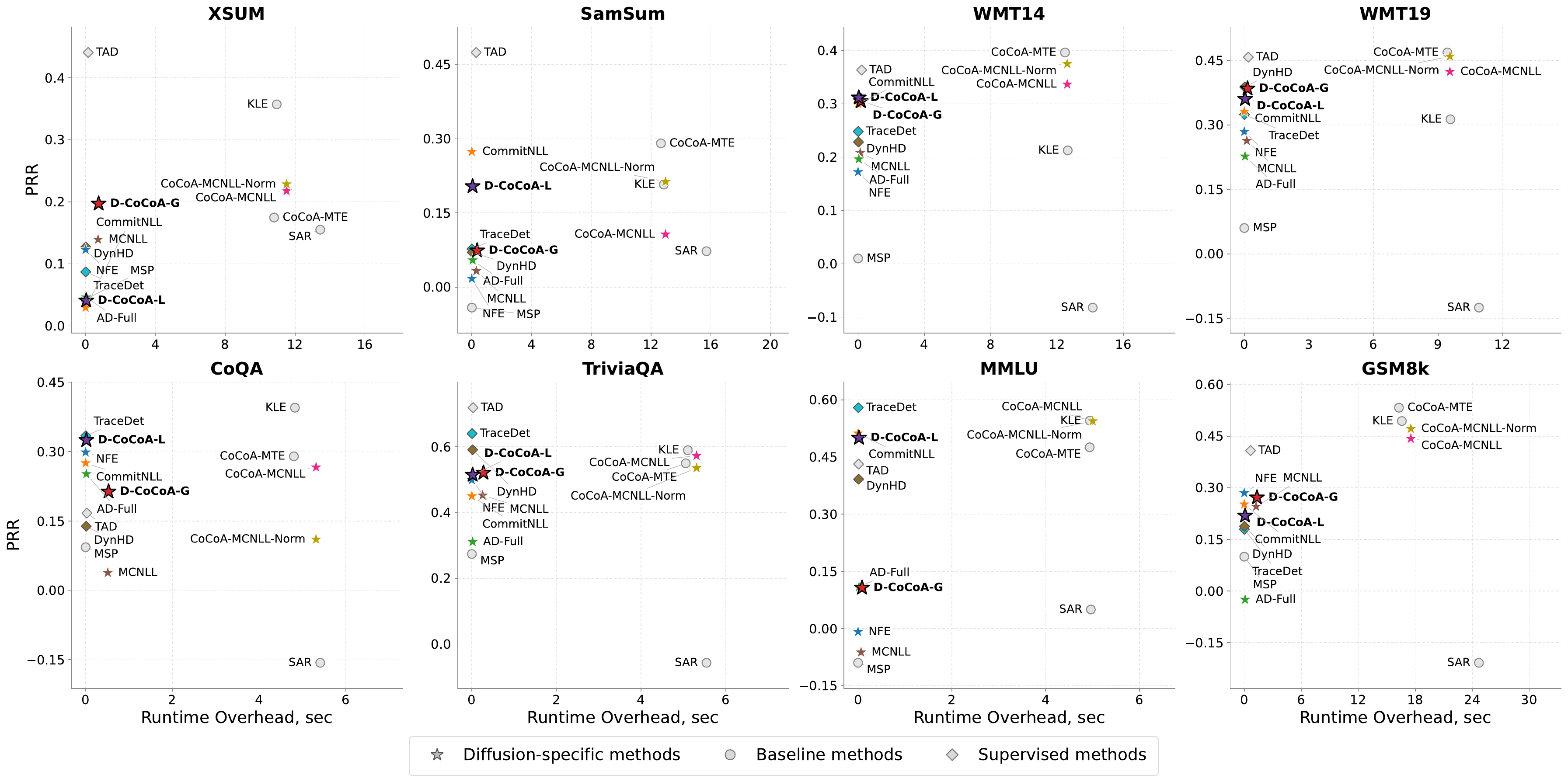}
    \caption{
    \textbf{ReFusion performance and runtime.} Comparison of uncertainty quantification performance (PRR$\uparrow$) versus computational complexity (Time in seconds$\downarrow$) across eight benchmark datasets for the ReFusion model (see Figures~\ref{fig:llada_prr_vs_comp}, \ref{fig:dream_prr_vs_comp}, and~\ref{fig:diffusiongemma_prr_vs_comp} for the LLaDA-1.5, Dream, and DiffusionGemma models, respectively).
    }
    \label{fig:refusion_prr_vs_comp}
  \end{figure*}

  Tables~\ref{tab:refusion_ar_baselines_results}, \ref{tab:refusion_diff_baselines_results}, \ref{tab:refusion_ar_improved_results}, \ref{tab:refusion_ar_traj_results}, and~\ref{tab:refusion_detailed_main_results} present the results for ReFusion with baselines from standard autoregressive LLMs, diffusion-specific methods, sampling-based baselines using MCNLL, sampling-based baselines using denoising trajectory states, and the best method from each category, respectively. Figure~\ref{fig:refusion_prr_vs_comp} presents the performance-efficiency trade-off for each dataset.

\begin{table*}[!ht] \resizebox{\textwidth}{!}{\begin{tabular}{ll|c|c|c|c|c|c|c|c|c}
\toprule
\textbf{Category} & \textbf{UQ Method} & \textbf{XSum} & \textbf{SamSum} & \textbf{WMT14} & \textbf{WMT19} & \textbf{CoQA} & \textbf{TriviaQA} & \textbf{MMLU} & \textbf{GSM8k} & \textbf{Mean} \\
\midrule
\multirow{6}{*}{\shortstack[l]{Information-\\based}} & MSP & \cellcolor[rgb]{0.7825907906117646,0.8497192224705883,0.9983175350588236} 0.04 & \cellcolor[rgb]{0.6240456089019608,0.6630065116588235,0.8888343256764706} -0.04 & \cellcolor[rgb]{0.7419271648,0.8110249248,0.9874041013} 0.01 & \cellcolor[rgb]{0.7798733627843137,0.8473838640392157,0.9980376545882352} 0.06 & \cellcolor[rgb]{0.9090282467058823,0.927794838772549,0.9573188082745099} 0.09 & \cellcolor[rgb]{0.9337138175431372,0.9321882998862745,0.9313012310098039} 0.27 & \cellcolor[rgb]{0.6741616707058824,0.7328555732549019,0.9441730814705882} -0.09 & \cellcolor[rgb]{0.9133921582352942,0.9291026777686274,0.9534763223137255} 0.10 & \cellcolor[rgb]{0.7961779297490197,0.861396014627451,0.9997169374117647} 0.06 \\
 & MTE & \cellcolor[rgb]{0.7690021078509803,0.8374507968235294,0.9958609467764705} 0.03 & \cellcolor[rgb]{0.9791396989627451,0.8021675484411765,0.7416485506941177} 0.34 & \cellcolor[rgb]{0.9844470791666666,0.8397397817137255,0.7814061455764706} 0.35 & \cellcolor[rgb]{0.9818859091411765,0.8745427552862746,0.824710021645098} 0.36 & \cellcolor[rgb]{0.7608481404156863,0.8297993031764705,0.9938680116235294} -0.04 & \cellcolor[rgb]{0.9845960255235294,0.8529178378588235,0.7968521304901961} 0.43 & \cellcolor[rgb]{0.7285232392627451,0.797002774964706,0.9815146148450979} -0.03 & \cellcolor[rgb]{0.9807976965156863,0.8111235437352942,0.7507756750235295} 0.32 & \cellcolor[rgb]{0.9613407263117647,0.9142840690588235,0.8885271948941176} 0.22 \\
 & Perplexity & \cellcolor[rgb]{0.8816813900509803,0.917546110909804,0.9778288382784314} 0.12 & \cellcolor[rgb]{0.6473592160588235,0.6967709648705882,0.9171920835607843} -0.02 & \cellcolor[rgb]{0.61490285,0.649358983,0.8768415764999999} -0.13 & \cellcolor[rgb]{0.6355521478235294,0.6800053309882352,0.9035475637176471} -0.11 & \cellcolor[rgb]{0.7961779297490197,0.861396014627451,0.9997169374117647} -0.01 & \cellcolor[rgb]{0.7717200969960785,0.8400012947058824,0.9965252584941177} 0.04 & \cellcolor[rgb]{0.7690021078509803,0.8374507968235294,0.9958609467764705} 0.01 & \cellcolor[rgb]{0.61490285,0.649358983,0.8768415764999999} -0.27 & \cellcolor[rgb]{0.6667452396901961,0.7231326097019608,0.9372260848549019} -0.05 \\
 & AttScore & \cellcolor[rgb]{0.61490285,0.649358983,0.8768415764999999} -0.10 & \cellcolor[rgb]{0.8543598448588235,0.9028606944647058,0.9905584045235294} 0.14 & \cellcolor[rgb]{0.7527113214117647,0.8219973367843137,0.9915787156372549} 0.02 & \cellcolor[rgb]{0.7988883877470588,0.8636648624411765,0.9998883658882353} 0.08 & \cellcolor[rgb]{0.8440942415960784,0.8965891896490196,0.9940190521784313} 0.03 & \cellcolor[rgb]{0.8620206859411765,0.9074551963235293,0.9878254853235294} 0.16 & \cellcolor[rgb]{0.8362689763627451,0.8914307235588235,0.9959909503117648} 0.08 & \cellcolor[rgb]{0.9736727018,0.8973477524,0.8584952529000001} 0.21 & \cellcolor[rgb]{0.8204138911862745,0.8803757532058823,0.9989228874411764} 0.08 \\
 & CCP & \cellcolor[rgb]{0.8336264621666667,0.8895882285,0.9964796065} 0.08 & \cellcolor[rgb]{0.61490285,0.649358983,0.8768415764999999} -0.05 & \cellcolor[rgb]{0.6944259359764706,0.7581492177882353,0.9606867415411765} -0.04 & \cellcolor[rgb]{0.7365350864941176,0.8055387188078431,0.9853167941313725} 0.01 & \cellcolor[rgb]{0.8569262456745098,0.9044285706686275,0.989693242609804} 0.04 & \cellcolor[rgb]{0.8492270432274511,0.8997249420568627,0.9922887283509804} 0.14 & \cellcolor[rgb]{0.61490285,0.649358983,0.8768415764999999} -0.16 & \cellcolor[rgb]{0.876805309,0.9151164254999999,0.9804355785000001} 0.05 & \cellcolor[rgb]{0.7365350864941176,0.8055387188078431,0.9853167941313725} 0.01 \\
 & RAUQ & \cellcolor[rgb]{0.7446232039529412,0.8137680277960784,0.9884477548843138} 0.01 & \cellcolor[rgb]{0.6473592160588235,0.6967709648705882,0.9171920835607843} -0.02 & \cellcolor[rgb]{0.6842533066588236,0.7455705968156863,0.9526215641058824} -0.05 & \cellcolor[rgb]{0.7022106452470588,0.7673217452235295,0.966000956317647} -0.03 & \cellcolor[rgb]{0.7473192431058824,0.8165111307921569,0.9894914084686275} -0.05 & \cellcolor[rgb]{0.7554121621254901,0.8246983074117646,0.9925393881882353} 0.02 & \cellcolor[rgb]{0.7581301512705882,0.8272488052941176,0.9932036999058824} 0.00 & \cellcolor[rgb]{0.8594926464901961,0.905996446872549,0.9888280806960784} 0.03 & \cellcolor[rgb]{0.7125994850980393,0.779529089882353,0.9730307285392157} -0.01 \\\midrule
\multirow{6}{*}{\shortstack[l]{Sampling,\\white-box}} & MCNSE & \cellcolor[rgb]{0.8933603506784313,0.9224036051843137,0.9699051924745098} 0.13 & \cellcolor[rgb]{0.7500152822588235,0.8192542337882354,0.9905350620529412} 0.06 & \cellcolor[rgb]{0.6309026780784314,0.6732421581529412,0.8978288875588235} -0.11 & \cellcolor[rgb]{0.61490285,0.649358983,0.8768415764999999} -0.14 & \cellcolor[rgb]{0.61490285,0.649358983,0.8768415764999999} -0.17 & \cellcolor[rgb]{0.61490285,0.649358983,0.8768415764999999} -0.18 & \cellcolor[rgb]{0.6473592160588235,0.6967709648705882,0.9171920835607843} -0.12 & \cellcolor[rgb]{0.6286169883529411,0.6698302759882353,0.8948307002647058} -0.25 & \cellcolor[rgb]{0.61490285,0.649358983,0.8768415764999999} -0.10 \\
 & SemanticEntropy & \cellcolor[rgb]{0.7825907906117646,0.8497192224705883,0.9983175350588236} 0.04 & \cellcolor[rgb]{0.6842533066588236,0.7455705968156863,0.9526215641058824} 0.01 & \cellcolor[rgb]{0.7527113214117647,0.8219973367843137,0.9915787156372549} 0.02 & \cellcolor[rgb]{0.7880256462666666,0.8543899393333334,0.9988772960000001} 0.07 & \cellcolor[rgb]{0.9727701494549019,0.8993028702666667,0.8615527086490196} 0.17 & \cellcolor[rgb]{0.9808223691882353,0.8790145912705882,0.830891189582353} 0.39 & \cellcolor[rgb]{0.8362689763627451,0.8914307235588235,0.9959909503117648} 0.08 & \cellcolor[rgb]{0.9704394715,0.9027982014117647,0.8675832782352941} 0.20 & \cellcolor[rgb]{0.866948893627451,0.9100089393823529,0.9853621830490196} 0.12 \\
 & SAR & \cellcolor[rgb]{0.9236824528058825,0.9312362411882353,0.9427702351823529} 0.16 & \cellcolor[rgb]{0.7635661295607843,0.8323498010588236,0.9945323233411765} 0.07 & \cellcolor[rgb]{0.6569731755647059,0.7100258308470588,0.927496270517647} -0.08 & \cellcolor[rgb]{0.6286169883529411,0.6698302759882353,0.8948307002647058} -0.12 & \cellcolor[rgb]{0.6240456089019608,0.6630065116588235,0.8888343256764706} -0.16 & \cellcolor[rgb]{0.6970208390666667,0.7612067269333334,0.9624581464666666} -0.06 & \cellcolor[rgb]{0.806966326382353,0.8699614911470589,0.9995711860294118} 0.05 & \cellcolor[rgb]{0.6594161915960784,0.7133025255607843,0.9299287241019608} -0.21 & \cellcolor[rgb]{0.6766845796941177,0.736034329145098,0.9462852021294117} -0.04 \\
 & EigenScore & \cellcolor[rgb]{0.8336264621666667,0.8895882285,0.9964796065} 0.08 & \cellcolor[rgb]{0.7635661295607843,0.8323498010588236,0.9945323233411765} 0.07 & \cellcolor[rgb]{0.9563825307352941,0.9183409471764705,0.8972560558823529} 0.25 & \cellcolor[rgb]{0.9627817115,0.9127586490352941,0.8855681539058824} 0.30 & \cellcolor[rgb]{0.9004151256333333,0.6254146054,0.6128478516333333} 0.36 & \cellcolor[rgb]{0.9696268857588235,0.769790744282353,0.7119501024647059} 0.53 & \cellcolor[rgb]{0.9423217193470588,0.7050085489411765,0.6612532746235295} 0.45 & \cellcolor[rgb]{0.9404481933235294,0.7011656391372549,0.658557593745098} 0.42 & \cellcolor[rgb]{0.9830083599196078,0.8230648707941176,0.7629451741294118} 0.31 \\
 & CoCoA-MTE & \cellcolor[rgb]{0.931695915645098,0.9325418979098039,0.9338169421313726} 0.17 & \cellcolor[rgb]{0.9834812190705882,0.867835001309804,0.8154382697392157} 0.29 & \cellcolor[rgb]{0.9747269947588235,0.7861939559392157,0.7267214884509804} 0.40 & \cellcolor[rgb]{0.9708639649117647,0.7732067385098039,0.7148535351862745} 0.47 & \cellcolor[rgb]{0.9634414899941177,0.752710773145098,0.6974329388568627} 0.29 & \cellcolor[rgb]{0.9622044108411765,0.7492947789176471,0.6945295061352941} 0.55 & \cellcolor[rgb]{0.9196824685392158,0.6609281104705882,0.632461990490196} 0.48 & \cellcolor[rgb]{0.852836579,0.50777808,0.575116406} 0.53 & \cellcolor[rgb]{0.9326956664686274,0.6855638360235294,0.6478844782078431} 0.40 \\
 & SemanticDensity & \cellcolor[rgb]{0.7554121621254901,0.8246983074117646,0.9925393881882353} 0.02 & \cellcolor[rgb]{0.9666105915000001,0.9077784252235295,0.8765757160705883} 0.25 & \cellcolor[rgb]{0.8517934440431373,0.9012928182607842,0.9914235664372548} 0.12 & \cellcolor[rgb]{0.8492270432274511,0.8997249420568627,0.9922887283509804} 0.14 & \cellcolor[rgb]{0.9003004236470589,0.9251791607803921,0.9650037801960785} 0.08 & \cellcolor[rgb]{0.9818859091411765,0.8745427552862746,0.824710021645098} 0.40 & \cellcolor[rgb]{0.7771559349568627,0.8450485056078432,0.9977577741176471} 0.02 & \cellcolor[rgb]{0.9256858185156862,0.9315626553686274,0.9405319119196078} 0.12 & \cellcolor[rgb]{0.888688766427451,0.9204606074745099,0.9730746507960784} 0.14 \\\midrule
\multirow{3}{*}{\shortstack[l]{Sampling,\\black-box}} & LUQ & \cellcolor[rgb]{0.9528917390058824,0.727592846082353,0.6776679419235294} 0.35 & \cellcolor[rgb]{0.931695915645098,0.9325418979098039,0.9338169421313726} 0.21 & \cellcolor[rgb]{0.9068462909411765,0.9271409192745098,0.959240051254902} 0.18 & \cellcolor[rgb]{0.9459084460607843,0.9259866385215686,0.914542637127451} 0.27 & \cellcolor[rgb]{0.864597965917647,0.5433394684235294,0.5836202017019608} 0.39 & \cellcolor[rgb]{0.9423217193470588,0.7050085489411765,0.6612532746235295} 0.59 & \cellcolor[rgb]{0.8616576191882352,0.5344491213176471,0.5814942527764706} 0.54 & \cellcolor[rgb]{0.8871684250764706,0.5998796340235294,0.6012672772176471} 0.49 & \cellcolor[rgb]{0.9497671903627452,0.7203459010784313,0.6720534316176471} 0.38 \\
 & KLE & \cellcolor[rgb]{0.9441952453705882,0.708851458745098,0.6639489555019608} 0.36 & \cellcolor[rgb]{0.931695915645098,0.9325418979098039,0.9338169421313726} 0.21 & \cellcolor[rgb]{0.9296925499352942,0.9322154837294118,0.9360552653941177} 0.21 & \cellcolor[rgb]{0.9666105915000001,0.9077784252235295,0.8765757160705883} 0.31 & \cellcolor[rgb]{0.852836579,0.50777808,0.575116406} 0.40 & \cellcolor[rgb]{0.9423217193470588,0.7050085489411765,0.6612532746235295} 0.59 & \cellcolor[rgb]{0.852836579,0.50777808,0.575116406} 0.55 & \cellcolor[rgb]{0.8871684250764706,0.5998796340235294,0.6012672772176471} 0.49 & \cellcolor[rgb]{0.9404481933235294,0.7011656391372549,0.658557593745098} 0.39 \\
 & DegMat & \cellcolor[rgb]{0.9763803588352942,0.8914823988,0.8493228856529411} 0.23 & \cellcolor[rgb]{0.9024823794117647,0.9258330802784314,0.9630825372156863} 0.18 & \cellcolor[rgb]{0.9514243351588236,0.9223978252941176,0.9059849168705882} 0.24 & \cellcolor[rgb]{0.9704394715,0.9027982014117647,0.8675832782352941} 0.32 & \cellcolor[rgb]{0.852836579,0.50777808,0.575116406} 0.40 & \cellcolor[rgb]{0.9423217193470588,0.7050085489411765,0.6612532746235295} 0.59 & \cellcolor[rgb]{0.8616576191882352,0.5344491213176471,0.5814942527764706} 0.54 & \cellcolor[rgb]{0.8979687697431373,0.6209226426705883,0.6104148745666667} 0.48 & \cellcolor[rgb]{0.9560162876490197,0.7348397910862745,0.6832824522294118} 0.37 \\\midrule
\multirow{4}{*}{Supervised} & SAPLMA & \cellcolor[rgb]{0.9802905992117648,0.8812505092627452,0.8339817735509804} 0.24 & \cellcolor[rgb]{0.9696268857588235,0.769790744282353,0.7119501024647059} 0.36 & \cellcolor[rgb]{0.8844643114450981,0.5946066788549019,0.5989826965745098} 0.53 & \cellcolor[rgb]{0.947942297417647,0.7165372783529411,0.6693403172588235} 0.52 & \cellcolor[rgb]{0.8336264621666667,0.8895882285,0.9964796065} 0.02 & \cellcolor[rgb]{0.947942297417647,0.7165372783529411,0.6693403172588235} 0.58 & \cellcolor[rgb]{0.9839369241588236,0.8629234540470588,0.8092446173921568} 0.31 & \cellcolor[rgb]{0.9842498738333334,0.8369886898862745,0.7783246280235294} 0.29 & \cellcolor[rgb]{0.9634414899941177,0.752710773145098,0.6974329388568627} 0.36 \\
 & SATRMD & \cellcolor[rgb]{0.9841016995,0.8604220499999999,0.8061464956666666} 0.26 & \cellcolor[rgb]{0.9836582578333333,0.8287354144039216,0.7690800753647059} 0.32 & \cellcolor[rgb]{0.9847608008647059,0.8504164338117647,0.7937540087647059} 0.34 & \cellcolor[rgb]{0.9575785619705882,0.7384632635882353,0.6860897073823529} 0.50 & \cellcolor[rgb]{0.9547297988764706,0.919693239882353,0.9001656762117647} 0.14 & \cellcolor[rgb]{0.9818859091411765,0.8745427552862746,0.824710021645098} 0.40 & \cellcolor[rgb]{0.8362689763627451,0.8914307235588235,0.9959909503117648} 0.08 & \cellcolor[rgb]{0.9849255762058824,0.8479150297647058,0.7906558870392157} 0.28 & \cellcolor[rgb]{0.9849255762058824,0.8479150297647058,0.7906558870392157} 0.29 \\
 & LookBackLens & \cellcolor[rgb]{0.9028614815235294,0.6299065681294118,0.6152808287} 0.40 & \cellcolor[rgb]{0.9236824528058825,0.9312362411882353,0.9427702351823529} 0.20 & \cellcolor[rgb]{0.852836579,0.50777808,0.575116406} 0.56 & \cellcolor[rgb]{0.852836579,0.50777808,0.575116406} 0.64 & \cellcolor[rgb]{0.9634414899941177,0.752710773145098,0.6974329388568627} 0.29 & \cellcolor[rgb]{0.8925766523392157,0.6104255443607843,0.6058364385039215} 0.67 & \cellcolor[rgb]{0.9102005491941176,0.643382456317647,0.6225797599} 0.49 & \cellcolor[rgb]{0.8871684250764706,0.5998796340235294,0.6012672772176471} 0.49 & \cellcolor[rgb]{0.852836579,0.50777808,0.575116406} 0.47 \\
 & TAD & \cellcolor[rgb]{0.852836579,0.50777808,0.575116406} 0.44 & \cellcolor[rgb]{0.852836579,0.50777808,0.575116406} 0.47 & \cellcolor[rgb]{0.9838554631666667,0.8314865062313725,0.7721615929176471} 0.36 & \cellcolor[rgb]{0.9747269947588235,0.7861939559392157,0.7267214884509804} 0.46 & \cellcolor[rgb]{0.9727701494549019,0.8993028702666667,0.8615527086490196} 0.17 & \cellcolor[rgb]{0.852836579,0.50777808,0.575116406} 0.72 & \cellcolor[rgb]{0.9544540133274511,0.7312163185843137,0.6804751970764706} 0.43 & \cellcolor[rgb]{0.9460687713941176,0.7126943685490196,0.6666446363803922} 0.41 & \cellcolor[rgb]{0.9004151256333333,0.6254146054,0.6128478516333333} 0.43 \\
\bottomrule
\end{tabular}
}\caption{\label{tab:refusion_ar_baselines_results} \textbf{AR baselines.} PRR$\uparrow$ for the ReFusion model across various tasks for baselines from standard autoregressive LLMs (see Tables~\ref{tab:ar_baselines_results}, \ref{tab:dream_ar_baselines_results}, and~\ref{tab:diffusiongemma_ar_baselines_results} for the LLaDA-1.5, Dream, and DiffusionGemma models, respectively). Warmer color indicates better results.}\end{table*}
\begin{table*}[!ht] \resizebox{\textwidth}{!}{\begin{tabular}{ll|c|c|c|c|c|c|c|c|c}
\toprule
\textbf{Category} & \textbf{UQ Method} & \textbf{XSum} & \textbf{SamSum} & \textbf{WMT14} & \textbf{WMT19} & \textbf{CoQA} & \textbf{TriviaQA} & \textbf{MMLU} & \textbf{GSM8k} & \textbf{Mean} \\
\midrule
\multirow{8}{*}{\shortstack[l]{Diffusion\\statistics}} & MCNLL & \cellcolor[rgb]{0.9175136022176471,0.6568221562117647,0.6298915758705882} 0.14 & \cellcolor[rgb]{0.6594161915960784,0.7133025255607843,0.9299287241019608} 0.03 & \cellcolor[rgb]{0.9848414898333333,0.8452419653686274,0.7875691806823529} 0.21 & \cellcolor[rgb]{0.9838554631666667,0.8314865062313725,0.7721615929176471} 0.26 & \cellcolor[rgb]{0.7258692802352942,0.794090494117647,0.9801006337941176} 0.04 & \cellcolor[rgb]{0.9829494490941176,0.8700709193019608,0.8185288537078431} 0.45 & \cellcolor[rgb]{0.61490285,0.649358983,0.8768415764999999} -0.06 & \cellcolor[rgb]{0.9102005491941176,0.643382456317647,0.6225797599} 0.25 & \cellcolor[rgb]{0.7125994850980393,0.779529089882353,0.9730307285392157} 0.16 \\
 & MCNLL-Norm & \cellcolor[rgb]{0.852836579,0.50777808,0.575116406} 0.16 & \cellcolor[rgb]{0.931695915645098,0.9325418979098039,0.9338169421313726} 0.14 & \cellcolor[rgb]{0.9404481933235294,0.7011656391372549,0.658557593745098} 0.26 & \cellcolor[rgb]{0.9102005491941176,0.643382456317647,0.6225797599} 0.31 & \cellcolor[rgb]{0.61490285,0.649358983,0.8768415764999999} -0.03 & \cellcolor[rgb]{0.61490285,0.649358983,0.8768415764999999} 0.19 & \cellcolor[rgb]{0.61490285,0.649358983,0.8768415764999999} -0.06 & \cellcolor[rgb]{0.9738270920764706,0.7829882170843137,0.7237390740196079} 0.20 & \cellcolor[rgb]{0.6766845796941177,0.736034329145098,0.9462852021294117} 0.15 \\
\cmidrule(lr){2-11}
 & NFE & \cellcolor[rgb]{0.9622044108411765,0.7492947789176471,0.6945295061352941} 0.12 & \cellcolor[rgb]{0.6355521478235294,0.6800053309882352,0.9035475637176471} 0.02 & \cellcolor[rgb]{0.9580352625941178,0.9169886544705883,0.8943464355529411} 0.17 & \cellcolor[rgb]{0.9683898066058824,0.766374750054902,0.7090466697431372} 0.28 & \cellcolor[rgb]{0.947942297417647,0.7165372783529411,0.6693403172588235} 0.30 & \cellcolor[rgb]{0.9729271893941176,0.7797824782294118,0.7207566595882353} 0.50 & \cellcolor[rgb]{0.6594161915960784,0.7133025255607843,0.9299287241019608} -0.01 & \cellcolor[rgb]{0.852836579,0.50777808,0.575116406} 0.28 & \cellcolor[rgb]{0.888688766427451,0.9204606074745099,0.9730746507960784} 0.21 \\
 & Remask & \cellcolor[rgb]{0.9836582578333333,0.8287354144039216,0.7690800753647059} 0.10 & \cellcolor[rgb]{0.9133921582352942,0.9291026777686274,0.9534763223137255} 0.13 & \cellcolor[rgb]{0.9781854635254902,0.8875721630666666,0.8432079741549019} 0.19 & \cellcolor[rgb]{0.9348276152529411,0.6896369097254902,0.6504705511098039} 0.30 & \cellcolor[rgb]{0.947942297417647,0.7165372783529411,0.6693403172588235} 0.30 & \cellcolor[rgb]{0.9729271893941176,0.7797824782294118,0.7207566595882353} 0.50 & \cellcolor[rgb]{0.6594161915960784,0.7133025255607843,0.9299287241019608} -0.01 & \cellcolor[rgb]{0.9423217193470588,0.7050085489411765,0.6612532746235295} 0.23 & \cellcolor[rgb]{0.9196757213862745,0.930583412827451,0.9472468817078432} 0.22 \\
 & FlipCount & \cellcolor[rgb]{0.9781854635254902,0.8875721630666666,0.8432079741549019} 0.08 & \cellcolor[rgb]{0.7099953545176471,0.7764942726588235,0.9713151710941177} 0.05 & \cellcolor[rgb]{0.9046643351764706,0.9264869997764706,0.9611612942352941} 0.14 & \cellcolor[rgb]{0.9841016995,0.8604220499999999,0.8061464956666666} 0.25 & \cellcolor[rgb]{0.9348276152529411,0.6896369097254902,0.6504705511098039} 0.31 & \cellcolor[rgb]{0.9791396989627451,0.8021675484411765,0.7416485506941177} 0.49 & \cellcolor[rgb]{0.6594161915960784,0.7133025255607843,0.9299287241019608} -0.01 & \cellcolor[rgb]{0.953077067017647,0.9210455325882353,0.9030752965411765} 0.12 & \cellcolor[rgb]{0.7853082184392157,0.8520545809019608,0.9985974155294117} 0.18 \\
\cmidrule(lr){2-11}
 & TrajNLL & \cellcolor[rgb]{0.61490285,0.649358983,0.8768415764999999} -0.05 & \cellcolor[rgb]{0.61490285,0.649358983,0.8768415764999999} 0.01 & \cellcolor[rgb]{0.61490285,0.649358983,0.8768415764999999} 0.01 & \cellcolor[rgb]{0.61490285,0.649358983,0.8768415764999999} 0.09 & \cellcolor[rgb]{0.947942297417647,0.7165372783529411,0.6693403172588235} 0.30 & \cellcolor[rgb]{0.852836579,0.50777808,0.575116406} 0.59 & \cellcolor[rgb]{0.852836579,0.50777808,0.575116406} 0.58 & \cellcolor[rgb]{0.7205613621803922,0.7882659324235295,0.9772726716921569} -0.01 & \cellcolor[rgb]{0.8204138911862745,0.8803757532058823,0.9989228874411764} 0.19 \\
 & TrajEntropy & \cellcolor[rgb]{0.8336264621666667,0.8895882285,0.9964796065} 0.02 & \cellcolor[rgb]{0.9053078374137256,0.6343985308588236,0.6177138057666667} 0.25 & \cellcolor[rgb]{0.9404481933235294,0.7011656391372549,0.658557593745098} 0.26 & \cellcolor[rgb]{0.852836579,0.50777808,0.575116406} 0.33 & \cellcolor[rgb]{0.852836579,0.50777808,0.575116406} 0.36 & \cellcolor[rgb]{0.9196824685392158,0.6609281104705882,0.632461990490196} 0.55 & \cellcolor[rgb]{0.852836579,0.50777808,0.575116406} 0.58 & \cellcolor[rgb]{0.9659156483,0.7595427616,0.7032398043} 0.21 & \cellcolor[rgb]{0.852836579,0.50777808,0.575116406} 0.32 \\
 & CommitNLL & \cellcolor[rgb]{0.8644847897843138,0.9087320678529411,0.9865938341862746} 0.03 & \cellcolor[rgb]{0.852836579,0.50777808,0.575116406} 0.27 & \cellcolor[rgb]{0.852836579,0.50777808,0.575116406} 0.30 & \cellcolor[rgb]{0.852836579,0.50777808,0.575116406} 0.33 & \cellcolor[rgb]{0.9671527274529412,0.762958755827451,0.7061432370215687} 0.28 & \cellcolor[rgb]{0.9829494490941176,0.8700709193019608,0.8185288537078431} 0.45 & \cellcolor[rgb]{0.924020201182353,0.6691400189882353,0.6376028197294118} 0.51 & \cellcolor[rgb]{0.9102005491941176,0.643382456317647,0.6225797599} 0.25 & \cellcolor[rgb]{0.9218513348607843,0.6650340647294117,0.635032405109804} 0.30 \\
\midrule
\multirow{8}{*}{\shortstack[l]{Trajectory\\dissimilarity}} & AD-Block & \cellcolor[rgb]{0.8336264621666667,0.8895882285,0.9964796065} 0.02 & \cellcolor[rgb]{0.6842533066588236,0.7455705968156863,0.9526215641058824} 0.04 & \cellcolor[rgb]{0.8840171821764706,0.9185176097647059,0.976244109117647} 0.13 & \cellcolor[rgb]{0.953077067017647,0.9210455325882353,0.9030752965411765} 0.22 & \cellcolor[rgb]{0.9834610525,0.8259843225764706,0.7659985578117647} 0.25 & \cellcolor[rgb]{0.8096589725941177,0.8720603673823529,0.9994654594098039} 0.31 & \cellcolor[rgb]{0.7853082184392157,0.8520545809019608,0.9985974155294117} 0.11 & \cellcolor[rgb]{0.9398111318019609,0.9290874692039215,0.9229219340686274} 0.11 & \cellcolor[rgb]{0.6766845796941177,0.736034329145098,0.9462852021294117} 0.15 \\
 & WeightedAD-Block & \cellcolor[rgb]{0.8644847897843138,0.9087320678529411,0.9865938341862746} 0.03 & \cellcolor[rgb]{0.6842533066588236,0.7455705968156863,0.9526215641058824} 0.04 & \cellcolor[rgb]{0.9704394715,0.9027982014117647,0.8675832782352941} 0.18 & \cellcolor[rgb]{0.9790880158705882,0.8856170452000001,0.8401505184058824} 0.24 & \cellcolor[rgb]{0.9796923648137255,0.8051528802058823,0.7446909254705882} 0.26 & \cellcolor[rgb]{0.8096589725941177,0.8720603673823529,0.9994654594098039} 0.31 & \cellcolor[rgb]{0.7853082184392157,0.8520545809019608,0.9985974155294117} 0.11 & \cellcolor[rgb]{0.9849255762058824,0.8479150297647058,0.7906558870392157} 0.17 & \cellcolor[rgb]{0.7473192431058824,0.8165111307921569,0.9894914084686275} 0.17 \\
\cmidrule(lr){2-11}
 & AD-LastBlock & \cellcolor[rgb]{0.8336264621666667,0.8895882285,0.9964796065} 0.02 & \cellcolor[rgb]{0.7365350864941176,0.8055387188078431,0.9853167941313725} 0.06 & \cellcolor[rgb]{0.9580352625941178,0.9169886544705883,0.8943464355529411} 0.17 & \cellcolor[rgb]{0.9691631781666668,0.9044582760156863,0.8705807575137254} 0.23 & \cellcolor[rgb]{0.9834610525,0.8259843225764706,0.7659985578117647} 0.25 & \cellcolor[rgb]{0.8096589725941177,0.8720603673823529,0.9994654594098039} 0.31 & \cellcolor[rgb]{0.7853082184392157,0.8520545809019608,0.9985974155294117} 0.11 & \cellcolor[rgb]{0.8389114905588235,0.8932732186176471,0.9955022941235294} 0.05 & \cellcolor[rgb]{0.6766845796941177,0.736034329145098,0.9462852021294117} 0.15 \\
 & WeightedAD-LastBlock & \cellcolor[rgb]{0.8933603506784313,0.9224036051843137,0.9699051924745098} 0.04 & \cellcolor[rgb]{0.7099953545176471,0.7764942726588235,0.9713151710941177} 0.05 & \cellcolor[rgb]{0.9580352625941178,0.9169886544705883,0.8943464355529411} 0.17 & \cellcolor[rgb]{0.9838554631666667,0.8314865062313725,0.7721615929176471} 0.26 & \cellcolor[rgb]{0.9796923648137255,0.8051528802058823,0.7446909254705882} 0.26 & \cellcolor[rgb]{0.8096589725941177,0.8720603673823529,0.9994654594098039} 0.31 & \cellcolor[rgb]{0.7853082184392157,0.8520545809019608,0.9985974155294117} 0.11 & \cellcolor[rgb]{0.8594926464901961,0.905996446872549,0.9888280806960784} 0.06 & \cellcolor[rgb]{0.7125994850980393,0.779529089882353,0.9730307285392157} 0.16 \\
\cmidrule(lr){2-11}
 & AD-LastBlockPrefix & \cellcolor[rgb]{0.8644847897843138,0.9087320678529411,0.9865938341862746} 0.03 & \cellcolor[rgb]{0.866948893627451,0.9100089393823529,0.9853621830490196} 0.11 & \cellcolor[rgb]{0.8415278407803921,0.8950213134450979,0.9948842140921569} 0.11 & \cellcolor[rgb]{0.8863529743019607,0.9194891086196079,0.9746593799568628} 0.19 & \cellcolor[rgb]{0.9834610525,0.8259843225764706,0.7659985578117647} 0.25 & \cellcolor[rgb]{0.8096589725941177,0.8720603673823529,0.9994654594098039} 0.31 & \cellcolor[rgb]{0.7853082184392157,0.8520545809019608,0.9985974155294117} 0.11 & \cellcolor[rgb]{0.61490285,0.649358983,0.8768415764999999} -0.07 & \cellcolor[rgb]{0.61490285,0.649358983,0.8768415764999999} 0.13 \\
 & WeightedAD-LastBlockPrefix & \cellcolor[rgb]{0.8933603506784313,0.9224036051843137,0.9699051924745098} 0.04 & \cellcolor[rgb]{0.815044265017647,0.8762581198529411,0.9992540061705882} 0.09 & \cellcolor[rgb]{0.8644847897843138,0.9087320678529411,0.9865938341862746} 0.12 & \cellcolor[rgb]{0.931695915645098,0.9325418979098039,0.9338169421313726} 0.21 & \cellcolor[rgb]{0.9796923648137255,0.8051528802058823,0.7446909254705882} 0.26 & \cellcolor[rgb]{0.8096589725941177,0.8720603673823529,0.9994654594098039} 0.31 & \cellcolor[rgb]{0.7853082184392157,0.8520545809019608,0.9985974155294117} 0.11 & \cellcolor[rgb]{0.6473592160588235,0.6967709648705882,0.9171920835607843} -0.05 & \cellcolor[rgb]{0.6449978024117646,0.6934178380941176,0.9144631795921568} 0.14 \\
\cmidrule(lr){2-11}
 & AD-Full & \cellcolor[rgb]{0.9196757213862745,0.930583412827451,0.9472468817078432} 0.05 & \cellcolor[rgb]{0.7099953545176471,0.7764942726588235,0.9713151710941177} 0.05 & \cellcolor[rgb]{0.9834812190705882,0.867835001309804,0.8154382697392157} 0.20 & \cellcolor[rgb]{0.9691631781666668,0.9044582760156863,0.8705807575137254} 0.23 & \cellcolor[rgb]{0.9834610525,0.8259843225764706,0.7659985578117647} 0.25 & \cellcolor[rgb]{0.8096589725941177,0.8720603673823529,0.9994654594098039} 0.31 & \cellcolor[rgb]{0.7853082184392157,0.8520545809019608,0.9985974155294117} 0.11 & \cellcolor[rgb]{0.7022106452470588,0.7673217452235295,0.966000956317647} -0.02 & \cellcolor[rgb]{0.6766845796941177,0.736034329145098,0.9462852021294117} 0.15 \\
 & WeightedAD-Full & \cellcolor[rgb]{0.9459084460607843,0.9259866385215686,0.914542637127451} 0.06 & \cellcolor[rgb]{0.61490285,0.649358983,0.8768415764999999} 0.01 & \cellcolor[rgb]{0.9834812190705882,0.867835001309804,0.8154382697392157} 0.20 & \cellcolor[rgb]{0.9838554631666667,0.8314865062313725,0.7721615929176471} 0.26 & \cellcolor[rgb]{0.9796923648137255,0.8051528802058823,0.7446909254705882} 0.26 & \cellcolor[rgb]{0.8096589725941177,0.8720603673823529,0.9994654594098039} 0.31 & \cellcolor[rgb]{0.7853082184392157,0.8520545809019608,0.9985974155294117} 0.11 & \cellcolor[rgb]{0.7798733627843137,0.8473838640392157,0.9980376545882352} 0.02 & \cellcolor[rgb]{0.6766845796941177,0.736034329145098,0.9462852021294117} 0.15 \\
\bottomrule
\end{tabular}
}\caption{\label{tab:refusion_diff_baselines_results} \textbf{Diffusion-specific UQ signals.} PRR$\uparrow$ for the ReFusion model across various tasks for diffusion statistics and trajectory dissimilarity UQ methods (see Tables~\ref{tab:diff_baselines_results}, \ref{tab:dream_diff_baselines_results}, and~\ref{tab:diffusiongemma_diff_baselines_results} for the LLaDA-1.5, Dream, and DiffusionGemma models, respectively). Warmer color indicates better results.}\end{table*}

\begin{table*}[!ht] \resizebox{\textwidth}{!}{\begin{tabular}{ll|c|c|c|c|c|c|c|c|c}
\toprule
\textbf{Category} & \textbf{UQ Method} & \textbf{XSum} & \textbf{SamSum} & \textbf{WMT14} & \textbf{WMT19} & \textbf{CoQA} & \textbf{TriviaQA} & \textbf{MMLU} & \textbf{GSM8k} & \textbf{Mean} \\
\midrule
\multirow{4}{*}{AR} & MCNSE & \cellcolor[rgb]{0.9563825307352941,0.9183409471764705,0.8972560558823529} 0.13 & \cellcolor[rgb]{0.7473192431058824,0.8165111307921569,0.9894914084686275} 0.06 & \cellcolor[rgb]{0.61490285,0.649358983,0.8768415764999999} -0.11 & \cellcolor[rgb]{0.61490285,0.649358983,0.8768415764999999} -0.14 & \cellcolor[rgb]{0.61490285,0.649358983,0.8768415764999999} -0.17 & \cellcolor[rgb]{0.61490285,0.649358983,0.8768415764999999} -0.18 & \cellcolor[rgb]{0.61490285,0.649358983,0.8768415764999999} -0.12 & \cellcolor[rgb]{0.61490285,0.649358983,0.8768415764999999} -0.25 & \cellcolor[rgb]{0.61490285,0.649358983,0.8768415764999999} -0.10 \\
 & SemanticEntropy & \cellcolor[rgb]{0.6741616707058824,0.7328555732549019,0.9441730814705882} 0.04 & \cellcolor[rgb]{0.61490285,0.649358983,0.8768415764999999} 0.01 & \cellcolor[rgb]{0.8015810339588235,0.8657637386764706,0.9997826392686274} 0.02 & \cellcolor[rgb]{0.8594926464901961,0.905996446872549,0.9888280806960784} 0.07 & \cellcolor[rgb]{0.9802905992117648,0.8812505092627452,0.8339817735509804} 0.17 & \cellcolor[rgb]{0.9765268001235294,0.7926054336490196,0.7326863173137255} 0.39 & \cellcolor[rgb]{0.8123516188058824,0.8741592436176471,0.9993597327901961} 0.08 & \cellcolor[rgb]{0.9829494490941176,0.8700709193019608,0.8185288537078431} 0.20 & \cellcolor[rgb]{0.9216790870960785,0.9309098270078431,0.945008558445098} 0.12 \\
 & SemanticDensity & \cellcolor[rgb]{0.61490285,0.649358983,0.8768415764999999} 0.02 & \cellcolor[rgb]{0.852836579,0.50777808,0.575116406} 0.25 & \cellcolor[rgb]{0.9377786937156862,0.9301210794313726,0.9257150330490196} 0.12 & \cellcolor[rgb]{0.931695915645098,0.9325418979098039,0.9338169421313726} 0.14 & \cellcolor[rgb]{0.9133921582352942,0.9291026777686274,0.9534763223137255} 0.08 & \cellcolor[rgb]{0.9738270920764706,0.7829882170843137,0.7237390740196079} 0.40 & \cellcolor[rgb]{0.7500152822588235,0.8192542337882354,0.9905350620529412} 0.02 & \cellcolor[rgb]{0.9514243351588236,0.9223978252941176,0.9059849168705882} 0.12 & \cellcolor[rgb]{0.9438760079745099,0.9270202487490196,0.9173357361078431} 0.14 \\
 & CoCoA-MSP & \cellcolor[rgb]{0.9196824685392158,0.6609281104705882,0.632461990490196} 0.20 & \cellcolor[rgb]{0.7179074031529412,0.7853536515764705,0.9758586906411765} 0.05 & \cellcolor[rgb]{0.9575785619705882,0.7384632635882353,0.6860897073823529} 0.26 & \cellcolor[rgb]{0.9423217193470588,0.7050085489411765,0.6612532746235295} 0.34 & \cellcolor[rgb]{0.852836579,0.50777808,0.575116406} 0.37 & \cellcolor[rgb]{0.8817601978137255,0.5893337236862745,0.5966981159313726} 0.54 & \cellcolor[rgb]{0.852836579,0.50777808,0.575116406} 0.54 & \cellcolor[rgb]{0.9196824685392158,0.6609281104705882,0.632461990490196} 0.37 & \cellcolor[rgb]{0.8844643114450981,0.5946066788549019,0.5989826965745098} 0.34 \\
\midrule
\multirow{4}{*}{\shortstack[l]{AR +\\MCNLL}} & MCNSE-MCNLL & \cellcolor[rgb]{0.9563825307352941,0.9183409471764705,0.8972560558823529} 0.13 & \cellcolor[rgb]{0.9841016995,0.8604220499999999,0.8061464956666666} 0.17 & \cellcolor[rgb]{0.9126469050843138,0.6478744190470588,0.6250127369666667} 0.30 & \cellcolor[rgb]{0.864597965917647,0.5433394684235294,0.5836202017019608} 0.41 & \cellcolor[rgb]{0.7179074031529412,0.7853536515764705,0.9758586906411765} -0.08 & \cellcolor[rgb]{0.7880256462666666,0.8543899393333334,0.9988772960000001} 0.02 & \cellcolor[rgb]{0.8440942415960784,0.8965891896490196,0.9940190521784313} 0.11 & \cellcolor[rgb]{0.9824556940686274,0.8200795390294118,0.7599027993529412} 0.25 & \cellcolor[rgb]{0.9627817115,0.9127586490352941,0.8855681539058824} 0.16 \\
 & SemanticEntropy-MCNLL & \cellcolor[rgb]{0.9068462909411765,0.9271409192745098,0.959240051254902} 0.11 & \cellcolor[rgb]{0.7744380861411764,0.8425517925882353,0.9971895702117648} 0.07 & \cellcolor[rgb]{0.9836582578333333,0.8287354144039216,0.7690800753647059} 0.21 & \cellcolor[rgb]{0.9646785691470589,0.756126767372549,0.7003363715784314} 0.31 & \cellcolor[rgb]{0.8790560841823529,0.5840607685176471,0.5944135352882353} 0.35 & \cellcolor[rgb]{0.8704786593764706,0.5611201626352941,0.5878720995529412} 0.55 & \cellcolor[rgb]{0.9547297988764706,0.919693239882353,0.9001656762117647} 0.24 & \cellcolor[rgb]{0.9196824685392158,0.6609281104705882,0.632461990490196} 0.37 & \cellcolor[rgb]{0.9560162876490197,0.7348397910862745,0.6832824522294118} 0.28 \\
 & SemanticDensity-MCNLL & \cellcolor[rgb]{0.6426363887647059,0.690064711317647,0.9117342756235294} 0.03 & \cellcolor[rgb]{0.852836579,0.50777808,0.575116406} 0.25 & \cellcolor[rgb]{0.9818859091411765,0.8745427552862746,0.824710021645098} 0.18 & \cellcolor[rgb]{0.9818859091411765,0.8745427552862746,0.824710021645098} 0.22 & \cellcolor[rgb]{0.8933603506784313,0.9224036051843137,0.9699051924745098} 0.06 & \cellcolor[rgb]{0.9738270920764706,0.7829882170843137,0.7237390740196079} 0.40 & \cellcolor[rgb]{0.7500152822588235,0.8192542337882354,0.9905350620529412} 0.02 & \cellcolor[rgb]{0.9497671903627452,0.7203459010784313,0.6720534316176471} 0.33 & \cellcolor[rgb]{0.9802905992117648,0.8812505092627452,0.8339817735509804} 0.19 \\
 & CoCoA-MCNLL & \cellcolor[rgb]{0.852836579,0.50777808,0.575116406} 0.22 & \cellcolor[rgb]{0.8863529743019607,0.9194891086196079,0.9746593799568628} 0.11 & \cellcolor[rgb]{0.852836579,0.50777808,0.575116406} 0.34 & \cellcolor[rgb]{0.852836579,0.50777808,0.575116406} 0.42 & \cellcolor[rgb]{0.9607031106137255,0.7457102085921569,0.6917042176882353} 0.27 & \cellcolor[rgb]{0.852836579,0.50777808,0.575116406} 0.57 & \cellcolor[rgb]{0.852836579,0.50777808,0.575116406} 0.54 & \cellcolor[rgb]{0.852836579,0.50777808,0.575116406} 0.44 & \cellcolor[rgb]{0.852836579,0.50777808,0.575116406} 0.36 \\
\bottomrule
\end{tabular}
}\caption{\label{tab:refusion_ar_improved_results} \textbf{Baselines with diffusion likelihoods.} PRR$\uparrow$ for the ReFusion model across various tasks for sampling-based baselines from standard autoregressive LLMs, where the diffusion likelihood (MCNLL) is used instead of the AR likelihood (see Tables~\ref{tab:ar_improved_results} and \ref{tab:dream_ar_improved_results} for the LLaDA-1.5 and Dream models, respectively). Warmer color indicates better results.}\end{table*}
\begin{table*}[!ht] \resizebox{\textwidth}{!}{\begin{tabular}{ll|c|c|c|c|c|c|c|c|c}
\toprule
\textbf{Category} & \textbf{UQ Method} & \textbf{XSum} & \textbf{SamSum} & \textbf{WMT14} & \textbf{WMT19} & \textbf{CoQA} & \textbf{TriviaQA} & \textbf{MMLU} & \textbf{GSM8k} & \textbf{Mean} \\
\midrule
\multirow{6}{*}{\shortstack[l]{AR +\\trajectory\\states}} & D-LexSim & \cellcolor[rgb]{0.852836579,0.50777808,0.575116406} 0.09 & \cellcolor[rgb]{0.6892991246352941,0.7519281085960785,0.9568458054235294} 0.03 & \cellcolor[rgb]{0.9659156483,0.7595427616,0.7032398043} 0.07 & \cellcolor[rgb]{0.9659156483,0.7595427616,0.7032398043} 0.13 & \cellcolor[rgb]{0.931695915645098,0.9325418979098039,0.9338169421313726} 0.26 & \cellcolor[rgb]{0.8015810339588235,0.8657637386764706,0.9997826392686274} 0.37 & \cellcolor[rgb]{0.61490285,0.649358983,0.8768415764999999} 0.00 & \cellcolor[rgb]{0.61490285,0.649358983,0.8768415764999999} -0.16 & \cellcolor[rgb]{0.931695915645098,0.9325418979098039,0.9338169421313726} 0.10 \\
 & D-DegMat & \cellcolor[rgb]{0.9024823794117647,0.9258330802784314,0.9630825372156863} 0.04 & \cellcolor[rgb]{0.9791396989627451,0.8021675484411765,0.7416485506941177} 0.08 & \cellcolor[rgb]{0.9736727018,0.8973477524,0.8584952529000001} 0.05 & \cellcolor[rgb]{0.852836579,0.50777808,0.575116406} 0.18 & \cellcolor[rgb]{0.852836579,0.50777808,0.575116406} 0.27 & \cellcolor[rgb]{0.8015810339588235,0.8657637386764706,0.9997826392686274} 0.37 & \cellcolor[rgb]{0.852836579,0.50777808,0.575116406} 0.09 & \cellcolor[rgb]{0.9423217193470588,0.7050085489411765,0.6612532746235295} -0.04 & \cellcolor[rgb]{0.852836579,0.50777808,0.575116406} 0.13 \\
 & D-EigVal & \cellcolor[rgb]{0.9024823794117647,0.9258330802784314,0.9630825372156863} 0.04 & \cellcolor[rgb]{0.9791396989627451,0.8021675484411765,0.7416485506941177} 0.08 & \cellcolor[rgb]{0.9736727018,0.8973477524,0.8584952529000001} 0.05 & \cellcolor[rgb]{0.8817601978137255,0.5893337236862745,0.5966981159313726} 0.17 & \cellcolor[rgb]{0.931695915645098,0.9325418979098039,0.9338169421313726} 0.26 & \cellcolor[rgb]{0.9423217193470588,0.7050085489411765,0.6612532746235295} 0.41 & \cellcolor[rgb]{0.9720272867117647,0.7765767393745098,0.7177742451568627} 0.07 & \cellcolor[rgb]{0.9028614815235294,0.6299065681294118,0.6152808287} -0.03 & \cellcolor[rgb]{0.852836579,0.50777808,0.575116406} 0.13 \\
 & D-Ecc & \cellcolor[rgb]{0.61490285,0.649358983,0.8768415764999999} 0.00 & \cellcolor[rgb]{0.61490285,0.649358983,0.8768415764999999} 0.02 & \cellcolor[rgb]{0.61490285,0.649358983,0.8768415764999999} -0.01 & \cellcolor[rgb]{0.9765268001235294,0.7926054336490196,0.7326863173137255} 0.12 & \cellcolor[rgb]{0.61490285,0.649358983,0.8768415764999999} 0.25 & \cellcolor[rgb]{0.852836579,0.50777808,0.575116406} 0.42 & \cellcolor[rgb]{0.61490285,0.649358983,0.8768415764999999} 0.00 & \cellcolor[rgb]{0.852836579,0.50777808,0.575116406} -0.02 & \cellcolor[rgb]{0.931695915645098,0.9325418979098039,0.9338169421313726} 0.10 \\
 & D-KLE & \cellcolor[rgb]{0.6817303976705882,0.7423918409254902,0.9505094434470589} 0.01 & \cellcolor[rgb]{0.931695915645098,0.9325418979098039,0.9338169421313726} 0.06 & \cellcolor[rgb]{0.7419271648,0.8110249248,0.9874041013} 0.01 & \cellcolor[rgb]{0.61490285,0.649358983,0.8768415764999999} -0.07 & \cellcolor[rgb]{0.852836579,0.50777808,0.575116406} 0.27 & \cellcolor[rgb]{0.61490285,0.649358983,0.8768415764999999} 0.35 & \cellcolor[rgb]{0.61490285,0.649358983,0.8768415764999999} 0.00 & \cellcolor[rgb]{0.9696268857588235,0.769790744282353,0.7119501024647059} -0.05 & \cellcolor[rgb]{0.61490285,0.649358983,0.8768415764999999} 0.07 \\
 & D-LUQ & \cellcolor[rgb]{0.9596879944529412,0.9156363617647059,0.8914368152235295} 0.05 & \cellcolor[rgb]{0.852836579,0.50777808,0.575116406} 0.10 & \cellcolor[rgb]{0.852836579,0.50777808,0.575116406} 0.09 & \cellcolor[rgb]{0.9305268001470588,0.6814578817647059,0.6453140635882353} 0.15 & \cellcolor[rgb]{0.852836579,0.50777808,0.575116406} 0.27 & \cellcolor[rgb]{0.8015810339588235,0.8657637386764706,0.9997826392686274} 0.37 & \cellcolor[rgb]{0.61490285,0.649358983,0.8768415764999999} 0.00 & \cellcolor[rgb]{0.852836579,0.50777808,0.575116406} -0.02 & \cellcolor[rgb]{0.852836579,0.50777808,0.575116406} 0.13 \\
\bottomrule
\end{tabular}
}\caption{\label{tab:refusion_ar_traj_results} \textbf{Baselines with trajectory states.} PRR$\uparrow$ for the ReFusion model across various tasks for sampling-based baselines from standard autoregressive LLMs, where denoising trajectory states are used instead of sampled generations (see Tables~\ref{tab:ar_traj_results}, \ref{tab:dream_ar_traj_results}, and~\ref{tab:diffusiongemma_ar_traj_results} for the LLaDA-1.5, Dream, and DiffusionGemma models, respectively). Warmer color indicates better results.}\end{table*}
\begin{table*}[!ht] \resizebox{\textwidth}{!}{\begin{tabular}{ll|c|c|c|c|c|c|c|c|c}
\toprule
\textbf{Category} & \textbf{UQ Method} & \textbf{XSum} & \textbf{SamSum} & \textbf{WMT14} & \textbf{WMT19} & \textbf{CoQA} & \textbf{TriviaQA} & \textbf{MMLU} & \textbf{GSM8k} & \textbf{Mean} \\
\midrule
\multirow{5}{*}{\shortstack[l]{AR\\baselines}} & MSP & \cellcolor[rgb]{0.6286169883529411,0.6698302759882353,0.8948307002647058} 0.04 & \cellcolor[rgb]{0.61490285,0.649358983,0.8768415764999999} -0.04 & \cellcolor[rgb]{0.7311771982901961,0.7999150558117647,0.9829285958960784} 0.01 & \cellcolor[rgb]{0.815044265017647,0.8762581198529411,0.9992540061705882} 0.06 & \cellcolor[rgb]{0.9046643351764706,0.9264869997764706,0.9611612942352941} 0.09 & \cellcolor[rgb]{0.8910245585529412,0.9214321063294117,0.9714899216352941} 0.27 & \cellcolor[rgb]{0.61490285,0.649358983,0.8768415764999999} -0.09 & \cellcolor[rgb]{0.888688766427451,0.9204606074745099,0.9730746507960784} 0.10 & \cellcolor[rgb]{0.7500152822588235,0.8192542337882354,0.9905350620529412} 0.06 \\
 & SAR & \cellcolor[rgb]{0.8230564053823529,0.8822182482647059,0.9984342312529412} 0.16 & \cellcolor[rgb]{0.7527113214117647,0.8219973367843137,0.9915787156372549} 0.07 & \cellcolor[rgb]{0.61490285,0.649358983,0.8768415764999999} -0.08 & \cellcolor[rgb]{0.61490285,0.649358983,0.8768415764999999} -0.12 & \cellcolor[rgb]{0.61490285,0.649358983,0.8768415764999999} -0.16 & \cellcolor[rgb]{0.61490285,0.649358983,0.8768415764999999} -0.06 & \cellcolor[rgb]{0.7473192431058824,0.8165111307921569,0.9894914084686275} 0.05 & \cellcolor[rgb]{0.61490285,0.649358983,0.8768415764999999} -0.21 & \cellcolor[rgb]{0.61490285,0.649358983,0.8768415764999999} -0.04 \\
 & CoCoA-MTE & \cellcolor[rgb]{0.8389114905588235,0.8932732186176471,0.9955022941235294} 0.17 & \cellcolor[rgb]{0.9824176791176471,0.8723068372941176,0.8216194376764705} 0.29 & \cellcolor[rgb]{0.852836579,0.50777808,0.575116406} 0.40 & \cellcolor[rgb]{0.852836579,0.50777808,0.575116406} 0.47 & \cellcolor[rgb]{0.9646785691470589,0.756126767372549,0.7003363715784314} 0.29 & \cellcolor[rgb]{0.9708639649117647,0.7732067385098039,0.7148535351862745} 0.55 & \cellcolor[rgb]{0.9460687713941176,0.7126943685490196,0.6666446363803922} 0.48 & \cellcolor[rgb]{0.852836579,0.50777808,0.575116406} 0.53 & \cellcolor[rgb]{0.8979687697431373,0.6209226426705883,0.6104148745666667} 0.40 \\
 & KLE & \cellcolor[rgb]{0.9634414899941177,0.752710773145098,0.6974329388568627} 0.36 & \cellcolor[rgb]{0.9276891842254902,0.9318890695490196,0.9382935886568627} 0.21 & \cellcolor[rgb]{0.974575254145098,0.8953926345333334,0.8554377971509803} 0.21 & \cellcolor[rgb]{0.9819030282176471,0.8170942072647058,0.7568604245764706} 0.31 & \cellcolor[rgb]{0.852836579,0.50777808,0.575116406} 0.40 & \cellcolor[rgb]{0.9528917390058824,0.727592846082353,0.6776679419235294} 0.59 & \cellcolor[rgb]{0.8844643114450981,0.5946066788549019,0.5989826965745098} 0.55 & \cellcolor[rgb]{0.8898725387078432,0.6051525891921569,0.6035518578607844} 0.49 & \cellcolor[rgb]{0.9102005491941176,0.643382456317647,0.6225797599} 0.39 \\
 & TAD & \cellcolor[rgb]{0.852836579,0.50777808,0.575116406} 0.44 & \cellcolor[rgb]{0.852836579,0.50777808,0.575116406} 0.47 & \cellcolor[rgb]{0.9102005491941176,0.643382456317647,0.6225797599} 0.36 & \cellcolor[rgb]{0.864597965917647,0.5433394684235294,0.5836202017019608} 0.46 & \cellcolor[rgb]{0.9704394715,0.9027982014117647,0.8675832782352941} 0.17 & \cellcolor[rgb]{0.852836579,0.50777808,0.575116406} 0.72 & \cellcolor[rgb]{0.9729271893941176,0.7797824782294118,0.7207566595882353} 0.43 & \cellcolor[rgb]{0.9513294646843138,0.7239693735803922,0.6748606867705882} 0.41 & \cellcolor[rgb]{0.852836579,0.50777808,0.575116406} 0.43 \\\midrule
\multirow{2}{*}{\shortstack[l]{Diffusion\\baselines}} & TraceDet & \cellcolor[rgb]{0.7048055483372548,0.7703792543686274,0.9677723612431373} 0.09 & \cellcolor[rgb]{0.7662841187058823,0.8349002989411765,0.9951966350588235} 0.08 & \cellcolor[rgb]{0.9849255762058824,0.8479150297647058,0.7906558870392157} 0.25 & \cellcolor[rgb]{0.9796923648137255,0.8051528802058823,0.7446909254705882} 0.32 & \cellcolor[rgb]{0.9348276152529411,0.6896369097254902,0.6504705511098039} 0.33 & \cellcolor[rgb]{0.9218513348607843,0.6650340647294117,0.635032405109804} 0.64 & \cellcolor[rgb]{0.852836579,0.50777808,0.575116406} 0.58 & \cellcolor[rgb]{0.9459084460607843,0.9259866385215686,0.914542637127451} 0.18 & \cellcolor[rgb]{0.9796923648137255,0.8051528802058823,0.7446909254705882} 0.31 \\
 & DynHD & \cellcolor[rgb]{0.7717200969960785,0.8400012947058824,0.9965252584941177} 0.13 & \cellcolor[rgb]{0.7527113214117647,0.8219973367843137,0.9915787156372549} 0.07 & \cellcolor[rgb]{0.9824176791176471,0.8723068372941176,0.8216194376764705} 0.23 & \cellcolor[rgb]{0.9385746672999999,0.6973227293333333,0.6558619128666667} 0.39 & \cellcolor[rgb]{0.9514243351588236,0.9223978252941176,0.9059849168705882} 0.14 & \cellcolor[rgb]{0.9528917390058824,0.727592846082353,0.6776679419235294} 0.59 & \cellcolor[rgb]{0.9834610525,0.8259843225764706,0.7659985578117647} 0.39 & \cellcolor[rgb]{0.953077067017647,0.9210455325882353,0.9030752965411765} 0.19 & \cellcolor[rgb]{0.9837721488176471,0.8654248580941176,0.812342739117647} 0.27 \\\midrule
\multirow{4}{*}{\shortstack[l]{Diffusion\\signals}} & MCNLL & \cellcolor[rgb]{0.7880256462666666,0.8543899393333334,0.9988772960000001} 0.14 & \cellcolor[rgb]{0.6996157421568627,0.7642642360784314,0.9642295513921568} 0.03 & \cellcolor[rgb]{0.974575254145098,0.8953926345333334,0.8554377971509803} 0.21 & \cellcolor[rgb]{0.9818859091411765,0.8745427552862746,0.824710021645098} 0.26 & \cellcolor[rgb]{0.8492270432274511,0.8997249420568627,0.9922887283509804} 0.04 & \cellcolor[rgb]{0.9834812190705882,0.867835001309804,0.8154382697392157} 0.45 & \cellcolor[rgb]{0.6402749751176471,0.6867115845411765,0.909005371654902} -0.06 & \cellcolor[rgb]{0.9790880158705882,0.8856170452000001,0.8401505184058824} 0.25 & \cellcolor[rgb]{0.8910245585529412,0.9214321063294117,0.9714899216352941} 0.16 \\
 & NFE & \cellcolor[rgb]{0.7554121621254901,0.8246983074117646,0.9925393881882353} 0.12 & \cellcolor[rgb]{0.6867762156470588,0.7487493527058824,0.9547336847647059} 0.02 & \cellcolor[rgb]{0.9438760079745099,0.9270202487490196,0.9173357361078431} 0.17 & \cellcolor[rgb]{0.9845960255235294,0.8529178378588235,0.7968521304901961} 0.28 & \cellcolor[rgb]{0.9575785619705882,0.7384632635882353,0.6860897073823529} 0.30 & \cellcolor[rgb]{0.9834610525,0.8259843225764706,0.7659985578117647} 0.50 & \cellcolor[rgb]{0.6867762156470588,0.7487493527058824,0.9547336847647059} -0.01 & \cellcolor[rgb]{0.9839369241588236,0.8629234540470588,0.8092446173921568} 0.28 & \cellcolor[rgb]{0.9497716033000001,0.923750118,0.9088945372} 0.21 \\
 & AD-Full & \cellcolor[rgb]{0.6426363887647059,0.690064711317647,0.9117342756235294} 0.05 & \cellcolor[rgb]{0.7258692802352942,0.794090494117647,0.9801006337941176} 0.05 & \cellcolor[rgb]{0.9691631781666668,0.9044582760156863,0.8705807575137254} 0.20 & \cellcolor[rgb]{0.9717157648333333,0.9011381268078431,0.8645857989568628} 0.23 & \cellcolor[rgb]{0.9813503623666666,0.8141088755,0.7538180498} 0.25 & \cellcolor[rgb]{0.9196757213862745,0.930583412827451,0.9472468817078432} 0.31 & \cellcolor[rgb]{0.8096589725941177,0.8720603673823529,0.9994654594098039} 0.11 & \cellcolor[rgb]{0.7798733627843137,0.8473838640392157,0.9980376545882352} -0.02 & \cellcolor[rgb]{0.8792694128431373,0.9163932970294117,0.9792039273627451} 0.15 \\
 & CommitNLL & \cellcolor[rgb]{0.61490285,0.649358983,0.8768415764999999} 0.03 & \cellcolor[rgb]{0.9754778064901961,0.8934375166666666,0.8523803414019608} 0.27 & \cellcolor[rgb]{0.9683898066058824,0.766374750054902,0.7090466697431372} 0.30 & \cellcolor[rgb]{0.9756268974411765,0.7893996947941176,0.729703902882353} 0.33 & \cellcolor[rgb]{0.9696268857588235,0.769790744282353,0.7119501024647059} 0.28 & \cellcolor[rgb]{0.9834812190705882,0.867835001309804,0.8154382697392157} 0.45 & \cellcolor[rgb]{0.9218513348607843,0.6650340647294117,0.635032405109804} 0.51 & \cellcolor[rgb]{0.9790880158705882,0.8856170452000001,0.8401505184058824} 0.25 & \cellcolor[rgb]{0.9824556940686274,0.8200795390294118,0.7599027993529412} 0.30 \\\midrule
\multirow{2}{*}{\shortstack[l]{AR +\\MCNLL}} & CoCoA-MCNLL & \cellcolor[rgb]{0.9133921582352942,0.9291026777686274,0.9534763223137255} 0.22 & \cellcolor[rgb]{0.806966326382353,0.8699614911470589,0.9995711860294118} 0.11 & \cellcolor[rgb]{0.9348276152529411,0.6896369097254902,0.6504705511098039} 0.34 & \cellcolor[rgb]{0.9102005491941176,0.643382456317647,0.6225797599} 0.42 & \cellcolor[rgb]{0.9747269947588235,0.7861939559392157,0.7267214884509804} 0.27 & \cellcolor[rgb]{0.9634414899941177,0.752710773145098,0.6974329388568627} 0.57 & \cellcolor[rgb]{0.8952807659705883,0.6156984995294118,0.6081210191470588} 0.54 & \cellcolor[rgb]{0.9326956664686274,0.6855638360235294,0.6478844782078431} 0.44 & \cellcolor[rgb]{0.9460687713941176,0.7126943685490196,0.6666446363803922} 0.36 \\
 & CoCoA-MCNLL-Norm & \cellcolor[rgb]{0.9256858185156862,0.9315626553686274,0.9405319119196078} 0.23 & \cellcolor[rgb]{0.9276891842254902,0.9318890695490196,0.9382935886568627} 0.21 & \cellcolor[rgb]{0.8979687697431373,0.6209226426705883,0.6104148745666667} 0.37 & \cellcolor[rgb]{0.864597965917647,0.5433394684235294,0.5836202017019608} 0.46 & \cellcolor[rgb]{0.9236824528058825,0.9312362411882353,0.9427702351823529} 0.11 & \cellcolor[rgb]{0.9747269947588235,0.7861939559392157,0.7267214884509804} 0.54 & \cellcolor[rgb]{0.8952807659705883,0.6156984995294118,0.6081210191470588} 0.54 & \cellcolor[rgb]{0.9077541933039216,0.6388904935882354,0.6201467828333334} 0.47 & \cellcolor[rgb]{0.9348276152529411,0.6896369097254902,0.6504705511098039} 0.37 \\\midrule
\multirow{2}{*}{\shortstack[l]{\textbf{D-CoCoA}\\\textbf{(ours)}}} & D-CoCoA-G & \cellcolor[rgb]{0.8863529743019607,0.9194891086196079,0.9746593799568628} 0.20 & \cellcolor[rgb]{0.7527113214117647,0.8219973367843137,0.9915787156372549} 0.07 & \cellcolor[rgb]{0.9683898066058824,0.766374750054902,0.7090466697431372} 0.30 & \cellcolor[rgb]{0.947942297417647,0.7165372783529411,0.6693403172588235} 0.38 & \cellcolor[rgb]{0.9839369241588236,0.8629234540470588,0.8092446173921568} 0.21 & \cellcolor[rgb]{0.9796923648137255,0.8051528802058823,0.7446909254705882} 0.52 & \cellcolor[rgb]{0.8096589725941177,0.8720603673823529,0.9994654594098039} 0.11 & \cellcolor[rgb]{0.9829494490941176,0.8700709193019608,0.8185288537078431} 0.27 & \cellcolor[rgb]{0.9813541391647058,0.8767786732784314,0.8278006056137255} 0.26 \\
 & D-CoCoA-L & \cellcolor[rgb]{0.6286169883529411,0.6698302759882353,0.8948307002647058} 0.04 & \cellcolor[rgb]{0.9176723556764705,0.9302569986470588,0.9494852049705882} 0.20 & \cellcolor[rgb]{0.9622044108411765,0.7492947789176471,0.6945295061352941} 0.31 & \cellcolor[rgb]{0.9607031106137255,0.7457102085921569,0.6917042176882353} 0.36 & \cellcolor[rgb]{0.9423217193470588,0.7050085489411765,0.6612532746235295} 0.32 & \cellcolor[rgb]{0.9813503623666666,0.8141088755,0.7538180498} 0.51 & \cellcolor[rgb]{0.9305268001470588,0.6814578817647059,0.6453140635882353} 0.50 & \cellcolor[rgb]{0.9678868848333333,0.9061183506196078,0.8735782367921568} 0.22 & \cellcolor[rgb]{0.9796923648137255,0.8051528802058823,0.7446909254705882} 0.31 \\
\bottomrule
\end{tabular}}\caption{\label{tab:refusion_detailed_main_results} \textbf{D-CoCoA comparison.} PRR$\uparrow$ for the ReFusion model across various tasks for D-CoCoA methods, comparing the best method from each category (see Tables~\ref{tab:detailed_main_results}, \ref{tab:dream_detailed_main_results}, and~\ref{tab:diffusiongemma_detailed_main_results} for the LLaDA-1.5, Dream, and DiffusionGemma models, respectively). Warmer color indicates better results.}\end{table*}

\clearpage
\subsection{Results for DiffusionGemma}
\label{app:res_diffusiongemma}
\begin{figure*}[h!]
    \centering
    \includegraphics[trim={0.cm 0.cm 0.cm 0.cm},clip,width=0.99\linewidth]{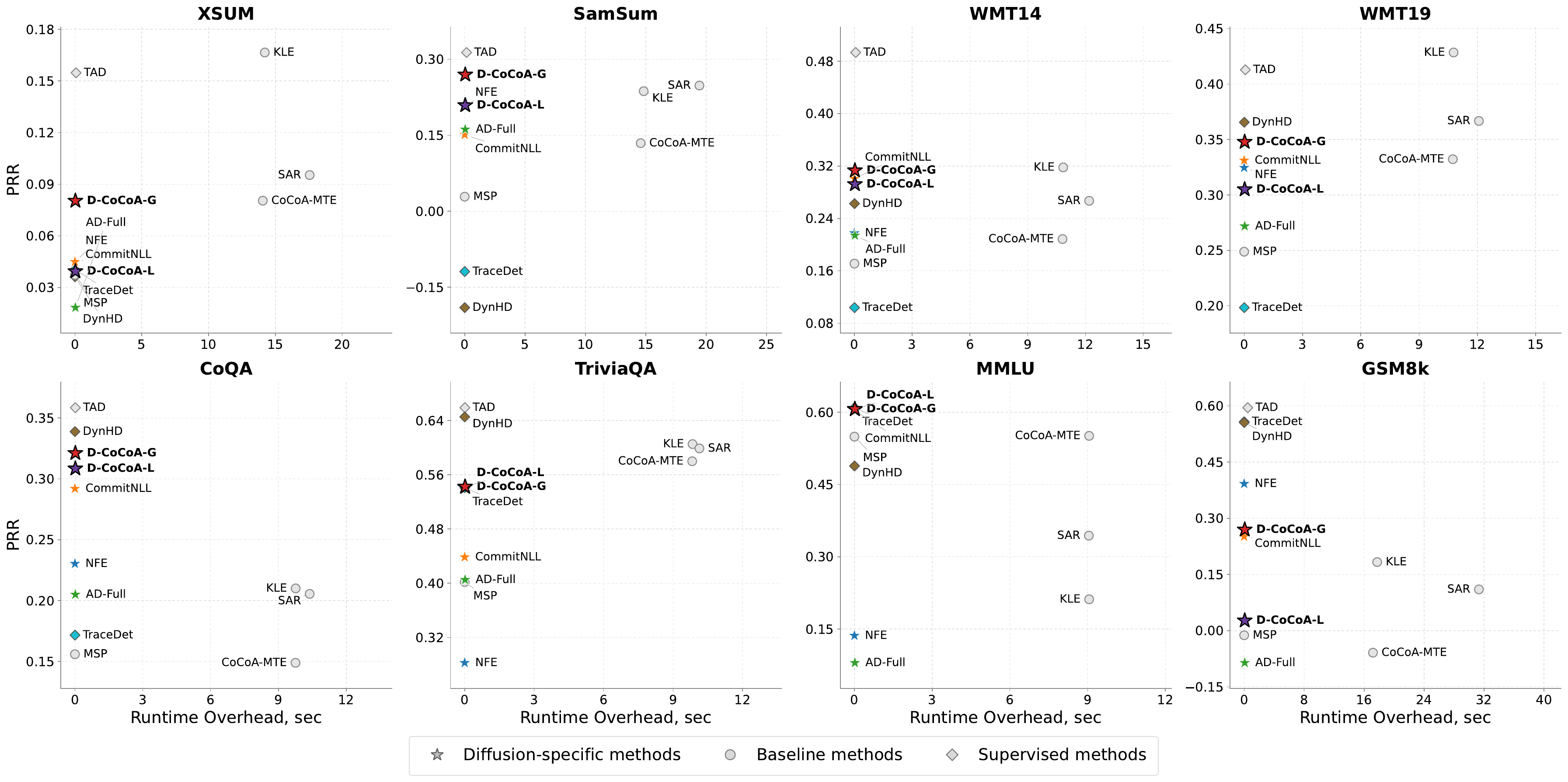}
    \caption{
    \textbf{DiffusionGemma performance and runtime.} Comparison of uncertainty quantification performance (PRR$\uparrow$) versus computational complexity (Time in seconds$\downarrow$) across eight benchmark datasets for the DiffusionGemma model (see Figures~\ref{fig:llada_prr_vs_comp}, \ref{fig:dream_prr_vs_comp}, and~\ref{fig:refusion_prr_vs_comp} for the LLaDA-1.5, Dream, and ReFusion models, respectively).
    }
    \label{fig:diffusiongemma_prr_vs_comp}
  \end{figure*}

  Tables~\ref{tab:diffusiongemma_ar_baselines_results}, \ref{tab:diffusiongemma_diff_baselines_results}, \ref{tab:diffusiongemma_ar_traj_results}, and~\ref{tab:diffusiongemma_detailed_main_results} present the results for DiffusionGemma with baselines from standard autoregressive LLMs, diffusion-specific methods, sampling-based baselines using denoising trajectory states, and the best method from each category, respectively. Figure~\ref{fig:diffusiongemma_prr_vs_comp} presents the performance-efficiency trade-off for each dataset. Since DiffusionGemma has no mask token, MCNLL and the methods based on it are not applicable to this model.

  \begin{table*}[!ht] \resizebox{\textwidth}{!}{\begin{tabular}{ll|c|c|c|c|c|c|c|c|c}
\toprule
\textbf{Category} & \textbf{UQ Method} & \textbf{XSum} & \textbf{SamSum} & \textbf{WMT14} & \textbf{WMT19} & \textbf{CoQA} & \textbf{TriviaQA} & \textbf{MMLU} & \textbf{GSM8k} & \textbf{Mean} \\
\midrule
\multirow{6}{*}{\shortstack[l]{Information-\\based}} & MSP & \cellcolor[rgb]{0.8096589725941177,0.8720603673823529,0.9994654594098039} 0.04 & \cellcolor[rgb]{0.7853082184392157,0.8520545809019608,0.9985974155294117} 0.03 & \cellcolor[rgb]{0.6918310328862745,0.7550917086431372,0.9589153366156863} 0.17 & \cellcolor[rgb]{0.7446232039529412,0.8137680277960784,0.9884477548843138} 0.25 & \cellcolor[rgb]{0.9024823794117647,0.9258330802784314,0.9630825372156863} 0.16 & \cellcolor[rgb]{0.9418435698882353,0.9280538589764706,0.9201288350882353} 0.40 & \cellcolor[rgb]{0.8898725387078432,0.6051525891921569,0.6035518578607844} 0.55 & \cellcolor[rgb]{0.6867762156470588,0.7487493527058824,0.9547336847647059} -0.01 & \cellcolor[rgb]{0.7258692802352942,0.794090494117647,0.9801006337941176} 0.20 \\
 & MTE & \cellcolor[rgb]{0.7880256462666666,0.8543899393333334,0.9988772960000001} 0.03 & \cellcolor[rgb]{0.7311771982901961,0.7999150558117647,0.9829285958960784} -0.00 & \cellcolor[rgb]{0.61490285,0.649358983,0.8768415764999999} 0.11 & \cellcolor[rgb]{0.6545301595333333,0.7067491361333333,0.9250638169333334} 0.19 & \cellcolor[rgb]{0.7179074031529412,0.7853536515764705,0.9758586906411765} 0.06 & \cellcolor[rgb]{0.9514243351588236,0.9223978252941176,0.9059849168705882} 0.41 & \cellcolor[rgb]{0.9004151256333333,0.6254146054,0.6128478516333333} 0.54 & \cellcolor[rgb]{0.61490285,0.649358983,0.8768415764999999} -0.11 & \cellcolor[rgb]{0.61490285,0.649358983,0.8768415764999999} 0.15 \\
 & Perplexity & \cellcolor[rgb]{0.7880256462666666,0.8543899393333334,0.9988772960000001} 0.03 & \cellcolor[rgb]{0.6944259359764706,0.7581492177882353,0.9606867415411765} -0.02 & \cellcolor[rgb]{0.626331298627451,0.6664183938235294,0.8918325129705882} 0.12 & \cellcolor[rgb]{0.6691882557215687,0.7264093044156863,0.9396585384392157} 0.20 & \cellcolor[rgb]{0.7179074031529412,0.7853536515764705,0.9758586906411765} 0.06 & \cellcolor[rgb]{0.9133921582352942,0.9291026777686274,0.9534763223137255} 0.37 & \cellcolor[rgb]{0.8898725387078432,0.6051525891921569,0.6035518578607844} 0.55 & \cellcolor[rgb]{0.6426363887647059,0.690064711317647,0.9117342756235294} -0.07 & \cellcolor[rgb]{0.61490285,0.649358983,0.8768415764999999} 0.15 \\
 & AttScore & \cellcolor[rgb]{0.9004151256333333,0.6254146054,0.6128478516333333} 0.23 & \cellcolor[rgb]{0.9837721488176471,0.8654248580941176,0.812342739117647} 0.18 & \cellcolor[rgb]{0.6520871435019608,0.7034724414196079,0.9226313633490196} 0.14 & \cellcolor[rgb]{0.61490285,0.649358983,0.8768415764999999} 0.16 & \cellcolor[rgb]{0.6817303976705882,0.7423918409254902,0.9505094434470589} 0.04 & \cellcolor[rgb]{0.61490285,0.649358983,0.8768415764999999} 0.12 & \cellcolor[rgb]{0.6545301595333333,0.7067491361333333,0.9250638169333334} 0.05 & \cellcolor[rgb]{0.9544540133274511,0.7312163185843137,0.6804751970764706} 0.58 & \cellcolor[rgb]{0.7022106452470588,0.7673217452235295,0.966000956317647} 0.19 \\
 & CCP & \cellcolor[rgb]{0.876805309,0.9151164254999999,0.9804355785000001} 0.07 & \cellcolor[rgb]{0.7853082184392157,0.8520545809019608,0.9985974155294117} 0.03 & \cellcolor[rgb]{0.6643022236588235,0.7198559149882353,0.9347936312705882} 0.15 & \cellcolor[rgb]{0.6691882557215687,0.7264093044156863,0.9396585384392157} 0.20 & \cellcolor[rgb]{0.9596879944529412,0.9156363617647059,0.8914368152235295} 0.20 & \cellcolor[rgb]{0.9720272867117647,0.7765767393745098,0.7177742451568627} 0.54 & \cellcolor[rgb]{0.8763519705509804,0.5787878133490196,0.5921289546450981} 0.56 & \cellcolor[rgb]{0.6944259359764706,0.7581492177882353,0.9606867415411765} -0.00 & \cellcolor[rgb]{0.7744380861411764,0.8425517925882353,0.9971895702117648} 0.22 \\
 & RAUQ & \cellcolor[rgb]{0.61490285,0.649358983,0.8768415764999999} -0.05 & \cellcolor[rgb]{0.8743412051568626,0.9138395539705882,0.9816672296372548} 0.08 & \cellcolor[rgb]{0.6379135614705882,0.6833584577647058,0.9062764676862745} 0.13 & \cellcolor[rgb]{0.6402749751176471,0.6867115845411765,0.909005371654902} 0.18 & \cellcolor[rgb]{0.9024823794117647,0.9258330802784314,0.9630825372156863} 0.16 & \cellcolor[rgb]{0.8336264621666667,0.8895882285,0.9964796065} 0.30 & \cellcolor[rgb]{0.8898725387078432,0.6051525891921569,0.6035518578607844} 0.55 & \cellcolor[rgb]{0.9236824528058825,0.9312362411882353,0.9427702351823529} 0.29 & \cellcolor[rgb]{0.7500152822588235,0.8192542337882354,0.9905350620529412} 0.21 \\ \midrule
\multirow{6}{*}{\shortstack[l]{Sampling,\\white-box}} & MCNSE & \cellcolor[rgb]{0.7179074031529412,0.7853536515764705,0.9758586906411765} 0.00 & \cellcolor[rgb]{0.7853082184392157,0.8520545809019608,0.9985974155294117} 0.03 & \cellcolor[rgb]{0.679207488682353,0.7392130850352943,0.9483973227882354} 0.16 & \cellcolor[rgb]{0.6691882557215687,0.7264093044156863,0.9396585384392157} 0.20 & \cellcolor[rgb]{0.7744380861411764,0.8425517925882353,0.9971895702117648} 0.09 & \cellcolor[rgb]{0.9763803588352942,0.8914823988,0.8493228856529411} 0.45 & \cellcolor[rgb]{0.8763519705509804,0.5787878133490196,0.5921289546450981} 0.56 & \cellcolor[rgb]{0.6643022236588235,0.7198559149882353,0.9347936312705882} -0.04 & \cellcolor[rgb]{0.679207488682353,0.7392130850352943,0.9483973227882354} 0.18 \\
 & SemanticEntropy & \cellcolor[rgb]{0.9736727018,0.8973477524,0.8584952529000001} 0.13 & \cellcolor[rgb]{0.9849255762058824,0.8479150297647058,0.7906558870392157} 0.19 & \cellcolor[rgb]{0.7635661295607843,0.8323498010588236,0.9945323233411765} 0.22 & \cellcolor[rgb]{0.8816813900509803,0.917546110909804,0.9778288382784314} 0.34 & \cellcolor[rgb]{0.9830083599196078,0.8230648707941176,0.7629451741294118} 0.26 & \cellcolor[rgb]{0.9261890675039215,0.6732459732470588,0.6401732343490196} 0.60 & \cellcolor[rgb]{0.852836579,0.50777808,0.575116406} 0.58 & \cellcolor[rgb]{0.7527113214117647,0.8219973367843137,0.9915787156372549} 0.07 & \cellcolor[rgb]{0.9514243351588236,0.9223978252941176,0.9059849168705882} 0.30 \\
 & SAR & \cellcolor[rgb]{0.9337138175431372,0.9321882998862745,0.9313012310098039} 0.10 & \cellcolor[rgb]{0.9497671903627452,0.7203459010784313,0.6720534316176471} 0.25 & \cellcolor[rgb]{0.8389114905588235,0.8932732186176471,0.9955022941235294} 0.27 & \cellcolor[rgb]{0.9216790870960785,0.9309098270078431,0.945008558445098} 0.37 & \cellcolor[rgb]{0.9691631781666668,0.9044582760156863,0.8705807575137254} 0.21 & \cellcolor[rgb]{0.9261890675039215,0.6732459732470588,0.6401732343490196} 0.60 & \cellcolor[rgb]{0.9678868848333333,0.9061183506196078,0.8735782367921568} 0.34 & \cellcolor[rgb]{0.7853082184392157,0.8520545809019608,0.9985974155294117} 0.11 & \cellcolor[rgb]{0.9133921582352942,0.9291026777686274,0.9534763223137255} 0.28 \\
 & EigenScore & \cellcolor[rgb]{0.8956961428039216,0.9233751040392157,0.9683204633137255} 0.08 & \cellcolor[rgb]{0.9607031106137255,0.7457102085921569,0.6917042176882353} 0.24 & \cellcolor[rgb]{0.8230564053823529,0.8822182482647059,0.9984342312529412} 0.26 & \cellcolor[rgb]{0.8956961428039216,0.9233751040392157,0.9683204633137255} 0.35 & \cellcolor[rgb]{0.9783266054882354,0.7990169113588235,0.7386511461764707} 0.27 & \cellcolor[rgb]{0.9802905992117648,0.8812505092627452,0.8339817735509804} 0.46 & \cellcolor[rgb]{0.7853082184392157,0.8520545809019608,0.9985974155294117} 0.16 & \cellcolor[rgb]{0.852836579,0.50777808,0.575116406} 0.72 & \cellcolor[rgb]{0.9754778064901961,0.8934375166666666,0.8523803414019608} 0.32 \\
 & CoCoA-MTE & \cellcolor[rgb]{0.8956961428039216,0.9233751040392157,0.9683204633137255} 0.08 & \cellcolor[rgb]{0.9459084460607843,0.9259866385215686,0.914542637127451} 0.13 & \cellcolor[rgb]{0.7500152822588235,0.8192542337882354,0.9905350620529412} 0.21 & \cellcolor[rgb]{0.866948893627451,0.9100089393823529,0.9853621830490196} 0.33 & \cellcolor[rgb]{0.8863529743019607,0.9194891086196079,0.9746593799568628} 0.15 & \cellcolor[rgb]{0.9441952453705882,0.708851458745098,0.6639489555019608} 0.58 & \cellcolor[rgb]{0.8898725387078432,0.6051525891921569,0.6035518578607844} 0.55 & \cellcolor[rgb]{0.6497206297058824,0.7001240916470588,0.9199209875294118} -0.06 & \cellcolor[rgb]{0.8492270432274511,0.8997249420568627,0.9922887283509804} 0.25 \\
 & SemanticDensity & \cellcolor[rgb]{0.9528917390058824,0.727592846082353,0.6776679419235294} 0.20 & \cellcolor[rgb]{0.9756268974411765,0.7893996947941176,0.729703902882353} 0.22 & \cellcolor[rgb]{0.9377786937156862,0.9301210794313726,0.9257150330490196} 0.35 & \cellcolor[rgb]{0.9791396989627451,0.8021675484411765,0.7416485506941177} 0.49 & \cellcolor[rgb]{0.8694129974705882,0.9112858109117647,0.9841305319117647} 0.14 & \cellcolor[rgb]{0.9261890675039215,0.6732459732470588,0.6401732343490196} 0.60 & \cellcolor[rgb]{0.61490285,0.649358983,0.8768415764999999} 0.01 & \cellcolor[rgb]{0.9580352625941178,0.9169886544705883,0.8943464355529411} 0.35 & \cellcolor[rgb]{0.9514243351588236,0.9223978252941176,0.9059849168705882} 0.30 \\\midrule
\multirow{3}{*}{\shortstack[l]{Sampling,\\black-box}} & LUQ & \cellcolor[rgb]{0.9640580048333334,0.9110985744313725,0.882570674627451} 0.12 & \cellcolor[rgb]{0.9807976965156863,0.8111235437352942,0.7507756750235295} 0.21 & \cellcolor[rgb]{0.915574114,0.9297565972666666,0.9515550793333334} 0.33 & \cellcolor[rgb]{0.9547297988764706,0.919693239882353,0.9001656762117647} 0.40 & \cellcolor[rgb]{0.9691631781666668,0.9044582760156863,0.8705807575137254} 0.21 & \cellcolor[rgb]{0.9261890675039215,0.6732459732470588,0.6401732343490196} 0.60 & \cellcolor[rgb]{0.7961779297490197,0.861396014627451,0.9997169374117647} 0.17 & \cellcolor[rgb]{0.9112102024705881,0.9284487582705883,0.9553975652941177} 0.27 & \cellcolor[rgb]{0.931695915645098,0.9325418979098039,0.9338169421313726} 0.29 \\
 & KLE & \cellcolor[rgb]{0.9813503623666666,0.8141088755,0.7538180498} 0.17 & \cellcolor[rgb]{0.9607031106137255,0.7457102085921569,0.6917042176882353} 0.24 & \cellcolor[rgb]{0.9046643351764706,0.9264869997764706,0.9611612942352941} 0.32 & \cellcolor[rgb]{0.9772829111803922,0.8895272809333333,0.8462654299039215} 0.43 & \cellcolor[rgb]{0.9691631781666668,0.9044582760156863,0.8705807575137254} 0.21 & \cellcolor[rgb]{0.9150932609745098,0.6523663817764705,0.6274457140333334} 0.61 & \cellcolor[rgb]{0.8440942415960784,0.8965891896490196,0.9940190521784313} 0.21 & \cellcolor[rgb]{0.8440942415960784,0.8965891896490196,0.9940190521784313} 0.18 & \cellcolor[rgb]{0.9514243351588236,0.9223978252941176,0.9059849168705882} 0.30 \\
 & DegMat & \cellcolor[rgb]{0.9841016995,0.8604220499999999,0.8061464956666666} 0.15 & \cellcolor[rgb]{0.9683898066058824,0.766374750054902,0.7090466697431372} 0.23 & \cellcolor[rgb]{0.8910245585529412,0.9214321063294117,0.9714899216352941} 0.31 & \cellcolor[rgb]{0.9772829111803922,0.8895272809333333,0.8462654299039215} 0.43 & \cellcolor[rgb]{0.9596879944529412,0.9156363617647059,0.8914368152235295} 0.20 & \cellcolor[rgb]{0.9261890675039215,0.6732459732470588,0.6401732343490196} 0.60 & \cellcolor[rgb]{0.8440942415960784,0.8965891896490196,0.9940190521784313} 0.21 & \cellcolor[rgb]{0.8440942415960784,0.8965891896490196,0.9940190521784313} 0.18 & \cellcolor[rgb]{0.931695915645098,0.9325418979098039,0.9338169421313726} 0.29 \\\midrule
\multirow{4}{*}{Supervised} & SAPLMA & \cellcolor[rgb]{0.9196824685392158,0.6609281104705882,0.632461990490196} 0.22 & \cellcolor[rgb]{0.852836579,0.50777808,0.575116406} 0.31 & \cellcolor[rgb]{0.8979687697431373,0.6209226426705883,0.6104148745666667} 0.55 & \cellcolor[rgb]{0.852836579,0.50777808,0.575116406} 0.60 & \cellcolor[rgb]{0.61490285,0.649358983,0.8768415764999999} 0.00 & \cellcolor[rgb]{0.9671527274529412,0.762958755827451,0.7061432370215687} 0.55 & \cellcolor[rgb]{0.9459084460607843,0.9259866385215686,0.914542637127451} 0.31 & \cellcolor[rgb]{0.9367011412764705,0.6934798195294117,0.6531662319882353} 0.61 & \cellcolor[rgb]{0.9423217193470588,0.7050085489411765,0.6612532746235295} 0.39 \\
 & SATRMD & \cellcolor[rgb]{0.915574114,0.9297565972666666,0.9515550793333334} 0.09 & \cellcolor[rgb]{0.9802905992117648,0.8812505092627452,0.8339817735509804} 0.17 & \cellcolor[rgb]{0.8389114905588235,0.8932732186176471,0.9955022941235294} 0.27 & \cellcolor[rgb]{0.7446232039529412,0.8137680277960784,0.9884477548843138} 0.25 & \cellcolor[rgb]{0.9830083599196078,0.8230648707941176,0.7629451741294118} 0.26 & \cellcolor[rgb]{0.9528917390058824,0.727592846082353,0.6776679419235294} 0.57 & \cellcolor[rgb]{0.9196757213862745,0.930583412827451,0.9472468817078432} 0.28 & \cellcolor[rgb]{0.8871684250764706,0.5998796340235294,0.6012672772176471} 0.68 & \cellcolor[rgb]{0.9754778064901961,0.8934375166666666,0.8523803414019608} 0.32 \\
 & LookBackLens & \cellcolor[rgb]{0.852836579,0.50777808,0.575116406} 0.25 & \cellcolor[rgb]{0.61490285,0.649358983,0.8768415764999999} -0.07 & \cellcolor[rgb]{0.852836579,0.50777808,0.575116406} 0.58 & \cellcolor[rgb]{0.8675383126470588,0.5522298155294117,0.585746150627451} 0.59 & \cellcolor[rgb]{0.8925766523392157,0.6104255443607843,0.6058364385039215} 0.34 & \cellcolor[rgb]{0.864597965917647,0.5433394684235294,0.5836202017019608} 0.65 & \cellcolor[rgb]{0.6766845796941177,0.736034329145098,0.9462852021294117} 0.07 & \cellcolor[rgb]{0.924020201182353,0.6691400189882353,0.6376028197294118} 0.63 & \cellcolor[rgb]{0.9575785619705882,0.7384632635882353,0.6860897073823529} 0.38 \\
 & TAD & \cellcolor[rgb]{0.9841016995,0.8604220499999999,0.8061464956666666} 0.15 & \cellcolor[rgb]{0.852836579,0.50777808,0.575116406} 0.31 & \cellcolor[rgb]{0.9634414899941177,0.752710773145098,0.6974329388568627} 0.49 & \cellcolor[rgb]{0.9640580048333334,0.9110985744313725,0.882570674627451} 0.41 & \cellcolor[rgb]{0.852836579,0.50777808,0.575116406} 0.36 & \cellcolor[rgb]{0.852836579,0.50777808,0.575116406} 0.66 & - & \cellcolor[rgb]{0.9441952453705882,0.708851458745098,0.6639489555019608} 0.60 & \cellcolor[rgb]{0.852836579,0.50777808,0.575116406} 0.43 \\
\bottomrule
\end{tabular}
}\caption{\label{tab:diffusiongemma_ar_baselines_results} \textbf{AR baselines.} PRR$\uparrow$ for the DiffusionGemma model across various tasks for baselines from standard autoregressive LLMs (see Tables~\ref{tab:ar_baselines_results}, \ref{tab:dream_ar_baselines_results}, and~\ref{tab:refusion_ar_baselines_results} for the LLaDA-1.5, Dream, and ReFusion models, respectively). Warmer color indicates better results.}\end{table*}
  \begin{table*}[!ht] \resizebox{\textwidth}{!}{\begin{tabular}{ll|c|c|c|c|c|c|c|c|c}
\toprule
\textbf{Category} & \textbf{UQ Method} & \textbf{XSum} & \textbf{SamSum} & \textbf{WMT14} & \textbf{WMT19} & \textbf{CoQA} & \textbf{TriviaQA} & \textbf{MMLU} & \textbf{GSM8k} & \textbf{Mean} \\
\midrule
\multirow{6}{*}{\shortstack[l]{Diffusion\\statistics}} & NFE & \cellcolor[rgb]{0.8015810339588235,0.8657637386764706,0.9997826392686274} 0.04 & \cellcolor[rgb]{0.9808223691882353,0.8790145912705882,0.830891189582353} 0.21 & \cellcolor[rgb]{0.6741616707058824,0.7328555732549019,0.9441730814705882} 0.22 & \cellcolor[rgb]{0.915574114,0.9297565972666666,0.9515550793333334} 0.32 & \cellcolor[rgb]{0.7635661295607843,0.8323498010588236,0.9945323233411765} 0.23 & \cellcolor[rgb]{0.61490285,0.649358983,0.8768415764999999} 0.28 & \cellcolor[rgb]{0.6944259359764706,0.7581492177882353,0.9606867415411765} 0.14 & \cellcolor[rgb]{0.9802450306647059,0.8081382119705882,0.7477333002470588} 0.39 & \cellcolor[rgb]{0.8569262456745098,0.9044285706686275,0.989693242609804} 0.23 \\
 & Remask & \cellcolor[rgb]{0.9836582578333333,0.8287354144039216,0.7690800753647059} 0.07 & \cellcolor[rgb]{0.7907430740941177,0.8567252977647059,0.9991571764705882} 0.17 & \cellcolor[rgb]{0.6741616707058824,0.7328555732549019,0.9441730814705882} 0.22 & \cellcolor[rgb]{0.7419271648,0.8110249248,0.9874041013} 0.28 & \cellcolor[rgb]{0.9133921582352942,0.9291026777686274,0.9534763223137255} 0.26 & \cellcolor[rgb]{0.6473592160588235,0.6967709648705882,0.9171920835607843} 0.30 & \cellcolor[rgb]{0.6569731755647059,0.7100258308470588,0.927496270517647} 0.11 & \cellcolor[rgb]{0.852836579,0.50777808,0.575116406} 0.56 & \cellcolor[rgb]{0.9196757213862745,0.930583412827451,0.9472468817078432} 0.25 \\
 & FlipCount & \cellcolor[rgb]{0.61490285,0.649358983,0.8768415764999999} 0.02 & \cellcolor[rgb]{0.9591408362921569,0.742086736090196,0.6888969625352941} 0.23 & \cellcolor[rgb]{0.9842498738333334,0.8369886898862745,0.7783246280235294} 0.28 & \cellcolor[rgb]{0.61490285,0.649358983,0.8768415764999999} 0.25 & \cellcolor[rgb]{0.9514243351588236,0.9223978252941176,0.9059849168705882} 0.27 & \cellcolor[rgb]{0.876805309,0.9151164254999999,0.9804355785000001} 0.42 & \cellcolor[rgb]{0.61490285,0.649358983,0.8768415764999999} 0.07 & \cellcolor[rgb]{0.9824556940686274,0.8200795390294118,0.7599027993529412} 0.38 & \cellcolor[rgb]{0.888688766427451,0.9204606074745099,0.9730746507960784} 0.24 \\
\cmidrule(lr){2-11}
 & TrajNLL & \cellcolor[rgb]{0.852836579,0.50777808,0.575116406} 0.09 & \cellcolor[rgb]{0.7285232392627451,0.797002774964706,0.9815146148450979} 0.16 & \cellcolor[rgb]{0.876805309,0.9151164254999999,0.9804355785000001} 0.25 & \cellcolor[rgb]{0.6970208390666667,0.7612067269333334,0.9624581464666666} 0.27 & \cellcolor[rgb]{0.9846442845,0.8424908735411765,0.7844876631294118} 0.29 & \cellcolor[rgb]{0.9836582578333333,0.8287354144039216,0.7690800753647059} 0.53 & \cellcolor[rgb]{0.852836579,0.50777808,0.575116406} 0.61 & \cellcolor[rgb]{0.8616576191882352,0.5344491213176471,0.5814942527764706} 0.55 & \cellcolor[rgb]{0.8898725387078432,0.6051525891921569,0.6035518578607844} 0.34 \\
 & TrajEntropy & \cellcolor[rgb]{0.7022106452470588,0.7673217452235295,0.966000956317647} 0.03 & \cellcolor[rgb]{0.61490285,0.649358983,0.8768415764999999} 0.14 & \cellcolor[rgb]{0.852836579,0.50777808,0.575116406} 0.31 & \cellcolor[rgb]{0.9497716033000001,0.923750118,0.9088945372} 0.33 & \cellcolor[rgb]{0.852836579,0.50777808,0.575116406} 0.33 & \cellcolor[rgb]{0.852836579,0.50777808,0.575116406} 0.63 & \cellcolor[rgb]{0.852836579,0.50777808,0.575116406} 0.61 & \cellcolor[rgb]{0.9747269947588235,0.7861939559392157,0.7267214884509804} 0.41 & \cellcolor[rgb]{0.852836579,0.50777808,0.575116406} 0.35 \\
 & CommitNLL & \cellcolor[rgb]{0.8015810339588235,0.8657637386764706,0.9997826392686274} 0.04 & \cellcolor[rgb]{0.6691882557215687,0.7264093044156863,0.9396585384392157} 0.15 & \cellcolor[rgb]{0.9196824685392158,0.6609281104705882,0.632461990490196} 0.30 & \cellcolor[rgb]{0.9497716033000001,0.923750118,0.9088945372} 0.33 & \cellcolor[rgb]{0.9846442845,0.8424908735411765,0.7844876631294118} 0.29 & \cellcolor[rgb]{0.9112102024705881,0.9284487582705883,0.9553975652941177} 0.44 & \cellcolor[rgb]{0.864597965917647,0.5433394684235294,0.5836202017019608} 0.60 & \cellcolor[rgb]{0.9438760079745099,0.9270202487490196,0.9173357361078431} 0.25 & \cellcolor[rgb]{0.9807976965156863,0.8111235437352942,0.7507756750235295} 0.30 \\
\midrule
\multirow{8}{*}{\shortstack[l]{Trajectory\\dissimilarity}} & AD-Block & \cellcolor[rgb]{0.61490285,0.649358983,0.8768415764999999} 0.02 & \cellcolor[rgb]{0.7285232392627451,0.797002774964706,0.9815146148450979} 0.16 & \cellcolor[rgb]{0.61490285,0.649358983,0.8768415764999999} 0.21 & \cellcolor[rgb]{0.61490285,0.649358983,0.8768415764999999} 0.25 & \cellcolor[rgb]{0.61490285,0.649358983,0.8768415764999999} 0.20 & \cellcolor[rgb]{0.8389114905588235,0.8932732186176471,0.9955022941235294} 0.40 & \cellcolor[rgb]{0.6240456089019608,0.6630065116588235,0.8888343256764706} 0.08 & \cellcolor[rgb]{0.61490285,0.649358983,0.8768415764999999} -0.09 & \cellcolor[rgb]{0.61490285,0.649358983,0.8768415764999999} 0.16 \\
 & WeightedAD-Block & \cellcolor[rgb]{0.8933603506784313,0.9224036051843137,0.9699051924745098} 0.05 & \cellcolor[rgb]{0.852836579,0.50777808,0.575116406} 0.25 & \cellcolor[rgb]{0.852836579,0.50777808,0.575116406} 0.31 & \cellcolor[rgb]{0.9385746672999999,0.6973227293333333,0.6558619128666667} 0.38 & \cellcolor[rgb]{0.9514243351588236,0.9223978252941176,0.9059849168705882} 0.27 & \cellcolor[rgb]{0.876805309,0.9151164254999999,0.9804355785000001} 0.42 & \cellcolor[rgb]{0.61490285,0.649358983,0.8768415764999999} 0.07 & \cellcolor[rgb]{0.7825907906117646,0.8497192224705883,0.9983175350588236} 0.08 & \cellcolor[rgb]{0.8569262456745098,0.9044285706686275,0.989693242609804} 0.23 \\
\cmidrule(lr){2-11}
 & AD-LastBlock & \cellcolor[rgb]{0.61490285,0.649358983,0.8768415764999999} 0.02 & \cellcolor[rgb]{0.7285232392627451,0.797002774964706,0.9815146148450979} 0.16 & \cellcolor[rgb]{0.61490285,0.649358983,0.8768415764999999} 0.21 & \cellcolor[rgb]{0.61490285,0.649358983,0.8768415764999999} 0.25 & \cellcolor[rgb]{0.61490285,0.649358983,0.8768415764999999} 0.20 & \cellcolor[rgb]{0.8389114905588235,0.8932732186176471,0.9955022941235294} 0.40 & \cellcolor[rgb]{0.6240456089019608,0.6630065116588235,0.8888343256764706} 0.08 & \cellcolor[rgb]{0.61490285,0.649358983,0.8768415764999999} -0.09 & \cellcolor[rgb]{0.61490285,0.649358983,0.8768415764999999} 0.16 \\
 & WeightedAD-LastBlock & \cellcolor[rgb]{0.8933603506784313,0.9224036051843137,0.9699051924745098} 0.05 & \cellcolor[rgb]{0.852836579,0.50777808,0.575116406} 0.25 & \cellcolor[rgb]{0.852836579,0.50777808,0.575116406} 0.31 & \cellcolor[rgb]{0.9385746672999999,0.6973227293333333,0.6558619128666667} 0.38 & \cellcolor[rgb]{0.9514243351588236,0.9223978252941176,0.9059849168705882} 0.27 & \cellcolor[rgb]{0.876805309,0.9151164254999999,0.9804355785000001} 0.42 & \cellcolor[rgb]{0.61490285,0.649358983,0.8768415764999999} 0.07 & \cellcolor[rgb]{0.7825907906117646,0.8497192224705883,0.9983175350588236} 0.08 & \cellcolor[rgb]{0.8569262456745098,0.9044285706686275,0.989693242609804} 0.23 \\
\cmidrule(lr){2-11}
 & AD-LastBlockPrefix & \cellcolor[rgb]{0.61490285,0.649358983,0.8768415764999999} 0.02 & \cellcolor[rgb]{0.7285232392627451,0.797002774964706,0.9815146148450979} 0.16 & \cellcolor[rgb]{0.61490285,0.649358983,0.8768415764999999} 0.21 & \cellcolor[rgb]{0.6970208390666667,0.7612067269333334,0.9624581464666666} 0.27 & \cellcolor[rgb]{0.61490285,0.649358983,0.8768415764999999} 0.20 & \cellcolor[rgb]{0.8594926464901961,0.905996446872549,0.9888280806960784} 0.41 & \cellcolor[rgb]{0.6240456089019608,0.6630065116588235,0.8888343256764706} 0.08 & \cellcolor[rgb]{0.61490285,0.649358983,0.8768415764999999} -0.09 & \cellcolor[rgb]{0.61490285,0.649358983,0.8768415764999999} 0.16 \\
 & WeightedAD-LastBlockPrefix & \cellcolor[rgb]{0.8933603506784313,0.9224036051843137,0.9699051924745098} 0.05 & \cellcolor[rgb]{0.852836579,0.50777808,0.575116406} 0.25 & \cellcolor[rgb]{0.852836579,0.50777808,0.575116406} 0.31 & \cellcolor[rgb]{0.852836579,0.50777808,0.575116406} 0.40 & \cellcolor[rgb]{0.9772829111803922,0.8895272809333333,0.8462654299039215} 0.28 & \cellcolor[rgb]{0.8933603506784313,0.9224036051843137,0.9699051924745098} 0.43 & \cellcolor[rgb]{0.61490285,0.649358983,0.8768415764999999} 0.07 & \cellcolor[rgb]{0.7825907906117646,0.8497192224705883,0.9983175350588236} 0.08 & \cellcolor[rgb]{0.8569262456745098,0.9044285706686275,0.989693242609804} 0.23 \\
\cmidrule(lr){2-11}
 & AD-Full & \cellcolor[rgb]{0.61490285,0.649358983,0.8768415764999999} 0.02 & \cellcolor[rgb]{0.7285232392627451,0.797002774964706,0.9815146148450979} 0.16 & \cellcolor[rgb]{0.61490285,0.649358983,0.8768415764999999} 0.21 & \cellcolor[rgb]{0.6970208390666667,0.7612067269333334,0.9624581464666666} 0.27 & \cellcolor[rgb]{0.61490285,0.649358983,0.8768415764999999} 0.20 & \cellcolor[rgb]{0.8594926464901961,0.905996446872549,0.9888280806960784} 0.41 & \cellcolor[rgb]{0.6240456089019608,0.6630065116588235,0.8888343256764706} 0.08 & \cellcolor[rgb]{0.61490285,0.649358983,0.8768415764999999} -0.09 & \cellcolor[rgb]{0.61490285,0.649358983,0.8768415764999999} 0.16 \\
 & WeightedAD-Full & \cellcolor[rgb]{0.8933603506784313,0.9224036051843137,0.9699051924745098} 0.05 & \cellcolor[rgb]{0.852836579,0.50777808,0.575116406} 0.25 & \cellcolor[rgb]{0.852836579,0.50777808,0.575116406} 0.31 & \cellcolor[rgb]{0.852836579,0.50777808,0.575116406} 0.40 & \cellcolor[rgb]{0.9772829111803922,0.8895272809333333,0.8462654299039215} 0.28 & \cellcolor[rgb]{0.8933603506784313,0.9224036051843137,0.9699051924745098} 0.43 & \cellcolor[rgb]{0.61490285,0.649358983,0.8768415764999999} 0.07 & \cellcolor[rgb]{0.7825907906117646,0.8497192224705883,0.9983175350588236} 0.08 & \cellcolor[rgb]{0.8569262456745098,0.9044285706686275,0.989693242609804} 0.23 \\
\bottomrule
\end{tabular}
}\caption{\label{tab:diffusiongemma_diff_baselines_results} \textbf{Diffusion-specific UQ signals.} PRR$\uparrow$ for the DiffusionGemma model across various tasks for diffusion statistics and trajectory dissimilarity UQ methods (see Tables~\ref{tab:diff_baselines_results}, \ref{tab:dream_diff_baselines_results}, and~\ref{tab:refusion_diff_baselines_results} for the LLaDA-1.5, Dream, and ReFusion models, respectively). Warmer color indicates better results. DiffusionGemma has no mask token, making MCNLL-based methods inapplicable; D-CoCoA-G instead uses CommitNLL.}\end{table*}

  \begin{table*}[!ht] \resizebox{\textwidth}{!}{\begin{tabular}{ll|c|c|c|c|c|c|c|c|c}
\toprule
\textbf{Category} & \textbf{UQ Method} & \textbf{XSum} & \textbf{SamSum} & \textbf{WMT14} & \textbf{WMT19} & \textbf{CoQA} & \textbf{TriviaQA} & \textbf{MMLU} & \textbf{GSM8k} & \textbf{Mean} \\
\midrule
\multirow{6}{*}{\shortstack[l]{AR +\\trajectory\\states}} & D-LexSim & \cellcolor[rgb]{0.9024823794117647,0.9258330802784314,0.9630825372156863} 0.01 & \cellcolor[rgb]{0.9133921582352942,0.9291026777686274,0.9534763223137255} 0.04 & \cellcolor[rgb]{0.9423217193470588,0.7050085489411765,0.6612532746235295} 0.24 & \cellcolor[rgb]{0.9791396989627451,0.8021675484411765,0.7416485506941177} 0.23 & \cellcolor[rgb]{0.9841016995,0.8604220499999999,0.8061464956666666} 0.27 & \cellcolor[rgb]{0.61490285,0.649358983,0.8768415764999999} 0.16 & \cellcolor[rgb]{0.61490285,0.649358983,0.8768415764999999} 0.07 & \cellcolor[rgb]{0.852836579,0.50777808,0.575116406} 0.02 & \cellcolor[rgb]{0.9678868848333333,0.9061183506196078,0.8735782367921568} 0.13 \\
 & D-DegMat & \cellcolor[rgb]{0.9337138175431372,0.9321882998862745,0.9313012310098039} 0.02 & \cellcolor[rgb]{0.9671527274529412,0.762958755827451,0.7061432370215687} 0.17 & \cellcolor[rgb]{0.9754778064901961,0.8934375166666666,0.8523803414019608} 0.17 & \cellcolor[rgb]{0.9497671903627452,0.7203459010784313,0.6720534316176471} 0.26 & \cellcolor[rgb]{0.61490285,0.649358983,0.8768415764999999} 0.21 & \cellcolor[rgb]{0.8790560841823529,0.5840607685176471,0.5944135352882353} 0.43 & \cellcolor[rgb]{0.852836579,0.50777808,0.575116406} 0.13 & \cellcolor[rgb]{0.8042736801705883,0.8678626149117648,0.9996769126490196} -0.20 & \cellcolor[rgb]{0.9849255762058824,0.8479150297647058,0.7906558870392157} 0.15 \\
 & D-EigVal & \cellcolor[rgb]{0.9337138175431372,0.9321882998862745,0.9313012310098039} 0.02 & \cellcolor[rgb]{0.9671527274529412,0.762958755827451,0.7061432370215687} 0.17 & \cellcolor[rgb]{0.9653342981666666,0.909438499827451,0.8795731953490196} 0.16 & \cellcolor[rgb]{0.9849255762058824,0.8479150297647058,0.7906558870392157} 0.21 & \cellcolor[rgb]{0.852836579,0.50777808,0.575116406} 0.30 & \cellcolor[rgb]{0.8790560841823529,0.5840607685176471,0.5944135352882353} 0.43 & \cellcolor[rgb]{0.8336264621666667,0.8895882285,0.9964796065} 0.09 & \cellcolor[rgb]{0.8466606424117646,0.8981570658529412,0.9931538902647059} -0.18 & \cellcolor[rgb]{0.9849255762058824,0.8479150297647058,0.7906558870392157} 0.15 \\
 & D-Ecc & \cellcolor[rgb]{0.852836579,0.50777808,0.575116406} 0.11 & \cellcolor[rgb]{0.852836579,0.50777808,0.575116406} 0.25 & \cellcolor[rgb]{0.852836579,0.50777808,0.575116406} 0.28 & \cellcolor[rgb]{0.852836579,0.50777808,0.575116406} 0.31 & \cellcolor[rgb]{0.9720272867117647,0.7765767393745098,0.7177742451568627} 0.28 & \cellcolor[rgb]{0.852836579,0.50777808,0.575116406} 0.44 & \cellcolor[rgb]{0.7179074031529412,0.7853536515764705,0.9758586906411765} 0.08 & \cellcolor[rgb]{0.9838554631666667,0.8314865062313725,0.7721615929176471} -0.07 & \cellcolor[rgb]{0.852836579,0.50777808,0.575116406} 0.21 \\
 & D-KLE & \cellcolor[rgb]{0.61490285,0.649358983,0.8768415764999999} -0.07 & \cellcolor[rgb]{0.61490285,0.649358983,0.8768415764999999} -0.14 & \cellcolor[rgb]{0.61490285,0.649358983,0.8768415764999999} -0.00 & \cellcolor[rgb]{0.61490285,0.649358983,0.8768415764999999} -0.01 & \cellcolor[rgb]{0.9596879944529412,0.9156363617647059,0.8914368152235295} 0.26 & \cellcolor[rgb]{0.9514243351588236,0.9223978252941176,0.9059849168705882} 0.31 & \cellcolor[rgb]{0.7179074031529412,0.7853536515764705,0.9758586906411765} 0.08 & \cellcolor[rgb]{0.61490285,0.649358983,0.8768415764999999} -0.29 & \cellcolor[rgb]{0.61490285,0.649358983,0.8768415764999999} 0.02 \\
 & D-LUQ & \cellcolor[rgb]{0.7554121621254901,0.8246983074117646,0.9925393881882353} -0.03 & \cellcolor[rgb]{0.9818859091411765,0.8745427552862746,0.824710021645098} 0.11 & \cellcolor[rgb]{0.9514243351588236,0.9223978252941176,0.9059849168705882} 0.15 & \cellcolor[rgb]{0.9479408841470589,0.9249530282941176,0.9117495381470588} 0.16 & \cellcolor[rgb]{0.9720272867117647,0.7765767393745098,0.7177742451568627} 0.28 & \cellcolor[rgb]{0.9696268857588235,0.769790744282353,0.7119501024647059} 0.38 & \cellcolor[rgb]{0.7179074031529412,0.7853536515764705,0.9758586906411765} 0.08 & \cellcolor[rgb]{0.7825907906117646,0.8497192224705883,0.9983175350588236} -0.21 & \cellcolor[rgb]{0.9196757213862745,0.930583412827451,0.9472468817078432} 0.11 \\
\bottomrule
\end{tabular}
}\caption{\label{tab:diffusiongemma_ar_traj_results} \textbf{Baselines with trajectory states.} PRR$\uparrow$ for the DiffusionGemma model across various tasks for sampling-based baselines from standard autoregressive LLMs, where denoising trajectory states are used instead of sampled generations (see Tables~\ref{tab:ar_traj_results}, \ref{tab:dream_ar_traj_results}, and~\ref{tab:refusion_ar_traj_results} for the LLaDA-1.5, Dream, and ReFusion models, respectively). Warmer color indicates better results.}\end{table*}

  \begin{table*}[!ht] \resizebox{\textwidth}{!}{\begin{tabular}{ll|c|c|c|c|c|c|c|c|c}
\toprule
\textbf{Category} & \textbf{UQ Method} & \textbf{XSum} & \textbf{SamSum} & \textbf{WMT14} & \textbf{WMT19} & \textbf{CoQA} & \textbf{TriviaQA} & \textbf{MMLU} & \textbf{GSM8k} & \textbf{Mean} \\
\midrule
\multirow{5}{*}{\shortstack[l]{AR\\baselines}} & MSP & \cellcolor[rgb]{0.6970208390666667,0.7612067269333334,0.9624581464666666} 0.04 & \cellcolor[rgb]{0.9003004236470589,0.9251791607803921,0.9650037801960785} 0.03 & \cellcolor[rgb]{0.7258692802352942,0.794090494117647,0.9801006337941176} 0.17 & \cellcolor[rgb]{0.7527113214117647,0.8219973367843137,0.9915787156372549} 0.25 & \cellcolor[rgb]{0.6426363887647059,0.690064711317647,0.9117342756235294} 0.16 & \cellcolor[rgb]{0.8204138911862745,0.8803757532058823,0.9989228874411764} 0.40 & \cellcolor[rgb]{0.9261890675039215,0.6732459732470588,0.6401732343490196} 0.55 & \cellcolor[rgb]{0.6842533066588236,0.7455705968156863,0.9526215641058824} -0.01 & \cellcolor[rgb]{0.7048055483372548,0.7703792543686274,0.9677723612431373} 0.20 \\
 & SAR & \cellcolor[rgb]{0.9497716033000001,0.923750118,0.9088945372} 0.10 & \cellcolor[rgb]{0.9305268001470588,0.6814578817647059,0.6453140635882353} 0.25 & \cellcolor[rgb]{0.8980319349294118,0.9243466028941176,0.9667357341529412} 0.27 & \cellcolor[rgb]{0.9802450306647059,0.8081382119705882,0.7477333002470588} 0.37 & \cellcolor[rgb]{0.8015810339588235,0.8657637386764706,0.9997826392686274} 0.21 & \cellcolor[rgb]{0.9497671903627452,0.7203459010784313,0.6720534316176471} 0.60 & \cellcolor[rgb]{0.9276891842254902,0.9318890695490196,0.9382935886568627} 0.34 & \cellcolor[rgb]{0.8042736801705883,0.8678626149117648,0.9996769126490196} 0.11 & \cellcolor[rgb]{0.9024823794117647,0.9258330802784314,0.9630825372156863} 0.28 \\
 & CoCoA-MTE & \cellcolor[rgb]{0.876805309,0.9151164254999999,0.9804355785000001} 0.08 & \cellcolor[rgb]{0.9813541391647058,0.8767786732784314,0.8278006056137255} 0.13 & \cellcolor[rgb]{0.7988883877470588,0.8636648624411765,0.9998883658882353} 0.21 & \cellcolor[rgb]{0.9627817115,0.9127586490352941,0.8855681539058824} 0.33 & \cellcolor[rgb]{0.61490285,0.649358983,0.8768415764999999} 0.15 & \cellcolor[rgb]{0.9683898066058824,0.766374750054902,0.7090466697431372} 0.58 & \cellcolor[rgb]{0.9261890675039215,0.6732459732470588,0.6401732343490196} 0.55 & \cellcolor[rgb]{0.6402749751176471,0.6867115845411765,0.909005371654902} -0.06 & \cellcolor[rgb]{0.8336264621666667,0.8895882285,0.9964796065} 0.25 \\
 & KLE & \cellcolor[rgb]{0.852836579,0.50777808,0.575116406} 0.17 & \cellcolor[rgb]{0.9404481933235294,0.7011656391372549,0.658557593745098} 0.24 & \cellcolor[rgb]{0.9627817115,0.9127586490352941,0.8855681539058824} 0.32 & \cellcolor[rgb]{0.852836579,0.50777808,0.575116406} 0.43 & \cellcolor[rgb]{0.8015810339588235,0.8657637386764706,0.9997826392686274} 0.21 & \cellcolor[rgb]{0.9367011412764705,0.6934798195294117,0.6531662319882353} 0.61 & \cellcolor[rgb]{0.7717200969960785,0.8400012947058824,0.9965252584941177} 0.21 & \cellcolor[rgb]{0.8718771013137254,0.9125626824411764,0.9828988807745098} 0.18 & \cellcolor[rgb]{0.9418435698882353,0.9280538589764706,0.9201288350882353} 0.30 \\
 & TAD & \cellcolor[rgb]{0.9385746672999999,0.6973227293333333,0.6558619128666667} 0.15 & \cellcolor[rgb]{0.852836579,0.50777808,0.575116406} 0.31 & \cellcolor[rgb]{0.852836579,0.50777808,0.575116406} 0.49 & \cellcolor[rgb]{0.9126469050843138,0.6478744190470588,0.6250127369666667} 0.41 & \cellcolor[rgb]{0.852836579,0.50777808,0.575116406} 0.36 & \cellcolor[rgb]{0.852836579,0.50777808,0.575116406} 0.66 & - & \cellcolor[rgb]{0.852836579,0.50777808,0.575116406} 0.60 & \cellcolor[rgb]{0.852836579,0.50777808,0.575116406} 0.43 \\\midrule
\multirow{2}{*}{\shortstack[l]{Diffusion\\baselines}} & TraceDet & \cellcolor[rgb]{0.6970208390666667,0.7612067269333334,0.9624581464666666} 0.04 & \cellcolor[rgb]{0.6996157421568627,0.7642642360784314,0.9642295513921568} -0.12 & \cellcolor[rgb]{0.61490285,0.649358983,0.8768415764999999} 0.10 & \cellcolor[rgb]{0.61490285,0.649358983,0.8768415764999999} 0.20 & \cellcolor[rgb]{0.6716387617176471,0.7296768173647059,0.9420609608117647} 0.17 & \cellcolor[rgb]{0.9849255762058824,0.8479150297647058,0.7906558870392157} 0.54 & \cellcolor[rgb]{0.852836579,0.50777808,0.575116406} 0.61 & \cellcolor[rgb]{0.8925766523392157,0.6104255443607843,0.6058364385039215} 0.56 & \cellcolor[rgb]{0.8569262456745098,0.9044285706686275,0.989693242609804} 0.26 \\
 & DynHD & \cellcolor[rgb]{0.6970208390666667,0.7612067269333334,0.9624581464666666} 0.04 & \cellcolor[rgb]{0.61490285,0.649358983,0.8768415764999999} -0.19 & \cellcolor[rgb]{0.8840171821764706,0.9185176097647059,0.976244109117647} 0.26 & \cellcolor[rgb]{0.9802450306647059,0.8081382119705882,0.7477333002470588} 0.37 & \cellcolor[rgb]{0.9175136022176471,0.6568221562117647,0.6298915758705882} 0.34 & \cellcolor[rgb]{0.8704786593764706,0.5611201626352941,0.5878720995529412} 0.65 & \cellcolor[rgb]{0.9729271893941176,0.7797824782294118,0.7207566595882353} 0.49 & \cellcolor[rgb]{0.8925766523392157,0.6104255443607843,0.6058364385039215} 0.56 & \cellcolor[rgb]{0.9596879944529412,0.9156363617647059,0.8914368152235295} 0.31 \\\midrule
\multirow{3}{*}{\shortstack[l]{Diffusion\\signals}} & NFE & \cellcolor[rgb]{0.6970208390666667,0.7612067269333334,0.9624581464666666} 0.04 & \cellcolor[rgb]{0.9659156483,0.7595427616,0.7032398043} 0.21 & \cellcolor[rgb]{0.815044265017647,0.8762581198529411,0.9992540061705882} 0.22 & \cellcolor[rgb]{0.9438760079745099,0.9270202487490196,0.9173357361078431} 0.32 & \cellcolor[rgb]{0.8644847897843138,0.9087320678529411,0.9865938341862746} 0.23 & \cellcolor[rgb]{0.61490285,0.649358983,0.8768415764999999} 0.28 & \cellcolor[rgb]{0.6817303976705882,0.7423918409254902,0.9505094434470589} 0.14 & \cellcolor[rgb]{0.9844470791666666,0.8397397817137255,0.7814061455764706} 0.39 & \cellcolor[rgb]{0.7825907906117646,0.8497192224705883,0.9983175350588236} 0.23 \\
 & AD-Full & \cellcolor[rgb]{0.61490285,0.649358983,0.8768415764999999} 0.02 & \cellcolor[rgb]{0.9842498738333334,0.8369886898862745,0.7783246280235294} 0.16 & \cellcolor[rgb]{0.7988883877470588,0.8636648624411765,0.9998883658882353} 0.21 & \cellcolor[rgb]{0.8123516188058824,0.8741592436176471,0.9993597327901961} 0.27 & \cellcolor[rgb]{0.7662841187058823,0.8349002989411765,0.9951966350588235} 0.20 & \cellcolor[rgb]{0.8389114905588235,0.8932732186176471,0.9955022941235294} 0.41 & \cellcolor[rgb]{0.61490285,0.649358983,0.8768415764999999} 0.08 & \cellcolor[rgb]{0.61490285,0.649358983,0.8768415764999999} -0.09 & \cellcolor[rgb]{0.61490285,0.649358983,0.8768415764999999} 0.16 \\
 & CommitNLL & \cellcolor[rgb]{0.6970208390666667,0.7612067269333334,0.9624581464666666} 0.04 & \cellcolor[rgb]{0.9847608008647059,0.8504164338117647,0.7937540087647059} 0.15 & \cellcolor[rgb]{0.9398111318019609,0.9290874692039215,0.9229219340686274} 0.30 & \cellcolor[rgb]{0.9627817115,0.9127586490352941,0.8855681539058824} 0.33 & \cellcolor[rgb]{0.9841016995,0.8604220499999999,0.8061464956666666} 0.29 & \cellcolor[rgb]{0.888688766427451,0.9204606074745099,0.9730746507960784} 0.44 & \cellcolor[rgb]{0.864597965917647,0.5433394684235294,0.5836202017019608} 0.60 & \cellcolor[rgb]{0.9296925499352942,0.9322154837294118,0.9360552653941177} 0.25 & \cellcolor[rgb]{0.9418435698882353,0.9280538589764706,0.9201288350882353} 0.30 \\\midrule
\multirow{2}{*}{\shortstack[l]{\textbf{D-CoCoA}\\\textbf{(ours)}}} & D-CoCoA-G & \cellcolor[rgb]{0.876805309,0.9151164254999999,0.9804355785000001} 0.08 & \cellcolor[rgb]{0.9077541933039216,0.6388904935882354,0.6201467828333334} 0.27 & \cellcolor[rgb]{0.9514243351588236,0.9223978252941176,0.9059849168705882} 0.31 & \cellcolor[rgb]{0.9829494490941176,0.8700709193019608,0.8185288537078431} 0.35 & \cellcolor[rgb]{0.9622044108411765,0.7492947789176471,0.6945295061352941} 0.32 & \cellcolor[rgb]{0.9849255762058824,0.8479150297647058,0.7906558870392157} 0.54 & \cellcolor[rgb]{0.852836579,0.50777808,0.575116406} 0.61 & \cellcolor[rgb]{0.9438760079745099,0.9270202487490196,0.9173357361078431} 0.27 & \cellcolor[rgb]{0.9841016995,0.8604220499999999,0.8061464956666666} 0.34 \\
 & D-CoCoA-L & \cellcolor[rgb]{0.6970208390666667,0.7612067269333334,0.9624581464666666} 0.04 & \cellcolor[rgb]{0.9659156483,0.7595427616,0.7032398043} 0.21 & \cellcolor[rgb]{0.9256858185156862,0.9315626553686274,0.9405319119196078} 0.29 & \cellcolor[rgb]{0.9216790870960785,0.9309098270078431,0.945008558445098} 0.31 & \cellcolor[rgb]{0.9756268974411765,0.7893996947941176,0.729703902882353} 0.31 & \cellcolor[rgb]{0.9849255762058824,0.8479150297647058,0.7906558870392157} 0.54 & \cellcolor[rgb]{0.852836579,0.50777808,0.575116406} 0.61 & \cellcolor[rgb]{0.7232153212078432,0.7911782132705882,0.9786866527431373} 0.03 & \cellcolor[rgb]{0.9236824528058825,0.9312362411882353,0.9427702351823529} 0.29 \\
\bottomrule
\end{tabular}}\caption{\label{tab:diffusiongemma_detailed_main_results} \textbf{D-CoCoA comparison.} PRR$\uparrow$ for the DiffusionGemma model across various tasks for D-CoCoA methods, comparing the best method from each category (see Tables~\ref{tab:detailed_main_results}, \ref{tab:dream_detailed_main_results}, and~\ref{tab:refusion_detailed_main_results} for the LLaDA-1.5, Dream, and ReFusion models, respectively). Warmer color indicates better results. DiffusionGemma has no mask token, making MCNLL-based methods inapplicable; D-CoCoA-G instead uses CommitNLL.}\end{table*}

\clearpage
\subsection{Results Using the ROC-AUC Metric}
\label{app:rocauc}

\begin{figure*}[h!]
    \centering
    \begin{minipage}[t]{0.49\linewidth}
      \centering
      \includegraphics[trim={0.cm 0.cm 0.cm 1cm},clip,width=\linewidth]{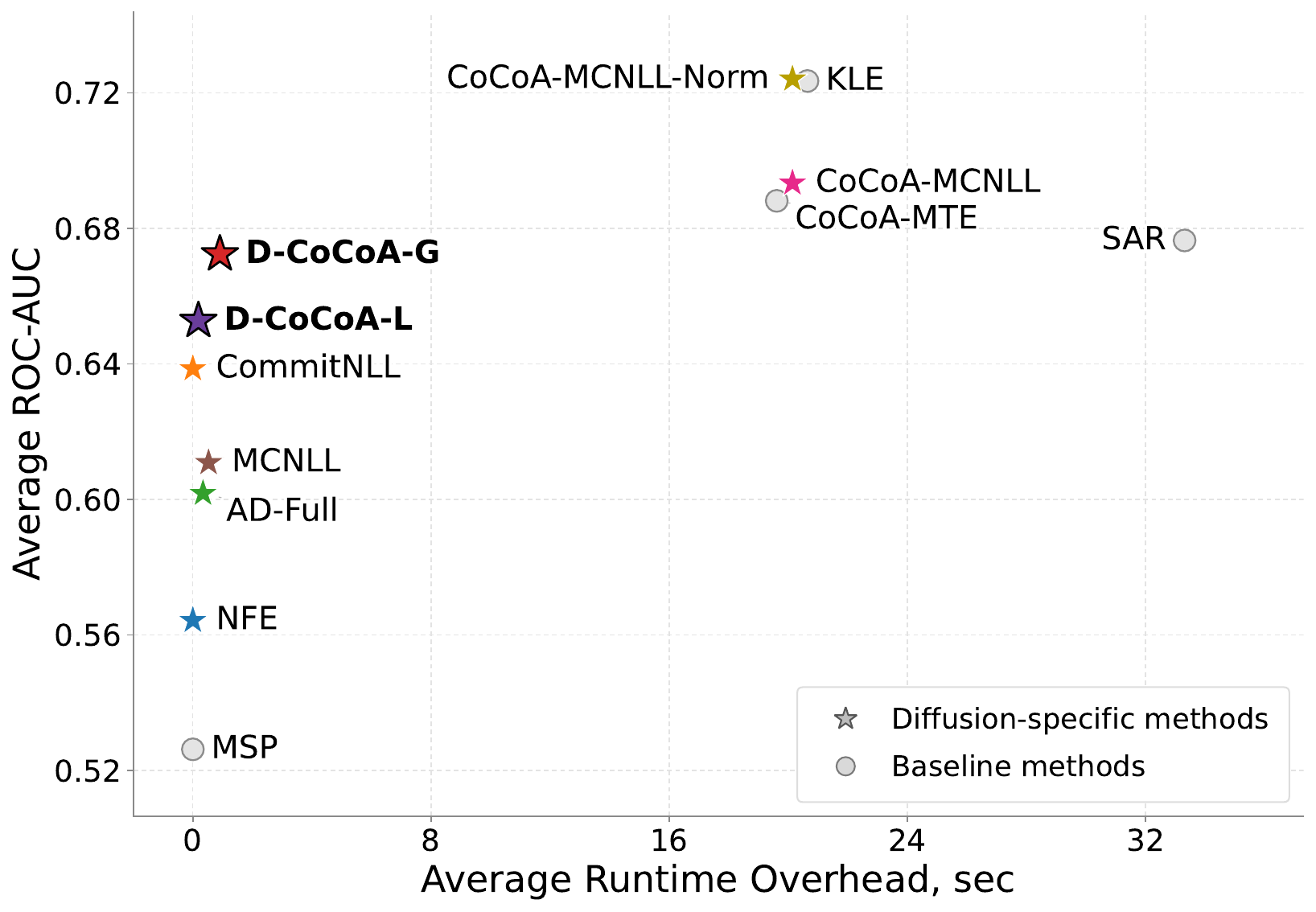}
      \subcaption{\textbf{LLaDA-1.5}}
      \label{fig:avg_auroc_vs_comp_llada}
    \end{minipage}
    \hfill
    \begin{minipage}[t]{0.49\linewidth}
      \centering
      \includegraphics[trim={0.cm 0.cm 0.cm 1cm},clip,width=\linewidth]{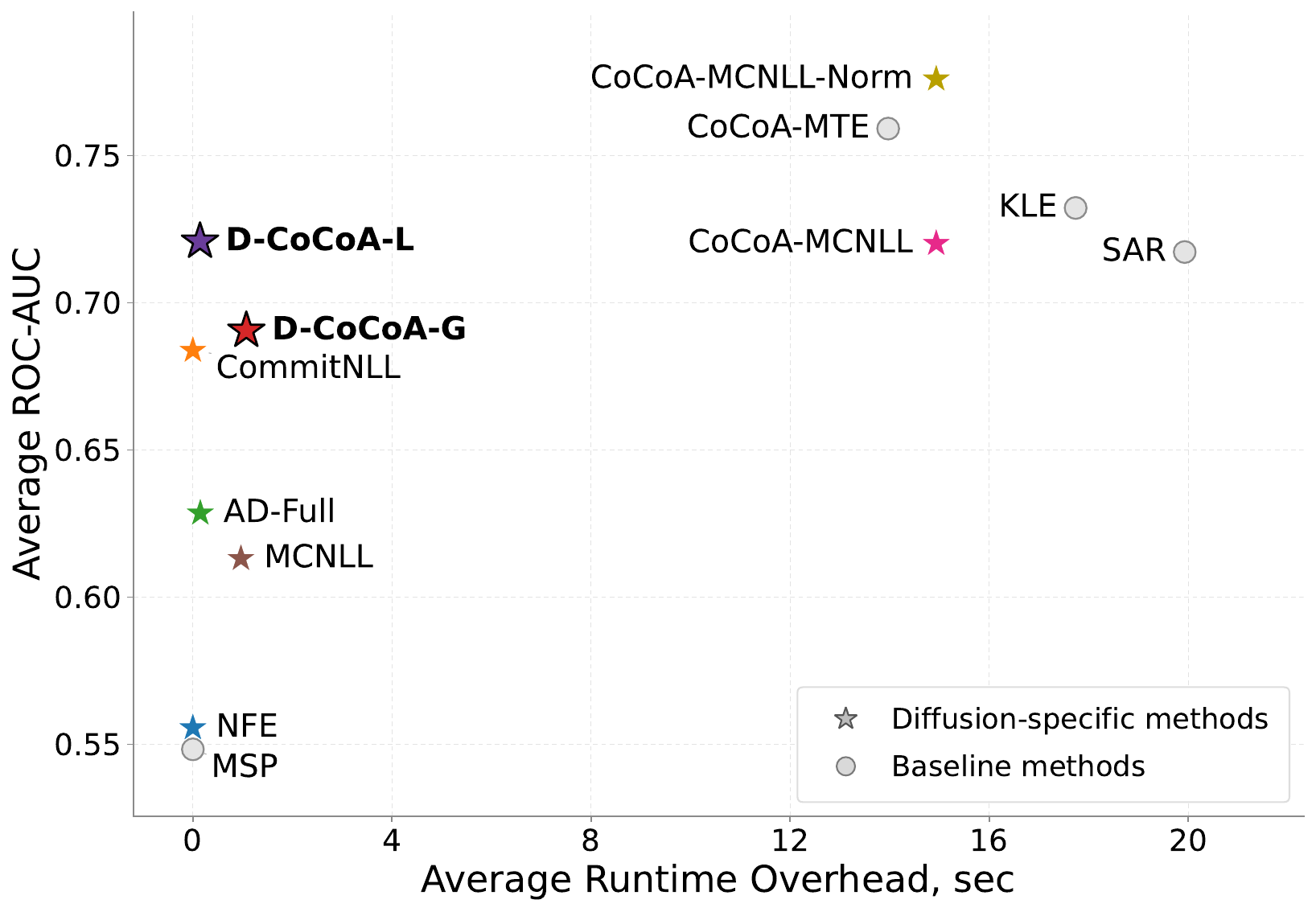}
      \subcaption{\textbf{Dream}}
      \label{fig:avg_auroc_vs_comp_dream}
    \end{minipage}

    \begin{minipage}[t]{0.49\linewidth}
      \centering
      \includegraphics[trim={0.cm 0.cm 0.cm 1cm},clip,width=\linewidth]{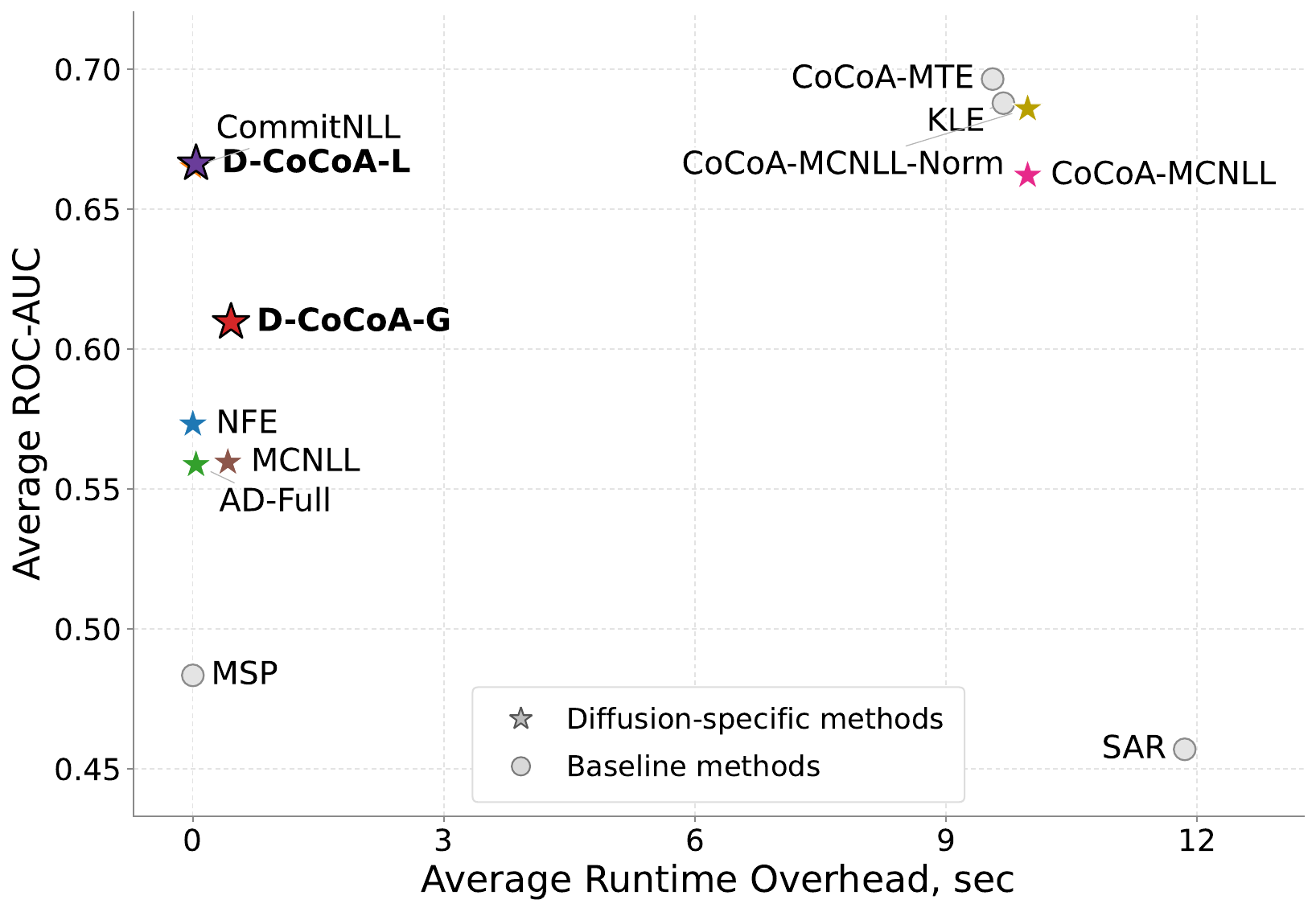}
      \subcaption{\textbf{ReFusion}}
      \label{fig:avg_auroc_vs_comp_refusion}
    \end{minipage}
    \hfill
    \begin{minipage}[t]{0.49\linewidth}
      \centering
      \includegraphics[trim={0.cm 0.cm 0.cm 1cm},clip,width=\linewidth]{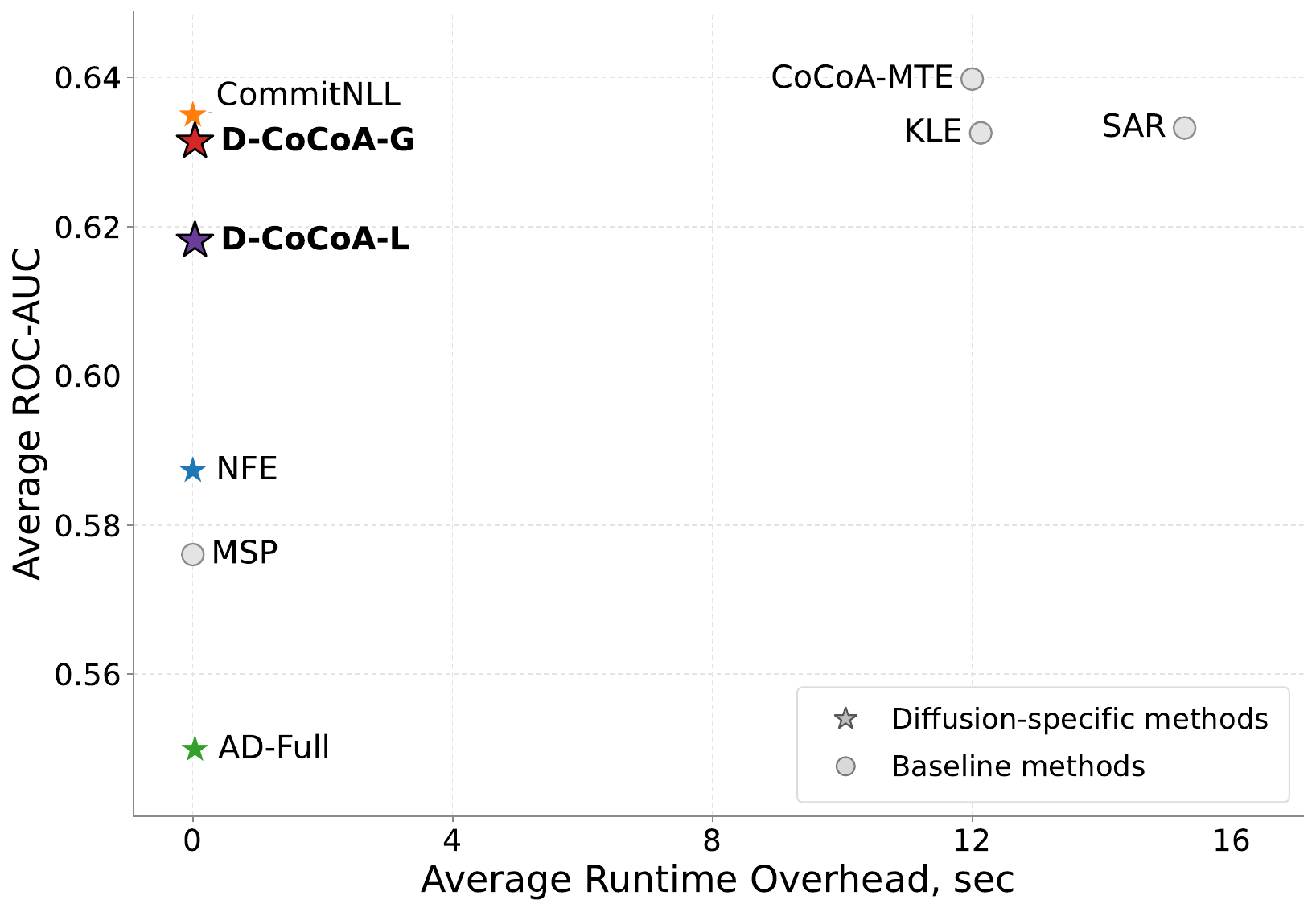}
      \subcaption{\textbf{DiffusionGemma}}
      \label{fig:avg_auroc_vs_comp_diffusiongemma}
    \end{minipage}
    \caption{
    \textbf{Average performance and runtime.} UQ performance (ROC-AUC$\uparrow$) versus computational overhead (runtime in seconds$\downarrow$) averaged across all datasets. DiffusionGemma has no mask token, so MCNLL-based methods are not applicable and D-CoCoA-G uses CommitNLL.
    }
    \label{fig:avg_auroc_vs_comp}
  \end{figure*}

\begin{figure*}[h!]
    \centering
    \includegraphics[trim={0.cm 0.cm 0.cm 0.cm},clip,width=0.99\linewidth]{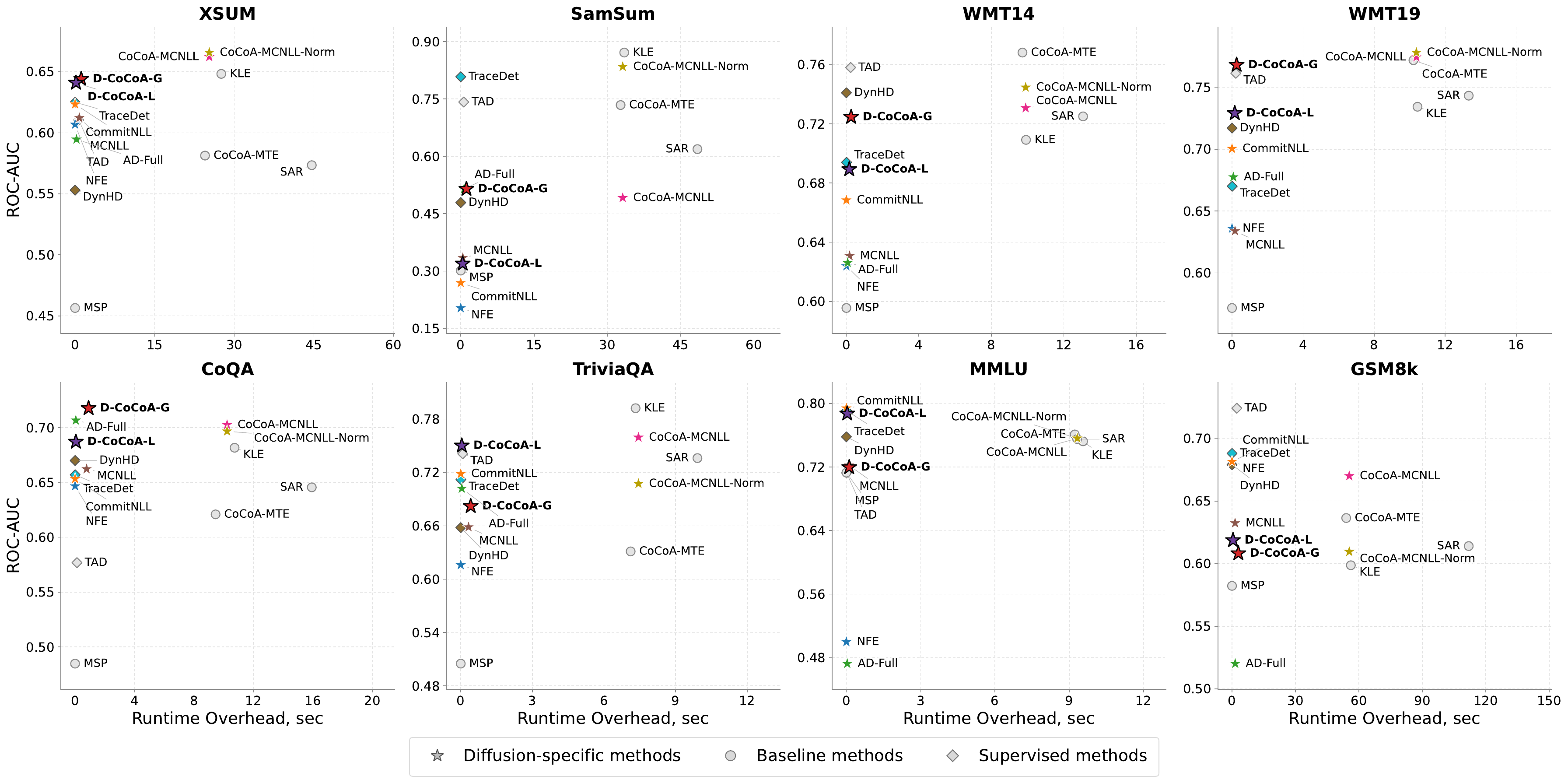}
    \caption{
    \textbf{LLaDA-1.5 performance and runtime.} Comparison of uncertainty quantification performance (ROC-AUC$\uparrow$) versus computational complexity (Time in seconds$\downarrow$) across eight benchmark datasets for the LLaDA-1.5 model (see Figures~\ref{fig:dream_auroc_vs_comp}, \ref{fig:refusion_auroc_vs_comp}, and~\ref{fig:diffusiongemma_auroc_vs_comp} for the Dream, ReFusion, and DiffusionGemma models, respectively).
    }
    \label{fig:llada_auroc_vs_comp}
  \end{figure*}

\begin{figure*}[h!]
    \centering
    \includegraphics[trim={0.cm 0.cm 0.cm 0.cm},clip,width=0.99\linewidth]{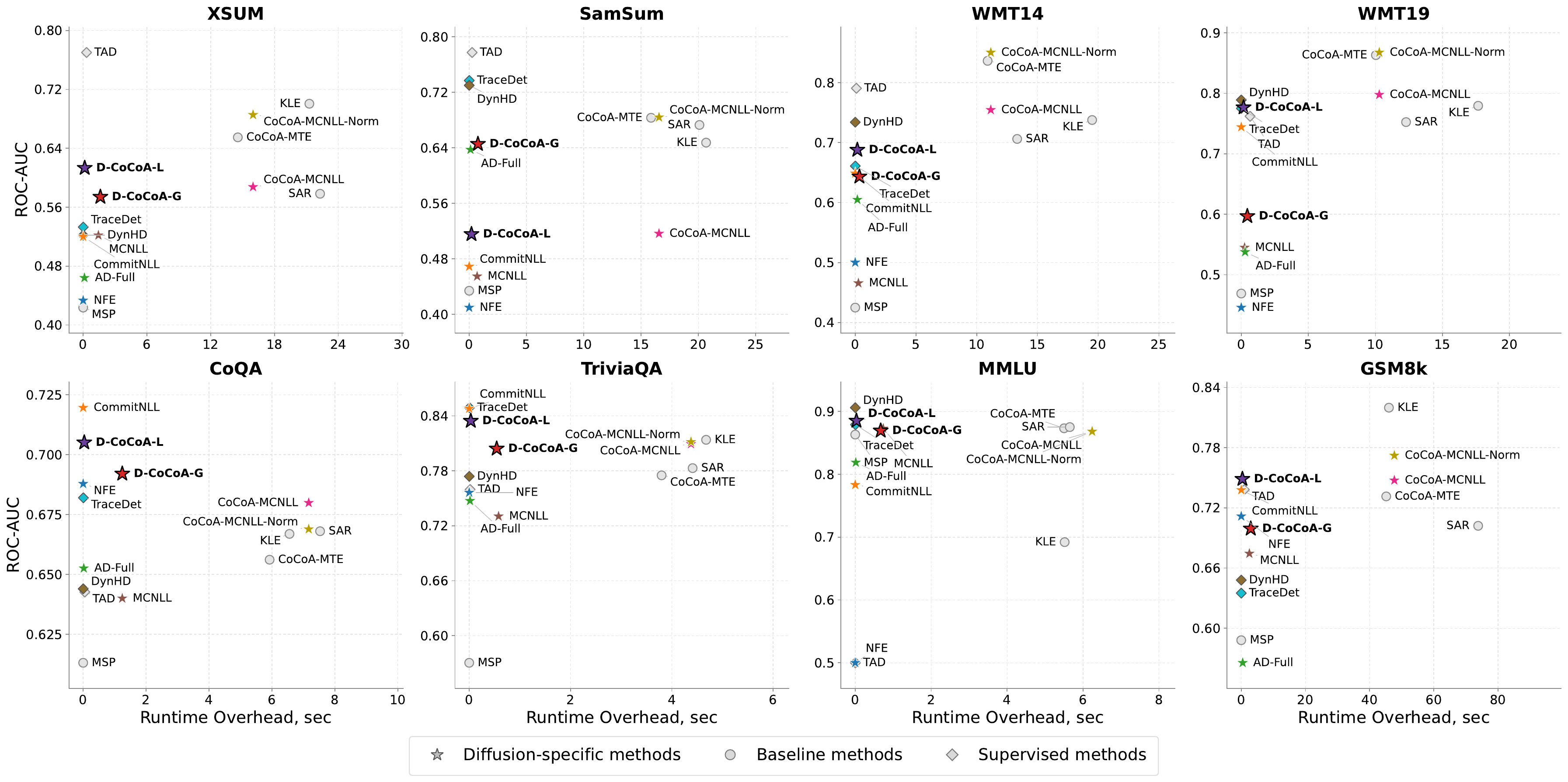}
    \caption{
    \textbf{Dream performance and runtime.} Comparison of uncertainty quantification performance (ROC-AUC$\uparrow$) versus computational complexity (Time in seconds$\downarrow$) across eight benchmark datasets for the Dream model (see Figures~\ref{fig:llada_auroc_vs_comp}, \ref{fig:refusion_auroc_vs_comp}, and~\ref{fig:diffusiongemma_auroc_vs_comp} for the LLaDA-1.5, ReFusion, and DiffusionGemma models, respectively).
    }
    \label{fig:dream_auroc_vs_comp}
  \end{figure*}

\begin{figure*}[h!]
    \centering
    \includegraphics[trim={0.cm 0.cm 0.cm 0.cm},clip,width=0.99\linewidth]{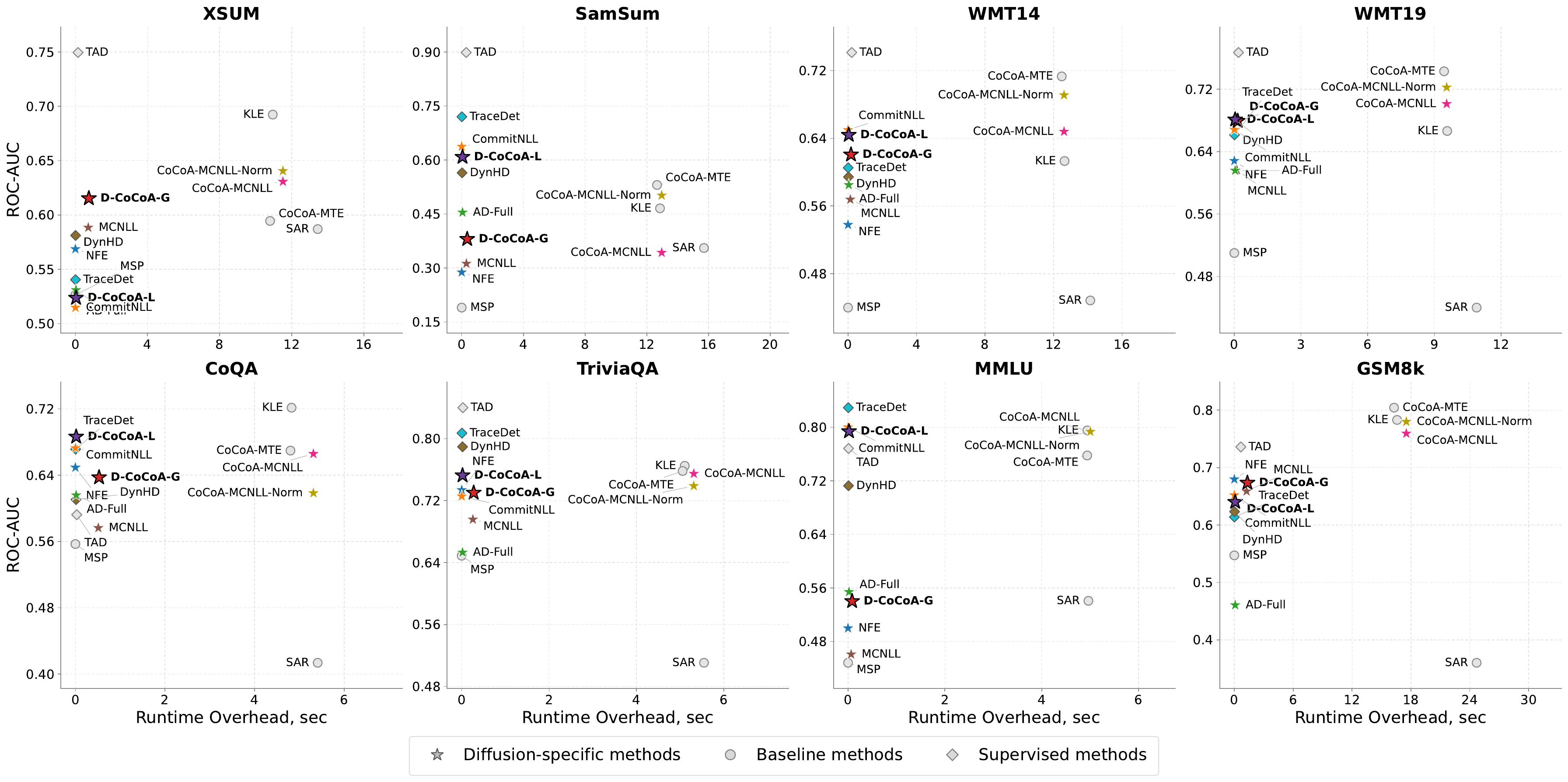}
    \caption{
    \textbf{ReFusion performance and runtime.} Comparison of uncertainty quantification performance (ROC-AUC$\uparrow$) versus computational complexity (Time in seconds$\downarrow$) across eight benchmark datasets for the ReFusion model (see Figures~\ref{fig:llada_auroc_vs_comp}, \ref{fig:dream_auroc_vs_comp}, and~\ref{fig:diffusiongemma_auroc_vs_comp} for the LLaDA-1.5, Dream, and DiffusionGemma models, respectively).
    }
    \label{fig:refusion_auroc_vs_comp}
  \end{figure*}

\begin{figure*}[h!]
    \centering
    \includegraphics[trim={0.cm 0.cm 0.cm 0.cm},clip,width=0.99\linewidth]{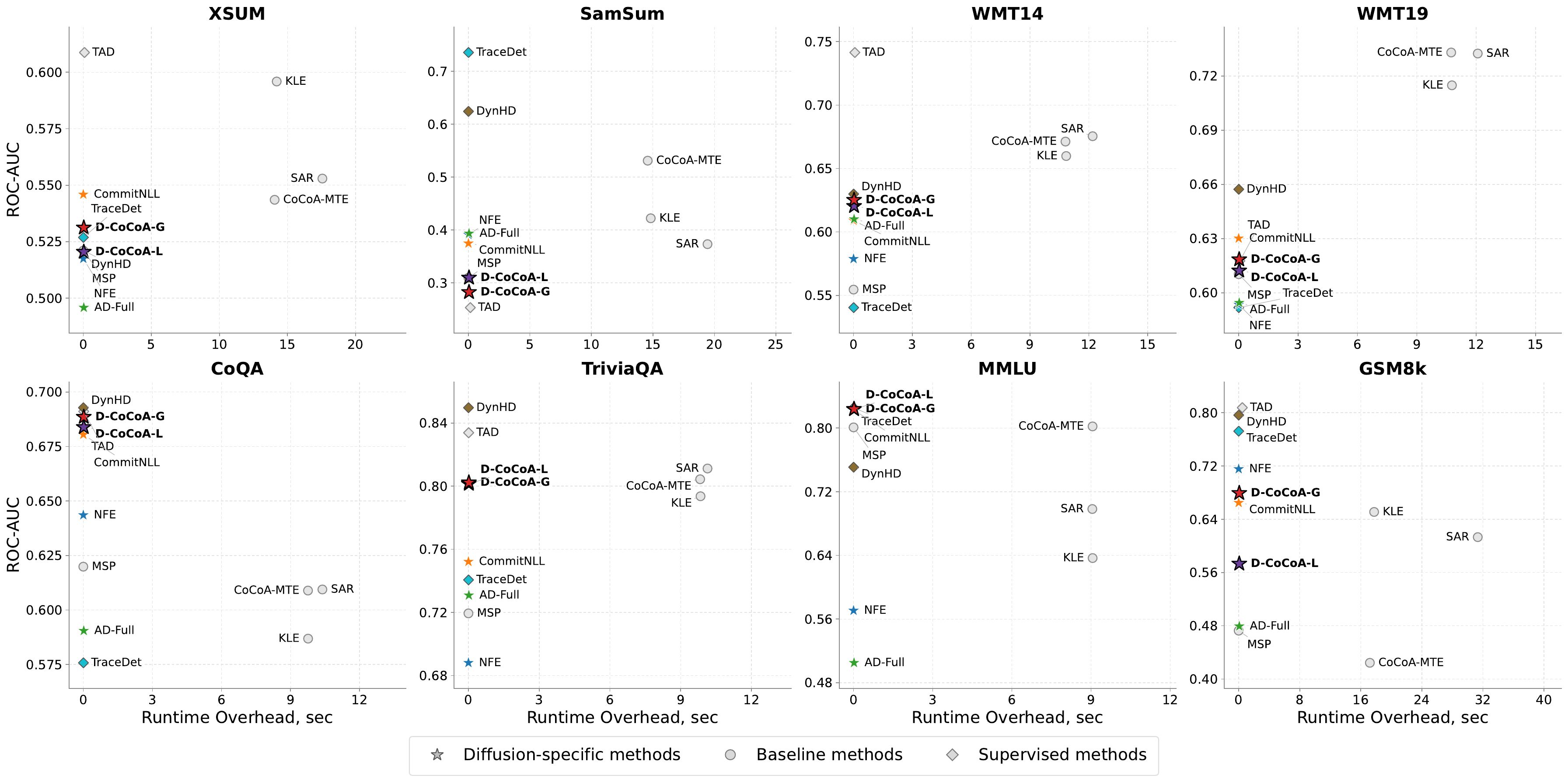}
    \caption{
    \textbf{DiffusionGemma performance and runtime.} Comparison of uncertainty quantification performance (ROC-AUC$\uparrow$) versus computational complexity (Time in seconds$\downarrow$) across eight benchmark datasets for the DiffusionGemma model (see Figures~\ref{fig:llada_auroc_vs_comp}, \ref{fig:dream_auroc_vs_comp}, and~\ref{fig:refusion_auroc_vs_comp} for the LLaDA-1.5, Dream, and ReFusion models, respectively).
    }
    \label{fig:diffusiongemma_auroc_vs_comp}
  \end{figure*}

The averaged results evaluated using the ROC-AUC metric are presented in Figure~\ref{fig:avg_auroc_vs_comp}. Detailed results for each dataset are presented in Figures~\ref{fig:llada_auroc_vs_comp}, \ref{fig:dream_auroc_vs_comp}, \ref{fig:refusion_auroc_vs_comp}, and~\ref{fig:diffusiongemma_auroc_vs_comp} for LLaDA, Dream, ReFusion, and DiffusionGemma, respectively.

For all generation quality metrics other than accuracy, we compute scores by thresholding the original continuous values to obtain discrete labels. The thresholds were empirically determined as follows: 0.3 for QA and Summ, and 0.8 for MT. We observe trends similar to those seen with the PRR metric. D-CoCoA methods show the best performance-efficiency trade-off. Among diffusion-based methods, D-CoCoA-G achieves the best average performance for LLaDA, and D-CoCoA-L for Dream and ReFusion, while for DiffusionGemma, CommitNLL performs slightly better than D-CoCoA-G (0.635 vs. 0.631). While absolute values differ slightly from PRR, these results again highlight the strong suitability of the proposed method for efficient LLDMs.

\clearpage
\section{Ablation Studies}
\label{sec:abls}
\label{app:abl}

  To understand the contributions of individual components of D-CoCoA methods, we conduct four ablation studies.

  \ecoparagraph{Sources of likelihood.} Table~\ref{tab:abl_cocoa_inf_results} reports ablation results in which the information-based component of D-CoCoA-G and D-CoCoA-L is replaced with alternative diffusion likelihood methods: trajectory likelihood (both variants), CommitNLL, and normalized/unnormalized MCNLL.
  
  The original D-CoCoA-G outperforms or ties with its variants on average, with particular strength in long-form tasks such as summarization and translation. Its TrajNLL variant provides the best results, surpassing the best sampling-based baseline on CoQA and MMLU, though MCNLL offers the most consistent performance overall. The original D-CoCoA-L likewise achieves the best performance among its variants on average, achieving top results on all datasets except CoQA.

  \ecoparagraph{Trajectory aggregation.} Table~\ref{tab:abl_cocoa_traj_results} evaluates D-CoCoA under different definitions of $\tilde{\yv}_t$ as stated in Section~\ref{sec:methods}. These results confirm that the full generation trajectory is optimal for D-CoCoA-G, and individual block-wise averaging for D-CoCoA-L. Overall, performance remains stable across all alternatives, with no significant degradation.

  \ecoparagraph{Trajectory similarity functions.} Table~\ref{tab:abl_cocoa_sim_results} presents D-CoCoA results under different similarity functions (DeBERTa NLI and Cross-Encoder with RoBERTa) and weighted or unweighted aggregation. These results demonstrate the overall consistency of the proposed methods. Replacing the Cross-Encoder with the NLI model reduces average performance by 3\% for D-CoCoA-G and does not change it for D-CoCoA-L. Progressive weighted scoring yields marginal WMT gains for D-CoCoA-G but degrades QA performance, leaving the overall unchanged, and slightly reduces performance across all datasets except MMLU and GSM8k for D-CoCoA-L.

  \ecoparagraph{Additional diffusion statistic.} Table~\ref{tab:abl_cocoa_nfe_results} evaluates the contribution of the NFE component in D-CoCoA. For D-CoCoA-G, incorporating NFE yields consistent and substantial improvements, confirming that it captures information distinct from other global trajectory-based signals. For D-CoCoA-L, however, it slightly reduces performance on all datasets except MMLU and GSM8k.

  \begin{table*}[!h] 
\centering
\resizebox{\textwidth}{!}{\begin{tabular}{l|c|c|c|c|c|c|c|c|c}
\toprule
\textbf{UQ Method} & \textbf{XSum} & \textbf{SamSum} & \textbf{WMT14} & \textbf{WMT19} & \textbf{CoQA} & \textbf{TriviaQA} & \textbf{MMLU} & \textbf{GSM8k} & \textbf{Mean} \\
\midrule
D-CoCoA-G (TrajNLL) & \cellcolor[rgb]{0.61490285,0.649358983,0.8768415764999999} 0.20 & \cellcolor[rgb]{0.9802905992117648,0.8812505092627452,0.8339817735509804} 0.19 & \cellcolor[rgb]{0.8718771013137254,0.9125626824411764,0.9828988807745098} 0.28 & \cellcolor[rgb]{0.7311771982901961,0.7999150558117647,0.9829285958960784} 0.31 & \cellcolor[rgb]{0.852836579,0.50777808,0.575116406} 0.35 & \cellcolor[rgb]{0.9790880158705882,0.8856170452000001,0.8401505184058824} 0.44 & \cellcolor[rgb]{0.852836579,0.50777808,0.575116406} 0.46 & \cellcolor[rgb]{0.9077541933039216,0.6388904935882354,0.6201467828333334} 0.25 & \cellcolor[rgb]{0.9783266054882354,0.7990169113588235,0.7386511461764707} 0.31 \\
D-CoCoA-G (TrajEntropy) & \cellcolor[rgb]{0.61490285,0.649358983,0.8768415764999999} 0.20 & \cellcolor[rgb]{0.9802905992117648,0.8812505092627452,0.8339817735509804} 0.19 & \cellcolor[rgb]{0.9216790870960785,0.9309098270078431,0.945008558445098} 0.30 & \cellcolor[rgb]{0.806966326382353,0.8699614911470589,0.9995711860294118} 0.34 & \cellcolor[rgb]{0.9326956664686274,0.6855638360235294,0.6478844782078431} 0.34 & \cellcolor[rgb]{0.9838554631666667,0.8314865062313725,0.7721615929176471} 0.46 & \cellcolor[rgb]{0.9819030282176471,0.8170942072647058,0.7568604245764706} 0.43 & \cellcolor[rgb]{0.9077541933039216,0.6388904935882354,0.6201467828333334} 0.25 & \cellcolor[rgb]{0.9783266054882354,0.7990169113588235,0.7386511461764707} 0.31 \\
D-CoCoA-G (MCNLL) & \cellcolor[rgb]{0.931695915645098,0.9325418979098039,0.9338169421313726} 0.21 & \cellcolor[rgb]{0.9678868848333333,0.9061183506196078,0.8735782367921568} 0.18 & \cellcolor[rgb]{0.9829494490941176,0.8700709193019608,0.8185288537078431} 0.34 & \cellcolor[rgb]{0.9024823794117647,0.9258330802784314,0.9630825372156863} 0.38 & \cellcolor[rgb]{0.852836579,0.50777808,0.575116406} 0.35 & \cellcolor[rgb]{0.9838554631666667,0.8314865062313725,0.7721615929176471} 0.46 & \cellcolor[rgb]{0.61490285,0.649358983,0.8768415764999999} 0.35 & \cellcolor[rgb]{0.924020201182353,0.6691400189882353,0.6376028197294118} 0.24 & \cellcolor[rgb]{0.9783266054882354,0.7990169113588235,0.7386511461764707} 0.31 \\
D-CoCoA-G (CommitNLL) & \cellcolor[rgb]{0.61490285,0.649358983,0.8768415764999999} 0.20 & \cellcolor[rgb]{0.9678868848333333,0.9061183506196078,0.8735782367921568} 0.18 & \cellcolor[rgb]{0.9844470791666666,0.8397397817137255,0.7814061455764706} 0.35 & \cellcolor[rgb]{0.9418435698882353,0.9280538589764706,0.9201288350882353} 0.40 & \cellcolor[rgb]{0.931695915645098,0.9325418979098039,0.9338169421313726} 0.31 & \cellcolor[rgb]{0.9528917390058824,0.727592846082353,0.6776679419235294} 0.49 & \cellcolor[rgb]{0.852836579,0.50777808,0.575116406} 0.46 & \cellcolor[rgb]{0.852836579,0.50777808,0.575116406} 0.28 & \cellcolor[rgb]{0.9326956664686274,0.6855638360235294,0.6478844782078431} 0.33 \\
D-CoCoA-G & \cellcolor[rgb]{0.931695915645098,0.9325418979098039,0.9338169421313726} 0.21 & \cellcolor[rgb]{0.852836579,0.50777808,0.575116406} 0.26 & \cellcolor[rgb]{0.852836579,0.50777808,0.575116406} 0.42 & \cellcolor[rgb]{0.852836579,0.50777808,0.575116406} 0.53 & \cellcolor[rgb]{0.9797588292352941,0.8834864272549019,0.8370723575196078} 0.32 & \cellcolor[rgb]{0.8042736801705883,0.8678626149117648,0.9996769126490196} 0.36 & \cellcolor[rgb]{0.61490285,0.649358983,0.8768415764999999} 0.35 & \cellcolor[rgb]{0.9774267028058823,0.7958111725039216,0.7356687317450981} 0.19 & \cellcolor[rgb]{0.9326956664686274,0.6855638360235294,0.6478844782078431} 0.33 \\
\midrule
D-CoCoA-L (TrajNLL) & \cellcolor[rgb]{0.61490285,0.649358983,0.8768415764999999} 0.20 & \cellcolor[rgb]{0.6449978024117646,0.6934178380941176,0.9144631795921568} 0.08 & \cellcolor[rgb]{0.6667452396901961,0.7231326097019608,0.9372260848549019} 0.21 & \cellcolor[rgb]{0.61490285,0.649358983,0.8768415764999999} 0.26 & \cellcolor[rgb]{0.8594926464901961,0.905996446872549,0.9888280806960784} 0.30 & \cellcolor[rgb]{0.9528917390058824,0.727592846082353,0.6776679419235294} 0.49 & \cellcolor[rgb]{0.852836579,0.50777808,0.575116406} 0.46 & \cellcolor[rgb]{0.6970208390666667,0.7612067269333334,0.9624581464666666} -0.04 & \cellcolor[rgb]{0.8594926464901961,0.905996446872549,0.9888280806960784} 0.25 \\
D-CoCoA-L (TrajEntropy) & \cellcolor[rgb]{0.61490285,0.649358983,0.8768415764999999} 0.20 & \cellcolor[rgb]{0.6449978024117646,0.6934178380941176,0.9144631795921568} 0.08 & \cellcolor[rgb]{0.8123516188058824,0.8741592436176471,0.9993597327901961} 0.26 & \cellcolor[rgb]{0.7554121621254901,0.8246983074117646,0.9925393881882353} 0.32 & \cellcolor[rgb]{0.7744380861411764,0.8425517925882353,0.9971895702117648} 0.29 & \cellcolor[rgb]{0.9348276152529411,0.6896369097254902,0.6504705511098039} 0.50 & \cellcolor[rgb]{0.9819030282176471,0.8170942072647058,0.7568604245764706} 0.43 & \cellcolor[rgb]{0.7152534441254902,0.7824413707294118,0.974444709590196} -0.03 & \cellcolor[rgb]{0.9003004236470589,0.9251791607803921,0.9650037801960785} 0.26 \\
D-CoCoA-L (MCNLL) & \cellcolor[rgb]{0.852836579,0.50777808,0.575116406} 0.22 & \cellcolor[rgb]{0.9807976965156863,0.8111235437352942,0.7507756750235295} 0.21 & \cellcolor[rgb]{0.8718771013137254,0.9125626824411764,0.9828988807745098} 0.28 & \cellcolor[rgb]{0.8816813900509803,0.917546110909804,0.9778288382784314} 0.37 & \cellcolor[rgb]{0.8594926464901961,0.905996446872549,0.9888280806960784} 0.30 & \cellcolor[rgb]{0.9838554631666667,0.8314865062313725,0.7721615929176471} 0.46 & \cellcolor[rgb]{0.61490285,0.649358983,0.8768415764999999} 0.35 & \cellcolor[rgb]{0.953077067017647,0.9210455325882353,0.9030752965411765} 0.11 & \cellcolor[rgb]{0.9790880158705882,0.8856170452000001,0.8401505184058824} 0.29 \\
D-CoCoA-L (MCNLL-Norm) & \cellcolor[rgb]{0.61490285,0.649358983,0.8768415764999999} 0.20 & \cellcolor[rgb]{0.61490285,0.649358983,0.8768415764999999} 0.07 & \cellcolor[rgb]{0.61490285,0.649358983,0.8768415764999999} 0.19 & \cellcolor[rgb]{0.61490285,0.649358983,0.8768415764999999} 0.26 & \cellcolor[rgb]{0.61490285,0.649358983,0.8768415764999999} 0.27 & \cellcolor[rgb]{0.61490285,0.649358983,0.8768415764999999} 0.29 & \cellcolor[rgb]{0.61490285,0.649358983,0.8768415764999999} 0.35 & \cellcolor[rgb]{0.61490285,0.649358983,0.8768415764999999} -0.09 & \cellcolor[rgb]{0.61490285,0.649358983,0.8768415764999999} 0.19 \\
D-CoCoA-L & \cellcolor[rgb]{0.852836579,0.50777808,0.575116406} 0.22 & \cellcolor[rgb]{0.9497671903627452,0.7203459010784313,0.6720534316176471} 0.23 & \cellcolor[rgb]{0.8844643114450981,0.5946066788549019,0.5989826965745098} 0.41 & \cellcolor[rgb]{0.9607031106137255,0.7457102085921569,0.6917042176882353} 0.48 & \cellcolor[rgb]{0.6918310328862745,0.7550917086431372,0.9589153366156863} 0.28 & \cellcolor[rgb]{0.852836579,0.50777808,0.575116406} 0.53 & \cellcolor[rgb]{0.852836579,0.50777808,0.575116406} 0.46 & \cellcolor[rgb]{0.9385746672999999,0.6973227293333333,0.6558619128666667} 0.23 & \cellcolor[rgb]{0.852836579,0.50777808,0.575116406} 0.35 \\

\bottomrule
\end{tabular}
}\caption{\label{tab:abl_cocoa_inf_results} \textbf{Likelihood component ablation.} PRR$\uparrow$ for the LLaDA-1.5 model across various tasks for D-CoCoA methods with various diffusion likelihood components. Warmer color indicates better results.}\end{table*}
  \begin{table*}[!h] 
\centering
\resizebox{\textwidth}{!}{\begin{tabular}{l|c|c|c|c|c|c|c|c|c}
\toprule
\textbf{UQ Method} & \textbf{XSum} & \textbf{SamSum} & \textbf{WMT14} & \textbf{WMT19} & \textbf{CoQA} & \textbf{TriviaQA} & \textbf{MMLU} & \textbf{GSM8k} & \textbf{Mean} \\
\midrule
D-CoCoA-G (Block-wise) & \cellcolor[rgb]{0.9841016995,0.8604220499999999,0.8061464956666666} 0.21 & \cellcolor[rgb]{0.9423217193470588,0.7050085489411765,0.6612532746235295} 0.25 & \cellcolor[rgb]{0.9337138175431372,0.9321882998862745,0.9313012310098039} 0.39 & \cellcolor[rgb]{0.9659156483,0.7595427616,0.7032398043} 0.51 & \cellcolor[rgb]{0.876805309,0.9151164254999999,0.9804355785000001} 0.30 & \cellcolor[rgb]{0.61490285,0.649358983,0.8768415764999999} 0.36 & \cellcolor[rgb]{0.61490285,0.649358983,0.8768415764999999} 0.35 & \cellcolor[rgb]{0.61490285,0.649358983,0.8768415764999999} 0.10 & \cellcolor[rgb]{0.7419271648,0.8110249248,0.9874041013} 0.31 \\
D-CoCoA-G (Last Block) & \cellcolor[rgb]{0.61490285,0.649358983,0.8768415764999999} 0.19 & \cellcolor[rgb]{0.8933603506784313,0.9224036051843137,0.9699051924745098} 0.22 & \cellcolor[rgb]{0.8336264621666667,0.8895882285,0.9964796065} 0.38 & \cellcolor[rgb]{0.9842498738333334,0.8369886898862745,0.7783246280235294} 0.50 & \cellcolor[rgb]{0.876805309,0.9151164254999999,0.9804355785000001} 0.30 & \cellcolor[rgb]{0.61490285,0.649358983,0.8768415764999999} 0.36 & \cellcolor[rgb]{0.61490285,0.649358983,0.8768415764999999} 0.35 & \cellcolor[rgb]{0.876805309,0.9151164254999999,0.9804355785000001} 0.16 & \cellcolor[rgb]{0.7419271648,0.8110249248,0.9874041013} 0.31 \\
D-CoCoA-G (Last Block with prefix) & \cellcolor[rgb]{0.8336264621666667,0.8895882285,0.9964796065} 0.20 & \cellcolor[rgb]{0.9653342981666666,0.909438499827451,0.8795731953490196} 0.23 & \cellcolor[rgb]{0.61490285,0.649358983,0.8768415764999999} 0.36 & \cellcolor[rgb]{0.931695915645098,0.9325418979098039,0.9338169421313726} 0.48 & \cellcolor[rgb]{0.9659156483,0.7595427616,0.7032398043} 0.32 & \cellcolor[rgb]{0.61490285,0.649358983,0.8768415764999999} 0.36 & \cellcolor[rgb]{0.61490285,0.649358983,0.8768415764999999} 0.35 & \cellcolor[rgb]{0.7880256462666666,0.8543899393333334,0.9988772960000001} 0.14 & \cellcolor[rgb]{0.61490285,0.649358983,0.8768415764999999} 0.30 \\
D-CoCoA-G & \cellcolor[rgb]{0.9841016995,0.8604220499999999,0.8061464956666666} 0.21 & \cellcolor[rgb]{0.852836579,0.50777808,0.575116406} 0.26 & \cellcolor[rgb]{0.852836579,0.50777808,0.575116406} 0.42 & \cellcolor[rgb]{0.852836579,0.50777808,0.575116406} 0.53 & \cellcolor[rgb]{0.9659156483,0.7595427616,0.7032398043} 0.32 & \cellcolor[rgb]{0.61490285,0.649358983,0.8768415764999999} 0.36 & \cellcolor[rgb]{0.61490285,0.649358983,0.8768415764999999} 0.35 & \cellcolor[rgb]{0.9736727018,0.8973477524,0.8584952529000001} 0.19 & \cellcolor[rgb]{0.9736727018,0.8973477524,0.8584952529000001} 0.33 \\
\midrule
D-CoCoA-L (Full Trace) & \cellcolor[rgb]{0.9841016995,0.8604220499999999,0.8061464956666666} 0.21 & \cellcolor[rgb]{0.61490285,0.649358983,0.8768415764999999} 0.19 & \cellcolor[rgb]{0.61490285,0.649358983,0.8768415764999999} 0.36 & \cellcolor[rgb]{0.61490285,0.649358983,0.8768415764999999} 0.43 & \cellcolor[rgb]{0.852836579,0.50777808,0.575116406} 0.33 & \cellcolor[rgb]{0.852836579,0.50777808,0.575116406} 0.53 & \cellcolor[rgb]{0.852836579,0.50777808,0.575116406} 0.46 & \cellcolor[rgb]{0.852836579,0.50777808,0.575116406} 0.25 & \cellcolor[rgb]{0.9659156483,0.7595427616,0.7032398043} 0.34 \\
D-CoCoA-L (Last Block) & \cellcolor[rgb]{0.8336264621666667,0.8895882285,0.9964796065} 0.20 & \cellcolor[rgb]{0.61490285,0.649358983,0.8768415764999999} 0.19 & \cellcolor[rgb]{0.8336264621666667,0.8895882285,0.9964796065} 0.38 & \cellcolor[rgb]{0.8096589725941177,0.8720603673823529,0.9994654594098039} 0.46 & \cellcolor[rgb]{0.61490285,0.649358983,0.8768415764999999} 0.28 & \cellcolor[rgb]{0.852836579,0.50777808,0.575116406} 0.53 & \cellcolor[rgb]{0.852836579,0.50777808,0.575116406} 0.46 & \cellcolor[rgb]{0.9813503623666666,0.8141088755,0.7538180498} 0.21 & \cellcolor[rgb]{0.9659156483,0.7595427616,0.7032398043} 0.34 \\
D-CoCoA-L (Last Block with prefix) & \cellcolor[rgb]{0.8336264621666667,0.8895882285,0.9964796065} 0.20 & \cellcolor[rgb]{0.8015810339588235,0.8657637386764706,0.9997826392686274} 0.21 & \cellcolor[rgb]{0.7179074031529412,0.7853536515764705,0.9758586906411765} 0.37 & \cellcolor[rgb]{0.876805309,0.9151164254999999,0.9804355785000001} 0.47 & \cellcolor[rgb]{0.852836579,0.50777808,0.575116406} 0.33 & \cellcolor[rgb]{0.852836579,0.50777808,0.575116406} 0.53 & \cellcolor[rgb]{0.852836579,0.50777808,0.575116406} 0.46 & \cellcolor[rgb]{0.915574114,0.9297565972666666,0.9515550793333334} 0.17 & \cellcolor[rgb]{0.9659156483,0.7595427616,0.7032398043} 0.34 \\
D-CoCoA-L & \cellcolor[rgb]{0.852836579,0.50777808,0.575116406} 0.22 & \cellcolor[rgb]{0.9653342981666666,0.909438499827451,0.8795731953490196} 0.23 & \cellcolor[rgb]{0.9528917390058824,0.727592846082353,0.6776679419235294} 0.41 & \cellcolor[rgb]{0.931695915645098,0.9325418979098039,0.9338169421313726} 0.48 & \cellcolor[rgb]{0.61490285,0.649358983,0.8768415764999999} 0.28 & \cellcolor[rgb]{0.852836579,0.50777808,0.575116406} 0.53 & \cellcolor[rgb]{0.852836579,0.50777808,0.575116406} 0.46 & \cellcolor[rgb]{0.9385746672999999,0.6973227293333333,0.6558619128666667} 0.23 & \cellcolor[rgb]{0.852836579,0.50777808,0.575116406} 0.35 \\

\bottomrule
\end{tabular}
}\caption{\label{tab:abl_cocoa_traj_results} \textbf{Trajectory aggregation ablation.} PRR$\uparrow$ for the LLaDA-1.5 model across various tasks for D-CoCoA methods with different trajectory block-wise aggregations. Warmer color indicates better results.}\end{table*}

  \begin{table*}[!ht] \resizebox{\textwidth}{!}{\begin{tabular}{l|c|c|c|c|c|c|c|c|c}
\toprule
\textbf{UQ Method} & \textbf{XSum} & \textbf{SamSum} & \textbf{WMT14} & \textbf{WMT19} & \textbf{CoQA} & \textbf{TriviaQA} & \textbf{MMLU} & \textbf{GSM8k} & \textbf{Mean} \\
\midrule
D-CoCoA-G (Deberta Progress) & \cellcolor[rgb]{0.8336264621666667,0.8895882285,0.9964796065} 0.20 & \cellcolor[rgb]{0.852836579,0.50777808,0.575116406} 0.28 & \cellcolor[rgb]{0.852836579,0.50777808,0.575116406} 0.45 & \cellcolor[rgb]{0.9196824685392158,0.6609281104705882,0.632461990490196} 0.53 & \cellcolor[rgb]{0.9596879944529412,0.9156363617647059,0.8914368152235295} 0.28 & \cellcolor[rgb]{0.6944259359764706,0.7581492177882353,0.9606867415411765} 0.33 & \cellcolor[rgb]{0.61490285,0.649358983,0.8768415764999999} 0.34 & \cellcolor[rgb]{0.9653342981666666,0.909438499827451,0.8795731953490196} 0.18 & \cellcolor[rgb]{0.876805309,0.9151164254999999,0.9804355785000001} 0.32 \\
D-CoCoA-G (Deberta) & \cellcolor[rgb]{0.9841016995,0.8604220499999999,0.8061464956666666} 0.21 & \cellcolor[rgb]{0.852836579,0.50777808,0.575116406} 0.28 & \cellcolor[rgb]{0.61490285,0.649358983,0.8768415764999999} 0.39 & \cellcolor[rgb]{0.8096589725941177,0.8720603673823529,0.9994654594098039} 0.47 & \cellcolor[rgb]{0.9720272867117647,0.7765767393745098,0.7177742451568627} 0.30 & \cellcolor[rgb]{0.8718771013137254,0.9125626824411764,0.9828988807745098} 0.39 & \cellcolor[rgb]{0.61490285,0.649358983,0.8768415764999999} 0.34 & \cellcolor[rgb]{0.61490285,0.649358983,0.8768415764999999} 0.02 & \cellcolor[rgb]{0.61490285,0.649358983,0.8768415764999999} 0.30 \\
D-CoCoA-G (Progress) & \cellcolor[rgb]{0.61490285,0.649358983,0.8768415764999999} 0.19 & \cellcolor[rgb]{0.9720272867117647,0.7765767393745098,0.7177742451568627} 0.26 & \cellcolor[rgb]{0.852836579,0.50777808,0.575116406} 0.45 & \cellcolor[rgb]{0.852836579,0.50777808,0.575116406} 0.54 & \cellcolor[rgb]{0.9596879944529412,0.9156363617647059,0.8914368152235295} 0.28 & \cellcolor[rgb]{0.61490285,0.649358983,0.8768415764999999} 0.30 & \cellcolor[rgb]{0.6643022236588235,0.7198559149882353,0.9347936312705882} 0.35 & \cellcolor[rgb]{0.9845960255235294,0.8529178378588235,0.7968521304901961} 0.21 & \cellcolor[rgb]{0.876805309,0.9151164254999999,0.9804355785000001} 0.32 \\
D-CoCoA-G & \cellcolor[rgb]{0.9841016995,0.8604220499999999,0.8061464956666666} 0.21 & \cellcolor[rgb]{0.9720272867117647,0.7765767393745098,0.7177742451568627} 0.26 & \cellcolor[rgb]{0.931695915645098,0.9325418979098039,0.9338169421313726} 0.42 & \cellcolor[rgb]{0.9196824685392158,0.6609281104705882,0.632461990490196} 0.53 & \cellcolor[rgb]{0.852836579,0.50777808,0.575116406} 0.32 & \cellcolor[rgb]{0.7825907906117646,0.8497192224705883,0.9983175350588236} 0.36 & \cellcolor[rgb]{0.6643022236588235,0.7198559149882353,0.9347936312705882} 0.35 & \cellcolor[rgb]{0.9754778064901961,0.8934375166666666,0.8523803414019608} 0.19 & \cellcolor[rgb]{0.9736727018,0.8973477524,0.8584952529000001} 0.33 \\\midrule
D-CoCoA-L (Deberta Progress) & \cellcolor[rgb]{0.8336264621666667,0.8895882285,0.9964796065} 0.20 & \cellcolor[rgb]{0.61490285,0.649358983,0.8768415764999999} 0.19 & \cellcolor[rgb]{0.7179074031529412,0.7853536515764705,0.9758586906411765} 0.40 & \cellcolor[rgb]{0.61490285,0.649358983,0.8768415764999999} 0.44 & \cellcolor[rgb]{0.6817303976705882,0.7423918409254902,0.9505094434470589} 0.24 & \cellcolor[rgb]{0.9802450306647059,0.8081382119705882,0.7477333002470588} 0.47 & \cellcolor[rgb]{0.852836579,0.50777808,0.575116406} 0.46 & \cellcolor[rgb]{0.852836579,0.50777808,0.575116406} 0.30 & \cellcolor[rgb]{0.9659156483,0.7595427616,0.7032398043} 0.34 \\
D-CoCoA-L (Deberta) & \cellcolor[rgb]{0.9841016995,0.8604220499999999,0.8061464956666666} 0.21 & \cellcolor[rgb]{0.7554121621254901,0.8246983074117646,0.9925393881882353} 0.21 & \cellcolor[rgb]{0.931695915645098,0.9325418979098039,0.9338169421313726} 0.42 & \cellcolor[rgb]{0.7419271648,0.8110249248,0.9874041013} 0.46 & \cellcolor[rgb]{0.61490285,0.649358983,0.8768415764999999} 0.23 & \cellcolor[rgb]{0.9367011412764705,0.6934798195294117,0.6531662319882353} 0.50 & \cellcolor[rgb]{0.852836579,0.50777808,0.575116406} 0.46 & \cellcolor[rgb]{0.8790560841823529,0.5840607685176471,0.5944135352882353} 0.29 & \cellcolor[rgb]{0.852836579,0.50777808,0.575116406} 0.35 \\
D-CoCoA-L (Progress) & \cellcolor[rgb]{0.61490285,0.649358983,0.8768415764999999} 0.19 & \cellcolor[rgb]{0.61490285,0.649358983,0.8768415764999999} 0.19 & \cellcolor[rgb]{0.7179074031529412,0.7853536515764705,0.9758586906411765} 0.40 & \cellcolor[rgb]{0.61490285,0.649358983,0.8768415764999999} 0.44 & \cellcolor[rgb]{0.7554121621254901,0.8246983074117646,0.9925393881882353} 0.25 & \cellcolor[rgb]{0.9560162876490197,0.7348397910862745,0.6832824522294118} 0.49 & \cellcolor[rgb]{0.852836579,0.50777808,0.575116406} 0.46 & \cellcolor[rgb]{0.852836579,0.50777808,0.575116406} 0.30 & \cellcolor[rgb]{0.9659156483,0.7595427616,0.7032398043} 0.34 \\
D-CoCoA-L & \cellcolor[rgb]{0.852836579,0.50777808,0.575116406} 0.22 & \cellcolor[rgb]{0.9024823794117647,0.9258330802784314,0.9630825372156863} 0.23 & \cellcolor[rgb]{0.8336264621666667,0.8895882285,0.9964796065} 0.41 & \cellcolor[rgb]{0.876805309,0.9151164254999999,0.9804355785000001} 0.48 & \cellcolor[rgb]{0.9596879944529412,0.9156363617647059,0.8914368152235295} 0.28 & \cellcolor[rgb]{0.852836579,0.50777808,0.575116406} 0.53 & \cellcolor[rgb]{0.852836579,0.50777808,0.575116406} 0.46 & \cellcolor[rgb]{0.9783266054882354,0.7990169113588235,0.7386511461764707} 0.23 & \cellcolor[rgb]{0.852836579,0.50777808,0.575116406} 0.35 \\

\bottomrule
\end{tabular}
}\caption{\label{tab:abl_cocoa_sim_results} \textbf{Similarity function ablation.} PRR$\uparrow$ for the LLaDA-1.5 model across various tasks for D-CoCoA methods with different similarity functions (DeBERTa NLI and Cross-Encoder with RoBERTa) and weighted or unweighted aggregation. Warmer color indicates better results.}\end{table*}
  \begin{table*}[!ht] \resizebox{\textwidth}{!}{\begin{tabular}{l|c|c|c|c|c|c|c|c|c}
\toprule
\textbf{UQ Method} & \textbf{XSum} & \textbf{SamSum} & \textbf{WMT14} & \textbf{WMT19} & \textbf{CoQA} & \textbf{TriviaQA} & \textbf{MMLU} & \textbf{GSM8k} & \textbf{Mean} \\
\midrule
D-CoCoA-G (w/o NFE) & \cellcolor[rgb]{0.61490285,0.649358983,0.8768415764999999} 0.19 & \cellcolor[rgb]{0.9528917390058824,0.727592846082353,0.6776679419235294} 0.25 & \cellcolor[rgb]{0.61490285,0.649358983,0.8768415764999999} 0.37 & \cellcolor[rgb]{0.9808223691882353,0.8790145912705882,0.830891189582353} 0.49 & \cellcolor[rgb]{0.7419271648,0.8110249248,0.9874041013} 0.28 & \cellcolor[rgb]{0.61490285,0.649358983,0.8768415764999999} 0.28 & \cellcolor[rgb]{0.61490285,0.649358983,0.8768415764999999} 0.35 & \cellcolor[rgb]{0.61490285,0.649358983,0.8768415764999999} -0.01 & \cellcolor[rgb]{0.61490285,0.649358983,0.8768415764999999} 0.28 \\
D-CoCoA-G & \cellcolor[rgb]{0.9841016995,0.8604220499999999,0.8061464956666666} 0.21 & \cellcolor[rgb]{0.852836579,0.50777808,0.575116406} 0.26 & \cellcolor[rgb]{0.852836579,0.50777808,0.575116406} 0.42 & \cellcolor[rgb]{0.852836579,0.50777808,0.575116406} 0.53 & \cellcolor[rgb]{0.852836579,0.50777808,0.575116406} 0.32 & \cellcolor[rgb]{0.8230564053823529,0.8822182482647059,0.9984342312529412} 0.36 & \cellcolor[rgb]{0.61490285,0.649358983,0.8768415764999999} 0.35 & \cellcolor[rgb]{0.9848414898333333,0.8452419653686274,0.7875691806823529} 0.19 & \cellcolor[rgb]{0.9836582578333333,0.8287354144039216,0.7690800753647059} 0.33 \\\midrule
D-CoCoA-L (with NFE) & \cellcolor[rgb]{0.9841016995,0.8604220499999999,0.8061464956666666} 0.21 & \cellcolor[rgb]{0.61490285,0.649358983,0.8768415764999999} 0.20 & \cellcolor[rgb]{0.61490285,0.649358983,0.8768415764999999} 0.37 & \cellcolor[rgb]{0.61490285,0.649358983,0.8768415764999999} 0.42 & \cellcolor[rgb]{0.61490285,0.649358983,0.8768415764999999} 0.27 & \cellcolor[rgb]{0.9497671903627452,0.7203459010784313,0.6720534316176471} 0.49 & \cellcolor[rgb]{0.852836579,0.50777808,0.575116406} 0.46 & \cellcolor[rgb]{0.852836579,0.50777808,0.575116406} 0.28 & \cellcolor[rgb]{0.9423217193470588,0.7050085489411765,0.6612532746235295} 0.34 \\
D-CoCoA-L & \cellcolor[rgb]{0.852836579,0.50777808,0.575116406} 0.22 & \cellcolor[rgb]{0.931695915645098,0.9325418979098039,0.9338169421313726} 0.23 & \cellcolor[rgb]{0.9659156483,0.7595427616,0.7032398043} 0.41 & \cellcolor[rgb]{0.9547297988764706,0.919693239882353,0.9001656762117647} 0.48 & \cellcolor[rgb]{0.7419271648,0.8110249248,0.9874041013} 0.28 & \cellcolor[rgb]{0.852836579,0.50777808,0.575116406} 0.53 & \cellcolor[rgb]{0.852836579,0.50777808,0.575116406} 0.46 & \cellcolor[rgb]{0.9560162876490197,0.7348397910862745,0.6832824522294118} 0.23 & \cellcolor[rgb]{0.852836579,0.50777808,0.575116406} 0.35 \\

\bottomrule
\end{tabular}
}\caption{\label{tab:abl_cocoa_nfe_results} \textbf{NFE component ablation.} PRR$\uparrow$ for the LLaDA-1.5 model across various tasks for D-CoCoA methods with and without the NFE component. Warmer color indicates better results.}\end{table*}

\section{Analysis}
\label{app:analyses}

  To further characterize D-CoCoA, we conduct four additional experiments on LLaDA-1.5: a hybrid variant with sampled generations, a comparison with generative perplexity from GPT-2~\citep{radford2019language}, a calibration analysis by expected calibration error (ECE; \citealp{pmlr-v70-guo17a}), and a hardware-independent estimate of the computational cost.

  \ecoparagraph{D-CoCoA hybrid with sampled generations.} Table~\ref{tab:abl_cocoa_hybrid_results} evaluates D-CoCoA-L-Hybrid, which combines the trajectory dissimilarity with the consistency over $M$ independently sampled generations:
  $$\uv_{\mathrm{D\text{-}CoCoA\text{-}L\text{-}Hybrid}}(\xv,\yv,M) = \uv_{\mathrm{CommitNLL}}(\xv,\yv) \cdot \big(\tfrac{1}{2}\uv_{\mathrm{AD\text{-}Block}}(\xv,\yv) + \tfrac{1}{2}\uv_{\mathrm{cons}}(\xv,\yv,M)\big)$$ 
  A single sampled generation already improves the average performance of D-CoCoA-L, with $M \ge 2$ the hybrid matches CoCoA-MCNLL, and with $M=10$ it slightly outperforms it. The largest gain is achieved on GSM8k, where PRR increases from 0.23 to 0.31. However, these gains come at the cost of additional generations, so D-CoCoA-L remains preferable for LLDMs.

  \ecoparagraph{Generative perplexity.} Table~\ref{tab:abl_genppl_results} compares D-CoCoA-L with the generative perplexity (Gen-PPL) of the GPT-2 model on answers generated by LLDMs. Gen-PPL performs close to random, with a mean PRR of 0.05 compared with 0.35 for D-CoCoA-L, and even ranks errors in the wrong direction on three datasets. Its Spearman correlation with the generation quality is weak on most datasets and moderate only for translation (WMT14: $\rho=-0.307$, WMT19: $\rho=-0.362$), where fluency is an important component of quality. This is expected, since Gen-PPL reflects how fluent the answer is rather than whether it is consistent with the input or the reference.

  \ecoparagraph{Calibration.} Table~\ref{tab:abl_ece_results} presents the ECE of the best method from each category, computed after min-max normalizing the uncertainty scores to $[0,1]$ on each dataset. Diffusion-specific methods provide the best overall calibration: CommitNLL achieves the lowest average ECE, and D-CoCoA-L is on par with the best-calibrated baseline, SAR. D-CoCoA-L performs best on XSum, while D-CoCoA-G performs best on WMT14 and WMT19.

  \ecoparagraph{Computational cost.} Table~\ref{tab:abl_tflops_results} presents the additional TFLOPs per instance, measured with PyTorch FlopCounterMode~\citep{NEURIPS2019_bdbca288} on TriviaQA, GSM8k, and SamSum and excluding the base generation. For $M$ sampled generations, KLE requires $M$ LLDM generations and $O(M^2)$ NLI comparisons; CoCoA-MCNLL requires $M$ LLDM generations, $O(M)$ cross-encoder comparisons, and the MC masked likelihood. In contrast, CommitNLL requires no additional forward passes, while D-CoCoA-L adds only $O(T)$ cross-encoder evaluations, where $T$ is the number of denoising steps. As a result, D-CoCoA-L requires 100--500 times fewer additional TFLOPs than KLE, which confirms that its efficiency does not depend on the hardware used for runtime measurements.

  \begin{table*}[!ht] \resizebox{\textwidth}{!}{\begin{tabular}{l|c|c|c|c|c|c|c|c|c}
\toprule
\textbf{UQ Method} & \textbf{XSum} & \textbf{SamSum} & \textbf{WMT14} & \textbf{WMT19} & \textbf{CoQA} & \textbf{TriviaQA} & \textbf{MMLU} & \textbf{GSM8k} & \textbf{Mean} \\
\midrule
D-CoCoA-L & \cellcolor[rgb]{0.61490285,0.649358983,0.8768415764999999} 0.22 & \cellcolor[rgb]{0.852836579,0.50777808,0.575116406} 0.23 & \cellcolor[rgb]{0.61490285,0.649358983,0.8768415764999999} 0.41 & \cellcolor[rgb]{0.61490285,0.649358983,0.8768415764999999} 0.48 & \cellcolor[rgb]{0.7771559349568627,0.8450485056078432,0.9977577741176471} 0.28 & \cellcolor[rgb]{0.9736727018,0.8973477524,0.8584952529000001} 0.53 & \cellcolor[rgb]{0.9836582578333333,0.8287354144039216,0.7690800753647059} 0.46 & \cellcolor[rgb]{0.61490285,0.649358983,0.8768415764999999} 0.23 & \cellcolor[rgb]{0.61490285,0.649358983,0.8768415764999999} 0.35 \\
D-CoCoA-L-Hybrid ($M$=1) & \cellcolor[rgb]{0.852836579,0.50777808,0.575116406} 0.23 & \cellcolor[rgb]{0.61490285,0.649358983,0.8768415764999999} 0.21 & \cellcolor[rgb]{0.876805309,0.9151164254999999,0.9804355785000001} 0.45 & \cellcolor[rgb]{0.9596879944529412,0.9156363617647059,0.8914368152235295} 0.53 & \cellcolor[rgb]{0.61490285,0.649358983,0.8768415764999999} 0.27 & \cellcolor[rgb]{0.9736727018,0.8973477524,0.8584952529000001} 0.53 & \cellcolor[rgb]{0.9653342981666666,0.909438499827451,0.8795731953490196} 0.45 & \cellcolor[rgb]{0.852836579,0.50777808,0.575116406} 0.31 & \cellcolor[rgb]{0.9841016995,0.8604220499999999,0.8061464956666666} 0.37 \\
D-CoCoA-L-Hybrid ($M$=2) & \cellcolor[rgb]{0.852836579,0.50777808,0.575116406} 0.23 & \cellcolor[rgb]{0.61490285,0.649358983,0.8768415764999999} 0.21 & \cellcolor[rgb]{0.9337138175431372,0.9321882998862745,0.9313012310098039} 0.46 & \cellcolor[rgb]{0.9841016995,0.8604220499999999,0.8061464956666666} 0.54 & \cellcolor[rgb]{0.7771559349568627,0.8450485056078432,0.9977577741176471} 0.28 & \cellcolor[rgb]{0.9659156483,0.7595427616,0.7032398043} 0.54 & \cellcolor[rgb]{0.8933603506784313,0.9224036051843137,0.9699051924745098} 0.44 & \cellcolor[rgb]{0.852836579,0.50777808,0.575116406} 0.31 & \cellcolor[rgb]{0.852836579,0.50777808,0.575116406} 0.38 \\
D-CoCoA-L-Hybrid ($M$=5) & \cellcolor[rgb]{0.852836579,0.50777808,0.575116406} 0.23 & \cellcolor[rgb]{0.61490285,0.649358983,0.8768415764999999} 0.21 & \cellcolor[rgb]{0.876805309,0.9151164254999999,0.9804355785000001} 0.45 & \cellcolor[rgb]{0.9841016995,0.8604220499999999,0.8061464956666666} 0.54 & \cellcolor[rgb]{0.931695915645098,0.9325418979098039,0.9338169421313726} 0.29 & \cellcolor[rgb]{0.852836579,0.50777808,0.575116406} 0.55 & \cellcolor[rgb]{0.9423217193470588,0.7050085489411765,0.6612532746235295} 0.47 & \cellcolor[rgb]{0.852836579,0.50777808,0.575116406} 0.31 & \cellcolor[rgb]{0.852836579,0.50777808,0.575116406} 0.38 \\
D-CoCoA-L-Hybrid ($M$=10) & \cellcolor[rgb]{0.852836579,0.50777808,0.575116406} 0.23 & \cellcolor[rgb]{0.931695915645098,0.9325418979098039,0.9338169421313726} 0.22 & \cellcolor[rgb]{0.9337138175431372,0.9321882998862745,0.9313012310098039} 0.46 & \cellcolor[rgb]{0.9720272867117647,0.7765767393745098,0.7177742451568627} 0.55 & \cellcolor[rgb]{0.931695915645098,0.9325418979098039,0.9338169421313726} 0.29 & \cellcolor[rgb]{0.852836579,0.50777808,0.575116406} 0.55 & \cellcolor[rgb]{0.852836579,0.50777808,0.575116406} 0.48 & \cellcolor[rgb]{0.852836579,0.50777808,0.575116406} 0.31 & \cellcolor[rgb]{0.852836579,0.50777808,0.575116406} 0.38 \\\midrule
CoCoA-MCNLL & \cellcolor[rgb]{0.852836579,0.50777808,0.575116406} 0.23 & \cellcolor[rgb]{0.61490285,0.649358983,0.8768415764999999} 0.21 & \cellcolor[rgb]{0.852836579,0.50777808,0.575116406} 0.51 & \cellcolor[rgb]{0.852836579,0.50777808,0.575116406} 0.57 & \cellcolor[rgb]{0.852836579,0.50777808,0.575116406} 0.31 & \cellcolor[rgb]{0.61490285,0.649358983,0.8768415764999999} 0.50 & \cellcolor[rgb]{0.61490285,0.649358983,0.8768415764999999} 0.41 & \cellcolor[rgb]{0.9797588292352941,0.8834864272549019,0.8370723575196078} 0.28 & \cellcolor[rgb]{0.852836579,0.50777808,0.575116406} 0.38 \\

\bottomrule
\end{tabular}
}\caption{\label{tab:abl_cocoa_hybrid_results} \textbf{Hybrid consistency ablation.} PRR$\uparrow$ for the LLaDA-1.5 model across various tasks for D-CoCoA-L-Hybrid, which complements the trajectory dissimilarity of D-CoCoA-L with the consistency over $M$ independently sampled generations, compared with D-CoCoA-L and the sampling-based CoCoA-MCNLL. Warmer color indicates better results.}\end{table*}

  \begin{table*}[!ht] \resizebox{\textwidth}{!}{\begin{tabular}{l|c|c|c|c|c|c|c|c|c}
\toprule
\textbf{UQ Method} & \textbf{XSum} & \textbf{SamSum} & \textbf{WMT14} & \textbf{WMT19} & \textbf{CoQA} & \textbf{TriviaQA} & \textbf{MMLU} & \textbf{GSM8k} & \textbf{Mean} \\
\midrule
Gen-PPL (GPT-2) & 0.01 & -0.08 & 0.36 & 0.15 & -0.01 & 0.05 & 0.03 & -0.16 & 0.05 \\
D-CoCoA-L & \textbf{0.22} & \textbf{0.23} & \textbf{0.41} & \textbf{0.48} & \textbf{0.28} & \textbf{0.53} & \textbf{0.46} & \textbf{0.23} & \textbf{0.35} \\

\bottomrule
\end{tabular}
}\caption{\label{tab:abl_genppl_results} \textbf{Generative perplexity comparison.} PRR$\uparrow$ for the LLaDA-1.5 model across various tasks for D-CoCoA-L and the generative perplexity (Gen-PPL) of the GPT-2 model computed on the generated answers. Best results are in bold.}\end{table*}

  \begin{table*}[!ht] \resizebox{\textwidth}{!}{\begin{tabular}{ll|c|c|c|c|c|c|c|c|c}
\toprule
\textbf{Category} & \textbf{UQ Method} & \textbf{XSum} & \textbf{SamSum} & \textbf{WMT14} & \textbf{WMT19} & \textbf{CoQA} & \textbf{TriviaQA} & \textbf{MMLU} & \textbf{GSM8k} & \textbf{Mean} \\
\midrule
\multirow{5}{*}{\shortstack[l]{AR\\baselines}} & MSP & \cellcolor[rgb]{0.9591408362921569,0.742086736090196,0.6888969625352941} 0.13 & \cellcolor[rgb]{0.8871684250764706,0.5998796340235294,0.6012672772176471} 0.06 & \cellcolor[rgb]{0.9841016995,0.8604220499999999,0.8061464956666666} 0.14 & \cellcolor[rgb]{0.9791396989627451,0.8021675484411765,0.7416485506941177} 0.14 & \cellcolor[rgb]{0.8015810339588235,0.8657637386764706,0.9997826392686274} 0.19 & \cellcolor[rgb]{0.6892991246352941,0.7519281085960785,0.9568458054235294} 0.46 & \cellcolor[rgb]{0.9818859091411765,0.8745427552862746,0.824710021645098} 0.28 & \cellcolor[rgb]{0.9634414899941177,0.752710773145098,0.6974329388568627} 0.12 & \cellcolor[rgb]{0.9772829111803922,0.8895272809333333,0.8462654299039215} 0.19 \\
 & SAR & \cellcolor[rgb]{0.8790560841823529,0.5840607685176471,0.5944135352882353} 0.05 & \cellcolor[rgb]{0.8283414337745099,0.8859032383823529,0.9974569188764706} 0.31 & \cellcolor[rgb]{0.9423217193470588,0.7050085489411765,0.6612532746235295} 0.10 & \cellcolor[rgb]{0.924020201182353,0.6691400189882353,0.6376028197294118} 0.10 & \cellcolor[rgb]{0.9841016995,0.8604220499999999,0.8061464956666666} 0.11 & \cellcolor[rgb]{0.9276891842254902,0.9318890695490196,0.9382935886568627} 0.28 & \cellcolor[rgb]{0.852836579,0.50777808,0.575116406} 0.07 & \cellcolor[rgb]{0.9846442845,0.8424908735411765,0.7844876631294118} 0.18 & \cellcolor[rgb]{0.9053078374137256,0.6343985308588236,0.6177138057666667} 0.15 \\
 & CoCoA-MTE & \cellcolor[rgb]{0.8898725387078432,0.6051525891921569,0.6035518578607844} 0.06 & \cellcolor[rgb]{0.852836579,0.50777808,0.575116406} 0.04 & \cellcolor[rgb]{0.9756268974411765,0.7893996947941176,0.729703902882353} 0.12 & \cellcolor[rgb]{0.9696268857588235,0.769790744282353,0.7119501024647059} 0.13 & \cellcolor[rgb]{0.61490285,0.649358983,0.8768415764999999} 0.25 & \cellcolor[rgb]{0.61490285,0.649358983,0.8768415764999999} 0.52 & \cellcolor[rgb]{0.9818859091411765,0.8745427552862746,0.824710021645098} 0.28 & \cellcolor[rgb]{0.9560162876490197,0.7348397910862745,0.6832824522294118} 0.11 & \cellcolor[rgb]{0.9772829111803922,0.8895272809333333,0.8462654299039215} 0.19 \\
 & KLE & \cellcolor[rgb]{0.61490285,0.649358983,0.8768415764999999} 0.58 & \cellcolor[rgb]{0.61490285,0.649358983,0.8768415764999999} 0.44 & \cellcolor[rgb]{0.9841016995,0.8604220499999999,0.8061464956666666} 0.14 & \cellcolor[rgb]{0.9653342981666666,0.909438499827451,0.8795731953490196} 0.19 & \cellcolor[rgb]{0.8933603506784313,0.9224036051843137,0.9699051924745098} 0.16 & \cellcolor[rgb]{0.9834812190705882,0.867835001309804,0.8154382697392157} 0.20 & \cellcolor[rgb]{0.9720272867117647,0.7765767393745098,0.7177742451568627} 0.20 & \cellcolor[rgb]{0.9813503623666666,0.8141088755,0.7538180498} 0.16 & \cellcolor[rgb]{0.6594161915960784,0.7133025255607843,0.9299287241019608} 0.26 \\
 & TAD & \cellcolor[rgb]{0.852836579,0.50777808,0.575116406} 0.03 & \cellcolor[rgb]{0.707400451427451,0.7734367635137256,0.9695437661686275} 0.38 & \cellcolor[rgb]{0.61490285,0.649358983,0.8768415764999999} 0.28 & \cellcolor[rgb]{0.61490285,0.649358983,0.8768415764999999} 0.35 & \cellcolor[rgb]{0.6426363887647059,0.690064711317647,0.9117342756235294} 0.24 & \cellcolor[rgb]{0.852836579,0.50777808,0.575116406} 0.03 & \cellcolor[rgb]{0.9818859091411765,0.8745427552862746,0.824710021645098} 0.28 & \cellcolor[rgb]{0.61490285,0.649358983,0.8768415764999999} 0.54 & \cellcolor[rgb]{0.61490285,0.649358983,0.8768415764999999} 0.27 \\\midrule
\multirow{4}{*}{\shortstack[l]{Diffusion\\signals}} & MCNLL & \cellcolor[rgb]{0.9848414898333333,0.8452419653686274,0.7875691806823529} 0.20 & \cellcolor[rgb]{0.9196757213862745,0.930583412827451,0.9472468817078432} 0.25 & \cellcolor[rgb]{0.9781854635254902,0.8875721630666666,0.8432079741549019} 0.15 & \cellcolor[rgb]{0.9754778064901961,0.8934375166666666,0.8523803414019608} 0.18 & \cellcolor[rgb]{0.852836579,0.50777808,0.575116406} 0.04 & \cellcolor[rgb]{0.9802905992117648,0.8812505092627452,0.8339817735509804} 0.21 & \cellcolor[rgb]{0.9717157648333333,0.9011381268078431,0.8645857989568628} 0.31 & \cellcolor[rgb]{0.9175136022176471,0.6568221562117647,0.6298915758705882} 0.07 & \cellcolor[rgb]{0.9846442845,0.8424908735411765,0.7844876631294118} 0.18 \\
 & NFE & \cellcolor[rgb]{0.8177369112294117,0.8783569960882354,0.9991482795509804} 0.41 & \cellcolor[rgb]{0.8096589725941177,0.8720603673823529,0.9994654594098039} 0.32 & \cellcolor[rgb]{0.9175136022176471,0.6568221562117647,0.6298915758705882} 0.09 & \cellcolor[rgb]{0.9575785619705882,0.7384632635882353,0.6860897073823529} 0.12 & \cellcolor[rgb]{0.8644847897843138,0.9087320678529411,0.9865938341862746} 0.17 & \cellcolor[rgb]{0.9802905992117648,0.8812505092627452,0.8339817735509804} 0.21 & \cellcolor[rgb]{0.61490285,0.649358983,0.8768415764999999} 0.66 & \cellcolor[rgb]{0.9283579338254901,0.6773519275058824,0.6427436489686275} 0.08 & \cellcolor[rgb]{0.6594161915960784,0.7133025255607843,0.9299287241019608} 0.26 \\
 & AD-Full & \cellcolor[rgb]{0.7527113214117647,0.8219973367843137,0.9915787156372549} 0.46 & \cellcolor[rgb]{0.876805309,0.9151164254999999,0.9804355785000001} 0.28 & \cellcolor[rgb]{0.9756268974411765,0.7893996947941176,0.729703902882353} 0.12 & \cellcolor[rgb]{0.9575785619705882,0.7384632635882353,0.6860897073823529} 0.12 & \cellcolor[rgb]{0.9836582578333333,0.8287354144039216,0.7690800753647059} 0.10 & \cellcolor[rgb]{0.9763803588352942,0.8914823988,0.8493228856529411} 0.22 & \cellcolor[rgb]{0.9839369241588236,0.8629234540470588,0.8092446173921568} 0.27 & \cellcolor[rgb]{0.9846442845,0.8424908735411765,0.7844876631294118} 0.18 & \cellcolor[rgb]{0.866948893627451,0.9100089393823529,0.9853621830490196} 0.22 \\
 & CommitNLL & \cellcolor[rgb]{0.9591408362921569,0.742086736090196,0.6888969625352941} 0.13 & \cellcolor[rgb]{0.9459084460607843,0.9259866385215686,0.914542637127451} 0.23 & \cellcolor[rgb]{0.8871684250764706,0.5998796340235294,0.6012672772176471} 0.08 & \cellcolor[rgb]{0.9028614815235294,0.6299065681294118,0.6152808287} 0.09 & \cellcolor[rgb]{0.9423217193470588,0.7050085489411765,0.6612532746235295} 0.07 & \cellcolor[rgb]{0.815044265017647,0.8762581198529411,0.9992540061705882} 0.37 & \cellcolor[rgb]{0.8898725387078432,0.6051525891921569,0.6035518578607844} 0.10 & \cellcolor[rgb]{0.852836579,0.50777808,0.575116406} 0.02 & \cellcolor[rgb]{0.852836579,0.50777808,0.575116406} 0.14 \\\midrule
\multirow{2}{*}{\shortstack[l]{AR +\\MCNLL}} & CoCoA-MCNLL & \cellcolor[rgb]{0.8898725387078432,0.6051525891921569,0.6035518578607844} 0.06 & \cellcolor[rgb]{0.9824556940686274,0.8200795390294118,0.7599027993529412} 0.15 & \cellcolor[rgb]{0.9175136022176471,0.6568221562117647,0.6298915758705882} 0.09 & \cellcolor[rgb]{0.9028614815235294,0.6299065681294118,0.6152808287} 0.09 & \cellcolor[rgb]{0.9196757213862745,0.930583412827451,0.9472468817078432} 0.15 & \cellcolor[rgb]{0.8283414337745099,0.8859032383823529,0.9974569188764706} 0.36 & \cellcolor[rgb]{0.9717157648333333,0.9011381268078431,0.8645857989568628} 0.31 & \cellcolor[rgb]{0.9283579338254901,0.6773519275058824,0.6427436489686275} 0.08 & \cellcolor[rgb]{0.947942297417647,0.7165372783529411,0.6693403172588235} 0.16 \\
 & CoCoA-MCNLL-Norm & \cellcolor[rgb]{0.8790560841823529,0.5840607685176471,0.5944135352882353} 0.05 & \cellcolor[rgb]{0.8704786593764706,0.5611201626352941,0.5878720995529412} 0.05 & \cellcolor[rgb]{0.9756268974411765,0.7893996947941176,0.729703902882353} 0.12 & \cellcolor[rgb]{0.9696268857588235,0.769790744282353,0.7119501024647059} 0.13 & \cellcolor[rgb]{0.7022106452470588,0.7673217452235295,0.966000956317647} 0.22 & \cellcolor[rgb]{0.7581301512705882,0.8272488052941176,0.9932036999058824} 0.41 & \cellcolor[rgb]{0.9717157648333333,0.9011381268078431,0.8645857989568628} 0.31 & \cellcolor[rgb]{0.9696268857588235,0.769790744282353,0.7119501024647059} 0.13 & \cellcolor[rgb]{0.9846442845,0.8424908735411765,0.7844876631294118} 0.18 \\\midrule
\multirow{2}{*}{\shortstack[l]{\textbf{D-CoCoA}\\\textbf{(ours)}}} & D-CoCoA-G & \cellcolor[rgb]{0.9756268974411765,0.7893996947941176,0.729703902882353} 0.16 & \cellcolor[rgb]{0.9844312501823529,0.8554192419058824,0.7999502522156863} 0.17 & \cellcolor[rgb]{0.852836579,0.50777808,0.575116406} 0.07 & \cellcolor[rgb]{0.852836579,0.50777808,0.575116406} 0.07 & \cellcolor[rgb]{0.6716387617176471,0.7296768173647059,0.9420609608117647} 0.23 & \cellcolor[rgb]{0.7446232039529412,0.8137680277960784,0.9884477548843138} 0.42 & \cellcolor[rgb]{0.9666105915000001,0.9077784252235295,0.8765757160705883} 0.32 & \cellcolor[rgb]{0.9560162876490197,0.7348397910862745,0.6832824522294118} 0.11 & \cellcolor[rgb]{0.9772829111803922,0.8895272809333333,0.8462654299039215} 0.19 \\
 & D-CoCoA-L & \cellcolor[rgb]{0.852836579,0.50777808,0.575116406} 0.03 & \cellcolor[rgb]{0.9563825307352941,0.9183409471764705,0.8972560558823529} 0.22 & \cellcolor[rgb]{0.8871684250764706,0.5998796340235294,0.6012672772176471} 0.08 & \cellcolor[rgb]{0.9028614815235294,0.6299065681294118,0.6152808287} 0.09 & \cellcolor[rgb]{0.9653342981666666,0.909438499827451,0.8795731953490196} 0.13 & \cellcolor[rgb]{0.7717200969960785,0.8400012947058824,0.9965252584941177} 0.40 & \cellcolor[rgb]{0.9385746672999999,0.6973227293333333,0.6558619128666667} 0.15 & \cellcolor[rgb]{0.9560162876490197,0.7348397910862745,0.6832824522294118} 0.11 & \cellcolor[rgb]{0.9053078374137256,0.6343985308588236,0.6177138057666667} 0.15 \\
\bottomrule
\end{tabular}
}\caption{\label{tab:abl_ece_results} \textbf{Calibration results.} ECE$\downarrow$ for the LLaDA-1.5 model across various tasks for the best method from each category. Uncertainty scores are min-max normalized to $[0,1]$ over the test instances of each dataset. Warmer color indicates better results.}\end{table*}

  \begin{table*}[!ht] \centering\resizebox{0.7\textwidth}{!}{
\begin{tabular}{l|c|c|c}
\toprule
\textbf{UQ Method} & \textbf{TriviaQA} & \textbf{GSM8k} & \textbf{SamSum} \\
\midrule
KLE & 94.85 $\pm$ 12.94 & 1870.26 $\pm$ 134.23 & 471.34 $\pm$ 203.44 \\
CoCoA-MCNLL & 153.02 $\pm$ 15.27 & 2112.06 $\pm$ 147.40 & 530.51 $\pm$ 237.63 \\
\midrule
D-CoCoA-L & \textbf{0.207 $\pm$ 0.226} & \textbf{3.886 $\pm$ 1.436} & \textbf{3.573 $\pm$ 2.520} \\

\bottomrule
\end{tabular}
}\caption{\label{tab:abl_tflops_results} \textbf{Computational overhead.} Additional TFLOPs$\downarrow$ per instance (mean $\pm$ standard deviation over 10 instances) introduced by UQ methods for the LLaDA-1.5 model, measured with PyTorch FlopCounterMode; the cost of the base generation is excluded. Best results are in bold.}\end{table*}

\section{Additional Details of Experimental Setup}
\label{app:exp_setup}

\subsection{Implementation Details}
\label{app:implementation}
  We evaluate four instruction-tuned LLDMs: the masked diffusion models LLaDA-1.5 (8B parameters)\footnote{\url{https://huggingface.co/GSAI-ML/LLaDA-1.5}}~\citep{zhu2025llada}, Dream-v0-Instruct-7B\footnote{\url{https://huggingface.co/Dream-org/Dream-v0-Instruct-7B}}~\citep{ye2025dream}, and ReFusion (8B parameters)\footnote{\url{https://huggingface.co/GSAI-ML/ReFusion}}~\citep{li2026refusion}, as well as the uniform diffusion model DiffusionGemma-26B-A4B-it (26B parameters with 4B active)\footnote{\url{https://huggingface.co/google/diffusiongemma-26B-A4B-it}}~\citep{DiffusionGemma}. To reflect practical inference conditions, generation is performed using KV cache and Parallel Decoding~\citep{wu2025fastdllmtrainingfreeaccelerationdiffusion}, following the default configuration of FastDLLM, except for DiffusionGemma, which uses its native generation procedure with the default configuration. We use a factor-based parallel decoding strategy with a factor of $1$ for the other models. Detailed generation hyperparameters are presented in Table~\ref{tab:datasets}.

  For semantic evaluation, we use two pretrained discriminative models: RoBERTa-large\footnote{\url{https://huggingface.co/cross-encoder/stsb-roberta-large}}, which produces a continuous similarity score in $[0, 1]$, and DeBERTa-large\footnote{\url{https://huggingface.co/microsoft/deberta-large-mnli}}, from which we extract the softmax probability assigned to the entailment class. RoBERTa-large serves as the default similarity function in the proposed D-CoCoA methods. Both models are used in inference-only mode without fine-tuning.

  All supervised methods are evaluated in an in-domain setup, where uncertainty predictors are trained on $1000$ instances ($500$ for GSM8k and XSum) from the training split of the same dataset. This establishes an upper bound on supervised performance. In practice, such methods tend to degrade substantially in out-of-domain scenarios~\citep{vazhentsev-etal-2025-unconditional}.

  For all uses of MCNLL and MCNLL-Norm, both as standalone methods and as components of other methods, we use $N_{mc}=16$. While previous work~\citep{llada} suggests using $N_{mc}=128$, we observe that it usually does not introduce additional performance gains, while requiring substantially more runtime. For MMLU, we do not perform masking since only one token is generated.

  All experiments are conducted on NVIDIA A100 40GB, RTX 6000 48GB, and RTX PRO 6000 96GB GPUs using single-batch inference. Figures~\ref{fig:llada_prr_vs_comp}, \ref{fig:dream_prr_vs_comp}, \ref{fig:refusion_prr_vs_comp}, and~\ref{fig:diffusiongemma_prr_vs_comp} report the average runtime overhead per instance for each UQ method relative to standard LLM inference without UQ. Figure~\ref{fig:avg_prr_vs_comp} reports the same metric averaged across datasets.

\subsection{Detailed Description of UQ Baselines}
\label{app:baselines}
  We evaluate eighteen unsupervised and four supervised UQ baselines, originally developed for standard autoregressive LLMs, on diffusion LLMs.

  We evaluate simple unsupervised baselines that rely on the probability distributions of the tokens in the final answer. These include Maximum Sequence Probability (MSP), Perplexity, and Mean Token Entropy (MTE; \citealp{fomicheva-etal-2020-unsupervised}). For LLDMs, these probabilities are derived from the tokens at the final denoising step. However, most diffusion models are not trained to provide sequence-level probabilities, which limits the effectiveness of these baselines in capturing uncertainty.

  Among single forward-pass state-of-the-art baselines for white-box LLMs, we evaluate Claim-Conditioned Probability (CCP; \citealp{fadeeva-etal-2024-fact}), which estimates token-level uncertainty based on the similarity of the top-$k$ predicted tokens in the output distribution; Attention Score~\citep{NEURIPS2024_LLM}, which quantifies uncertainty using the sum of the logarithms of the eigenvalues of the attention matrix, i.e., its log-determinant; and Recurrent Attention-based Uncertainty Quantification (RAUQ; \citealp{rauq}), which combines the attention from each generated token to the last preceding token with its probability.

  Among sampling-based state-of-the-art baselines for white-box LLMs, we evaluate Monte Carlo Normalized Sequence Entropy (MCNSE; \citealp{malinin2020uncertainty}), which estimates the entropy of the sequence-level predictive distribution from sampled generations; Semantic Entropy~\citep{kuhn2023semantic}, which clusters the sampled generations by meaning and computes the entropy over clusters using their summed probabilities; Shifting Attention to Relevance (SAR; \citealp{duan-etal-2024-shifting}), which reweights token-level and sequence-level probabilities according to their semantic relevance; Semantic Density~\citep{qiu2024semantic}, which estimates the probability density of the generated answer in the semantic space using the sampled generations and their probabilities; EigenScore~\citep{chen2024inside}, which measures the semantic diversity of the sampled generations via the eigenvalues of the covariance matrix of their sequence-level embeddings; and CoCoA with MTE~\citep{vashurin2025cocoa}, which multiplies the MTE score by the average dissimilarity between the sampled generations and the greedy generation.

  Moreover, we consider UQ methods for black-box LLMs, which rely only on the sampled generations. We evaluate lexical similarity~\citep{fomicheva-etal-2020-unsupervised}, which computes the average ROUGE-L between pairs of sampled generations; Long-text Uncertainty Quantification (LUQ; \citealp{zhang2024luq}), which measures how consistently the sentences of the generated answer are entailed by the other sampled generations; Kernel Language Entropy (KLE; \citealp{nikitin2024kernel}), which computes the von Neumann entropy of a semantic similarity kernel over the sampled generations; and three methods from~\citep{lin2024generating} based on the graph of pairwise semantic similarities between generations: Degree Matrix (DegMat), which uses the average node degree; Eccentricity (Ecc), which measures the spread of the generations in the spectral embedding of the graph; and the sum of eigenvalues of the graph Laplacian (EigVal), which estimates the effective number of semantic clusters.

  Additionally, we consider four supervised state-of-the-art baselines for white-box LLMs. These include two methods based on hidden states: Statement Accuracy Prediction based on Language Model Activations (SAPLMA; \citealp{azaria-mitchell-2023-internal}), which trains an MLP classifier on the hidden states of the final answer token from a single layer; and Supervised Average Token-level Relative Mahalanobis Distance (SATRMD; \citealp{vazhentsev-etal-2025-token}), which trains a linear model on the token-level relative Mahalanobis distances between the hidden states of each answer token and the mean hidden states of the training set. We also evaluate two methods based on attention weights: Lookback Lens~\citep{chuang-etal-2024-lookback}, which trains a linear classifier on the lookback ratios of the answer tokens, i.e., the ratio of attention to the prompt tokens relative to the generated tokens, from all layers and heads; and Trainable Attention-based Dependency (TAD; \citealp{vazhentsev-etal-2025-unconditional}), which trains a linear model on the attention weights from each answer token to its top-$k$ preceding tokens and their probabilities to predict the correctness of each token.

  Finally, we consider two supervised state-of-the-art baselines developed specifically for LLDMs, which are based on the denoising trajectory of a single generation: TraceDet~\citep{chang2026tracedet}, which trains a Transformer classifier on the per-step token entropies across denoising steps to identify the most informative steps; and DynHD~\citep{qian2026dynhdhallucinationdetectiondiffusion}, which trains a model on the denoising dynamics of the generation, including per-step entropies, token changes, and hidden states of intermediate generations, and uses their deviation from the dynamics of correct generations to predict hallucinations. Both methods are trained in the same in-domain setup as the other supervised baselines.

\subsection{Dataset and Generation Statistics}
\label{app:datasets}
  We perform experiments across three representative generation tasks: question answering (\texttt{QA}), text summarization (\texttt{Summ}), and machine translation (\texttt{MT}).
  For QA, we consider four datasets covering diverse domains and reasoning types: MMLU~\citep{hendryckstest2021}, TriviaQA~\citep{joshi-etal-2017-triviaqa}, CoQA~\citep{reddy-etal-2019-coqa}, and GSM8k~\citep{cobbe2021training}.
  For summarization, we use SamSum~\citep{gliwa-etal-2019-samsum} and XSum~\citep{narayan-etal-2018-dont}, representing dialog, news, and extreme summarization settings.
  For machine translation, we evaluate on WMT benchmarks: German--English (WMT19; \citealp{barrault-etal-2019-findings}) and French--English (WMT14; \citealp{bojar-etal-2014-findings}).

  We follow the default inference configuration of FastDLLM~\citep{wu2025fastdllmtrainingfreeaccelerationdiffusion} for all models except DiffusionGemma, for which we use its native default generation configuration. The datasets and prompts are taken from the LM-Polygraph framework\footnote{\url{https://huggingface.co/LM-Polygraph/datasets}}~\citep{fadeeva2023lm}. Dataset statistics and inference hyperparameters are presented in Table~\ref{tab:datasets}.

  \begin{table*}[!ht]
\centering
\resizebox{\textwidth}{!}{\begin{tabular}{c|c|c|c|c|c|c}
\toprule
\textbf{Task} & \textbf{Dataset} & \textbf{\multirowcell{Number of \\test samples}} & \textbf{N-shot} & \textbf{\multirowcell{Max generation \\length}} & \textbf{\multirowcell{Block \\length}} & \textbf{\multirowcell{Num. of diff. \\steps}} \\

\midrule
\multirow{5}{*}{QA} & MMLU & 2000 & 5 & 3 & 3 & \multirow{9}{*}{256} \\
& TriviaQA & 2000 & 5 & 20 & 5 & \\
& CoQA & 2000 & \multirowcell{all preceding\\ questions} & 20 & 5 & \\
& GSM8k & 1319 & 5 & 256 & 32 & \\
\cmidrule{1-3}
\multirow{2}{*}{Summ} & SamSum & 819 & 0 & 128 & 32 & \\
& XSum & 2000 & 0 & 128 & 32 & \\
\cmidrule{1-3}
\multirow{2}{*}{MT} & WMT19 (De-En) & 2000 & 0 & 128 & 32 & \\
& WMT14 (Fr-En) & 2000 & 0 & 128 & 32 & \\
\bottomrule
\end{tabular}
}\caption{\label{tab:datasets} \textbf{Datasets and generation settings.} Block length and the number of diffusion steps apply to LLaDA-1.5, Dream, and ReFusion; DiffusionGemma uses its native default generation configuration. For all datasets, we do not limit the maximum input length. For all texts, we use greedy generation (temperature = $0$); for sampling, we use temperature $1$ to obtain diverse outputs. }\end{table*}

\clearpage
\section{Limitations}
\label{app:limitations}
  While our evaluation covers the most prominent publicly available open-weighted discrete masked and uniform LLDMs at the time of submission, broader evaluation across other diffusion paradigms and model scales remains future work due to the currently limited availability of such models. Our theoretical results rely on standard assumptions that may not always hold exactly in practice, potentially loosening the stated bounds in some settings. Moreover, Theorem~\ref{thm:dissim-lower-bound} holds in expectation over data and random masks and does not directly provide guarantees for individual generations. Finally, the reported computational overhead is specific to the hardware configuration and inference algorithm used in our experiments and may vary across different setups.

\section{Broader Impact}
\label{app:broader_impact}
  This work presents a systematic study of uncertainty quantification in discrete masked and uniform LLDMs, enabling reliable hallucination detection without requiring task-specific labels or repeated sampling. We consider this a meaningful step toward more trustworthy deployment of diffusion-based language models, particularly in safety-critical applications. In our experiments, we use open-weight models and publicly available datasets that are not focused on harmful content. The explored approaches pose no negative societal impact, as they do not rely on sensitive data, user annotations, or other elements that could raise ethical concerns.

\clearpage

\newpage
\section*{NeurIPS Paper Checklist}

\begin{enumerate}

\item {\bf Claims}
    \item[] Question: Do the main claims made in the abstract and introduction accurately reflect the paper's contributions and scope?
    \item[] Answer: \answerYes{} %
    \item[] Justification: Yes, the claims made in the abstract and introduction are precisely supported by the description of the method in Section~\ref{sec:methods} and by the experimental evaluations in Section~\ref{sec:experiments}.

\item {\bf Limitations}
    \item[] Question: Does the paper discuss the limitations of the work performed by the authors?
    \item[] Answer: \answerYes{} %
    \item[] Justification: Yes, the limitations of the work are discussed in Appendix~\ref{app:limitations}.

\item {\bf Theory assumptions and proofs}
    \item[] Question: For each theoretical result, does the paper provide the full set of assumptions and a complete (and correct) proof?
    \item[] Answer: \answerYes{} %
    \item[] Justification: Yes, the paper provides theoretical results in Theorem~\ref{thm:dissim-lower-bound}, with the full set of assumptions in Section~\ref{sec:diff_signals} and complete proof in Appendix~\ref{app:proofs}.

    \item {\bf Experimental result reproducibility}
    \item[] Question: Does the paper fully disclose all the information needed to reproduce the main experimental results of the paper to the extent that it affects the main claims and/or conclusions of the paper (regardless of whether the code and data are provided or not)?
    \item[] Answer: \answerYes{} %
    \item[] Justification: All necessary details for reproducing the experimental results, including hyperparameters, dataset statistics, model checkpoints, hardware configuration, and implementation details, are provided in Section~\ref{sec:experiments} and in Appendix~\ref{app:exp_setup} to ensure full reproducibility.

\item {\bf Open access to data and code}
    \item[] Question: Does the paper provide open access to the data and code, with sufficient instructions to faithfully reproduce the main experimental results, as described in supplemental material?
    \item[] Answer: \answerYes{} %
    \item[] Justification: The implementation code is publicly available at \url{http://lldm-uncertainty.nlpresearch.group/}, and all datasets are publicly available through the LM-Polygraph framework.

\item {\bf Experimental setting/details}
    \item[] Question: Does the paper specify all the training and test details (e.g., data splits, hyperparameters, how they were chosen, type of optimizer) necessary to understand the results?
    \item[] Answer: \answerYes{} %
    \item[] Justification: All relevant details of the testing experiments are provided in Section~\ref{sec:experiments} and Appendix~\ref{app:exp_setup}. No LLM training or fine-tuning is performed in this work; only the supervised UQ baselines are trained, as described in Appendix~\ref{app:implementation}.

\item {\bf Experiment statistical significance}
    \item[] Question: Does the paper report error bars suitably and correctly defined or other appropriate information about the statistical significance of the experiments?
    \item[] Answer: \answerYes{} %
    \item[] Justification: Reliability of the results is ensured by fixing random seeds and using greedy decoding, which leads to deterministic generation. To further demonstrate the robustness of our findings, we conducted experiments across four LLDMs and eight datasets from diverse domains and tasks.

\item {\bf Experiments compute resources}
    \item[] Question: For each experiment, does the paper provide sufficient information on the computer resources (type of compute workers, memory, time of execution) needed to reproduce the experiments?
    \item[] Answer: \answerYes{} %
    \item[] Justification: Details about resources and implementation are provided in Appendix~\ref{app:implementation}.
    
\item {\bf Code of ethics}
    \item[] Question: Does the research conducted in the paper conform, in every respect, with the NeurIPS Code of Ethics \url{https://neurips.cc/public/EthicsGuidelines}?
    \item[] Answer: \answerYes{} %
    \item[] Justification: Yes, the research conducted in this paper conforms in every respect with the NeurIPS Code of Ethics.

\item {\bf Broader impacts}
    \item[] Question: Does the paper discuss both potential positive societal impacts and negative societal impacts of the work performed?
    \item[] Answer: \answerYes{} %
    \item[] Justification: The broader impact of our work is discussed in Appendix~\ref{app:broader_impact}.
    
\item {\bf Safeguards}
    \item[] Question: Does the paper describe safeguards that have been put in place for responsible release of data or models that have a high risk for misuse (e.g., pre-trained language models, image generators, or scraped datasets)?
    \item[] Answer: \answerYes{} %
    \item[] Justification: Our work focuses on hallucination detection in LLMs, which does not introduce new safety risks but rather contributes to mitigating them. No new models or datasets are released as part of this work.

\item {\bf Licenses for existing assets}
    \item[] Question: Are the creators or original owners of assets (e.g., code, data, models), used in the paper, properly credited and are the license and terms of use explicitly mentioned and properly respected?
    \item[] Answer: \answerYes{} %
    \item[] Justification: Yes, we appropriately credit the authors and licenses for all models, methods, and datasets utilized in this paper.

\item {\bf New assets}
    \item[] Question: Are new assets introduced in the paper well documented and is the documentation provided alongside the assets?
    \item[] Answer: \answerNA{} %
    \item[] Justification: The paper does not release new datasets or models.

\item {\bf Crowdsourcing and research with human subjects}
    \item[] Question: For crowdsourcing experiments and research with human subjects, does the paper include the full text of instructions given to participants and screenshots, if applicable, as well as details about compensation (if any)? 
    \item[] Answer: \answerNA{} %
    \item[] Justification: This work does not involve crowdsourcing or research with human subjects.

\item {\bf Institutional review board (IRB) approvals or equivalent for research with human subjects}
    \item[] Question: Does the paper describe potential risks incurred by study participants, whether such risks were disclosed to the subjects, and whether Institutional Review Board (IRB) approvals (or an equivalent approval/review based on the requirements of your country or institution) were obtained?
    \item[] Answer: \answerNA{} %
    \item[] Justification: Not applicable, as this work does not involve human subjects or study participants.

\item {\bf Declaration of LLM usage}
    \item[] Question: Does the paper describe the usage of LLMs if it is an important, original, or non-standard component of the core methods in this research? Note that if the LLM is used only for writing, editing, or formatting purposes and does \emph{not} impact the core methodology, scientific rigor, or originality of the research, declaration is not required.
    \item[] Answer: \answerNA{} %
    \item[] Justification: LLMs are the subject of analysis in this work but are not used as part of the research methodology in a novel or non-standard way, in accordance with the LLM policy in the NeurIPS handbook. Their usage is limited to evaluation and does not constitute an original component of the core methodology.

\end{enumerate}

\end{document}